\documentclass[letterpaper]{article} % DO NOT CHANGE THIS
\usepackage{aaai2026}  % DO NOT CHANGE THIS
\usepackage{times}  % DO NOT CHANGE THIS
\usepackage{helvet}  % DO NOT CHANGE THIS
\usepackage{courier}  % DO NOT CHANGE THIS
\usepackage[hyphens]{url}  % DO NOT CHANGE THIS
\usepackage{graphicx} % DO NOT CHANGE THIS
\usepackage{natbib}  % DO NOT CHANGE THIS AND DO NOT ADD ANY OPTIONS TO IT  
\usepackage{caption} % DO NOT CHANGE THIS AND DO NOT ADD ANY OPTIONS TO IT
\usepackage{algorithm}
\usepackage{algorithmic}

\usepackage{newfloat}
\usepackage{listings}
\DeclareCaptionStyle{ruled}{labelfont=normalfont,labelsep=colon,strut=off} % DO NOT CHANGE THIS
\floatstyle{ruled}
\newfloat{listing}{tb}{lst}{}
\floatname{listing}{Listing}
\title{A\(^2\)LC: Active and Automated Label Correction for Semantic Segmentation}
\author{
    Youjin Jeon\equalcontrib, 
    Kyusik Cho\equalcontrib, 
    Suhan Woo, 
    Euntai Kim\thanks{Corresponding author.}
}
\affiliations{
    Yonsei University\\%, Seoul, Republic of Korea\\
    \{ngbrjyj, ks.cho, wsh112, etkim\}@yonsei.ac.kr
}

\usepackage{bibentry}
\usepackage{amsmath}
\usepackage{tikz}
\usepackage{amssymb}
\usepackage{booktabs}
\usepackage{multirow}
\usepackage{pifont}
\usepackage{subcaption}
\usepackage{makecell}
\usepackage{kotex} 
\usepackage{pgfplots} 

\begin{document}

\maketitle

\begin{abstract} 
Active Label Correction (ALC) has emerged as a promising solution to the high cost and error-prone nature of manual pixel-wise annotation in semantic segmentation, by actively identifying and correcting mislabeled data. Although recent work has improved correction efficiency by generating pseudo-labels using foundation models, substantial inefficiencies still remain. In this paper, we introduce A$^2$LC, an Active and Automated Label Correction framework for semantic segmentation, where manual and automatic correction stages operate in a cascaded manner. Specifically, the automatic correction stage leverages human feedback to extend label corrections beyond the queried samples, thereby maximizing cost efficiency. In addition, we introduce an adaptively balanced acquisition function that emphasizes underrepresented tail classes, working in strong synergy with the automatic correction stage. Extensive experiments on Cityscapes and PASCAL VOC 2012 demonstrate that A$^2$LC significantly outperforms previous state-of-the-art methods. Notably, A$^2$LC exhibits high efficiency by outperforming previous methods with only 20\% of their budget, and shows strong effectiveness by achieving a 27.23\% performance gain under the same budget on Cityscapes.
\end{abstract}

% Uncomment the following to link to your code, datasets, an extended version or similar.
% You must keep this block between (not within) the abstract and the main body of the paper.
\begin{links}
    \link{Code}{https://github.com/ngbrjyj/A2LC}
    % \link{Datasets}{https://aaai.org/example/datasets}
    % \link{Extended version}{https://aaai.org/example/extended-version}
\end{links}

\section{Introduction}

\begin{figure}[t]
\centering
\begin{tabular}{c @{\hskip +0.5em} c @{\hskip -0.15em} c}
 & \multicolumn{2}{c}{%
 \hspace{1em}%
    \begin{tikzpicture}
        \node[draw=none, fill=none] {
            \begin{tabular}{l@{\hskip 0.6em} l@{\hskip 0.9em} l@{\hskip 0.6em} l@{\hskip 0.6em} l@{\hskip 0.45em} l}  
            % ALC
            {\tikz{\filldraw[draw=red, fill=red, thick] (90:2.2pt) -- (162:2.2pt) -- 
            (234:2.2pt) -- (306:2.2pt) -- (18:2.2pt) -- cycle;}} & \small \textbf{ALC} &
            % A2LC
            {\tikz{\filldraw[draw=blue, fill=blue, thick] (0,0) rectangle (3.5pt,3.5pt);}} 
            & \small \textbf{A\(^2\)LC} &
            % Fully supervised
            {\tikz{\draw[violet, densely dashed, ultra thick] (0,0) -- (6pt,0);}} 
            & \small \textbf{Fully Supervised}
            \end{tabular}
        };
    \end{tikzpicture}
} \\
\raisebox{+.63\height}{\rotatebox{90}{\small \textbf{Cityscapes}}} &
% ---------------- Cityscapes Data ----------------
\begin{tikzpicture}
    \begin{axis}[
        xlabel={\# of Clicks ($\times 10^4$)},
        ylabel={Data mIoU (\%)},
        width=4.45cm, height=3.4cm,
        ymin=45, ymax=95, xmin=0, xmax=10,
        xtick={0,...,10}, ytick={45,55,...,95},
        xlabel style={yshift=0.27cm}, ylabel style={xshift=0.05cm, yshift=-0.65cm}, 
        label style={font=\scriptsize}, tick label style={font=\scriptsize}
    ]
    \addplot[red, thick, mark=pentagon*, mark size=1.65pt] table[col sep=comma, x=x, y=alc_data]{csv/third_figure.csv};
    \addplot[blue, thick, mark=square*, mark size=1.25pt] table[col sep=comma, x=x, y=ours_data]{csv/third_figure.csv};
    \end{axis}
\end{tikzpicture} &
% ---------------- Cityscapes Model ----------------
\begin{tikzpicture}
    \begin{axis}[
        xlabel={\# of Clicks ($\times 10^4$)},
        ylabel={Model mIoU (\%)},
        width=4.45cm, height=3.4cm,
        ymin=45, ymax=79, xmin=0, xmax=10,
        xtick={0,...,10}, ytick={45,55,...,75,85},
        xlabel style={yshift=0.27cm}, ylabel style={xshift=0.05cm, yshift=-0.65cm}, 
        label style={font=\scriptsize}, tick label style={font=\scriptsize}
    ]
    \addplot[red, thick, mark=pentagon*, mark size=1.65pt] table[col sep=comma, x=x, y=alc_model]{csv/third_figure.csv};
    \addplot[blue, thick, mark=square*, mark size=1.25pt] table[col sep=comma, x=x, y=ours_model]{csv/third_figure.csv};
    \addplot[violet, densely dashed, ultra thick] coordinates {(0,76.31) (10,76.31)};
    \end{axis}
\end{tikzpicture}
\\[-0.25em]
\raisebox{+0.85\height}{\rotatebox{90}{\small \textbf{PASCAL}}} &
% ---------------- Pascal Data ----------------
\begin{tikzpicture}
    \begin{axis}[
        xlabel={\# of Clicks ($\times 10^3$)},
        ylabel={Data mIoU (\%)},
        width=4.45cm, height=3.4cm,
        ymin=55, ymax=95, xmin=0, xmax=10,
        xtick={0,...,10}, ytick={55,65,...,95},
        xlabel style={yshift=0.27cm}, ylabel style={xshift=0.05cm, yshift=-0.65cm}, 
        label style={font=\scriptsize}, tick label style={font=\scriptsize}
    ]
    \addplot[red, thick, mark=pentagon*, mark size=1.65pt] table[col sep=comma, x=x, y=alc_data]{csv/second_figure.csv};
    \addplot[blue, thick, mark=square*, mark size=1.25pt] table[col sep=comma, x=x, y=ours_data]{csv/second_figure.csv};
    \end{axis}
\end{tikzpicture} &
% ---------------- Pascal Model ----------------
\begin{tikzpicture}
    \begin{axis}[
        xlabel={\# of Clicks ($\times 10^3$)},
        ylabel={Model mIoU (\%)},
        width=4.45cm, height=3.4cm,
        ymin=55, ymax=75, xmin=0, xmax=10,
        xtick={0,...,10}, ytick={55,60,...,75},
        xlabel style={yshift=0.27cm}, ylabel style={xshift=0.05cm, yshift=-0.65cm}, 
        label style={font=\scriptsize}, tick label style={font=\scriptsize}
    ]
    \addplot[red, thick, mark=pentagon*, mark size=1.65pt] table[col sep=comma, x=x, y=alc_model]{csv/second_figure.csv};
    \addplot[blue, thick, mark=square*, mark size=1.25pt] table[col sep=comma, x=x, y=ours_model]{csv/second_figure.csv};
    \addplot[violet, densely dashed, ultra thick] coordinates {(0,72.95) (10,72.95)};
    \end{axis}
\end{tikzpicture}
\end{tabular}

\caption{Effectiveness of A\(^2\)LC framework. 
Our model demonstrates clear effectiveness, saturating by round 6 on Cityscapes and round 5 on PASCAL.
At these points, it outperforms the baseline by 11.81 and 4.44 mIoU and reaches 94\% and 95\% of full supervision.
}
\label{fig:10r_figures}
\end{figure}

The high cost and labor-intensive nature of pixel-wise annotation pose a significant challenge in semantic segmentation, where large scale labeled datasets are essential for deep learning models~\cite{edlund2021livecell, li2023systematic}. Furthermore, such datasets frequently contain noisy labels due to manual annotation errors, which hinder the learning process of deep neural networks~\cite{li2023semi, plank2022problem}. In response to this challenge, active label correction methods~\cite{kim2024active, Kim2022ActiveLC, kremer2018robust} offer a promising solution, as they iteratively identify and refine noisy labels through human review.

While active label correction has been actively explored in other vision tasks, its application to semantic segmentation remains limited due to the inherent difficulty of correcting pixel-wise labels. Recent advancements in foundation models~\cite{luddecke2022image, ren2024grounded, zou2023segment} have helped mitigate these challenges by enabling zero-shot prediction, generating high quality pseudo-labels that can serve as a better starting point for label refinement. Building on these advancements, Kim et al.~\shortcite{kim2024active} developed a framework that generates pseudo-labels using a foundation model for semantic segmentation and actively selects uncertain pixels for correction by annotators.

This framework effectively incorporates a pretrained deep neural network into the conventional active learning paradigm, establishing an efficient pipeline that significantly reduces the need for human intervention. Nevertheless, we found that their integration of deep learning techniques remained incomplete, particularly in addressing redundancy during correction. Specifically, prior methods rely exclusively on annotators for label correction, and the corrected labels are not generalizable beyond the selected samples. Consequently, many similar pixels are repeatedly sampled and corrected, leading to inefficient querying and increased annotation costs. This redundancy ultimately slows down performance improvement, as illustrated in Figure~\ref{fig:10r_figures}.

To address this limitation, we propose an additional automatic correction stage that exploits annotator feedback to perform label correction beyond the queried samples. Unlike conventional pipelines that rely solely on manual correction, our method incorporates a secondary correction source, the Label Correction Module (LCM), forming a dual-source correction framework. The primary advantage of the LCM lies in its ability to maximize the utility of human-provided corrections without incurring additional annotation overhead, thereby improving cost-efficiency. Furthermore, we introduce an adaptively balanced acquisition function, designed based on the pseudo-label statistics. This function prioritizes tail classes during the querying process, thereby enabling efficient mitigation of the class imbalance problem. Notably, our proposed acquisition function synergizes well with the LCM, as it directs annotator effort toward a subset of samples targeted for tail classes. This, in turn, guides the LCM's automatic label correction toward improved class balance, enhancing overall annotation efficiency. Extensive experiments on Cityscapes~\cite{cordts2016cityscapes} and PASCAL VOC 2012~\cite{everingham2015pascal} demonstrate that A\(^2\)LC significantly outperforms previous state-of-the-art methods, as shown in Figure~\ref{fig:10r_figures}. 

Our contributions can be summarized as follows:
\begin{itemize}
    \item
    We develop a label correction module that enables automatic correction and maximizes the utility of human effort, significantly improving cost efficiency.
    \item
    By introducing an adaptive class weight and integrating it into the acquisition function, we effectively address the class imbalance challenge.
    \item
    A\(^2\)LC achieves high efficiency by outperforming previous methods while using only 20\% and 60\% of their budget on Cityscapes and PASCAL, respectively.  
    \item 
    A\(^2\)LC demonstrates strong effectiveness by yielding 27.23\% and 14.30\% performance improvements under equivalent budget constraints on Cityscapes and PASCAL, respectively. 
\end{itemize}

\section{Related Works}
\subsection{Active Label Correction}
Active label correction aims to construct a clean labeled dataset with minimal cost by selectively querying and correcting samples that are likely to be mislabeled.
As cost is directly tied to the query unit, extensive research has explored its optimal label query unit. 
Early studies used image-level querying~\cite{chen2024think, yang2023towards}, but this led to excessive labeling overhead as entire images required annotation regardless of the informativeness of specific regions. 
To improve efficiency, pixel-level querying~\cite{didari2024bayesian, ma2025integrating, ruckin2024active, ruckin2024semi, schachtsiek2023class} was introduced, though it still incurred substantial costs. 
A two-step querying approach~\cite{ribeiro2024uncertainty, van2024active} refined this by first selecting uncertain images and then querying only the most informative pixels. 
More recently, superpixel-level querying~\cite{cai2021revisiting, ge2024esa, hwang2023active, kim2023adaptive, Wu_2025_WACV} has been explored to balance granularity and efficiency. 
Due to the practical challenge of obtaining perfectly clean labels, active label correction has been increasingly applied across various tasks, e.g., text classification~\cite{hou2025co, taneja2024can, wang2024simulation}, image classification~\cite{ekambaram2016active, beck2024beyond, bernhardt2022active, khanal2024active, Kim2022ActiveLC, kremer2018robust, li2022improving, 6413805, wang2024simulation}, and semantic segmentation~\cite{kim2024active}. 
Despite the integration of ALC into semantic segmentation, prior studies have not optimized annotation cost efficiency, as they assign individual annotation costs to each similar pixel. 
We effectively address this issue by proposing a module that performs automatic correction for similar pixels.
To the best of our knowledge, this is the first study to introduce automatic label correction in semantic segmentation. 

\subsection{Acquisition Function}
Since accurately identifying mislabeled data is essential to avoid unnecessary expenditure, the acquisition function is crucial for the success of active label correction. 
It can be broadly categorized into three types. 
Uncertainty-based functions prioritize sampling data with high uncertainty, e.g., BALD~\cite{gal2017deep}, Confidence~\cite{wang2014new}, Entropy~\cite{safaei2024entropic}, Margin~\cite{roth2006margin},   MeanSTD~\cite{7789580}. 
Diversity-based functions aim to sample data that effectively represent the overall dataset, e.g., BADGE~\cite{Ash2020Deep}, Cluster-Margin~\cite{Citovsky2021BatchAL}, CoreSet~\cite{sener2018active}. 
Hybrid functions consider both uncertainty and diversity, e.g., CLAUS~\cite{rana2023hybrid}, which performs frame selection based on model uncertainty and diverse video sampling through deep clustering. 
In this work, we effectively sample informative data by incorporating our newly defined adaptive class weight into the acquisition function, which dynamically modulates diversity consideration based on dataset imbalance.

\section{A\(^2\)LC Framework}

\subsection{Overview}
\begin{figure*}[!t]
\centering
\includegraphics[width=0.87\linewidth]{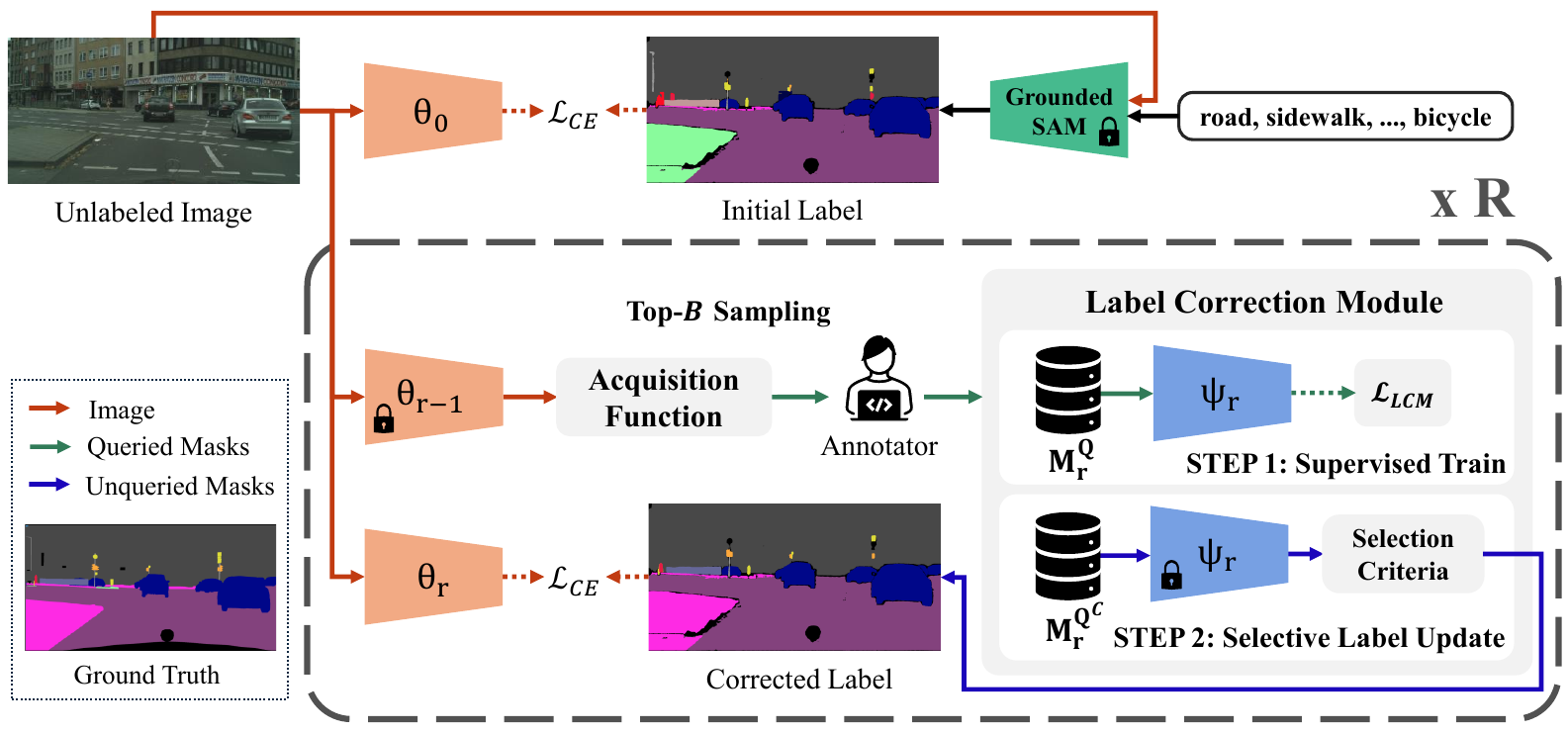}
\caption{Overview of A\(^2\)LC framework.
We execute Grounded SAM on unlabeled images to generate initial pseudo-labels. For each of the \(R\) rounds, the model \(\theta_{r-1}\) trained with pseudo-labels selects \(B\) masks via the acquisition function to query for manual correction. Then, the model \(\psi_r\) in Label Correction Module (LCM) is trained using the queried masks and corrects the labels of unqueried masks. After both manual and automatic corrections, the pseudo-labels and model are updated, completing a single round of the correction cycle.
}
\label{fig:overview_figure}
\end{figure*}
An overview of our proposed framework is shown in Figure~\ref{fig:overview_figure}.
We construct the initial dataset \mbox{\(\mathcal{D}_0 = \{(X_i, \hat{Y_i})\}_{i=1}^{N}\)} using zero-shot predictions from the foundation model~\cite{ren2024grounded} as pseudo-labels, and initialize the model \(\theta_0\).
The set of masks generated by SAM, denoted as \(\mathcal{M}_0 = \{(m_i, \hat{y_i})\}_{i=1}^{M}\), serves as the correction unit. 
At round \(r\), all masks are scored based on an acquisition function computed by the model \(\theta_{r-1}\).
The top-\(B\) most informative masks are sampled and their true labels are obtained by querying an annotator.
Whereas prior research concluded the correction process at this manual stage, we cascade an additional automatic correction stage for further refinement. 
Specifically, the model \(\psi_r\) in Label Correction Module (LCM) is trained using the set of queried masks \(\mathcal{M}_r^{Q}\), which contains only clean labels obtained from the manual correction stage.
Subsequently, the trained \(\psi_r\) performs automatic label correction by inferring the unqueried masks \(\mathcal{M}_r^{Q^c} = \mathcal{M}_{r-1} \setminus \mathcal{M}_r^Q\). 
This secondary correction is applied only to the masks that satisfy all selection criteria.
Once both the manual and automatic correction stages are completed, the model \(\theta_r\) is updated using the corrected dataset \(\mathcal{D}_r\).
The overall procedure is then iteratively repeated until the allocated budget is exhausted. 

The following sections introduce two key techniques that constitute our Active and Automated Label Correction (A\(^2\)LC) framework.  
First, we describe the automatic label correction stage, which plays a crucial role in enhancing cost efficiency.
Next, we introduce a novel acquisition function that strengthens both correction stages through data-driven balanced sampling.

\subsection{Label Correction Module}

\begin{figure}[!t]
\centering
\includegraphics[width=0.9\linewidth]{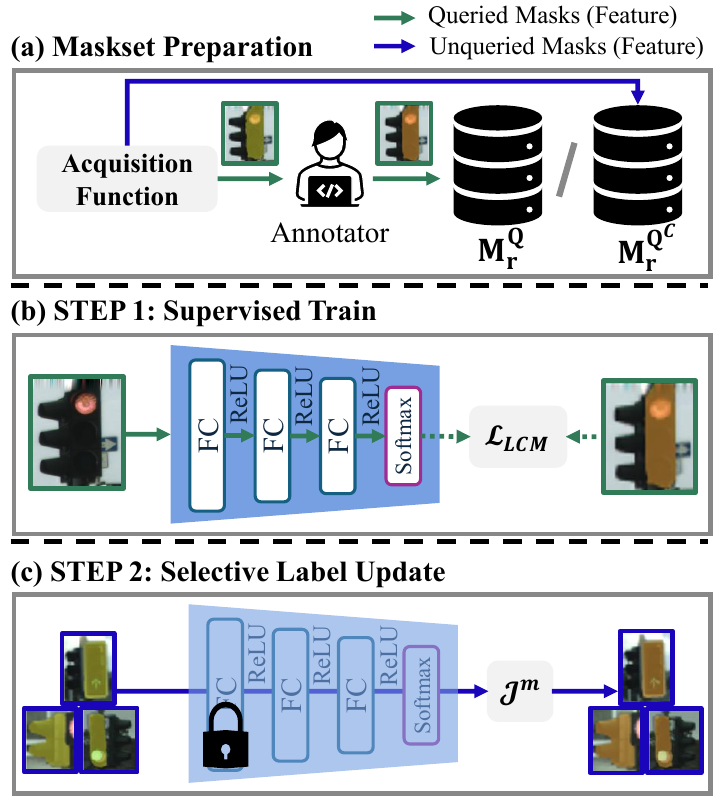}
\caption{LCM pipeline. 
% LCM operates in two sequential stages:
LCM operates in two sequential steps: the model is first trained with accurately labeled masks queried in the current round, followed by correcting potentially mislabeled masks based on the model’s predictions. 
Here, a \textit{traffic light (orange)} mask, mislabeled as a \textit{traffic sign (yellow)}, was corrected by the annotator. This manually corrected mask is then used to train the model, enabling automatic refinement of similar cases. As a result, three additional similar masks were automatically corrected.
}
\label{fig:lcm_architecture}
\end{figure}

An analysis of prior studies~\cite{xu2023progressive, xu2021instance} demonstrates that mislabeling predominantly arises between semantically similar categories, such as \textit{traffic sign} and \textit{traffic light}, rather than between disparate classes.
This occurs because the foundation model, trained on broad and diverse domains, often fails to capture the subtle distinctions required in fine-grained domains.
As a result, it inevitably generates biased pseudo-labels, and pixels with similar features naturally cluster within comparable acquisition score ranges.
This leads to redundant queries of similar masks across rounds, resulting in inefficient use of annotation resources. 
To address this issue, we propose a module that utilizes annotator-provided true labels to automatically correct noisy labels exhibiting similar features, enabling the correction of a more diverse set of noisy labels in the following rounds. 
This module is executed through two sequential steps in each round: the learning phase and the correction phase. 
The overall pipeline of our proposed LCM is illustrated in Figure~\ref{fig:lcm_architecture}.

\subsubsection{Components of LCM.} 
The LCM is built upon two core components: a learnable model and selection criteria.
The model \(\psi_r\) at round \(r\) takes the mask feature \(f_{\theta_{r-1}}(m)\) as input, defined as
\begin{equation}
f_{\theta_{r-1}}(m) = \frac{1}{|m|} \sum_{x \in m} f_{\theta_{r-1}}(x),
\end{equation}
where the feature \(f_{\theta_{r-1}}(x)\) of pixel \(x\) is extracted by the model \(\theta\) trained at round \(r-1\), and \(f_{\theta_{r-1}}(m)\) denotes the average feature over all pixels \( x \in m \). 
For simplicity, we refer to mask feature \(f_{\theta_{r-1}}(m)\) as \(m\).
A four-layer fully connected network with ReLU activations reduces input features to 256, 128, and 64 dimensions, followed by a final classification layer with softmax activation to produce class probabilities.
The selection criteria are designed to identify predictions with high reliability, so that automatic label updates are applied exclusively to the subset of confidently inferred labels.

\subsubsection{STEP 1: Supervised Train with Queried Masks.}
In the learning phase, the masks queried in round \(r\), denoted as \(m \in \mathcal{M}_r^Q\) are used to train \(\psi_r\) under supervision, with a weighted cross-entropy loss: 
\begin{equation}
\mathcal{L}_{LCM} := \lambda_{y(m)} \cdot \mathcal{L}_{CE}(y(m), \psi_r(m)), 
\label{eq:ce_loss3}
\end{equation}
where \( y(m) \) denotes the annotator-corrected label of mask \( m \), and \( \lambda_{y(m)} \) represents the corresponding normalized class weight.
The weight \( \lambda_{k} \) for class \(k\) is defined as \(\lambda_k := \frac{N / N_k}{\sum_{c \in \mathcal{C}} N / N_{c}}\), where \(N_k\) and \(N\) denote the number of training samples in class \(k\) and the total number of training samples, respectively.

\subsubsection{STEP 2: Selective Label Update for Unqueried Masks.}
\begin{figure}[!t]
\centering
\begin{tabular}{c  c @{\hskip 0.17em} c @{\hskip 0.17em} c @{\hskip 0.17em} c @{\hskip 0.17em} c}
\raisebox{+.6\height}{\rotatebox{90}{\small \textit{\textbf{Car}}}} &
% ---------- 1st Row ----------
\begin{subfigure}[t]{1.4cm}
    \centering
    \includegraphics[width=1.4cm, height=1cm]{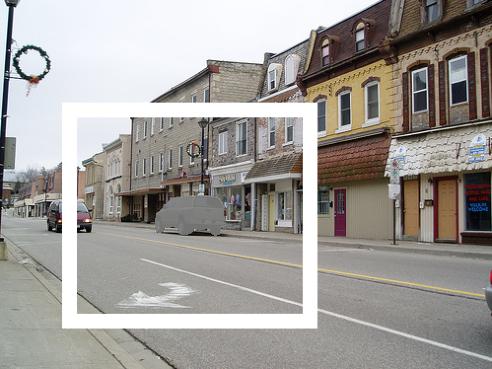}
\end{subfigure} &
\begin{subfigure}[t]{1.4cm}
    \centering
    \includegraphics[width=1.4cm, height=1cm]{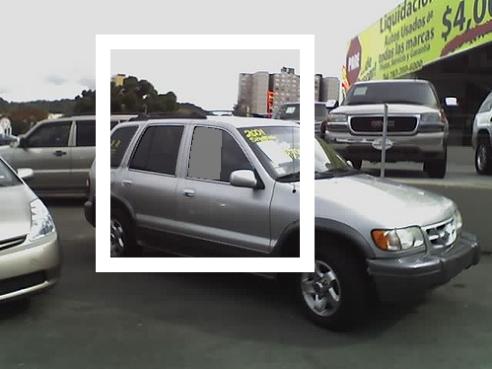}
\end{subfigure} &
\begin{subfigure}[t]{1.4cm}
    \centering
    \includegraphics[width=1.4cm, height=1cm]{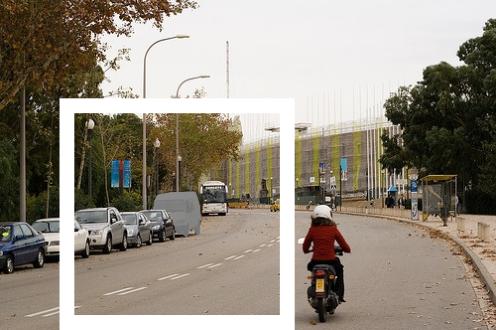}
\end{subfigure} &
\begin{subfigure}[t]{1.4cm}
    \centering
    \includegraphics[width=1.4cm, height=1cm]{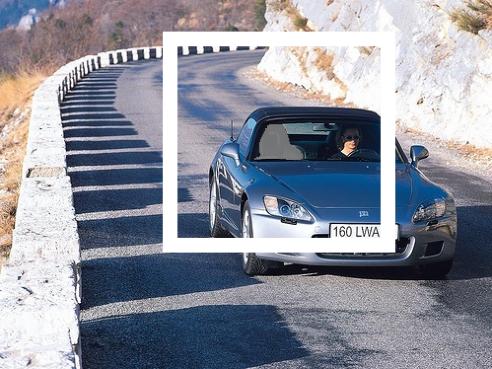}
\end{subfigure} &
\begin{subfigure}[t]{1.4cm}
    \centering
    \includegraphics[width=1.4cm, height=1cm]{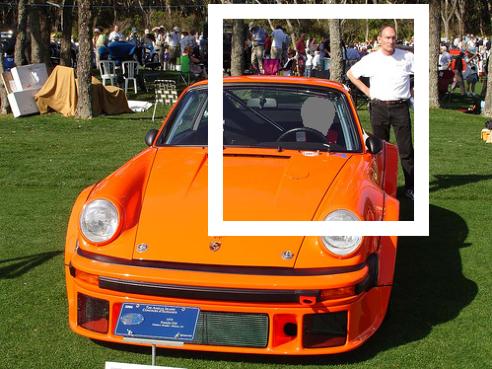}
\end{subfigure}
\\[0em]
\raisebox{+.3\height}{\rotatebox{90}{\small \textit{\textbf{Train}}}} &
% ---------- 2nd Row ----------
\begin{subfigure}[t]{1.4cm}
    \centering
    \includegraphics[width=1.4cm, height=1cm]{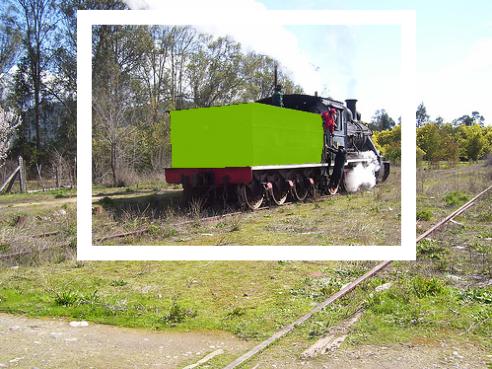}
\end{subfigure} &
\begin{subfigure}[t]{1.4cm}
    \centering
    \includegraphics[width=1.4cm, height=1cm]{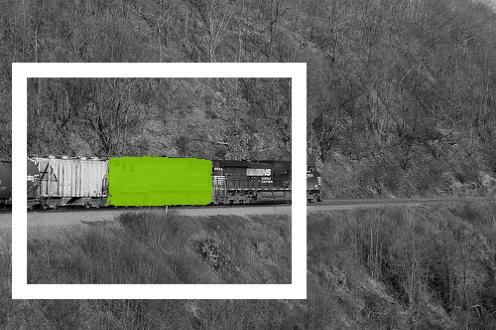}
\end{subfigure} &
\begin{subfigure}[t]{1.4cm}
    \centering
    \includegraphics[width=1.4cm, height=1cm]{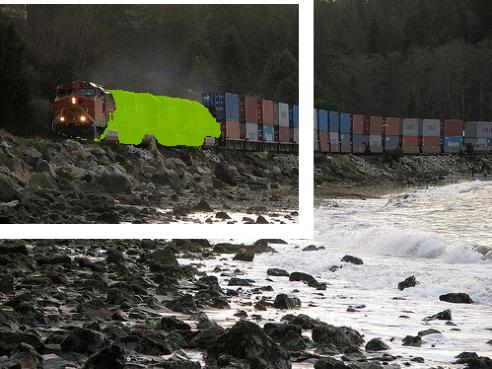}
\end{subfigure} &
\begin{subfigure}[t]{1.4cm}
    \centering
    \includegraphics[width=1.4cm, height=1cm]{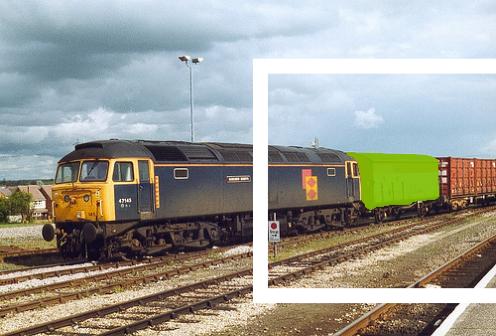}
\end{subfigure} &
\begin{subfigure}[t]{1.4cm}
    \centering
    \includegraphics[width=1.4cm, height=1cm]{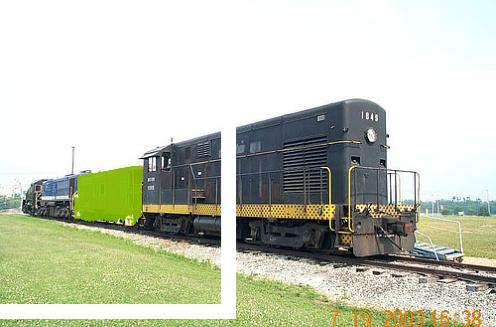}
\end{subfigure}
\end{tabular}
\caption{Visualization of automatically corrected masks for the \textit{car} and \textit{train} classes.}
\label{fig:lcm_masks_vis}
\end{figure}

In the correction phase, the model trained during the learning phase attempts to correct all unqueried masks. 
Label prediction is performed solely on the masks \(m \in \mathcal{M}_r^{Q^c}\) unqueried in round \(r\), using model \(\psi_r\) trained in the previous step. 
However, direct label updates to predicted classes pose a risk of erroneous correction, as class imbalance in \(\mathcal{M}_r^Q\) can lead to biased learning. 
To mitigate this, we introduce a conservative label update strategy based on three selection criteria. 
A mask is eligible for automatic correction only if all three of the following conditions are satisfied: (1) the model's prediction confidence exceeds a confidence threshold \(\tau\), (2) the predicted class is not among the tail classes, and (3) the original pseudo-label is also not from a tail class. 
These three constraints jointly serve to minimize the risk of incorrect label updates. 
We define the overall selection criteria \(\mathcal{J}^m\) for mask \(m\) as:
\begin{equation}
\mathcal{J}^m 
= \mathcal{J}_1^m \land \mathcal{J}_2^m \land \mathcal{J}_3^m,
\label{eq:criteria}
\end{equation}
\begin{align}
\mathcal{J}_1^m  
&= \mathbb{I} \Big( \max_{c \in \mathcal{C}} \psi_r(c; m) \geq \tau \Big), \label{eq:first_judge} \\
\mathcal{J}_2^m  
&= \mathbb{I} \Big( \hat{y}_{\psi_r}(m) \notin 
\{ c  \: | \: \text{rank}(c) \geq \left( 1 - \alpha \right) \cdot  |\mathcal{C}|, \: c \in \mathcal{C} \} \Big), \label{eq:second_judge} \\
\mathcal{J}_3^m
&= \mathbb{I} \Big( \hat{y}(m) \neq \underset{c \in \mathcal{C}}{\arg\max} \: \text{rank}(c) \Big), \label{eq:third_judge}
\end{align}
where \(\mathbb{I}(\cdot)\) is the indicator function, \(\text{rank}(c)\) represents the index of class \(c\) after sorting all classes by sample count in descending order, and \(\alpha\) is a hyperparameter specifying the threshold ratio for tail classes (e.g., \(\alpha = 0.5\) corresponds to the bottom 50\% of classes). 
We set \(\alpha = 0.5\), and \(\tau\) is initialized at 0.99 and incrementally increased for careful correction over time. 
Figure~\ref{fig:lcm_masks_vis} visualizes the masks corrected by LCM.

\subsection{Adaptively Balanced Acquisition Function}

\begin{figure*}
\centering
\begin{tikzpicture}
    \node[draw=none, fill=none] at (0,0) {
        \begin{tabular}{l l l l}  
        {\tikz{\draw[red, fill=red!90] (0,0) rectangle (6pt,6pt);}} & \textbf{ALC} &
        {\tikz{\draw[blue, fill=blue!90] (0,0) rectangle (6pt,6pt);}} & \textbf{A\(^2\)LC}
    \end{tabular}
    };
\end{tikzpicture}

\vspace{-0.5em}

\begin{tikzpicture}
    \begin{axis}[
        width=\textwidth, 
        height=0.26\textwidth,
        ybar,
        symbolic x coords={
            light, rider, {motor.}, train, bicycle,
            bus, truck, sign, pole, terrain, person, fence,
            wall, sky, {side.}, car, {veg.}, {buil.}, road},
        xtick=data,
        x tick label style={rotate=45, anchor=east},
        ymin=0, ymax=12500, 
        bar width=6pt, 
        x=0.78cm,
        enlarge x limits=0.04,
        ymajorgrids=true,
        grid style=dashed,
        after end axis/.code={
            \node[anchor=north west, font=\itshape] at (rel axis cs:0.02, -0.35) {← Tail};
            \node[anchor=north east, font=\itshape] at (rel axis cs:0.98, -0.35) {Head →};
        }
    ]
    \addplot[color=red, fill=red!90] coordinates {
        (light, 176) (rider, 626) ({motor.}, 466) (train, 192) (bicycle, 629)
        (bus, 656) (truck, 1740) (sign, 2228) (pole, 4167) (terrain, 1314)
        (person, 2062) (fence, 2939) (wall, 2121) (sky, 1161) ({side.}, 5664)
        (car, 4202) ({veg.}, 5132) ({buil.}, 10511) (road, 4014)
    };
    \addplot[color=blue, fill=blue!90] coordinates {
        (light, 3858) (rider, 3144) ({motor.}, 2593) (train, 215) (bicycle, 4987)
        (bus, 1280) (truck, 2794) (sign, 5350) (pole, 7599) (terrain, 1471)
        (person, 4311) (fence, 3353) (wall, 2188) (sky, 285) ({side.}, 3087)
        (car, 973) ({veg.}, 1179) ({buil.}, 978) (road, 355)
    };            
    \end{axis}
\end{tikzpicture}
\caption{Class distribution of sampled data. 
The x-axis shows classes sorted by pseudo-label frequency, and the y-axis shows the number of sampled masks. 
Unlike the baseline, which largely concentrates on head classes, our proposed ABC acquisition function substantially increases the sampling of tail classes}
\label{fig:city_class_samples}
\end{figure*}

Based on the assumption that the label of a pixel predicted with low confidence by the model is more likely to be incorrect, numerous acquisition functions~\cite {kim2024active, roth2006margin, safaei2024entropic, wang2014new} have been proposed. However, directly applying the conventional acquisition functions results in marginal performance improvements due to a sampling bias toward head classes, as illustrated in Figure~\ref{fig:city_class_samples}. 
Moreover, since the model within the LCM is trained on the data sampled by the acquisition function, class imbalance in the sampled data can also degrade the overall effectiveness of the LCM.
To address this, we introduce an adaptive class weight \(w(x)\), applied multiplicatively to the base acquisition function.
This factor dynamically adjusts the degree of balanced sampling according to the pseudo-label statistics at each round. The weight consists of two components: the class rarity score and dataset imbalance score. For simplicity, we refer to \(\mathcal{M}_{r-1}\) as \(\mathcal{M}\) and \(\theta_{r-1}\) as \(\theta\).

\subsubsection{Class Rarity Score.}
The class rarity score \(\hat{w}(x)\) is designed to prioritize tail classes and is defined as follows:
\begin{equation}
\hat{w}(x) := \frac{\min_{c \in \mathcal{C}} \left| \{ x' \in \mathcal{M} : \hat{y}(x') = c \} \right|}{\left| \{ x' \in \mathcal{M} : \hat{y}(x') =\hat{y}(x) \} \right|},
\label{eq:tail}
\end{equation}
where \(\hat{y}(x)\) is the pseudo-label assigned to pixel \(x\), and \(\mathcal{C}\) denotes total classes.
\(\hat{w}(x)\) is a class-wise value assigned to each pixel within the range (0, 1], where pixels labeled with less frequent classes in the dataset receive values closer to 1, and those with more frequent classes receive values closer to 0. In particular, a pixel labeled with the rarest class is assigned a class rarity score of 1.

\subsubsection{Dataset Imbalance Score.}
While the previously defined 
\(\hat{w}(x)\) effectively emphasizes tail classes by reflecting their relative frequency within the dataset, it does not consider the overall distributional characteristics of the dataset.
As a result, relying solely on \(\hat{w}(x)\) may lead to suboptimal weighting, particularly in datasets with low levels of class imbalance.
To address this limitation, we introduce a dataset imbalance score that modulates the strength of \(\hat{w}(x)\) based on the dataset class distribution.
We measure dataset imbalance using the KL divergence between the dataset’s pixel distribution \(\mathbb{P}_{\text{dist}}\) and a uniform distribution \(\mathbb{U}_{\text{dist}}\):
\begin{equation}
\text{KL}(\mathbb{P}_{\text{dist}} \parallel \mathbb{U}_{\text{dist}}) = \sum_{c \in C} \mathbb{P}(c) \log \frac{\mathbb{P}(c)}{\mathbb{U}(c)},
\label{eq:kl}
\end{equation}
where \(\mathbb{P}(c) = \frac{\left| \{ x \in \mathcal{M} : \hat{y}(x) = c \} \right|}{\left| \{ x \in \mathcal{M} \} \right|}\) quantifies the proportion of pixels  \(x \in \mathcal{M}\) whose pseudo-label \(\hat{y}(x)\) belongs to class \(c\), \(\mathbb{U}(c) = \frac{1}{|C|}\) represents the uniform class distribution, and \(C\) denotes the total classes.

\subsubsection{Adaptive Class Weight.}
Finally, we define an adaptive class weight 
\(w(x)\) that emphasizes tail class pixels based on current pseudo-label statistics updated at every round:
\begin{equation}
w(x) 
:= \hat{w}(x) ^{ \text{KL}^3(\mathbb{P}_{\text{dist}} \parallel \mathbb{U}_{\text{dist}})},
\label{eq:classweight}
\end{equation} 
ensuring \(w(x) \in (0, 1]\).
The magnitude of \(\text{KL}(\mathbb{P}_{\text{dist}} \parallel \mathbb{U}_{\text{dist}})\) controls the strength of \(\hat{w}(x)\), where its larger value in the highly imbalanced datasets amplifies \(\hat{w}(x)\), guiding the correction process toward tail classes, whereas the smaller value in mildly imbalanced datasets suppresses \(\hat{w}(x)\), reducing its effect.

\subsubsection{Adaptively Balanced CIL.}
Confidence In Label (CIL) formulates the label quality of each pixel as the model's predictive confidence for a given label~\cite{lad2023estimating}. 
To encode the CIL of a mask into a single representative pixel, Kim et al.~\shortcite{kim2024active} introduce a SIM acquisition function by computing a weighted sum of the CILs of pixels within each mask based on their cosine similarity. 
Building upon this, we propose Adaptively Balanced CIL (ABC), which addresses the class imbalance issue by incorporating the adaptive class weight \(w(x)\) formulated in Eq.~(\ref{eq:classweight}).
\begin{align}
a_{\text{CIL}}(x;\theta) 
&:= 1 - f_\theta(\hat{y}(x); x), \label{cil} \\[0.4em]
a_{\text{ABC}}(x;\theta)
&:= w(x) \cdot a_{\text{CIL}}(x;\theta). \label{eq:abc}
\end{align}
The acquisition function for each mask is defined as the weighted sum of pixel-level acquisition scores \(a_{\text{ABC}}(x;\theta)\).
\begin{align}
a_{\text{ABC}}(m;\theta) 
:= \sum_{x \in m} \frac{f_{\theta}(x) \cdot f_{\theta}(m')}
{\|f_{\theta}(x)\| \|f_{\theta}(m')\|} \cdot a_{\text{ABC}}(x;\theta),
\label{eq:ABC}
\end{align}
where the \mbox{\(m' := \{ x \in m : y_\theta(x) = \mathcal{D}_\theta(m) \}\)} denotes the subset of pixels \(x\) in mask \(m\) whose predicted label \(y_\theta(x)\) matches the pseudo dominant label \(\mathcal{D}_\theta(m)\). 
The pseudo dominant label \(\mathcal{D}_\theta(m) = \arg\max_{c \in \mathcal{C}} \left| \{ x \in m : y_\theta(x) = c \} \right|\) is defined as the most frequently predicted class by the model \(\theta\) within mask \(m\)~\cite{kim2023adaptive, kim2024active}. 
The feature \(f_{\theta}(x)\) is the model representation at pixel \(x\), and \(f_{\theta}(m')\) is the average feature over \(m'\). 
Finally, the top-\(B\) most informative  masks, \(\mathcal{M}_r^Q\), are sampled through the ABC acquisition function as follows:
\begin{equation}
\mathcal{M}_r^Q := \arg\max\nolimits^{B}_{m \in \mathcal{M}} a_{\text{ABC}}(m;\theta),
\label{eq:topb}
\end{equation}
where the superscript \(B\) denotes the budget assigned for the current round \(r\).

\section{Experiments}
\begin{table*}[!t]
\centering
\begin{tabular}{c @{\hskip 7pt} c @{\hskip 5pt} c c @{\hskip 8pt} c @{\hskip 8pt} c @{\hskip 8pt} c @{\hskip 8pt} c @{\hskip 8pt} c}
\toprule
\textbf{Dataset} & \textbf{mIoU (\%)} & \textbf{Methods} & \textbf{Init.} & \textbf{1R} & \textbf{2R} & \textbf{3R} & \textbf{4R} & \textbf{5R} \\
\midrule
\raisebox{-3pt}{\multirow{4}{*}{Cityscapes}}
 &  \multirow{2}{*}{Data}
 & ALC & 50.68\(_{\pm 0.00}\) & 62.28\(_{\pm 1.25}\) & 65.36\(_{\pm 0.56}\) & 66.19\(_{\pm 0.60}\) & 66.87\(_{\pm 0.29}\) & 67.01\(_{\pm 0.30}\) \\
                             &                      & \textbf{A\(^2\)LC (Ours)} & 
                             50.68\(_{\pm 0.00}\) & 
                             \textbf{70.01\(_{\pm 1.05}\)} & 
                             \textbf{75.60\(_{\pm 0.58}\)} & 
                             \textbf{80.84\(_{\pm 0.53}\)} & 
                             \textbf{82.69\(_{\pm 0.92}\)} & 
                             \textbf{85.26\(_{\pm 0.55}\)} \\
\cmidrule(lr){2-9}
                             & \multirow{2}{*}{Model} & ALC & 51.55\(_{\pm 0.71}\) & 56.61\(_{\pm 0.42}\) & 58.27\(_{\pm 0.63}\) & 58.58\(_{\pm 0.08}\) & 58.52\(_{\pm 0.26}\) & 58.59\(_{\pm 0.05}\) \\
                             &                        & \textbf{A\(^2\)LC (Ours)} & 
                             51.55\(_{\pm 0.71}\) & 
                             \textbf{60.83\(_{\pm 0.66}\)} & 
                             \textbf{63.89\(_{\pm 0.44}\)} & 
                             \textbf{67.50\(_{\pm 0.31}\)} & 
                             \textbf{68.87\(_{\pm 0.78}\)} & 
                             \textbf{70.51\(_{\pm 0.32}\)} \\
\midrule
\raisebox{-3pt}{\multirow{4}{*}{PASCAL}} & \multirow{2}{*}{Data} & ALC & 58.63\(_{\pm 0.00}\) & \textbf{68.19\(_{\pm 0.34}\)} & 72.72\(_{\pm 0.16}\) & 74.84\(_{\pm 0.29}\) & 76.41\(_{\pm 0.70}\) & 77.06\(_{\pm 0.73}\) \\
                         &                      & \textbf{A\(^2\)LC (Ours)} & 
                         58.63\(_{\pm 0.00}\)  & 
                         67.49\(_{\pm 2.03}\) & 
                         \textbf{74.88\(_{\pm 1.45}\)} & 
                         \textbf{80.88\(_{\pm 0.83}\)} & 
                         \textbf{84.81\(_{\pm 0.48}\)} & 
                         \textbf{88.08\(_{\pm 0.44}\)} \\
\cmidrule(lr){2-9}
                         & \multirow{2}{*}{Model} & ALC & 56.94\(_{\pm 0.44}\) & \textbf{62.11\(_{\pm 0.61}\)} & \textbf{64.12\(_{\pm 0.31}\)} & 64.15\(_{\pm 0.68}\) & 65.00\(_{\pm 0.33}\) & 65.48\(_{\pm 0.83}\) \\
                         &                        & \textbf{A\(^2\)LC (Ours)} & 
                         56.94\(_{\pm 0.44}\) & 
                         60.87\(_{\pm 2.93}\) & 
                         64.08\(_{\pm 2.43}\) & 
                         \textbf{66.45\(_{\pm 0.95}\)} & 
                         \textbf{67.76\(_{\pm 0.06}\)} & 
                         \textbf{68.42\(_{\pm 0.87}\)} \\
\bottomrule
\end{tabular}
\caption{Quantitative results across multiple rounds. `Init.' represents the performance of the initial pseudo-labels generated by Grounded SAM, before any label correction is performed.}
\label{tab:ss_5r}
\end{table*}

\subsection{Experimental Setup}

\subsubsection{Baselines.}
We compare our framework primarily with ALC~\cite{kim2024active}, the first state-of-the-art approach to introduce active label correction for semantic segmentation, as well as with state-of-the-art segmentation methods incorporating active learning, namely Spx~\cite{cai2021revisiting}, MerSpx~\cite{kim2023adaptive}, and MulSpx~\cite{hwang2023active}. 
Our proposed acquisition function is evaluated with a broad range of acquisition functions, including Random, Entropy~\cite{safaei2024entropic}, Margin~\cite{roth2006margin}, CIL~\cite{lad2023estimating}, LCIL~\cite{lad2023estimating}, AIoU~\cite{rottmann2023automated}, BvSB, and ClassBal from~\cite{cai2021revisiting}, MerSpx~\cite{kim2023adaptive}, and SIM~\cite{kim2024active}.

\begin{table}[!t]
\centering
\begin{tabular}{cc}
        \toprule
        \textbf{Methods} & \textbf{Model mIoU (\%)} \\
        \midrule
        Fully Supervised & 76.31 \\
        \midrule
        Spx~\cite{cai2021revisiting} & 63.77 \\ 
        MerSpx~\cite{kim2023adaptive} & 66.53 \\ 
        MulSpx~\cite{hwang2023active} & 66.60 \\ 
        ALC~\cite{kim2024active} & 70.71 \\ 
        \textbf {A\(^2\)LC (Ours)} & \textbf{72.37} \\ 
        \bottomrule
\end{tabular}
\caption{Quantitative comparison with state-of-the-art superpixel-based active learning methods.}
\label{tab:ss_base}
\end{table}

\subsubsection{Evaluation Metrics.}

Following previous work~\cite{kim2024active}, we adopt mean Intersection over Union (mIoU) as the primary evaluation metric.
This metric is used in two distinct contexts: Data mIoU and Model mIoU.
Data mIoU denotes the mIoU between pseudo-labels and the ground truth, whereas Model mIoU refers to the mIoU between model predictions and the ground truth on the validation set.
In addition, we employ Overall Accuracy (OA) to evaluate the accuracy of LCM by measuring the correctness of its automatically corrected masks.

\subsubsection{Correction Details.}
A\(^2\)LC achieves further efficiency through two key strategies: (1) mask-level correction and (2) non-redundant correction. 
Previous studies~\cite{kim2024active} perform pixel-level correction by sampling highly unconfident pixels, querying them for ground truth labels, and propagating the labels to similar pixels, referred to as label expansion. 
However, this label expansion technique can lead to erroneous corrections, as many pixels depend on a single queried pixel.
We address this limitation in a simple but effective manner by performing mask-level correction, where the mask is the prediction unit of the foundation model~\cite{ren2024grounded} employed in generating initial pseudo-labels. 
We further enhance efficiency through non-redundant correction, by excluding previously queried masks from the set of candidates for label correction. 
Non-redundant correction is highly synergistic with mask-level correction in enhancing cost efficiency.
As mask-level correction aggregates multiple pixels into a single query, the number of unqueried pixels becomes relatively small after several rounds of label correction, which maximizes the effect of non-redundant correction.

\begin{table}[!t]
\centering
\begin{tabular}{c @{\hskip 5pt} c @{\hskip 11pt} c @{\hskip 12pt} c}

\toprule
 &  & \multicolumn{2}{c}{\textbf{OA (\%)}} \\
\textbf{Dataset} & \textbf{\# of Corrected Masks} & Before & After \\

\midrule
Cityscapes & 22,105 & 19.13 & \textbf{60.99} \\  
PASCAL & 1,629 & 4.85 & \textbf{82.26} \\  

\bottomrule
\end{tabular}
\caption{Accuracy analysis of label correction module.
}
\label{tab:LCM}
\end{table}

\begin{table}[!t]
\centering
\begin{tabular}{c @{\hskip -3pt} c @{\hskip 3pt} c}

\toprule
\textbf{Acquisition Function} & \textbf{Cityscapes} & \textbf{PASCAL} \\

\midrule
Init.  & 50.68 & 58.63 \\

\midrule
Random  & 63.78 & 68.84 \\
Entropy~\cite{safaei2024entropic}  & 58.12 & 59.95 \\
Margin~\cite{roth2006margin}  & 62.68 & 58.72 \\
CIL~\cite{lad2023estimating}  & 61.51 & 59.29 \\
LCIL~\cite{kim2024active}  & 62.63 & 59.80 \\
AIoU~\cite{rottmann2023automated}  & 60.59 & 59.15 \\
BvSB~\cite{cai2021revisiting}  & 64.35 & 59.60 \\
ClassBal~\cite{cai2021revisiting}  & 65.65 & 60.56 \\
MerSpx~\cite{kim2023adaptive}  & 65.82 & 60.13 \\
SIM~\cite{kim2024active}  & 78.98 & \textbf{87.32} \\

\textbf{ABC (Ours)} & \textbf{84.92} & 87.24 \\ 

\bottomrule
\end{tabular}
\caption{Comparative analysis of adaptively balanced acquisition function (Data mIoU, \%).}
\label{tab:ABC}
\end{table}

\begin{figure}[!t]
\centering
\begin{tabular}{c @{\hskip 0.1em} c @{\hskip 0.1em} c @{\hskip 0.1em} c}
% ---------- 1st Row ----------
\begin{subfigure}[t]{2.0cm}
    \centering
    \includegraphics[width=2.0cm, height=1.2cm]
    {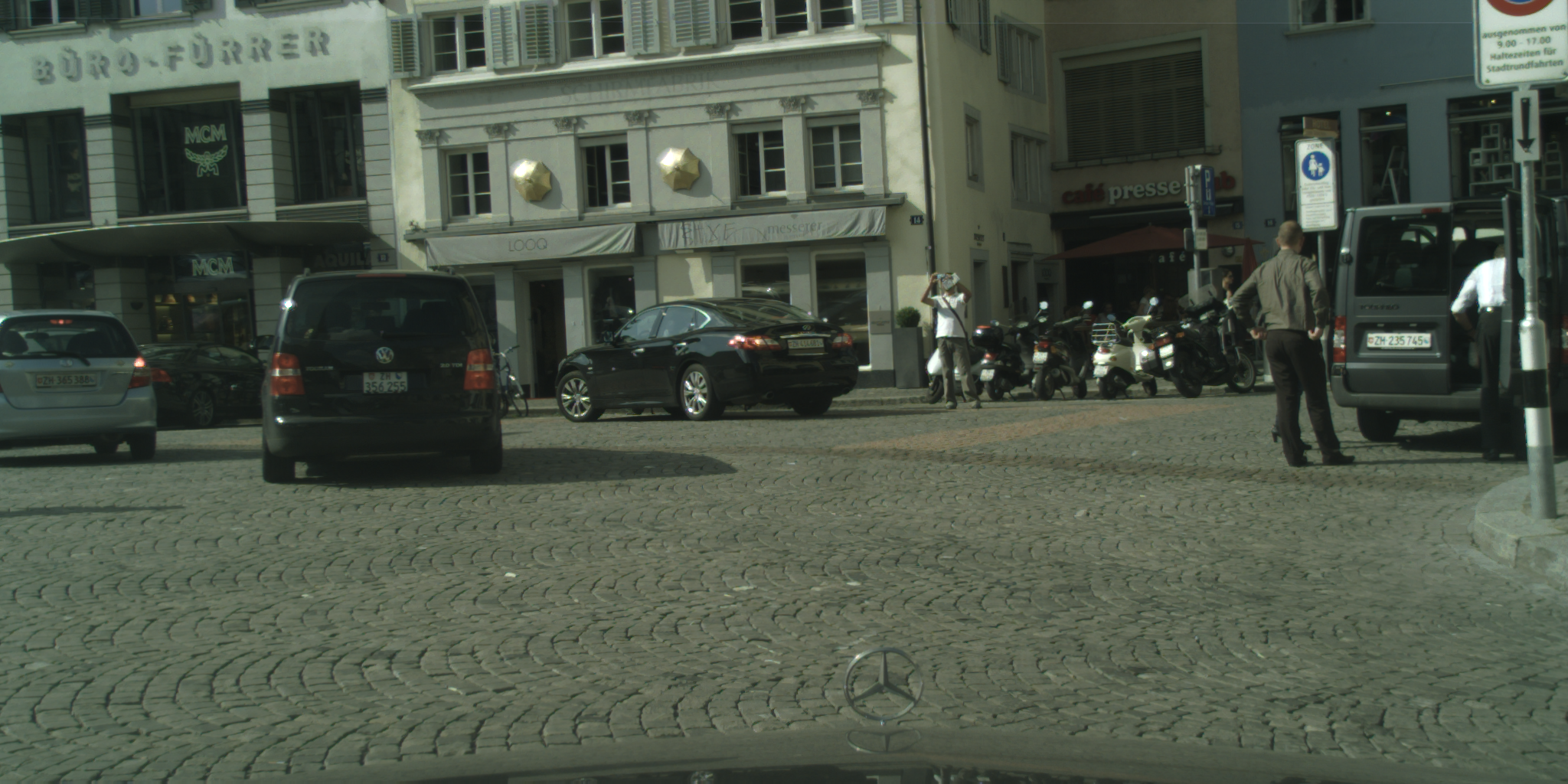}
\end{subfigure} &
\begin{subfigure}[t]{2.0cm}
    \centering
    \includegraphics[width=2.0cm, height=1.2cm]{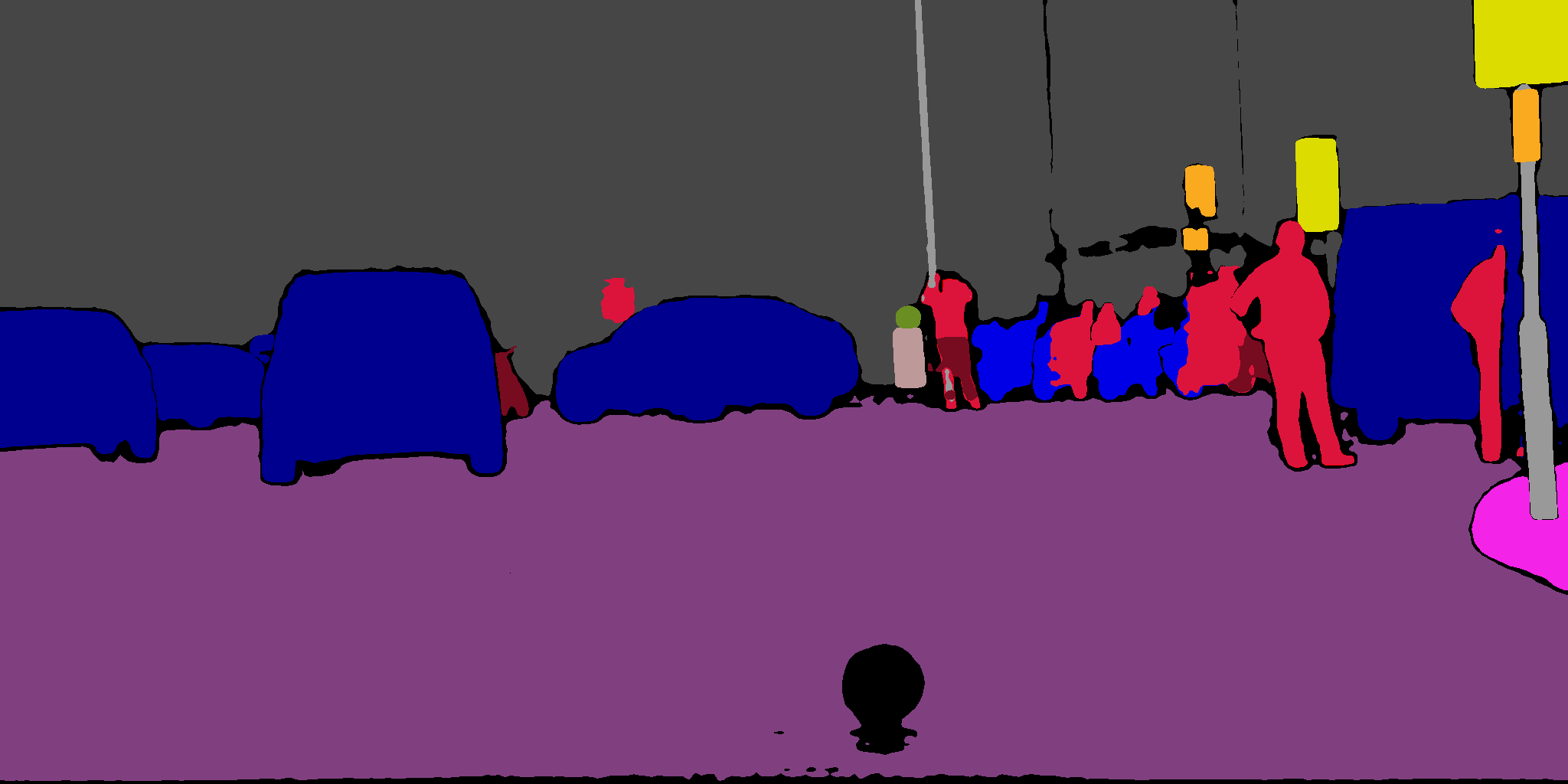}
\end{subfigure} &
\begin{subfigure}[t]{2.0cm}
    \centering
    \includegraphics[width=2.0cm, height=1.2cm]{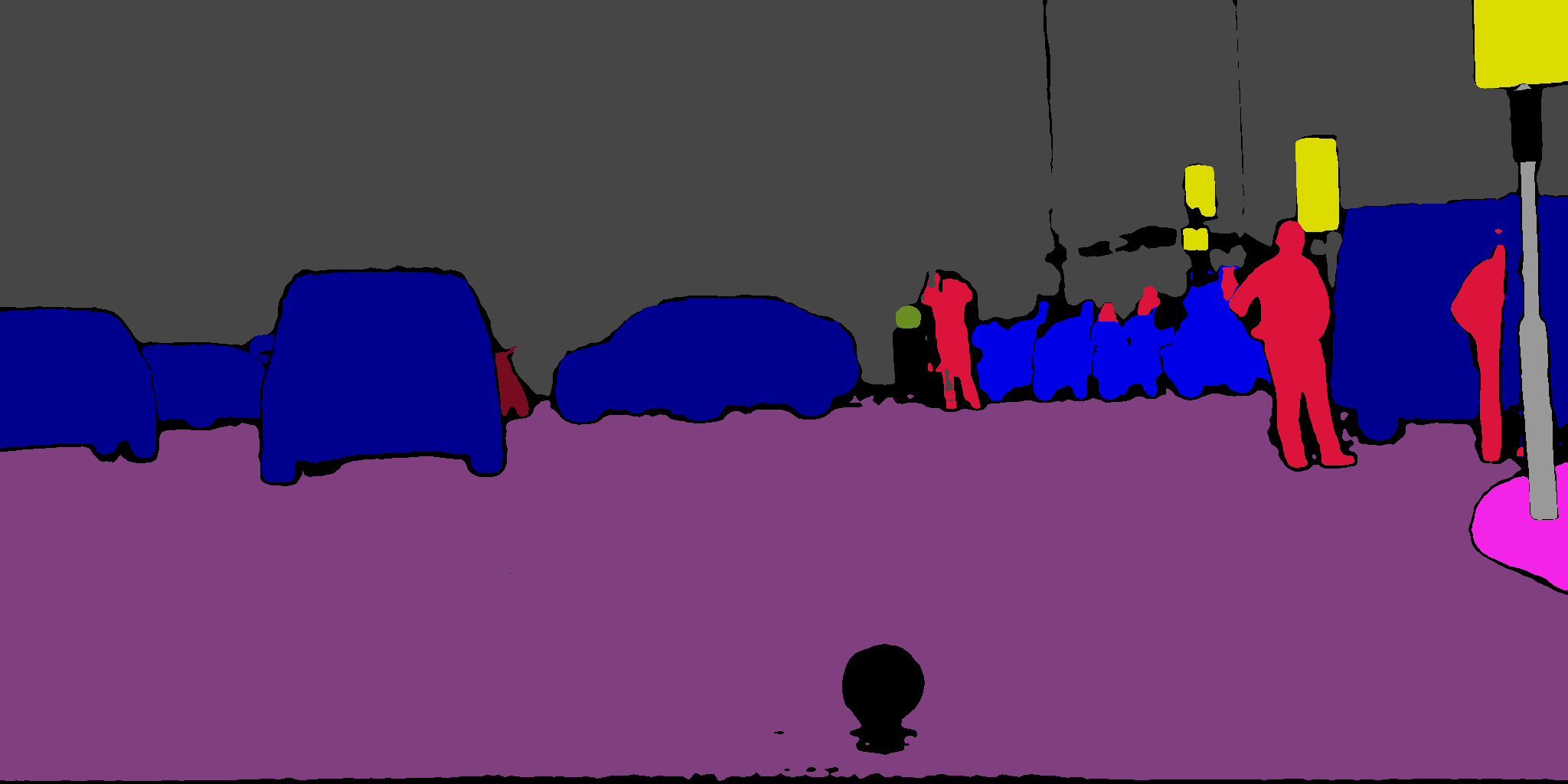}
\end{subfigure} &
\begin{subfigure}[t]{2.0cm}
    \centering
    \includegraphics[width=2.0cm, height=1.2cm]{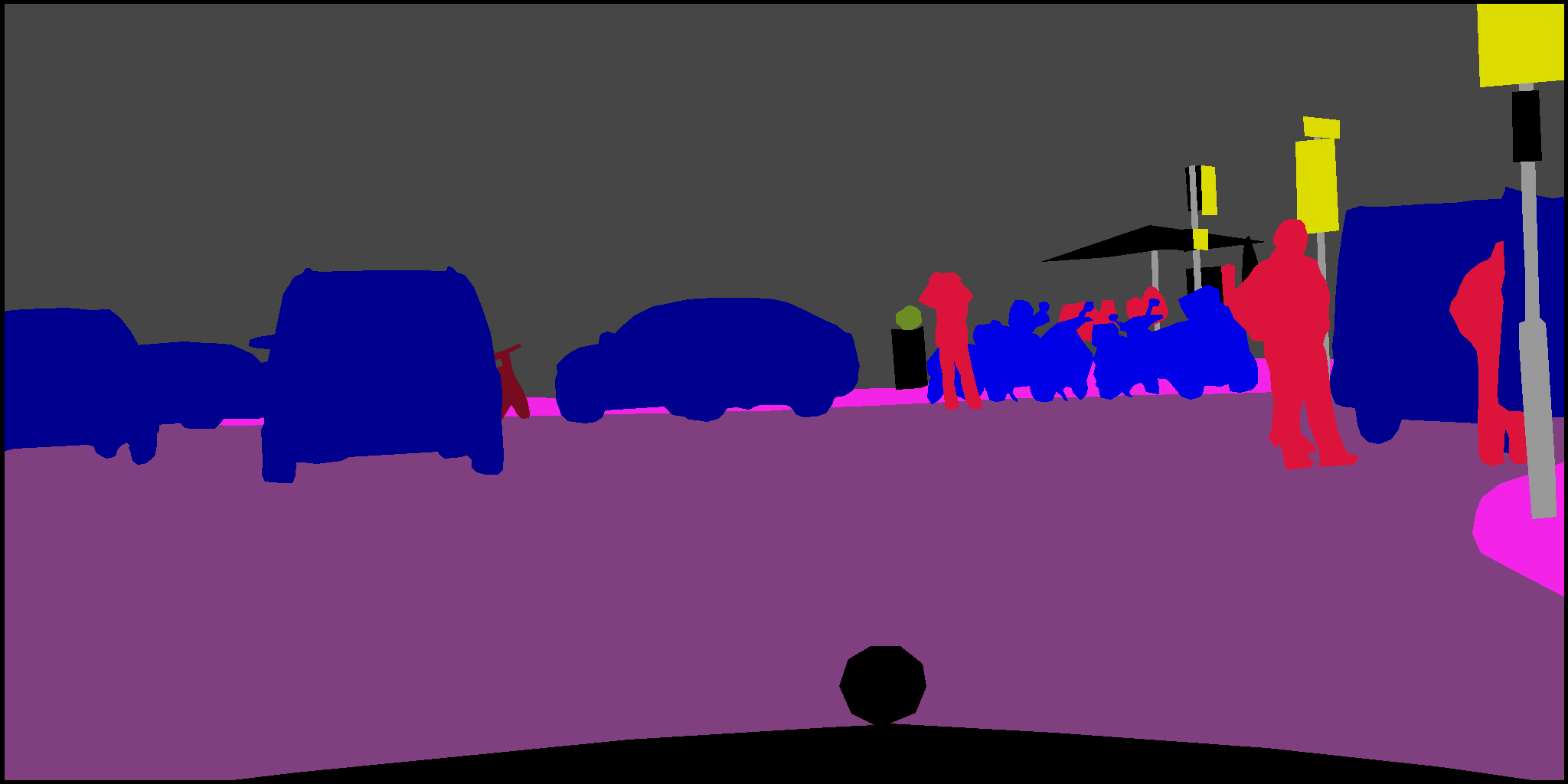}
\end{subfigure} 
\\[0em]
% ---------- 2nd Row ----------
\begin{subfigure}[t]{2.0cm}
    \centering
    \includegraphics[width=2.0cm, height=1.2cm]
    {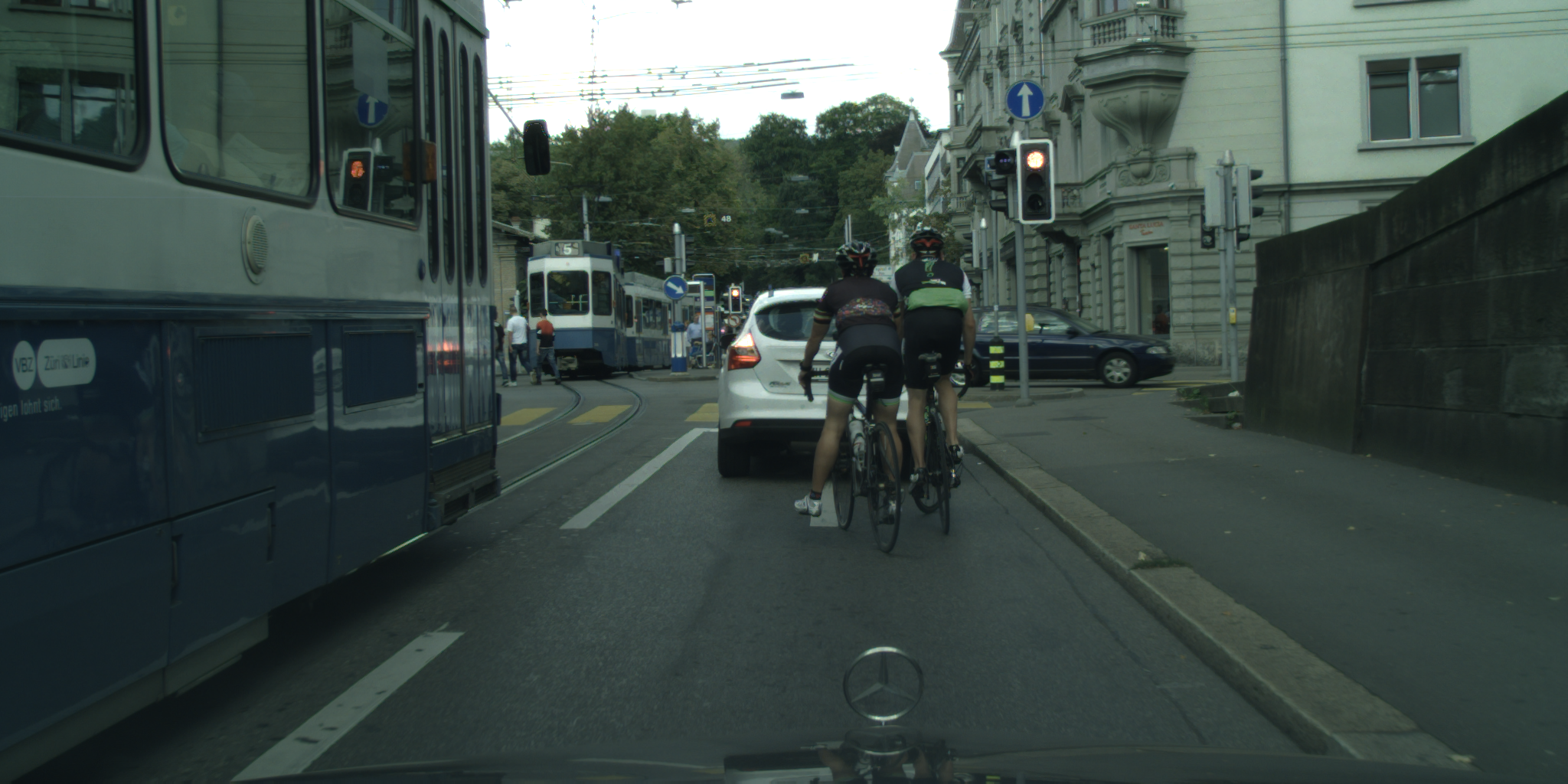}
    \captionsetup{skip=2pt}
    \caption*{(a)}
\end{subfigure} &
\begin{subfigure}[t]{2.0cm}
    \centering
    \includegraphics[width=2.0cm, height=1.2cm]{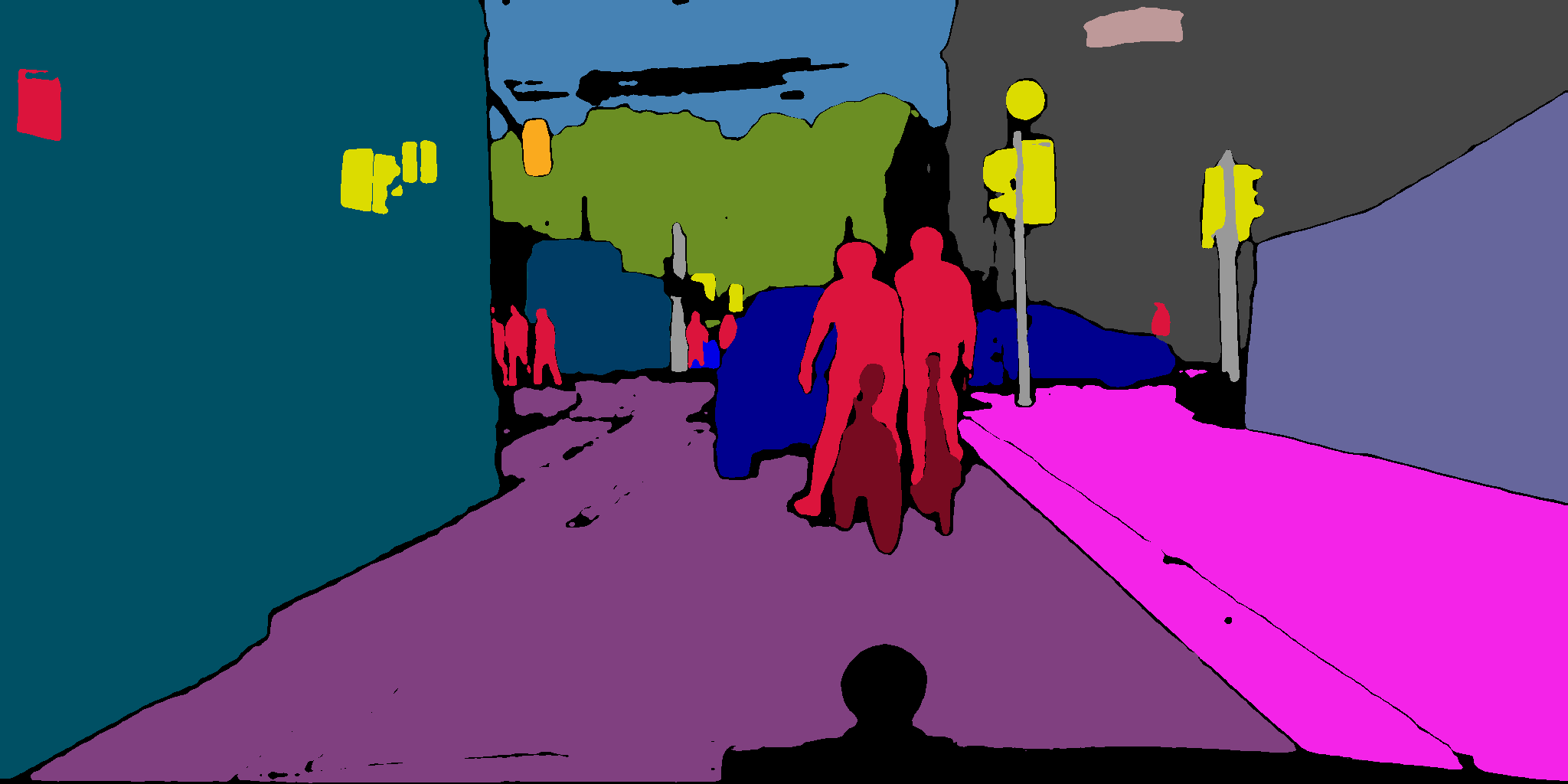}
    \captionsetup{skip=2pt}
    \caption*{(b)}
\end{subfigure} &
\begin{subfigure}[t]{2.0cm}
    \centering
    \includegraphics[width=2.0cm, height=1.2cm]{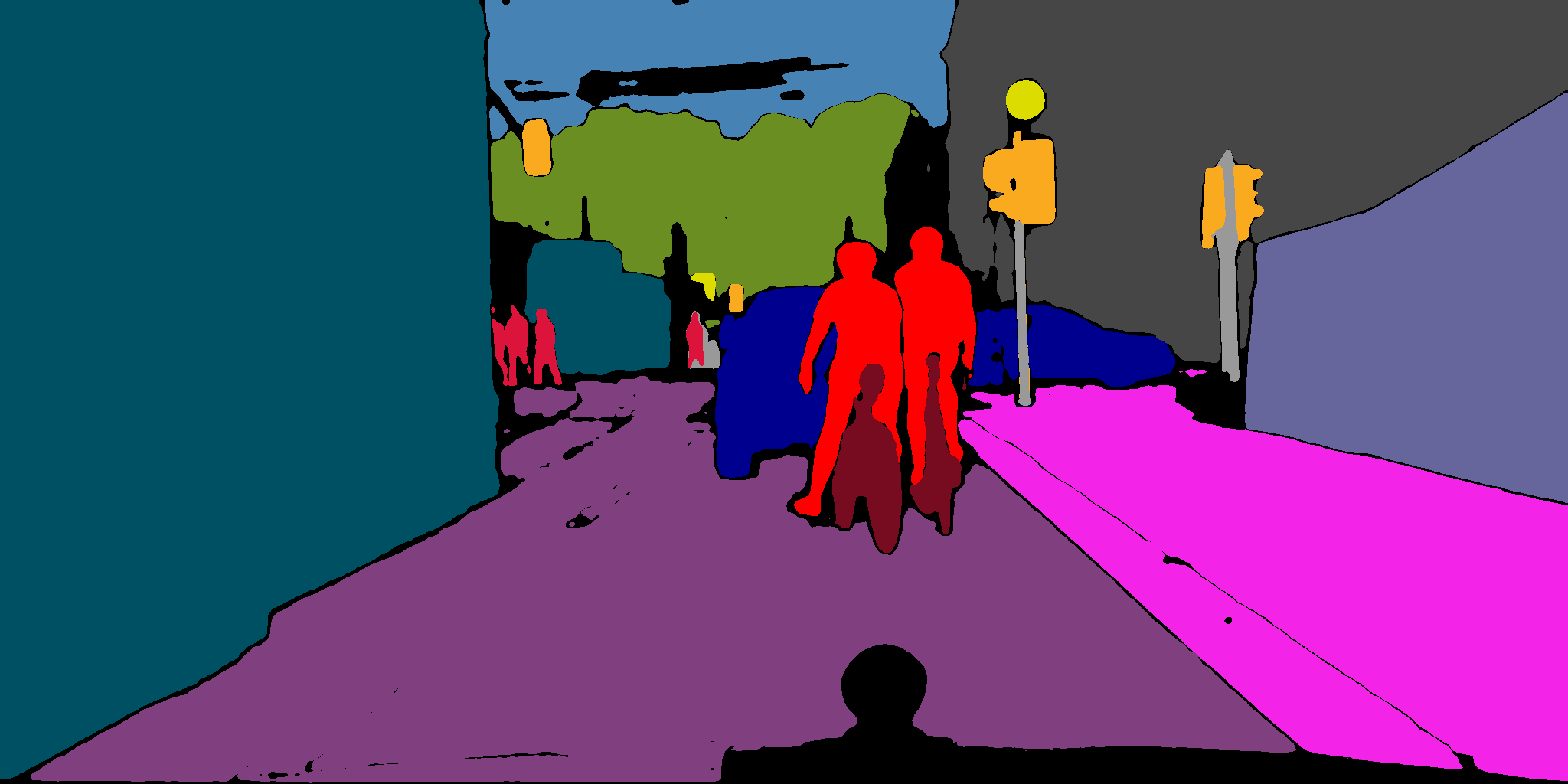}
    \captionsetup{skip=2pt}
    \caption*{(c)}
\end{subfigure} &
\begin{subfigure}[t]{2.0cm}
    \centering
    \includegraphics[width=2.0cm, height=1.2cm]{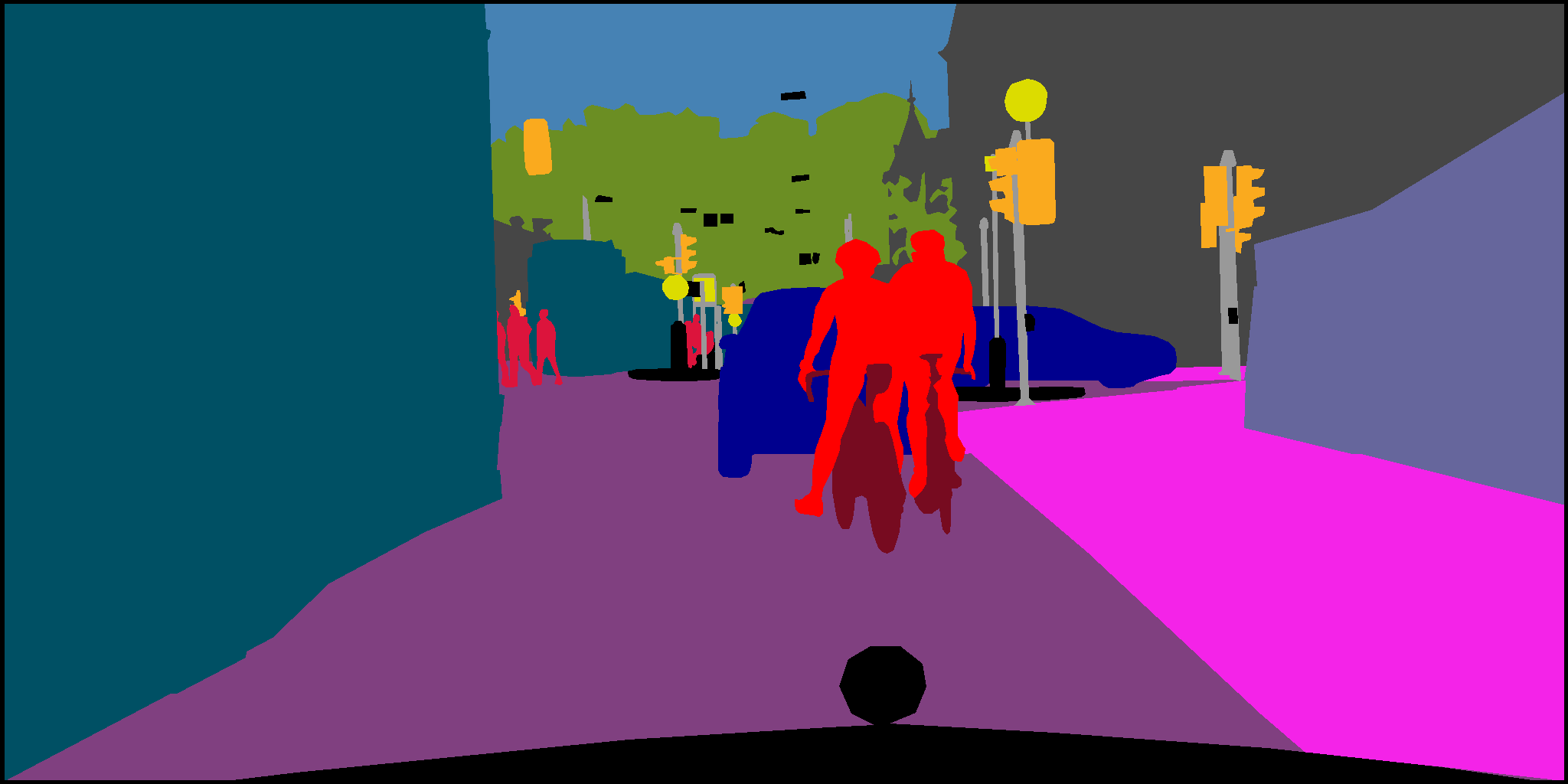}
    \captionsetup{skip=2pt}
    \caption*{(d)}
\end{subfigure}
\end{tabular}
\caption{
Qualitative results: (a) Image, (b) ALC based pseudo-label~\cite{kim2024active}, 
(c) A\(^2\)LC based pseudo-label (Ours), (d) Ground Truth.
A\(^2\)LC provides more fine-grained corrections particularly for tail classes (e.g., \textit{traffic light}, \textit{rider}, \textit{traffic sign}), which were largely ignored in the baseline.}
\label{fig:qualitative_results}
\end{figure}

\subsection{Semantic Segmentation Performance}

\begin{table*}[!t]
    \centering    
\begin{tabular}{ccccccccccccc}
\toprule
\multicolumn{3}{c}{\textbf{Methods}}
& \multicolumn{5}{c}{\textbf{ Data mIoU (\%) }} 
& \multicolumn{5}{c}{\textbf{Model mIoU (\%)}} \\ 
\cmidrule(lr){1-3} \cmidrule(lr){4-8} \cmidrule(lr){9-13}
LCM & ABC & Mask 
& 1R & 2R & 3R & 4R & 5R 
& 1R & 2R & 3R & 4R & 5R \\
\midrule
  &   &  & 62.60 & 65.20 & 66.03 & 66.85 & 66.86 & 56.86 & 58.93 & 58.63 & 58.59 & 58.59 \\
\ding{51} &   &  & 65.66 & 71.53 & 76.00 & 78.38 & 80.59 & 59.02 & 61.39 & 64.19 & 65.58 & 66.94 \\
& \ding{51} &  & 67.31 & 70.06 & 74.49 & 78.32 & 81.58 & 59.09 & 60.86 & 62.05 & 65.27 & 67.75 \\
\ding{51} & \ding{51} &   
& 70.62 & \textbf{77.29} & 80.39 & 82.80 & 84.15 & 60.32 & \textbf{65.09} & 67.36 & 68.71 & 70.11 \\
\ding{51} & \ding{51} & \ding{51} & \textbf{71.04} & 76.08 & \textbf{81.13} & \textbf{83.53} & \textbf{85.57} & \textbf{61.06} & 64.39 & \textbf{67.83} & \textbf{69.63} & \textbf{70.88} \\
\bottomrule
\end{tabular}
\caption{Ablation analysis of each component. 
Columns LCM, ABC, and Mask denote the Label Correction Module, Adaptively Balanced CIL, and Mask-level correction, respectively.
}
\label{tab:city_ablation_5round}
\end{table*}
In Table~\ref{tab:ss_5r}, we benchmark our proposed method under extremely restricted budget constraints across multiple rounds. 
Label correction is performed over five rounds with a budget of 10k and 1k per round on Cityscapes and PASCAL, respectively. 
To ensure statistical reliability, each reported result is averaged over three independent runs and presented with its corresponding standard deviation.
Our method outperforms the baseline by a considerable margin in every round on both datasets, except for the early rounds on PASCAL.
Notably, A\(^2\)LC achieves high efficiency by outperforming the baseline while using only 20\% and 60\% of its budget on Cityscapes and PASCAL, respectively. 
It also demonstrates strong effectiveness, yielding 27.23\% and 14.30\% performance improvements under the same budget constraints.
Figure~\ref{fig:qualitative_results} demonstrates that our method produces pseudo-labels of substantially higher quality compared to the baseline.
In Table~\ref{tab:ss_base}, we compare our proposed method not only with the baseline~\cite{kim2024active} but also with state-of-the-art superpixel-based active learning methods~\cite{cai2021revisiting, hwang2023active, kim2023adaptive}.
The evaluation is conducted on a single round with a fixed annotation budget of 100k on Cityscapes. 
A\(^2\)LC achieves the highest mIoU, outperforming the baseline by 1.66 and reaching 95\% of the fully supervised performance.

\subsection{Effectiveness of Label Correction Module} 

In Table~\ref{tab:LCM}, we evaluate the correction accuracy of LCM on Cityscapes and PASCAL. On Cityscapes, LCM improves mask accuracy by 41.86 by correcting 22,105 masks, increasing clean masks from 4,229 to 13,481. On PASCAL, it improves accuracy by 77.41 by correcting 1,629 masks, where clean masks increased from 79 to 1,340. 
While such manual corrections would require multiple rounds, our A\(^2\)LC framework, empowered by LCM, achieves them instantly without human intervention.

\subsection{Effectiveness of Adaptively Balanced Acquisition Function}
In Table~\ref{tab:ABC}, we compare the pseudo-labels obtained using our ABC function with those produced by other acquisition functions, including active learning baselines.
Our method achieves strong performance on both datasets.
This improvement stems from the pixel-wise adaptive class weight, which assigns higher acquisition scores to false positive pixels belonging to tail classes. 
By integrating this weight into the acquisition function, our method prioritizes the correction of masks misclassified as tail class labels. 
Notably, the performance gain is more pronounced on Cityscapes, likely due to its more severe class imbalance.

\subsection{Ablation Studies}

In Table~\ref{tab:city_ablation_5round}, we perform an ablation study to demonstrate the impact of each component in our method.
Label correction is conducted over five rounds with a 10k budget per round on Cityscapes, and all experiments follow the non-redundant correction setting.
Each component individually improves performance, while the best results are obtained when all are combined.
Notably, the combination of LCM and ABC yields greater gains than either alone, highlighting the complementary and synergistic nature of these two components.

\section{Conclusion}
In this study, we introduce A\(^2\)LC, a semi-automated label correction framework for semantic segmentation built upon cascading stages of manual and automatic correction.
The A\(^2\)LC framework incorporates two key components: an LCM that performs automatic correction by propagating human-provided corrections beyond the queried samples, and an ABC acquisition function that strengthens both correction stages through pixel-wise adaptive class weighting.
Extensive experiments on Cityscapes and PASCAL VOC 2012 show that our method achieves state-of-the-art performance with superior cost efficiency, highlighting its potential for real-world deployment.

\section{Acknowledgments}
This work was supported by Institute of Information \& communications Technology Planning \& Evaluation (IITP) grant funded by the Korea government(MSIT) (No.2022-0-01025, Development of core technology for mobile manipulator for 5G edge-based transportation and manipulation)

\bibliography{aaai2026}

@article{zou2023segment,
  title={Segment everything everywhere all at once},
  author={Zou, Xueyan and Yang, Jianwei and Zhang, Hao and Li, Feng and Li, Linjie and Wang, Jianfeng and Wang, Lijuan and Gao, Jianfeng and Lee, Yong Jae},
  journal={Advances in neural information processing systems},
  volume={36},
  pages={19769--19782},
  year={2023}
}

@inproceedings{luddecke2022image,
  title={Image segmentation using text and image prompts},
  author={L{\"u}ddecke, Timo and Ecker, Alexander},
  booktitle={Proceedings of the IEEE/CVF conference on computer vision and pattern recognition},
  pages={7086--7096},
  year={2022}
}

@article{edlund2021livecell,
  title={LIVECell—A large-scale dataset for label-free live cell segmentation},
  author={Edlund, Christoffer and Jackson, Timothy R and Khalid, Nabeel and Bevan, Nicola and Dale, Timothy and Dengel, Andreas and Ahmed, Sheraz and Trygg, Johan and Sj{\"o}gren, Rickard},
  journal={Nature methods},
  volume={18},
  number={9},
  pages={1038--1045},
  year={2021},
  publisher={Nature Publishing Group US New York}
}

@article{li2023systematic,
  title={A systematic collection of medical image datasets for deep learning},
  author={Li, Johann and Zhu, Guangming and Hua, Cong and Feng, Mingtao and Bennamoun, Basheer and Li, Ping and Lu, Xiaoyuan and Song, Juan and Shen, Peiyi and Xu, Xu and others},
  journal={ACM Computing Surveys},
  volume={56},
  number={5},
  pages={1--51},
  year={2023},
  publisher={ACM New York, NY}
}

@article{plank2022problem,
  title={The'Problem'of Human Label Variation: On Ground Truth in Data, Modeling and Evaluation},
  author={Plank, Barbara},
  journal={arXiv preprint arXiv:2211.02570},
  year={2022}
}

@inproceedings{li2023semi,
  title={Semi-supervised semantic segmentation under label noise via diverse learning groups},
  author={Li, Peixia and Purkait, Pulak and Ajanthan, Thalaiyasingam and Abdolshah, Majid and Garg, Ravi and Husain, Hisham and Xu, Chenchen and Gould, Stephen and Ouyang, Wanli and Van Den Hengel, Anton},
  booktitle={Proceedings of the IEEE/CVF International Conference on Computer Vision},
  pages={1229--1238},
  year={2023}
}

@inproceedings{rottmann2023automated,
  title={Automated detection of label errors in semantic segmentation datasets via deep learning and uncertainty quantification},
  author={Rottmann, Matthias and Reese, Marco},
  booktitle={Proceedings of the IEEE/CVF Winter Conference on Applications of Computer Vision},
  pages={3214--3223},
  year={2023}
}

@inproceedings{xu2023progressive,
  title={Progressive purification for instance-dependent partial label learning},
  author={Xu, Ning and Liu, Biao and Lv, Jiaqi and Qiao, Congyu and Geng, Xin},
  booktitle={International Conference on Machine Learning},
  pages={38551--38565},
  year={2023},
  organization={PMLR}
}

@article{xu2021instance,
  title={Instance-dependent partial label learning},
  author={Xu, Ning and Qiao, Congyu and Geng, Xin and Zhang, Min-Ling},
  journal={Advances in Neural Information Processing Systems},
  volume={34},
  pages={27119--27130},
  year={2021}
}

@article{ekambaram2016active,
  title={Active cleaning of label noise},
  author={Ekambaram, Rajmadhan and Fefilatyev, Sergiy and Shreve, Matthew and Kramer, Kurt and Hall, Lawrence O and Goldgof, Dmitry B and Kasturi, Rangachar},
  journal={Pattern Recognition},
  volume={51},
  pages={463--480},
  year={2016},
  publisher={Elsevier}
}

@article{wang2024simulation,
  title={A simulation-based approach for quantifying the impact of interactive label correction for machine learning},
  author={Wang, Yixuan and Zhao, Jieqiong and Hong, Jiayi and Askin, Ronald G and Maciejewski, Ross},
  journal={IEEE Transactions on Visualization and Computer Graphics},
  year={2024},
  publisher={IEEE}
}

@inproceedings{beck2024beyond,
  title={Beyond active learning: Leveraging the full potential of human interaction via auto-labeling, human correction, and human verification},
  author={Beck, Nathan and Killamsetty, Krishnateja and Kothawade, Suraj and Iyer, Rishabh},
  booktitle={Proceedings of the IEEE/CVF Winter Conference on Applications of Computer Vision},
  pages={2881--2889},
  year={2024}
}

@article{li2022improving,
  title={Improving deep label noise learning with dual active label correction},
  author={Li, Shao-Yuan and Shi, Ye and Huang, Sheng-Jun and Chen, Songcan},
  journal={Machine Learning},
  pages={1--22},
  year={2022},
  publisher={Springer}
}

@INPROCEEDINGS{6413805,
  author={Rebbapragada, Umaa and Brodley, Carla E. and Sulla-Menashe, Damien and Friedl, Mark A.},
  booktitle={2012 IEEE 12th International Conference on Data Mining}, 
  title={Active Label Correction}, 
  year={2012},
  volume={},
  number={},
  pages={1080-1085},
  doi={10.1109/ICDM.2012.162}}

@inproceedings{kremer2018robust,
  title={Robust active label correction},
  author={Kremer, Jan and Sha, Fei and Igel, Christian},
  booktitle={International conference on artificial intelligence and statistics},
  pages={308--316},
  year={2018},
  organization={PMLR}
}

@article{hou2025co,
  title={Co-active: an efficient selective relabeling model for resource constrained edge AI},
  author={Hou, Chenyu and Jiang, Kai and Li, Tiantian and Zhou, Meng and Jiang, Jun},
  journal={Wireless Networks},
  pages={1--14},
  year={2025},
  publisher={Springer}
}

@inproceedings{khanal2024active,
  title={Active Label Refinement for Robust Training of Imbalanced Medical Image Classification Tasks in the Presence of High Label Noise},
  author={Khanal, Bidur and Dai, Tianhong and Bhattarai, Binod and Linte, Cristian},
  booktitle={International Conference on Medical Image Computing and Computer-Assisted Intervention},
  pages={37--47},
  year={2024},
  organization={Springer}
}

@article{bernhardt2022active,
  title={Active label cleaning for improved dataset quality under resource constraints},
  author={Bernhardt, M{\'e}lanie and Castro, Daniel C and Tanno, Ryutaro and Schwaighofer, Anton and Tezcan, Kerem C and Monteiro, Miguel and Bannur, Shruthi and Lungren, Matthew P and Nori, Aditya and Glocker, Ben and others},
  journal={Nature communications},
  volume={13},
  number={1},
  pages={1161},
  year={2022},
  publisher={Nature Publishing Group UK London}
}

@article{taneja2024can,
  title={Can Active Label Correction Improve LLM-based Modular AI Systems?},
  author={Taneja, Karan and Goel, Ashok},
  journal={arXiv preprint arXiv:2401.05467},
  year={2024}
}

@article{schachtsiek2023class,
  title={Class Balanced Dynamic Acquisition for Domain Adaptive Semantic Segmentation using Active Learning},
  author={Schachtsiek, Marc and Rossi, Simone and Hannagan, Thomas},
  journal={arXiv preprint arXiv:2311.14146},
  year={2023}
}

@article{ruckin2024semi,
  title={Semi-supervised active learning for semantic segmentation in unknown environments using informative path planning},
  author={R{\"u}ckin, Julius and Magistri, Federico and Stachniss, Cyrill and Popovi{\'c}, Marija},
  journal={IEEE Robotics and Automation Letters},
  volume={9},
  number={3},
  pages={2662--2669},
  year={2024},
  publisher={IEEE}
}

@inproceedings{ribeiro2024uncertainty,
  title={Uncertainty driven active learning for image segmentation in underwater inspection},
  author={Ribeiro Marnet, Luiza and Brodskiy, Yury and Grasshof, Stella and Wasowski, Andrzej},
  booktitle={International Conference on Robotics, Computer Vision and Intelligent Systems},
  pages={66--81},
  year={2024},
  organization={Springer}
}

@article{van2024active,
  title={Active learning for efficient annotation in precision agriculture: a use-case on crop-weed semantic segmentation},
  author={van Marrewijk, Bart M and Dandjinou, Charbel and Rustia, Dan Jeric Arcega and Gonzalez, Nicolas Franco and Diallo, Boubacar and Dias, J{\'e}r{\^o}me and Melki, Paul and Blok, Pieter M},
  journal={arXiv preprint arXiv:2404.02580},
  year={2024}
}

@article{didari2024bayesian,
  title={Bayesian Active Learning for Semantic Segmentation},
  author={Didari, Sima and Hu, Wenjun and Woo, Jae Oh and Hao, Heng and Moon, Hankyu and Min, Seungjai},
  journal={arXiv preprint arXiv:2408.01694},
  year={2024}
}

@article{ge2024esa,
  title={Esa: Annotation-efficient active learning for semantic segmentation},
  author={Ge, Jinchao and Zhang, Zeyu and Phan, Minh Hieu and Zhang, Bowen and Liu, Akide and Zhao, Yang},
  journal={arXiv preprint arXiv:2408.13491},
  year={2024}
}

@article{ruckin2024active,
  title={Active Learning of Robot Vision Using Adaptive Path Planning},
  author={R{\"u}ckin, Julius and Magistri, Federico and Stachniss, Cyrill and Popovi{\'c}, Marija},
  journal={arXiv preprint arXiv:2410.10684},
  year={2024}
}

@article{ma2025integrating,
  title={Integrating Semi-Supervised and Active Learning for Semantic Segmentation},
  author={Ma, Wanli and Karakus, Oktay and Rosin, Paul L},
  journal={arXiv preprint arXiv:2501.19227},
  year={2025}
}

@inproceedings{yang2023towards,
  title={Towards controlled data augmentations for active learning},
  author={Yang, Jianan and Wang, Haobo and Wu, Sai and Chen, Gang and Zhao, Junbo},
  booktitle={International Conference on Machine Learning},
  pages={39524--39542},
  year={2023},
  organization={PMLR}
}

@inproceedings{sener2018active,
title={Active Learning for Convolutional Neural Networks: A Core-Set Approach},
author={Ozan Sener and Silvio Savarese},
booktitle={International Conference on Learning Representations},
year={2018},
url={https://openreview.net/forum?id=H1aIuk-RW},
}

@inproceedings{Citovsky2021BatchAL,
  title={Batch Active Learning at Scale},
  author={Gui Citovsky and Giulia DeSalvo and Claudio Gentile and Lazaros Karydas and Anand Rajagopalan and Afshin Rostamizadeh and Sanjiv Kumar},
  booktitle={Neural Information Processing Systems},
  year={2021},
  url={https://api.semanticscholar.org/CorpusID:236635499}
}

@inproceedings{Ash2020Deep,
title={Deep Batch Active Learning by Diverse, Uncertain Gradient Lower Bounds},
author={Jordan T. Ash and Chicheng Zhang and Akshay Krishnamurthy and John Langford and Alekh Agarwal},
booktitle={International Conference on Learning Representations},
year={2020},
url={https://openreview.net/forum?id=ryghZJBKPS}
}

@INPROCEEDINGS{7789580,
  author={Kampffmeyer, Michael and Salberg, Arnt-Børre and Jenssen, Robert},
  booktitle={2016 IEEE Conference on Computer Vision and Pattern Recognition Workshops (CVPRW)}, 
  title={Semantic Segmentation of Small Objects and Modeling of Uncertainty in Urban Remote Sensing Images Using Deep Convolutional Neural Networks}, 
  year={2016}
}

@inproceedings{gal2017deep,
  title={Deep bayesian active learning with image data},
  author={Gal, Yarin and Islam, Riashat and Ghahramani, Zoubin},
  booktitle={International conference on machine learning},
  pages={1183--1192},
  year={2017},
  organization={PMLR}
}

@inproceedings{wang2014new,
  title={A new active labeling method for deep learning},
  author={Wang, Dan and Shang, Yi},
  booktitle={2014 International joint conference on neural networks (IJCNN)},
  pages={112--119},
  year={2014},
  organization={IEEE}
}

@inproceedings{roth2006margin,
  title={Margin-based active learning for structured output spaces},
  author={Roth, Dan and Small, Kevin},
  booktitle={Machine Learning: ECML 2006: 17th European Conference on Machine Learning Berlin, Germany, September 18-22, 2006 Proceedings 17},
  pages={413--424},
  year={2006},
  organization={Springer}
}

@inproceedings{safaei2024entropic,
  title={Entropic open-set active learning},
  author={Safaei, Bardia and Vibashan, VS and de Melo, Celso M and Patel, Vishal M},
  booktitle={Proceedings of the AAAI conference on artificial intelligence},
  volume={38},
  number={5},
  pages={4686--4694},
  year={2024}
}

@InProceedings{Wu_2025_WACV,
    title={Active Learning with Context Sampling and One-vs-Rest Entropy for Semantic Segmentation},
    author={Wu, Fei and Marquez-Neila, Pablo and Rafi-Tarii, Hedyeh and Sznitman, Raphael},
    booktitle = {Proceedings of the Winter Conference on Applications of Computer Vision (WACV)},
    month     = {February},
    year      = {2025},
}

@article{hwang2023active,
  title={Active learning for semantic segmentation with multi-class label query},
  author={Hwang, Sehyun and Lee, Sohyun and Kim, Hoyoung and Oh, Minhyeon and Ok, Jungseul and Kwak, Suha},
  journal={Advances in Neural Information Processing Systems},
  volume={36},
  pages={27020--27039},
  year={2023}
}

@inproceedings{kim2023adaptive,
  title={Adaptive superpixel for active learning in semantic segmentation},
  author={Kim, Hoyoung and Oh, Minhyeon and Hwang, Sehyun and Kwak, Suha and Ok, Jungseul},
  booktitle={Proceedings of the IEEE/CVF International Conference on Computer Vision},
  pages={943--953},
  year={2023}
}

@inproceedings{cai2021revisiting,
  title={Revisiting superpixels for active learning in semantic segmentation with realistic annotation costs},
  author={Cai, Lile and Xu, Xun and Liew, Jun Hao and Foo, Chuan Sheng},
  booktitle={Proceedings of the IEEE/CVF conference on computer vision and pattern recognition},
  pages={10988--10997},
  year={2021}
}

@inproceedings{rana2023hybrid,
  title={Hybrid active learning via deep clustering for video action detection},
  author={Rana, Aayush J and Rawat, Yogesh S},
  booktitle={Proceedings of the IEEE/CVF Conference on Computer Vision and Pattern Recognition},
  pages={18867--18877},
  year={2023}
}

@inproceedings{chen2024think,
  title={Think twice before selection: Federated evidential active learning for medical image analysis with domain shifts},
  author={Chen, Jiayi and Ma, Benteng and Cui, Hengfei and Xia, Yong},
  booktitle={Proceedings of the IEEE/CVF Conference on Computer Vision and Pattern Recognition},
  pages={11439--11449},
  year={2024}
}

@inproceedings{kim2024active,
  title={Active Label Correction for Semantic Segmentation with Foundation Models},
  author={Kim, Hoyoung and Hwang, Sehyun and Kwak, Suha and Ok, Jungseul},
  booktitle={Forty-first International Conference
on Machine Learning},
  year={2024}
}

@article{lad2023estimating,
  title={Estimating label quality and errors in semantic segmentation data via any model},
  author={Lad, Vedang and Mueller, Jonas},
  journal={arXiv preprint arXiv:2307.05080},
  year={2023}
}

@inproceedings{cordts2016cityscapes,
  title={The cityscapes dataset for semantic urban scene understanding},
  author={Cordts, Marius and Omran, Mohamed and Ramos, Sebastian and Rehfeld, Timo and Enzweiler, Markus and Benenson, Rodrigo and Franke, Uwe and Roth, Stefan and Schiele, Bernt},
  booktitle={Proceedings of the IEEE conference on computer vision and pattern recognition},
  pages={3213--3223},
  year={2016}
}

@article{everingham2015pascal,
  title={The pascal visual object classes challenge: A retrospective},
  author={Everingham, Mark and Eslami, SM Ali and Van Gool, Luc and Williams, Christopher KI and Winn, John and Zisserman, Andrew},
  journal={International journal of computer vision},
  volume={111},
  pages={98--136},
  year={2015},
  publisher={Springer}
}

@article{ren2024grounded,
  title={Grounded sam: Assembling open-world models for diverse visual tasks},
  author={Ren, Tianhe and Liu, Shilong and Zeng, Ailing and Lin, Jing and Li, Kunchang and Cao, He and Chen, Jiayu and Huang, Xinyu and Chen, Yukang and Yan, Feng and others},
  journal={arXiv preprint arXiv:2401.14159},
  year={2024}
}

@inproceedings{liu2025grounding,
  title={Grounding dino: Marrying dino with grounded pre-training for open-set object detection},
  author={Liu, Shilong and Zeng, Zhaoyang and Ren, Tianhe and Li, Feng and Zhang, Hao and Yang, Jie and Jiang, Qing and Li, Chunyuan and Yang, Jianwei and Su, Hang and others},
  booktitle={European Conference on Computer Vision},
  pages={38--55},
  year={2025},
  organization={Springer}
}

@inproceedings{kirillov2023segment,
  title={Segment anything},
  author={Kirillov, Alexander and Mintun, Eric and Ravi, Nikhila and Mao, Hanzi and Rolland, Chloe and Gustafson, Laura and Xiao, Tete and Whitehead, Spencer and Berg, Alexander C and Lo, Wan-Yen and others},
  booktitle={Proceedings of the IEEE/CVF International Conference on Computer Vision},
  pages={4015--4026},
  year={2023}
}

@inproceedings{chen2018encoder,
  title={Encoder-decoder with atrous separable convolution for semantic image segmentation},
  author={Chen, Liang-Chieh and Zhu, Yukun and Papandreou, George and Schroff, Florian and Adam, Hartwig},
  booktitle={Proceedings of the European conference on computer vision (ECCV)},
  pages={801--818},
  year={2018}
}

@inproceedings{deng2009imagenet,
  title={Imagenet: A large-scale hierarchical image database},
  author={Deng, Jia and Dong, Wei and Socher, Richard and Li, Li-Jia and Li, Kai and Fei-Fei, Li},
  booktitle={2009 IEEE conference on computer vision and pattern recognition},
  pages={248--255},
  year={2009},
  organization={Ieee}
}

@inproceedings{he2016deep,
  title={Deep residual learning for image recognition},
  author={He, Kaiming and Zhang, Xiangyu and Ren, Shaoqing and Sun, Jian},
  booktitle={Proceedings of the IEEE conference on computer vision and pattern recognition},
  pages={770--778},
  year={2016}
}

@inproceedings{Kim2022ActiveLC,
  title={Active Label Correction Using Robust Parameter Update and Entropy Propagation},
  author={Kwang In Kim},
  booktitle={European Conference on Computer Vision},
  year={2022},
  url={https://api.semanticscholar.org/CorpusID:253121291}
}

% Technical Appendix
\newpage
\twocolumn[
\vbox{
  \vskip 0.26in
  \centering
  \LARGE \textbf{A\(^2\)LC: Active and Automated Label Correction for Semantic Segmentation}
  \vskip 0.13in 
  \LARGE Technical Appendix 
  \vskip 0.26in
}]

\appendix

\section{Experimental Details}
\subsection{Datasets}
We use two semantic segmentation datasets, Cityscapes~\cite{cordts2016cityscapes} and PASCAL VOC 2012~\cite{everingham2015pascal}, following the baseline settings to ensure a fair comparison.
Cityscapes is a dataset for urban scene understanding, comprising 19 object classes with 2,975 training images, 500 validation images, and 1,525 test images. 
PASCAL is a dataset consisting of diverse real-world images with 20 object classes, comprising 1,464 training images and 1,449 validation images.

\subsection{Models}
The initial pseudo-labels are generated using the Grounded SAM~\cite{ren2024grounded}, following the previous work~\cite{kim2024active}.
In Grounded SAM, objects are detected using Grounding DINO~\cite{liu2025grounding} with text prompts, and the predicted bounding boxes are subsequently utilized as prompts by SAM~\cite{kirillov2023segment} to generate segmentation masks. 
We use DeepLab-v3+~\cite{chen2018encoder} with a ResNet-101 backbone~\cite{he2016deep} pre-trained on ImageNet~\cite{deng2009imagenet} as the segmentation model in our experiments to ensure a fair comparison with the baseline~\cite{kim2024active}.

\subsection{Implementation Details}
For both Cityscapes and PASCAL, we set the box threshold of 0.2 and specify class names in the text prompt when generating the initial pseudo-labels using Grounded SAM. 
Also, we train a model for 30K iterations in each round using the SGD optimizer with a momentum of 0.9 and apply polynomial learning rate decay with a power of 0.9. 
For Cityscapes, training images are resized to 768 × 768 with a mini-batch size of 8, and the base learning rate is set to 0.05. 
For PASCAL, training images are resized to 513 × 513 with a mini-batch size of 16, and the base learning rate is set to 0.1. 
All experiments are conducted using a single NVIDIA RTX A5000 GPU and an Intel Xeon Gold 6226R CPU. 
For Cityscapes and PASCAL, each round of correction with 10K and 1K clicks, respectively, takes approximately 15 hours and 10 hours.

\section{Algorithms}
\begin{algorithm}[!t]
\caption{A\(^2\)LC Framework}
\label{alg:algorithm1}
\begin{algorithmic}[1]
\REQUIRE Budget size $B$, Final round $R$
\ENSURE $\mathcal{D}_R$ and $\theta_R$
\STATE Prepare initial dataset $\mathcal{D}_0$ and maskset $\mathcal{M}_0$
\STATE Train model $\theta_0$ with $\mathcal{D}_0$
\FOR{$r = 1$ to $R$}
    \FOR{each $(m, \hat{y}(m)) \in \mathcal{M}_{r-1}$}
        \STATE Score all masks based on Eq. (12)
    \ENDFOR
    \STATE Sample the top-$B$ masks $\mathcal{M}_r^Q$ based on Eq. (13)
    \STATE
    $\mathcal{M}_r^Q \gets 
    \left\{
    \begin{aligned}
    & (m, y(m)) \mid \\
    & m \in \mathcal{M}_r^Q, y(m) = \text{Oracle}(m)
    \end{aligned}
    \right\}$
    \STATE $\mathcal{M}_r^{Q^c} \gets \mathcal{M}_{r-1} \setminus \mathcal{M}_r^Q$
    \STATE $\mathcal{D}_r \gets \texttt{\textbf{LCM}\:}(\mathcal{D}_{r-1}, \mathcal{M}_r^Q, \mathcal{M}_r^{Q^c})$ 
    \STATE Train model $\theta_r$ with corrected $\mathcal{D}_r$
    \STATE $\mathcal{M}_{r} \gets \mathcal{M}_{r-1} \setminus \mathcal{M}_r^Q$
\ENDFOR
\end{algorithmic}
\end{algorithm}
\begin{algorithm}[!t]
\caption{Label Correction Module (LCM)}
\label{alg:LCM}
\begin{algorithmic}[1]
\REQUIRE Previous round dataset $\mathcal{D}_{r-1}$, Queried maskset \(\mathcal{M}_r^Q\), Unqueried maskset \(\mathcal{M}_r^{Q^c}\), Epochs $E$
\ENSURE Updated dataset $\mathcal{D}_r$
\STATE $\triangleright$ \textbf{Supervised train with queried masks}
\FOR{$e = 1, 2, \dots, E$}
    \FOR{each $(m, y(m)) \in \mathcal{M}_r^Q$}
        \STATE Train the model $\psi_r$ based on Eq. (2)
    \ENDFOR
\ENDFOR

\STATE $\triangleright$ \textbf{Selective label update for unqueried masks}
\FOR{each $(m, \hat{y}(m)) \in \mathcal{M}_r^{Q^c}$}
    \STATE $\hat{y}_{\psi_r}(m) \gets \arg\max_{c \in \mathcal{C}} \psi_r(c; m)$
    \IF{$m$ satisfies \textit{Selection\_Criteria} (Eq. (3))}
        \STATE $\hat{y}(m) \gets \hat{y}_{\psi_r}(m)$
    \ENDIF
\ENDFOR
\end{algorithmic}
\end{algorithm}
\subsection{A\(^2\)LC Framework}
Algorithm~\ref{alg:algorithm1} presents the pseudo-code of our proposed A\(^2\)LC framework.
\subsection{Label Correction Module}
Algorithm~\ref{alg:LCM} presents the pseudo-code of our proposed label correction module.

\section{Analysis of Label Correction Module}

\subsection{Complementary Property Analysis}
\begin{table*}[!h]
\centering
% ------------------ (a) Cityscapes ------------------
\newcolumntype{P}[1]{>{\centering\arraybackslash}p{#1}}
\begin{tabular}{cc|*{9}{P{0.75cm}}|c}

\toprule
\multicolumn{2}{c|}{\textbf{ LCM Accuracy (\%)}}& \multicolumn{9}{c|}{\textbf{\# of Successfully Corrected Masks}} & \textbf{Correction Rate} \\
Round & Before $\to$ After & 2R & 3R & 4R & 5R & 6R & 7R & 8R & 9R & 10R & M (\%) / CR (\%) \\
\midrule
1R                    &      15.5\% $\to$ \textbf{50.3\%}                             & 3,046 & 4,894 & 5,398 & 5,982 & 6,182 & 6,575 & 6,709 & 7,001 & 7,276 &      {49.7\%} / {\textbf{56.9\%}} \\
2R                    &      16.4\% $\to$ \textbf{57.1\%}          &                  & 1,715 & 2,213 & 2,913 & 3,047 & 3,193 & 3,238 & 3,354 & 3,503 &              {42.9\%} / {\textbf{32.0\%}} \\
3R                    &      10.9\% $\to$ \textbf{61.8\%}          & &                & 645 & 827 & 946 & 1,301 & 1,479 & 1,661 & 1,752 &                            {38.2\%} / {\textbf{64.7\%}} \\
4R                    &      22.0\% $\to$ \textbf{47.0\%}          & & &              & 237 & 340 & 440 & 512 & 612 & 612 &                                          {53.0\%} / {\textbf{90.8\%}} \\
5R                    &      28.2\% $\to$ \textbf{56.2\%}          & & & &            & 242 & 242 & 242 & 242 & 265 &                                                {43.8\%} / {\textbf{68.8\%}} \\ \midrule \midrule
Average              &      18.6\% $\to$ \textbf{54.5\%}          & & & &            & & & & & &                                                                    {45.5\%} / {\textbf{62.6\%}} \\\bottomrule
\end{tabular}
\caption*{\makebox[\linewidth][c]{\textbf{(a) Cityscapes}}\newline} 

% ------------------ (b) PASCAL VOC 2012 --------------------------
\begin{tabular}{cc|*{9}{P{0.75cm}}|c}
\toprule
\multicolumn{2}{c|}{\textbf{ LCM Accuracy (\%)}}& \multicolumn{9}{c|}{\textbf{\# of Successfully Corrected Masks}} & \textbf{Correction Rate} \\
Round & Before $\to$ After & 2R & 3R & 4R & 5R & 6R & 7R & 8R & 9R & 10R & M (\%) / CR (\%) \\
\midrule
1R                             &     3.0\% $\to$ \textbf{62.1\%}                 & 3 & 19 & 31 & 42 & 72 & 88 & 129 & 166 & 218 &    {37.9\%} / {\textbf{65.3\%}} \\
2R                             &     4.3\% $\to$ \textbf{44.5\%}  &              & 2 & 2 & 16 & 19 & 33 & 57 & 74 & 103 &    {55.5\%} / {\textbf{72.5\%}} \\
3R                             &     2.3\% $\to$ \textbf{68.4\%}  & &            & 4 & 12 & 19 & 31 & 55 & 72 & 118 &    {31.6\%} / {\textbf{45.7\%}} \\
4R                             &     3.3\% $\to$ \textbf{53.8\%}  & & &          & 6 & 13 & 24 & 47 & 84 & 137 &    {46.2\%} / {\textbf{57.3\%}} \\
5R                             &     3.5\% $\to$ \textbf{51.3\%}  & & & &        & 1 & 9 & 16 & 25 & 31 &    {48.7\%} / {\textbf{55.4\%}} \\ \midrule \midrule
Average                       &     3.3\% $\to$ \textbf{56.0\%}  & & & &        & & & & & &    {44.0\%} / {\textbf{59.2\%}} \\
\bottomrule
\end{tabular}

\caption*{\textbf{(b) PASCAL VOC 2012}}

\caption{Analysis of synergy with human corrections.
Our dual-source correction framework exhibits strong complementarity between LCM and human annotators.
In the table, the first column reports accuracy before and after LCM  correction, the second column shows the cumulative number of masks initially missed by LCM but later corrected, and the last column presents the initial LCM miss rate along with the proportion of these misses that are eventually corrected.
On average, LCM corrects the subset of human missed masks with initial accuracies of 18.6\% on Cityscapes and 3.3\% on PASCAL, improving these to 54.5\% and 56.0\%, respectively. 
Moreover, masks initially missed by LCM are subsequently corrected by either humans or LCM itself in later rounds. 
Specifically, 62.6\% of the missed masks (miss rate $\text{M} = 1 - \text{Accuracy}$) on Cityscapes and 59.2\% on PASCAL are eventually resolved. 
These results highlight the complementary synergy in which humans correct masks missed by LCM, and LCM corrects those missed by humans.
}
\label{tab:lcm_error_propagation}
\end{table*}

In Table~\ref{tab:lcm_error_propagation}, we analyze the synergistic and complementary interaction between human annotators and LCM within our proposed dual-source correction framework.
Label correction is performed for ten rounds on Cityscapes and PASCAL, with each round budget of 10k and 1k, respectively.
Here, we present three key statistics:
(1) LCM Accuracy (\%) (Before $\to$ After): Accuracy before and after correction for the masks corrected by LCM, (2) \# of Successfully Corrected Masks: the cumulative number of masks initially missed by LCM but successfully corrected in later rounds, and (3) Correction Rate (M (\%) / CR (\%)): initial LCM missed rate and the percentage of these misses successfully corrected later.

Our proposed dual-source correction framework demonstrates a strong complementary interaction between LCM and human annotators. 
Masks missed by humans are corrected by LCM, whereas masks missed by LCM are subsequently corrected by humans, leading to a synergistic correction process.
As shown in the first row of Table~\ref{tab:lcm_error_propagation} (b), LCM corrects masks that are not assigned to humans due to limited resources, improving their accuracy from only 3.0\% to 62.1\%. 
Furthermore, of the remaining 37.9\% of masks missed by LCM, 65.3\% are successfully corrected in subsequent rounds, where either humans or LCM progressively correct 3, 16, 12, \(\dots\), and 52 masks.

On Cityscapes, LCM corrects a subset of masks initially missed by humans, increasing their accuracy from an average of 18.6\% to 54.5\%. 
Of the masks that LCM initially misses which account for 45.5\% on average, 62.6\% are subsequently corrected by humans or LCM in later rounds. 
Similarly, on PASCAL, LCM increases the accuracy of the subset of human missed masks from 3.3\% to 56.0\%. 
Of the masks that LCM initially misses which account for 44.0\% on average, 59.2\% are subsequently corrected in later rounds.
This analysis highlights a complementary synergy in which LCM and humans mutually compensate for each other’s missed masks.

\subsection{Detailed Quantitative Results}
In Table~\ref{tab:lcm_cls_acc}, we evaluate the class-wise number of masks with accurate label before and after correction for those masks processed by LCM.
The overall count of accurately labeled masks increases by approximately 17 times, demonstrating the effectiveness of LCM.
However, the class-wise analysis also reveals cases where originally correct labels are altered incorrectly, indicating the need for careful hyperparameter tuning in LCM.

\begin{table*}[t]
\centering
\begin{tabular}{c 
c@{\hskip 1.1em}c@{\hskip 1.1em}c@{\hskip 1.1em}c@{\hskip 1.1em}c@{\hskip 1.1em}
c@{\hskip 1.1em}c@{\hskip 1.1em}c@{\hskip 1.1em}c@{\hskip 1.1em}c@{\hskip 1.1em}
c@{\hskip 1.1em}c@{\hskip 1.1em}c@{\hskip 1.1em}c@{\hskip 1.1em}c@{\hskip 1.1em}
c@{\hskip 1.1em}c@{\hskip 1.1em}c@{\hskip 1.1em}c@{\hskip 1.1em}c@{\hskip 1.1em}
c@{\hskip 1.1em}c}
\toprule
\textbf{LCM}
& \rotatebox{90}{\textit{back.}}
& \rotatebox{90}{\textit{aero.}}
& \rotatebox{90}{\textit{bicycle}~}
& \rotatebox{90}{\textit{bird}}
& \rotatebox{90}{\textit{boat}}
& \rotatebox{90}{\textit{bottle}}
& \rotatebox{90}{\textit{bus}}
& \rotatebox{90}{\textit{car}}
& \rotatebox{90}{\textit{cat}}
& \rotatebox{90}{\textit{chair}}
& \rotatebox{90}{\textit{cow}}
& \rotatebox{90}{\textit{table}}
& \rotatebox{90}{\textit{dog}}
& \rotatebox{90}{\textit{horse}}
& \rotatebox{90}{\textit{motor.}}
& \rotatebox{90}{\textit{person}}
& \rotatebox{90}{\textit{plant}}
& \rotatebox{90}{\textit{sheep}}
& \rotatebox{90}{\textit{sofa}}
& \rotatebox{90}{\textit{train}}
& \rotatebox{90}{\textit{tv/mon.}}
& \textbf{\# Acc} \\
\midrule
\ding{55}
& 0 & 0 & 0 & 2 & 1
& 6 & 0 & 6 & 3 & 2
& 1 & 1 & 0 & 0 & 2
& 40 & 13 & 0 & 1 & 0
& 1 & 79 \\
\ding{51}
& 574 & 0 & 0 & 0 & 0
& 0 & 434 & 151 & 2 & 4
& 0 & 12 & 0 & 0 & 28
& 58 & 0 & 0 & 3 & 74
& 0 & 1,340 \\
\bottomrule
\end{tabular}
\caption{Analysis of LCM class-wise accuracy.}
\label{tab:lcm_cls_acc}
\end{table*}

\subsection{Extended Architecture Variation Results}

We evaluate the LCM under configurations ranging from one to seven hidden layers, combined with three distinct activation function variants.
We provide a detailed description of the activation functions utilized in our experiments in Table~\ref{tab:lcm_activations}. 
In Table~\ref{tab:lcm_layer_act_results}, we observe that the performance variation among different activation functions and layer configurations is marginal.
Notably, models with one or two layers consistently underperform relative to deeper architectures, whereas using three or more layers yields comparable performance.
Therefore, we use three layers for the LCM model to reduce computational cost and set ReLU as the default activation in our overall experiments.
It is noteworthy that, given the baseline mIoU scores of 76.93 and 83.04 on Cityscapes and PASCAL, respectively, our proposed A\(^2\)LC framework consistently achieves higher performance across different configurations, clearly demonstrating its robustness and effectiveness.
\begin{table}[t]
\centering
\begin{tabular}{cc}
\toprule
\textbf{A.Function} & \textbf{Description} \\
\midrule
Mish & $f(x) = x \cdot \tanh(\ln(1+e^{x}))$ \\
ReLU & $f(x) = \max(0, x)$ \\
GELU & $f(x) = x \, \Phi(x) = \frac{x}{2}\bigg[1+\operatorname{erf}\Big(\frac{x}{\sqrt{2}}\Big)\bigg]$ \\
\bottomrule
\end{tabular}
\caption{Activation functions explored in our experiments.}
\label{tab:lcm_activations}
\end{table}

\begin{table}[t]
\centering
\begin{tabular}{cc ccc}
\toprule
 & & \multicolumn{3}{c}{\textbf{Activation Functions}} \\
\textbf{Dataset} &\textbf{Hidden Layer} & Mish & ReLU & GELU \\
\midrule
\multirow{7}{*}{Cityscapes} &
1  & 84.5 & 84.5 & 84.5 \\
 & 2  & 84.8 & 84.8 & 84.8 \\
 & 3  & 84.5 & \textbf{84.9} & \textbf{85.0} \\
 & 4  & 84.6 & \textbf{84.9} & 84.9 \\
 & 5  & \textbf{85.0} & \textbf{84.9} & 84.8 \\
 & 6  & 84.8 & 84.7 & 84.9 \\
 & 7  & \textbf{85.0} & 84.6 & \textbf{85.0} \\
\midrule
\multirow{7}{*}{PASCAL} &
 1  & 87.1 & 87.1 & 87.1 \\
 & 2  & 87.2 & 87.1 & 87.1 \\
 & 3  & 87.3 & \textbf{87.4} & \textbf{87.3} \\
 & 4  & 87.3 & 87.3 & \textbf{87.3} \\
 & 5  & \textbf{87.4} & \textbf{87.4} & 87.3 \\
 & 6  & 87.3 & 87.3 & 87.2 \\
 & 7  & 87.1 & \textbf{87.4} & 86.9 \\
\bottomrule
\end{tabular}
\caption{Analysis of LCM performance across model variants (Data mIoU, \%).}
\label{tab:lcm_layer_act_results}
\end{table}

\subsection{Ablation Study on Selection Criteria}
In Table~\ref{tab:criteria_ablation}, we evaluate individual contributions of each LCM selection criterion to overall performance.
Adding single $\mathcal{J}_1$ already improves the performance by 2.13 and 0.46 on Cityscapes and PASCAL, respectively.
On PASCAL, applying additional $\mathcal{J}_3$ provides an extra 0.01 improvement, and using all three criteria gives a further 0.08 gain.
On Cityscapes, performance is slightly reduced when $\mathcal{J}_3$ is additionally applied to $\mathcal{J}_1$, yet it still outperforms the configuration without any selection criteria, and is further improved by including $\mathcal{J}_2$.
It is noteworthy that configuring all the proposed selection criteria leads to overall performance improvements on both datasets.

\begin{table}[t]
    \centering
    \begin{tabular}{@{\hskip 10pt} c @{\hskip 16.5pt} cccc} 
        \toprule
        \multicolumn{3}{c}{\textbf{Selection Criteria}} & \multicolumn{2}{c}{\textbf{Dataset}} \\ 
        \(\mathcal{J}_1\) & \(\mathcal{J}_2\) & \(\mathcal{J}_3\) & Cityscapes & PASCAL \\
        \midrule
         &  &   & 82.82 & 86.83 \\
        \ding{51} &   &   & \textbf{84.95} & 87.29 \\
        \ding{51} &   & \ding{51} & 83.80 & 87.30 \\
        \ding{51} & \ding{51} & \ding{51} & 84.92 & \textbf{87.38} \\
        \bottomrule
    \end{tabular}
\caption{Ablation study on three selection criteria of LCM (Data mIoU, \%).}
\label{tab:criteria_ablation}
\end{table}

\begin{table}[t]
    \centering
    \begin{tabular}{cccccc} 
        \toprule
        \multicolumn{4}{c}{\textbf{Adaptive Class Weight}} & \multicolumn{2}{c}{\textbf{Dataset}} \\
        $\hat{w}(x)$ & $\text{KL}^1$ & $\text{KL}^2$ & $\text{KL}^3$ & Cityscapes
        & PASCAL \\
        \midrule
        \ding{51} &   &   &   & 84.67 & 82.82 \\
        \ding{51} & \ding{51} &   &   & 84.91 & 87.31 \\
        \ding{51} &   & \ding{51} &   & 84.67 & 87.33 \\
        \ding{51} &   &   & \ding{51} & \textbf{84.92} & \textbf{87.34} \\
        \bottomrule
    \end{tabular}
\caption{Ablation study on adaptive class weight (Data mIoU, \%).}
\label{tab:ablation_acq}
\end{table}

\section{Analysis of Adaptively Balanced Acquisition Function}

\begin{table*}[!t]
\centering
\setlength{\tabcolsep}{2.2pt}
\begin{tabular}{l ccccccccccccccccccc c}

\toprule
 & \multicolumn{9}{l}{\hspace{5pt} \(\leftarrow\) \small \textit{Tail}} 
 & \multicolumn{10}{r}{\small \textit{Head} \(\rightarrow\) \hspace{5pt}} &  \\
{\textbf{Acquisition}}
& \rotatebox{90}{\textit{light}}
& \rotatebox{90}{\textit{rider}}
& \rotatebox{90}{\textit{motor.}}
& \rotatebox{90}{\textit{train}}
& \rotatebox{90}{\textit{bicycle}}
& \rotatebox{90}{\textit{bus}}
& \rotatebox{90}{\textit{truck}}
& \rotatebox{90}{\textit{sign}}
& \rotatebox{90}{\textit{pole}}
& \rotatebox{90}{\textit{terrain}}
& \rotatebox{90}{\textit{person}}
& \rotatebox{90}{\textit{fence}}
& \rotatebox{90}{\textit{wall}}
& \rotatebox{90}{\textit{sky}}
& \rotatebox{90}{\textit{side.}}
& \rotatebox{90}{\textit{car}}
& \rotatebox{90}{\textit{veg.}}
& \rotatebox{90}{\textit{buil.}}
& \rotatebox{90}{\textit{road}}
& \multicolumn{1}{c}{\textbf{mIoU}}  \\

\midrule
Init.
& 2.4 & 8.8 & 53.2 & 16.1 & 68.5 & 27.9 & 27.1 & 29.5 & 57.7 & 42.4 & 69.6 & 39.9 & 17.6 & 90.5 & 68.0 & 87.2 & 77.6 & 82.9 & 95.8 & 50.7 \\

\midrule
Random 
& 51.2 & 31.4 & 68.5 & 37.1 & 77.3 & 36.8 & 39.4 & 42.1 & 62.8 & 55.7 & 82.6 & 54.6 & 36.4 & 94.3 & 74.9 & 91.5 & 87.5 & 90.9 & 96.8 & 63.8 \\
Entropy
& 33.3 & 16.8 & 72.1 & 30.2 & 78.4 & 35.2 & 41.4 & 37.7 & 61.4 & 48.8 & 82.0 & 53.5 & 33.8 & 90.7 & 39.8 & 91.7 & 84.9 & 90.1 & 82.5 & 58.1 \\
Margin
 & 42.5 & 34.2 & 74.4 & 40.4 & 78.5 & 35.9 & 42.9 & 38.9 & 61.3 & 52.7 & 83.8 & 53.0 & 40.7 & 91.5 & 58.9 & 91.1 & 86.3 & 91.5 & 92.3 & 62.7 \\
CIL 
& 43.3 & 24.3 & 73.5 & 20.9 & 79.4 & 31.9 & 42.7 & 43.7 & 67.6 & 52.6 & 83.2 & 51.4 & 28.7 & 91.3 & 69.8 & 91.4 & 86.1 & 90.6 & 96.2 & 61.5 \\
LCIL
& 48.7 & 27.3 & 74.4 & 21.0 & 78.9 & 32.1 & 47.6 & 44.6 & 67.3 & 52.4 & 83.5 & 51.6 & 33.9 & 91.2 & 69.8 & 92.3 & 85.8 & 91.4 & 96.2 & 62.6 \\
AIoU
 & 45.9 & 17.5 & 74.0 & 25.0 & 78.8 & 32.6 & 28.0 & 42.5 & 64.9 & 52.3 & 83.5 & 51.0 & 29.8 & 90.9 & 70.8 & 91.5 & 85.5 & 90.4 & 96.2 & 60.6 \\
BvSB
& 46.1 & 31.4 & 74.6 & 47.2 & 78.7 & 37.9 & 43.6 & 30.4 & 62.1 & 55.6 & 81.7 & 55.8 & 46.0 & 92.0 & 73.5 & 91.1 & 86.6 & 91.9 & 96.3 & 64.4 \\
ClassBal
 & 48.0 & 30.0 & 74.9 & 54.3 & 78.5 & 39.5 & 45.7 & 39.9 & 62.1 & 57.6 & 82.6 & 55.0 & 46.0 & 92.2 & 74.6 & 91.6 & 86.7 & 91.7 & 96.5 & 65.7 \\
MerSpx
& 49.6 & 34.8 & 75.5 & 54.3 & 78.7 & 39.4 & 44.8 & 39.9 & 62.4 & 54.6 & 83.7 & 55.8 & 46.5 & 92.0 & 72.7 & 91.4 & 86.6 & 91.8 & 96.1 & 65.8 \\
SIM
& 18.2 & 45.7 & 71.6 & 93.0 & 77.6 & 89.9 & 87.6 & 48.2 & 71.0 & \textbf{86.5} & 85.0 & 75.8 & \textbf{85.5} & \textbf{97.2} & \textbf{85.3} & 97.7 & \textbf{91.6} & \textbf{95.0} & \textbf{98.2} & 79.0 \\

\textbf{ABC (Ours)} 
& \textbf{66.6} & \textbf{66.8} & \textbf{85.5} & \textbf{93.9} & \textbf{82.7} & \textbf{95.4} & \textbf{94.1} & \textbf{66.2} & \textbf{73.0} & 80.8 & \textbf{88.5} & \textbf{76.9} & 84.9 & 96.5 & 81.8 & \textbf{97.7} & 89.9 & 94.4 & 97.7 & \textbf{84.9} \\

\bottomrule
\end{tabular}
\caption{Analysis of class-wise performance of different acquisition functions (Data mIoU, \%).
Label correction is performed in a single round with 50k budgets on Cityscapes. 
`Init.' represents performance of the initial pseudo-labels generated by Grounded SAM. Our proposed acquisition function exhibits superior performance on tail classes, which have been largely overlooked in prior studies, demonstrating its effectiveness in mitigating class imbalance.
}
\label{tab:acq_city_claswise_mIoU}
\end{table*}

\subsection{Ablation Study on Adaptive Class Weight}

Adaptive class weight is composed of two components: the class rarity score $\hat{w}(x)$, which prioritizes pixels belonging to tail classes during sampling, and the dataset imbalance score $\text{KL}^3(\mathbb{P}_{\text{dist}} \parallel \mathbb{U}_{\text{dist}})$, which adaptively emphasizes $\hat{w}(x)$ based on the overall imbalance level of the dataset. 
In Table~\ref{tab:ablation_acq}, we present an ablation study to analyze their individual contributions. 
The results show that applying both the class rarity score and the dataset imbalance score achieves better performance than using the class rarity score alone.
Notably, our proposed $\hat{w}(x)^{\text{KL}^3(\mathbb{P}{\text{dist}} \parallel \mathbb{U}{\text{dist}})}$ achieves the most significant improvement on both datasets, demonstrating the effectiveness of our weighting formulation.

\subsection{Detailed Comparative Analysis}

In Table~\ref{tab:acq_city_claswise_mIoU}, we present a class-wise quantitative comparison of different acquisition functions to analyze the effectiveness of our proposed ABC function.
The superior performance on tail classes demonstrates that ABC effectively addresses class imbalance challenge.
Although there is a slight decrease in head class performance compared to the baseline, which can be attributed to allocating a larger sampling proportion to tail classes under a limited annotation budget, this trade-off is minor and is outweighed by the substantial improvements in tail class accuracy.

\subsection{Detailed Performance Analysis}

In Figure~\ref{fig:cityscapes19}, we provide a more fine-grained analysis of the experiment shown in Figure~\ref{fig:10r_figures}, by tracking class-wise performance gains across multiple correction rounds to evaluate the effectiveness of our framework.
Label correction is performed for ten rounds with a budget of 10k clicks per round on Cityscapes.
The baseline shows limited updates for tail classes (e.g., \textit{traffic light}, \textit{rider}, \textit{motorcycle}) even after multiple correction rounds, suggesting that its budget is not effectively allocated to underrepresented categories.
In contrast, our framework yields substantial improvements on these tail classes, demonstrating that the proposed ABC function effectively directs corrections toward them.
Although this budget reallocation introduces a temporary trade-off, where head classes (e.g., \textit{sky}, \textit{vegetation}) underperform slightly compared to the baseline during the early rounds, its impact is negligible relative to the gains achieved for tail classes, and the performance of head classes eventually surpasses the baseline in later rounds.

\begin{figure*}[t]
\centering
% ---------------- Legend ----------------
\begin{tikzpicture}
\node {
    \begin{tabular}{llll}
        % ALC
        {\tikz{\filldraw[draw=red, fill=red, thick] 
            (90:2.2pt) -- (162:2.2pt) --
            (234:2.2pt) -- (306:2.2pt) --
            (18:2.2pt) -- cycle;}} & \textbf{ALC} &
        % A2LC
        {\tikz{\filldraw[draw=blue, fill=blue, thick]
            (0,0) rectangle (3.5pt,3.5pt);}} &
        \textbf{A\(^2\)LC}
    \end{tabular}
};
\end{tikzpicture}

\vspace{0.5em}

% ------------------- Main 4-column Table -------------------
\begin{tabular}{c @{\hskip -2em} c @{\hskip -2em} c @{\hskip -2em} c}

\begin{subfigure}{0.249\linewidth}
\centering
% ---------- Row 1 ----------
% Traffic Light
\begin{tikzpicture}
    \begin{axis}[        
        width=4.45cm, height=3.4cm,
        ymin=0, ymax=90, xmin=0, xmax=10,
        xtick={0,...,10}, ytick={0,30,...,100},
        xlabel style={yshift=0.27cm}, ylabel style={xshift=0.05cm, yshift=-0.6cm}, 
        label style={font=\scriptsize}, tick label style={font=\scriptsize}
    ]
    \addplot[red, line width=1.2pt, mark=pentagon*, mark size=1pt] table[col sep=comma, x=x, y=Traffic_Light]{csv/supp_city_19_alc.csv};
    \addplot[blue, line width=1.2pt, mark=square*, mark size=1pt] table[col sep=comma, x=x, y=Traffic_Light]{csv/supp_city_19_a2lc.csv};
    \end{axis}
\end{tikzpicture}
\caption*{(a) \textit{Traffic light} }
\end{subfigure} &
% Rider
\begin{subfigure}{0.249\linewidth}
\centering
\begin{tikzpicture}
    \begin{axis}[        
        width=4.45cm, height=3.4cm,
        ymin=0, ymax=90, xmin=0, xmax=10,
        xtick={0,...,10}, ytick={0,30,...,100},
        xlabel style={yshift=0.27cm}, ylabel style={xshift=0.05cm, yshift=-0.6cm}, 
        label style={font=\scriptsize}, tick label style={font=\scriptsize}
    ]
    \addplot[red, line width=1.2pt, mark=pentagon*, mark size=1pt] table[col sep=comma, x=x, y=Rider]{csv/supp_city_19_alc.csv};
    \addplot[blue, line width=1.2pt, mark=square*, mark size=1pt] table[col sep=comma, x=x, y=Rider]{csv/supp_city_19_a2lc.csv};
    \end{axis}
\end{tikzpicture}
\caption*{(b) \textit{Rider}}
\end{subfigure} &
% Motorcycle
\begin{subfigure}{0.249\linewidth}
\centering
\begin{tikzpicture}
   \begin{axis}[        
        width=4.45cm, height=3.4cm,
        ymin=50, ymax=90, xmin=0, xmax=10,
        xtick={0,...,10}, ytick={0,10,...,100},
        xlabel style={yshift=0.27cm}, ylabel style={xshift=0.05cm, yshift=-0.6cm}, 
        label style={font=\scriptsize}, tick label style={font=\scriptsize}
    ]
    \addplot[red, line width=1.2pt, mark=pentagon*, mark size=1pt] table[col sep=comma, x=x, y=Motorcycle]{csv/supp_city_19_alc.csv};
    \addplot[blue, line width=1.2pt, mark=square*, mark size=1pt] table[col sep=comma, x=x, y=Motorcycle]{csv/supp_city_19_a2lc.csv};
    \end{axis}
\end{tikzpicture}
\caption*{(c) \textit{Motorcycle}}
\end{subfigure} &
% Train
\begin{subfigure}{0.249\linewidth}
\centering
\begin{tikzpicture}
   \begin{axis}[        
        width=4.45cm, height=3.4cm,
        ymin=10, ymax=100, xmin=0, xmax=10,
        xtick={0,...,10}, ytick={0,20,...,100},
        xlabel style={yshift=0.27cm}, ylabel style={xshift=0.05cm, yshift=-0.6cm}, 
        label style={font=\scriptsize}, tick label style={font=\scriptsize}
    ]
    \addplot[red, line width=1.2pt, mark=pentagon*, mark size=1pt] table[col sep=comma, x=x, y=Train]{csv/supp_city_19_alc.csv};
    \addplot[blue, line width=1.2pt, mark=square*, mark size=1pt] table[col sep=comma, x=x, y=Train]{csv/supp_city_19_a2lc.csv};
    \end{axis}
\end{tikzpicture}
\caption*{(d) \textit{Train}}
\end{subfigure} \\
% ---------- Row 2 ----------
% Bicycle
\begin{subfigure}{0.249\linewidth}
\centering
\begin{tikzpicture}
    \begin{axis}[        
        width=4.45cm, height=3.4cm,
        ymin=60, ymax=90, xmin=0, xmax=10,
        xtick={0,...,10}, ytick={0,10,...,100},
        xlabel style={yshift=0.27cm}, ylabel style={xshift=0.05cm, yshift=-0.6cm}, 
        label style={font=\scriptsize}, tick label style={font=\scriptsize}
    ]
    \addplot[red, line width=1.2pt, mark=pentagon*, mark size=1pt] table[col sep=comma, x=x, y=Bicycle]{csv/supp_city_19_alc.csv};
    \addplot[blue, line width=1.2pt, mark=square*, mark size=1pt] table[col sep=comma, x=x, y=Bicycle]{csv/supp_city_19_a2lc.csv};
    \end{axis}
\end{tikzpicture}
\caption*{(e) \textit{Bicycle} }
\end{subfigure} &
% Bus
\begin{subfigure}{0.249\linewidth}
\centering
\begin{tikzpicture}
    \begin{axis}[        
        width=4.45cm, height=3.4cm,
        ymin=20, ymax=100, xmin=0, xmax=10,
        xtick={0,...,10}, ytick={0,20,...,100},
        xlabel style={yshift=0.27cm}, ylabel style={xshift=0.05cm, yshift=-0.6cm}, 
        label style={font=\scriptsize}, tick label style={font=\scriptsize}
    ]
    \addplot[red, line width=1.2pt, mark=pentagon*, mark size=1pt] table[col sep=comma, x=x, y=Bus]{csv/supp_city_19_alc.csv};
    \addplot[blue, line width=1.2pt, mark=square*, mark size=1pt] table[col sep=comma, x=x, y=Bus]{csv/supp_city_19_a2lc.csv};
    \end{axis}
\end{tikzpicture}
\caption*{(f) \textit{Bus}}
\end{subfigure} &
% Truck
\begin{subfigure}{0.249\linewidth}
\centering
\begin{tikzpicture}
   \begin{axis}[        
        width=4.45cm, height=3.4cm,
        ymin=20, ymax=100, xmin=0, xmax=10,
        xtick={0,...,10}, ytick={0,20,...,100},
        xlabel style={yshift=0.27cm}, ylabel style={xshift=0.05cm, yshift=-0.6cm}, 
        label style={font=\scriptsize}, tick label style={font=\scriptsize}
    ]
    \addplot[red, line width=1.2pt, mark=pentagon*, mark size=1pt] table[col sep=comma, x=x, y=Truck]{csv/supp_city_19_alc.csv};
    \addplot[blue, line width=1.2pt, mark=square*, mark size=1pt] table[col sep=comma, x=x, y=Truck]{csv/supp_city_19_a2lc.csv};
    \end{axis}
\end{tikzpicture}
\caption*{(g) \textit{Truck}}
\end{subfigure} &
% Traffic Sign
\begin{subfigure}{0.249\linewidth}
\centering
\begin{tikzpicture}
   \begin{axis}[        
        width=4.45cm, height=3.4cm,
        ymin=20, ymax=80, xmin=0, xmax=10,
        xtick={0,...,10}, ytick={0,20,...,100},
        xlabel style={yshift=0.27cm}, ylabel style={xshift=0.05cm, yshift=-0.6cm}, 
        label style={font=\scriptsize}, tick label style={font=\scriptsize}
    ]
    \addplot[red, line width=1.2pt, mark=pentagon*, mark size=1pt] table[col sep=comma, x=x, y=Traffic_Sign]{csv/supp_city_19_alc.csv};
    \addplot[blue, line width=1.2pt, mark=square*, mark size=1pt] table[col sep=comma, x=x, y=Traffic_Sign]{csv/supp_city_19_a2lc.csv};
    \end{axis}
\end{tikzpicture}
\caption*{(h) \textit{Traffic sign}}
\end{subfigure} \\
% ---------- Row 3 ----------
% Pole
\begin{subfigure}{0.249\linewidth}
\centering
\begin{tikzpicture}
    \begin{axis}[        
        width=4.45cm, height=3.4cm,
        ymin=50, ymax=80, xmin=0, xmax=10,
        xtick={0,...,10}, ytick={0,10,...,100},
        xlabel style={yshift=0.27cm}, ylabel style={xshift=0.05cm, yshift=-0.6cm}, 
        label style={font=\scriptsize}, tick label style={font=\scriptsize}
    ]
    \addplot[red, line width=1.2pt, mark=pentagon*, mark size=1pt] table[col sep=comma, x=x, y=Pole]{csv/supp_city_19_alc.csv};
    \addplot[blue, line width=1.2pt, mark=square*, mark size=1pt] table[col sep=comma, x=x, y=Pole]{csv/supp_city_19_a2lc.csv};
    \end{axis}
\end{tikzpicture}
\caption*{(i) \textit{Pole} }
\end{subfigure} &
% Terrain
\begin{subfigure}{0.249\linewidth}
\centering
\begin{tikzpicture}
    \begin{axis}[        
        width=4.45cm, height=3.4cm,
        ymin=30, ymax=90, xmin=0, xmax=10,
        xtick={0,...,10}, ytick={0,20,...,100},
        xlabel style={yshift=0.27cm}, ylabel style={xshift=0.05cm, yshift=-0.6cm}, 
        label style={font=\scriptsize}, tick label style={font=\scriptsize}
    ]
    \addplot[red, line width=1.2pt, mark=pentagon*, mark size=1pt] table[col sep=comma, x=x, y=Terrain]{csv/supp_city_19_alc.csv};
    \addplot[blue, line width=1.2pt, mark=square*, mark size=1pt] table[col sep=comma, x=x, y=Terrain]{csv/supp_city_19_a2lc.csv};
    \end{axis}
\end{tikzpicture}
\caption*{(j) \textit{Terrain}}
\end{subfigure} &
% Person
\begin{subfigure}{0.249\linewidth}
\centering
\begin{tikzpicture}
   \begin{axis}[        
        width=4.45cm, height=3.4cm,
        ymin=60, ymax=100, xmin=0, xmax=10,
        xtick={0,...,10}, ytick={0,10,...,100},
        xlabel style={yshift=0.27cm}, ylabel style={xshift=0.05cm, yshift=-0.6cm}, 
        label style={font=\scriptsize}, tick label style={font=\scriptsize}
    ]
    \addplot[red, line width=1.2pt, mark=pentagon*, mark size=1pt] table[col sep=comma, x=x, y=Person]{csv/supp_city_19_alc.csv};
    \addplot[blue, line width=1.2pt, mark=square*, mark size=1pt] table[col sep=comma, x=x, y=Person]{csv/supp_city_19_a2lc.csv};
    \end{axis}
\end{tikzpicture}
\caption*{(k) \textit{Person}}
\end{subfigure} &
% Fence
\begin{subfigure}{0.249\linewidth}
\centering
\begin{tikzpicture}
   \begin{axis}[        
        width=4.45cm, height=3.4cm,
        ymin=30, ymax=90, xmin=0, xmax=10,
        xtick={0,...,10}, ytick={30,50,...,100},
        xlabel style={yshift=0.27cm}, ylabel style={xshift=0.05cm, yshift=-0.6cm}, 
        label style={font=\scriptsize}, tick label style={font=\scriptsize}
    ]
    \addplot[red, line width=1.2pt, mark=pentagon*, mark size=1pt] table[col sep=comma, x=x, y=Fence]{csv/supp_city_19_alc.csv};
    \addplot[blue, line width=1.2pt, mark=square*, mark size=1pt] table[col sep=comma, x=x, y=Fence]{csv/supp_city_19_a2lc.csv};
    \end{axis}
\end{tikzpicture}
\caption*{(l) \textit{Fence}}
\end{subfigure} \\
% ---------- Row 4 ----------
% Wall
\begin{subfigure}{0.249\linewidth}
\centering
\begin{tikzpicture}
    \begin{axis}[        
        width=4.45cm, height=3.4cm,
        ymin=0, ymax=100, xmin=0, xmax=10,
        xtick={0,...,10}, ytick={0,20,...,100},
        xlabel style={yshift=0.27cm}, ylabel style={xshift=0.05cm, yshift=-0.6cm}, 
        label style={font=\scriptsize}, tick label style={font=\scriptsize}
    ]
    \addplot[red, line width=1.2pt, mark=pentagon*, mark size=1pt] table[col sep=comma, x=x, y=Wall]{csv/supp_city_19_alc.csv};
    \addplot[blue, line width=1.2pt, mark=square*, mark size=1pt] table[col sep=comma, x=x, y=Wall]{csv/supp_city_19_a2lc.csv};
    \end{axis}
\end{tikzpicture}
\caption*{(m) \textit{Wall}}
\end{subfigure} &
% Sky
\begin{subfigure}{0.249\linewidth}
\centering
\begin{tikzpicture}
    \begin{axis}[        
        width=4.45cm, height=3.4cm,
        ymin=90, ymax=100, xmin=0, xmax=10,
        xtick={0,...,10}, ytick={0,2,...,100},
        xlabel style={yshift=0.27cm}, ylabel style={xshift=0.05cm, yshift=-0.6cm}, 
        label style={font=\scriptsize}, tick label style={font=\scriptsize}
    ]
    \addplot[red, line width=1.2pt, mark=pentagon*, mark size=1pt] table[col sep=comma, x=x, y=Sky]{csv/supp_city_19_alc.csv};
    \addplot[blue, line width=1.2pt, mark=square*, mark size=1pt] table[col sep=comma, x=x, y=Sky]{csv/supp_city_19_a2lc.csv};
    \end{axis}
\end{tikzpicture}
\caption*{(n) \textit{Sky}}
\end{subfigure} &
% Sidewalk
\begin{subfigure}{0.249\linewidth}
\centering
\begin{tikzpicture}
   \begin{axis}[        
        width=4.45cm, height=3.4cm,
        ymin=65, ymax=85, xmin=0, xmax=10,
        xtick={0,...,10}, ytick={0,5,...,100},
        xlabel style={yshift=0.27cm}, ylabel style={xshift=0.05cm, yshift=-0.6cm}, 
        label style={font=\scriptsize}, tick label style={font=\scriptsize}
    ]
    \addplot[red, line width=1.2pt, mark=pentagon*, mark size=1pt] table[col sep=comma, x=x, y=Sidewalk]{csv/supp_city_19_alc.csv};
    \addplot[blue, line width=1.2pt, mark=square*, mark size=1pt] table[col sep=comma, x=x, y=Sidewalk]{csv/supp_city_19_a2lc.csv};
    \end{axis}
\end{tikzpicture}
\caption*{(o) \textit{Sidewalk}}
\end{subfigure} &
% Car
\begin{subfigure}{0.249\linewidth}
\centering
\begin{tikzpicture}
   \begin{axis}[        
        width=4.45cm, height=3.4cm,
        ymin=85, ymax=100, xmin=0, xmax=10,
        xtick={0,...,10}, ytick={0,5,...,100},
        xlabel style={yshift=0.27cm}, ylabel style={xshift=0.05cm, yshift=-0.6cm}, 
        label style={font=\scriptsize}, tick label style={font=\scriptsize}
    ]
    \addplot[red, line width=1.2pt, mark=pentagon*, mark size=1pt] table[col sep=comma, x=x, y=Car]{csv/supp_city_19_alc.csv};
    \addplot[blue, line width=1.2pt, mark=square*, mark size=1pt] table[col sep=comma, x=x, y=Car]{csv/supp_city_19_a2lc.csv};
    \end{axis}
\end{tikzpicture}
\caption*{(p) \textit{Car}}
\end{subfigure} \\
% ---------- Row 5 ----------
% Vegetation
\begin{subfigure}{0.249\linewidth}
\centering
\begin{tikzpicture}
    \begin{axis}[        
        width=4.45cm, height=3.4cm,
        ymin=75, ymax=95, xmin=0, xmax=10,
        xtick={0,...,10}, ytick={0,5,...,100},
        xlabel style={yshift=0.27cm}, ylabel style={xshift=0.05cm, yshift=-0.6cm}, 
        label style={font=\scriptsize}, tick label style={font=\scriptsize}
    ]
    \addplot[red, line width=1.2pt, mark=pentagon*, mark size=1pt] table[col sep=comma, x=x, y=Vegetation]{csv/supp_city_19_alc.csv};
    \addplot[blue, line width=1.2pt, mark=square*, mark size=1pt] table[col sep=comma, x=x, y=Vegetation]{csv/supp_city_19_a2lc.csv};
    \end{axis}
\end{tikzpicture}
\caption*{(q) \textit{Vegetation}}
\end{subfigure} &
% Building
\begin{subfigure}{0.249\linewidth}
\centering
\begin{tikzpicture}
    \begin{axis}[        
        width=4.45cm, height=3.4cm,
        ymin=80, ymax=95, xmin=0, xmax=10,
        xtick={0,...,10}, ytick={0,5,...,100},
        xlabel style={yshift=0.27cm}, ylabel style={xshift=0.05cm, yshift=-0.6cm}, 
        label style={font=\scriptsize}, tick label style={font=\scriptsize}
    ]
    \addplot[red, line width=1.2pt, mark=pentagon*, mark size=1pt] table[col sep=comma, x=x, y=Building]{csv/supp_city_19_alc.csv};
    \addplot[blue, line width=1.2pt, mark=square*, mark size=1pt] table[col sep=comma, x=x, y=Building]{csv/supp_city_19_a2lc.csv};
    \end{axis}
\end{tikzpicture}
\caption*{(r) \textit{Building}}
\end{subfigure} &
% Road
\begin{subfigure}{0.249\linewidth}
\centering
\begin{tikzpicture}
   \begin{axis}[        
        width=4.45cm, height=3.4cm,
        ymin=90, ymax=100, xmin=0, xmax=10,
        xtick={0,...,10}, ytick={0,2,...,100},
        xlabel style={yshift=0.27cm}, ylabel style={xshift=0.05cm, yshift=-0.6cm}, 
        label style={font=\scriptsize}, tick label style={font=\scriptsize}
    ]
    \addplot[red, line width=1.2pt, mark=pentagon*, mark size=1pt] table[col sep=comma, x=x, y=Road]{csv/supp_city_19_alc.csv};
    \addplot[blue, line width=1.2pt, mark=square*, mark size=1pt] table[col sep=comma, x=x, y=Road]{csv/supp_city_19_a2lc.csv};
    \end{axis}
\end{tikzpicture}
\caption*{(s) \textit{Road}}
\end{subfigure} & \\
\end{tabular}
\caption{Analysis of class-wise performance across multiple rounds (Data mIoU, \%). 
Figures (a)–(s) are ordered by the initial pixel frequency of each class in ascending order, where the x axis denotes the number of clicks ($\times 10^4$) and the y axis indicates the data mIoU (\%).
}
\label{fig:cityscapes19}
\end{figure*}

\begin{table}[t]
    \centering
    \begin{tabular}{ccccc} 
        \toprule
        \multicolumn{3}{c}{\textbf{Methods}} 
        & \textbf{Cityscapes} 
        & \textbf{PASCAL} \\ 
        LCM & ABC & Mask 
        & 1R
        & 1R \\
        \midrule
          &   &  
        & 76.93 & 83.04 \\
        \ding{51} &   &   
        & 78.28 & 87.07 \\
        \ding{51} & \ding{51} &   
        & 84.36 & 87.08 \\
        \ding{51} & \ding{51} & \ding{51} 
        & \textbf{84.92} & \textbf{87.38} \\
        \bottomrule
    \end{tabular}
\caption{Ablation study on our proposed method (Data mIoU, \%).}
    \label{tab:mIoU_city_pascal_1R_ablation}
\end{table}

\begin{table}[t]
    \centering
    \setlength\tabcolsep{5.5pt} 
    \begin{tabular}{ccccccc} 
        \toprule
        \multicolumn{3}{c}{\textbf{Methods}} 
        & \multicolumn{2}{c}{\textbf{Cityscapes}} 
        & \multicolumn{2}{c}{\textbf{PASCAL}} \\ 
        LCM & ABC & Mask 
        & 1R & 2R
        & 1R & 2R \\
        \midrule
          &   &  
        & 69.29 & 72.38 & 78.55 & 82.96 \\
        \ding{51} &   &   
        & 72.04 & 79.85 & 78.88 & 87.42 \\
        \ding{51} & \ding{51} &   
        & 78.50 & 83.84 & 78.90 & 87.45 \\
        \ding{51} & \ding{51} & \ding{51} 
        & \textbf{79.30} & \textbf{84.38} & \textbf{79.10} & \textbf{87.82} \\
        \bottomrule
    \end{tabular}
\caption{Ablation study on our proposed method (Data mIoU, \%).}
\label{tab:mIoU_city_pascal_2R_ablation}
\end{table}

\section{Additional Ablation Studies}

While the main paper presents ablation results over five correction rounds for the Cityscapes dataset, we further conduct ablation studies under different round settings on both Cityscapes and PASCAL.
Table~\ref{tab:mIoU_city_pascal_1R_ablation} presents results with label correction performed in a single round, whereas Table~\ref{tab:mIoU_city_pascal_2R_ablation} shows results with two rounds of correction.
For both settings, 50k budgets are used for Cityscapes and 5k budgets for PASCAL, evenly distributed across rounds in the two round case.
We observe that applying all components of our proposed method consistently yields the best performance across all settings.

\section{Motivation of Mask-Level Correction}

\begin{figure*}[t]
    \centering     
    \begin{minipage}{0.9\textwidth}
        \begin{subfigure}[h!]{0.24\linewidth}
            \centering
            \includegraphics[width=\linewidth, height=0.7\linewidth]{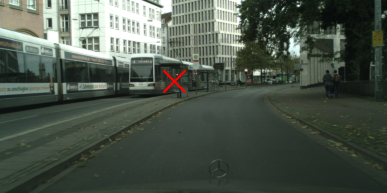}
        \end{subfigure}
        \begin{subfigure}[h!]{0.24\linewidth}
            \centering
            \includegraphics[width=\linewidth, height=0.7\linewidth]{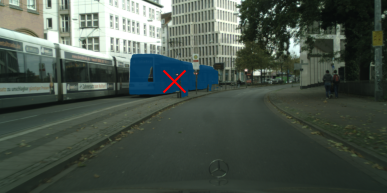}
        \end{subfigure}
        \begin{subfigure}[h!]{0.24\linewidth}
            \centering
            \includegraphics[width=\linewidth, height=0.7\linewidth]{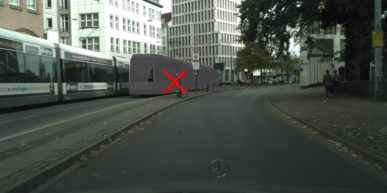}
        \end{subfigure}
        \begin{subfigure}[h!]{0.24\linewidth}
            \centering
            \includegraphics[width=\linewidth, height=0.7\linewidth]{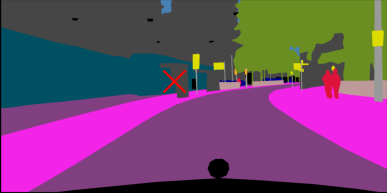}
        \end{subfigure}        
        
        \begin{subfigure}[h!]{0.24\linewidth}
            \centering
            \includegraphics[width=\linewidth, height=0.7\linewidth]{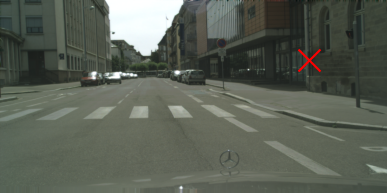}
        \end{subfigure}
        \begin{subfigure}[h!]{0.24\linewidth}
            \centering
            \includegraphics[width=\linewidth, height=0.7\linewidth]{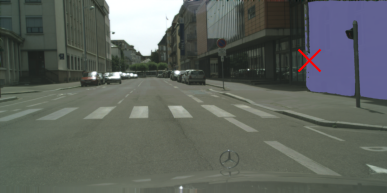}
        \end{subfigure}
        \begin{subfigure}[h!]{0.24\linewidth}
            \centering
            \includegraphics[width=\linewidth, height=0.7\linewidth]{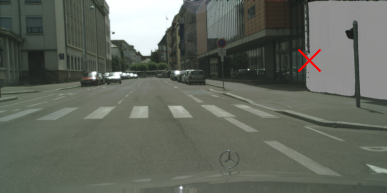}
        \end{subfigure}
        \begin{subfigure}[h!]{0.24\linewidth}
            \centering
            \includegraphics[width=\linewidth, height=0.7\linewidth]{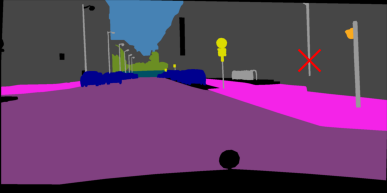}
        \end{subfigure}        
        
        \begin{subfigure}[h!]{0.24\linewidth}
            \centering
            \includegraphics[width=\linewidth, height=0.7\linewidth]{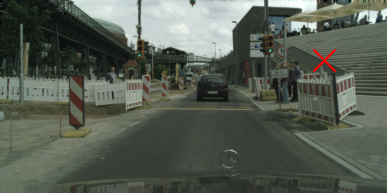}
            \caption*{(a) Unlabeled Image\newline}
        \end{subfigure}
        \begin{subfigure}[h!]{0.24\linewidth}
            \centering
            \includegraphics[width=\linewidth, height=0.7\linewidth]{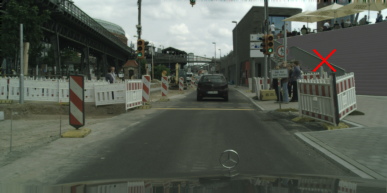}
            \caption*{(b) Before Correction\newline}
        \end{subfigure}
        \begin{subfigure}[h!]{0.24\linewidth}
            \centering
            \includegraphics[width=\linewidth, height=0.7\linewidth]{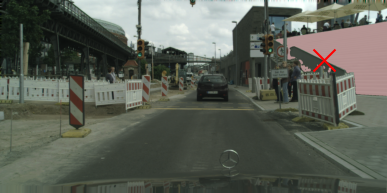}
            \caption*{(c) After Correction\newline}
        \end{subfigure}
        \begin{subfigure}[h!]{0.24\linewidth}
            \centering
            \includegraphics[width=\linewidth, height=0.7\linewidth]{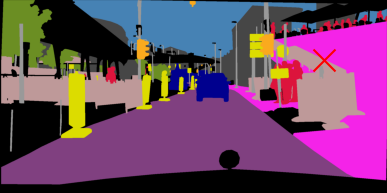}
            \caption*{(d) Ground Truth\newline}
        \end{subfigure}
    \end{minipage}
    
    \centering \makebox[\linewidth][c]{\textbf{(a) Cityscapes}}\newline \par

    \begin{minipage}{0.9\textwidth}
        \begin{subfigure}[h!]{0.24\linewidth}
            \centering
            \includegraphics[width=\linewidth, height=0.7\linewidth]{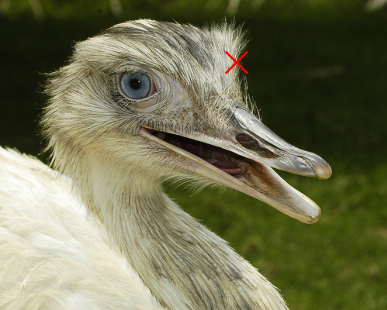}
        \end{subfigure}
        \begin{subfigure}[h!]{0.24\linewidth}
            \centering
            \includegraphics[width=\linewidth, height=0.7\linewidth]{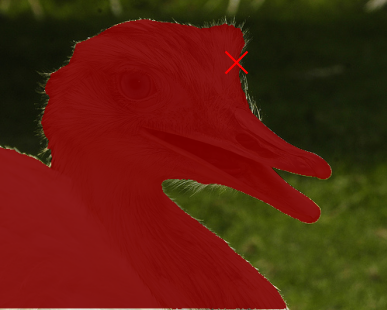}
        \end{subfigure}
        \begin{subfigure}[h!]{0.24\linewidth}
            \centering
            \includegraphics[width=\linewidth, height=0.7\linewidth]{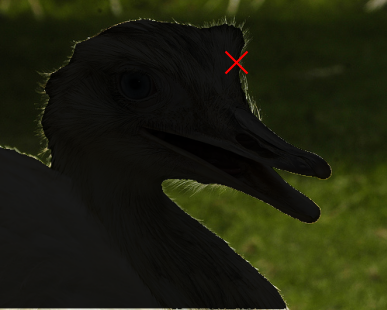}
        \end{subfigure}
        \begin{subfigure}[h!]{0.24\linewidth}
            \centering
            \includegraphics[width=\linewidth, height=0.7\linewidth]{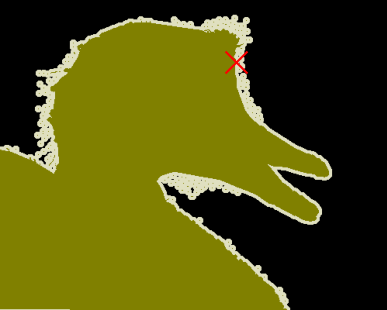}
        \end{subfigure}

        \begin{subfigure}[h!]{0.24\linewidth}
            \centering
            \includegraphics[width=\linewidth, height=0.7\linewidth]{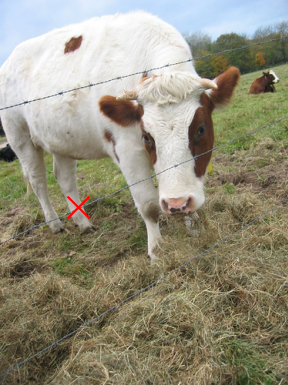}
        \end{subfigure}
        \begin{subfigure}[h!]{0.24\linewidth}
            \centering
            \includegraphics[width=\linewidth, height=0.7\linewidth]{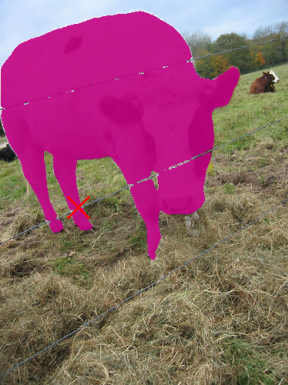}
        \end{subfigure}
        \begin{subfigure}[h!]{0.24\linewidth}
            \centering
            \includegraphics[width=\linewidth, height=0.7\linewidth]{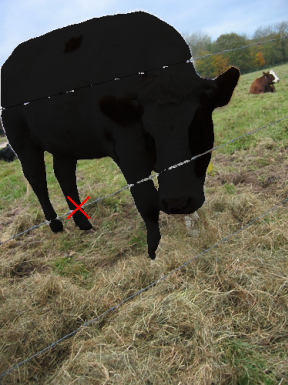}
        \end{subfigure}
        \begin{subfigure}[h!]{0.24\linewidth}
            \centering
            \includegraphics[width=\linewidth, height=0.7\linewidth]{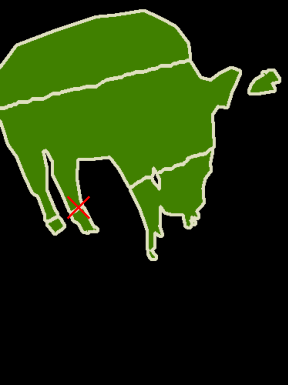}
        \end{subfigure}
        
        \begin{subfigure}[h!]{0.24\linewidth}
            \centering
            \includegraphics[width=\linewidth, height=0.7\linewidth]{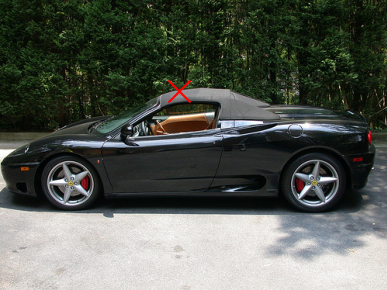}
            \caption*{(a) Unlabeled Image\newline}
        \end{subfigure}
        \begin{subfigure}[h!]{0.24\linewidth}
            \centering
            \includegraphics[width=\linewidth, height=0.7\linewidth]{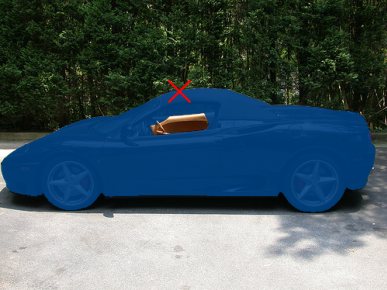}
            \caption*{(b) Before Correction\newline}
        \end{subfigure}
        \begin{subfigure}[h!]{0.24\linewidth}
            \centering
            \includegraphics[width=\linewidth, height=0.7\linewidth]{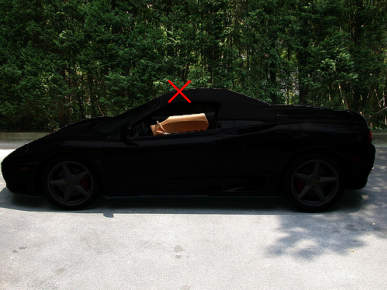}
            \caption*{(c) After Correction\newline}
        \end{subfigure}
        \begin{subfigure}[h!]{0.24\linewidth}
            \centering
            \includegraphics[width=\linewidth, height=0.7\linewidth]{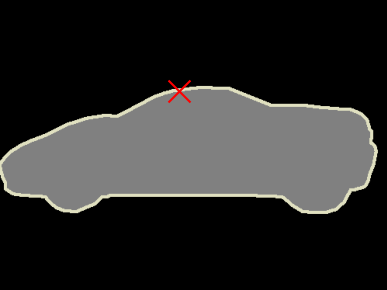}
            \caption*{(d) Ground Truth\newline}
        \end{subfigure}
    \end{minipage}
    \centering \makebox[\linewidth][c]{\textbf{(b) PASCAL VOC 2012}} \par
    
\caption{Limitation of pixel-level query.
For Cityscapes, the first, second, and third rows respectively show miscorrections where incorrect labels \textit{bus}, \textit{wall}, and \textit{building} are miscorrected to \textit{building}, \textit{pole}, and \textit{fence}, instead of their correct labels \textit{train}, \textit{building}, and \textit{sidewalk}. For  PASCAL, all three rows exhibit miscorrections where incorrect labels \textit{aeroplane}, \textit{horse}, and \textit{tv/monitor} are all miscorrected to \textit{ignore\_index} instead of their true labels \textit{bird}, \textit{cow}, and \textit{car}, respectively.}
\label{fig:visualize_pixelcorrection}
\end{figure*}

Querying annotators for a single pixel (i.e., representative pixel) label per mask, as performed in previous studies (i.e., pixel-level query), frequently leads to erroneous corrections.
This, in turn, results in substantial waste of annotation time and human resources.
Figure~\ref{fig:visualize_pixelcorrection} illustrates failure cases resulting from the pixel-level query adopted in prior work, where multiple pixels rely on the label of a single queried pixel, leading to incorrect corrections. This limitation motivated us to develop a mask-level query and correction approach.
We denote the queried representative pixel with red `X' marker, while the second and third columns illustrate the masks, colored by their respective class labels before and after correction.

\section{Qualitative Results}
\subsection{Automatically Corrected Masks}
\begin{figure*}[t]
\centering    
\begin{minipage}{0.94\textwidth}
    \centering
    \begin{minipage}{\textwidth}
        \begin{minipage}{0.02\textwidth}
            \rotatebox{90}{\small \textbf{\textit{sidewalk}}}
        \end{minipage}\hspace{-0.5em}%
        \begin{minipage}{0.98\textwidth}
            \centering
            \begin{subfigure}[h!]{0.14\linewidth}
                \centering
                \includegraphics[width=\linewidth, height=0.7\linewidth]{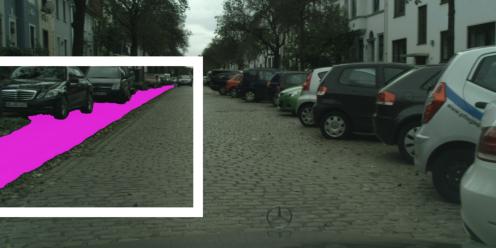}                
            \end{subfigure}
            \begin{subfigure}[h!]{0.14\linewidth}
                \centering
                \includegraphics[width=\linewidth, height=0.7\linewidth]{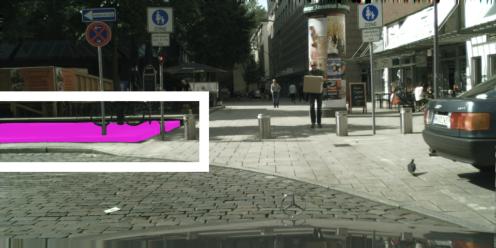}
            \end{subfigure}
            \begin{subfigure}[h!]{0.14\linewidth}
                \centering
                \includegraphics[width=\linewidth, height=0.7\linewidth]{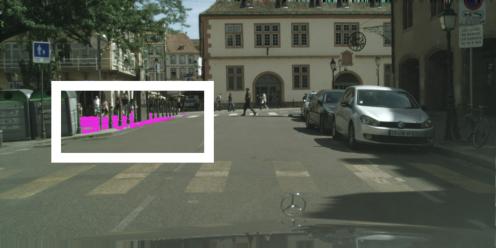}
            \end{subfigure}
            \begin{subfigure}[h!]{0.14\linewidth}
                \centering
                \includegraphics[width=\linewidth, height=0.7\linewidth]{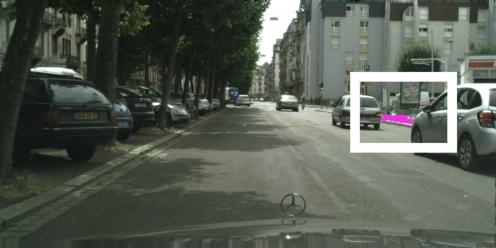}
            \end{subfigure}
            \begin{subfigure}[h!]{0.14\linewidth}
                \centering
                \includegraphics[width=\linewidth, height=0.7\linewidth]{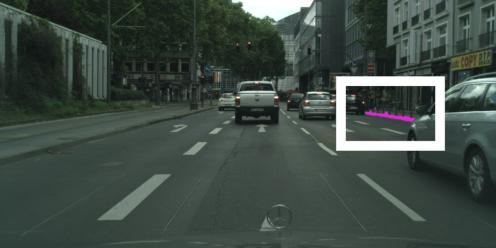}
            \end{subfigure}
            \begin{subfigure}[h!]{0.14\linewidth}
                \centering
                \includegraphics[width=\linewidth, height=0.7\linewidth]{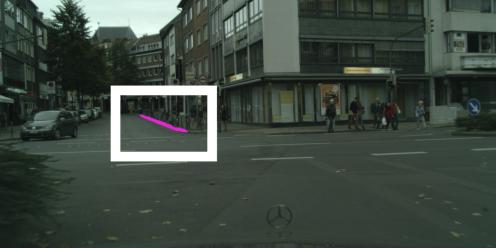}
            \end{subfigure}
        \end{minipage}

        \vspace{0.25em}
    
        \begin{minipage}{0.02\textwidth}
            \rotatebox{90}{\small \textbf{\textit{fence}}}
        \end{minipage}\hspace{-0.5em}%
        \begin{minipage}{0.98\textwidth}
            \centering
            \begin{subfigure}[h!]{0.14\linewidth}
                \centering
                \includegraphics[width=\linewidth, height=0.7\linewidth]{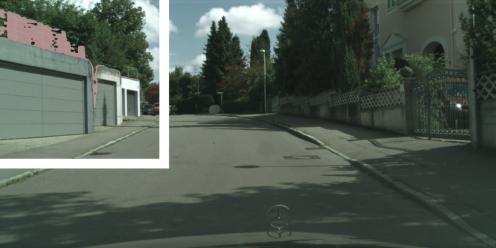}
            \end{subfigure}
            \begin{subfigure}[h!]{0.14\linewidth}
                \centering
                \includegraphics[width=\linewidth, height=0.7\linewidth]{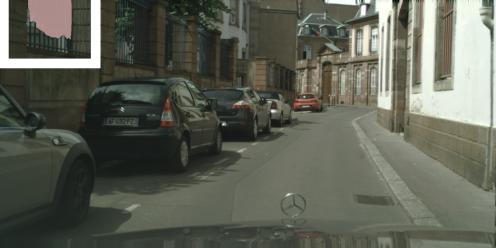}
            \end{subfigure}
            \begin{subfigure}[h!]{0.14\linewidth}
                \centering
                \includegraphics[width=\linewidth, height=0.7\linewidth]{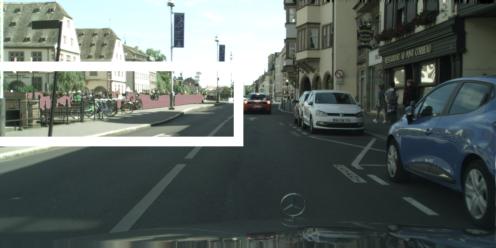}
            \end{subfigure}
            \begin{subfigure}[h!]{0.14\linewidth}
                \centering
                \includegraphics[width=\linewidth, height=0.7\linewidth]{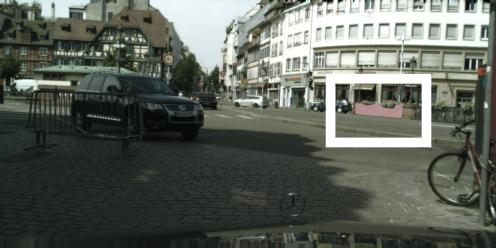}
            \end{subfigure}
            \begin{subfigure}[h!]{0.14\linewidth}
                \centering
                \includegraphics[width=\linewidth, height=0.7\linewidth]{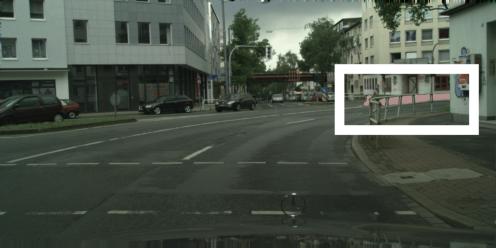}
            \end{subfigure}
            \begin{subfigure}[h!]{0.14\linewidth}
                \centering
                \includegraphics[width=\linewidth, height=0.7\linewidth]{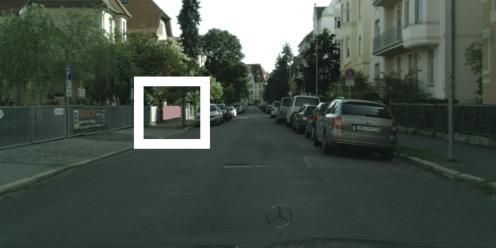}
            \end{subfigure}
        \end{minipage}

        \vspace{0.25em}

        \begin{minipage}{0.02\textwidth}
            \rotatebox{90}{\small \textbf{\textit{veget.}}}
        \end{minipage}\hspace{-0.5em}%
        \begin{minipage}{0.98\textwidth}
            \centering
            \begin{subfigure}[h!]{0.14\linewidth}
                \centering
                \includegraphics[width=\linewidth, height=0.7\linewidth]{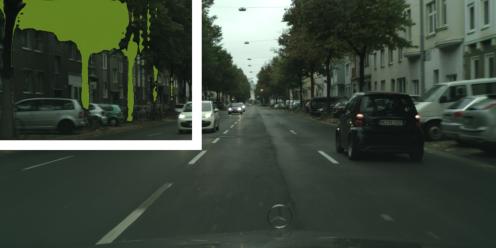}
            \end{subfigure}
            \begin{subfigure}[h!]{0.14\linewidth}
                \centering
                \includegraphics[width=\linewidth, height=0.7\linewidth]{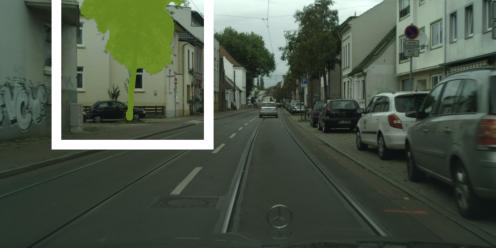}
            \end{subfigure}
            \begin{subfigure}[h!]{0.14\linewidth}
                \centering
                \includegraphics[width=\linewidth, height=0.7\linewidth]{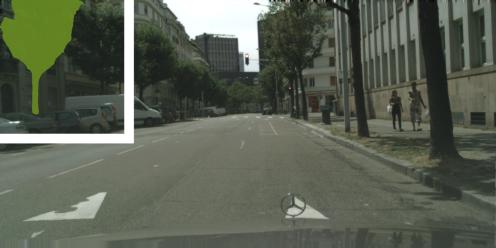}
            \end{subfigure}
            \begin{subfigure}[h!]{0.14\linewidth}
                \centering
                \includegraphics[width=\linewidth, height=0.7\linewidth]{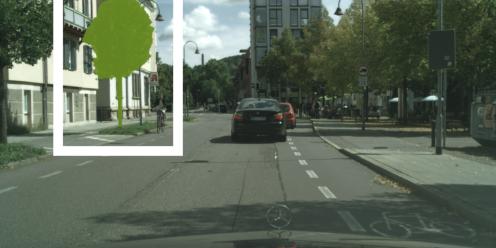}
            \end{subfigure}
            \begin{subfigure}[h!]{0.14\linewidth}
                \centering
                \includegraphics[width=\linewidth, height=0.7\linewidth]{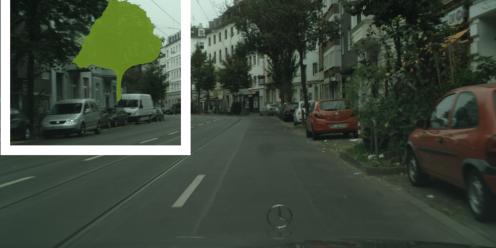}
            \end{subfigure}
            \begin{subfigure}[h!]{0.14\linewidth}
                \centering
                \includegraphics[width=\linewidth, height=0.7\linewidth]{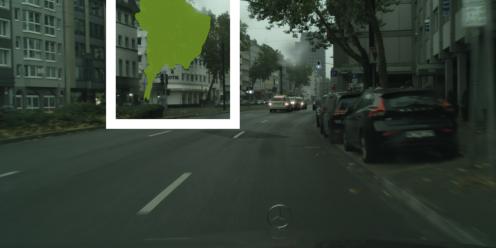}
            \end{subfigure}
        \end{minipage}

        \vspace{0.25em}        

        \begin{minipage}{0.02\textwidth}
            \rotatebox{90}{\small \textbf{\textit{person}}}
        \end{minipage}\hspace{-0.5em}%
        \begin{minipage}{0.98\textwidth}
            \centering
            \begin{subfigure}[h!]{0.14\linewidth}
                \centering
                \includegraphics[width=\linewidth, height=0.7\linewidth]{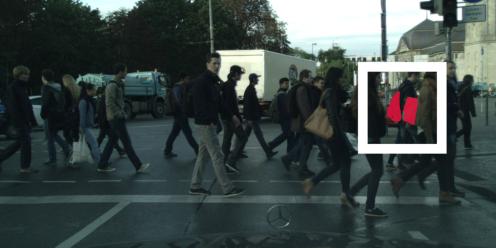}
            \end{subfigure}
            \begin{subfigure}[h!]{0.14\linewidth}
                \centering
                \includegraphics[width=\linewidth, height=0.7\linewidth]{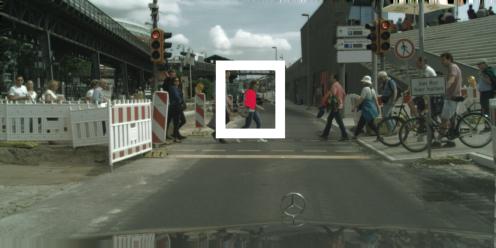}
            \end{subfigure}
            \begin{subfigure}[h!]{0.14\linewidth}
                \centering
                \includegraphics[width=\linewidth, height=0.7\linewidth]{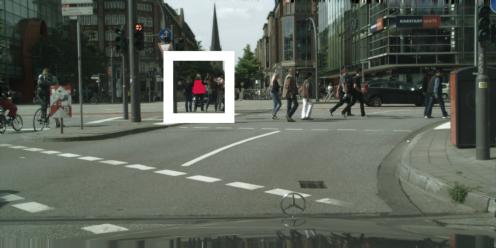}
            \end{subfigure}
            \begin{subfigure}[h!]{0.14\linewidth}
                \centering
                \includegraphics[width=\linewidth, height=0.7\linewidth]{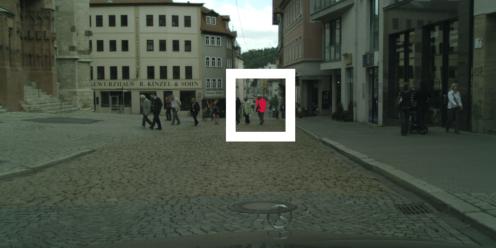}
            \end{subfigure}
            \begin{subfigure}[h!]{0.14\linewidth}
                \centering
                \includegraphics[width=\linewidth, height=0.7\linewidth]{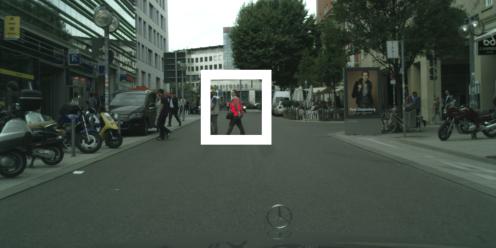}
            \end{subfigure}
            \begin{subfigure}[h!]{0.14\linewidth}
                \centering
                \includegraphics[width=\linewidth, height=0.7\linewidth]{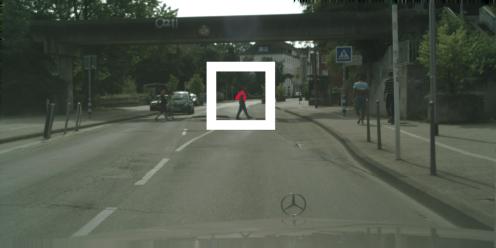}
            \end{subfigure}
        \end{minipage}

        \vspace{0.25em}

        \begin{minipage}{0.02\textwidth}
            \rotatebox{90}{\small \textbf{\textit{car}}}
        \end{minipage}\hspace{-0.5em}%
        \begin{minipage}{0.98\textwidth}
            \centering
            \begin{subfigure}[h!]{0.14\linewidth}
                \centering
                \includegraphics[width=\linewidth, height=0.7\linewidth]{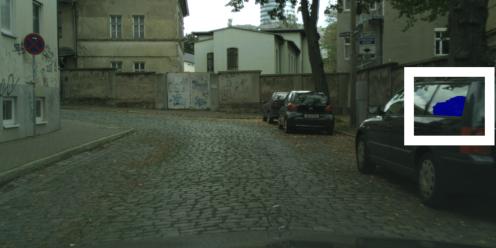}
            \end{subfigure}
            \begin{subfigure}[h!]{0.14\linewidth}
                \centering
                \includegraphics[width=\linewidth, height=0.7\linewidth]{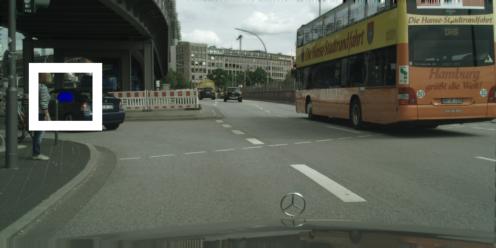}
            \end{subfigure}
            \begin{subfigure}[h!]{0.14\linewidth}
                \centering
                \includegraphics[width=\linewidth, height=0.7\linewidth]{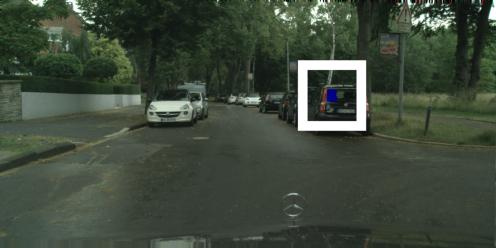}
            \end{subfigure}
            \begin{subfigure}[h!]{0.14\linewidth}
                \centering
                \includegraphics[width=\linewidth, height=0.7\linewidth]{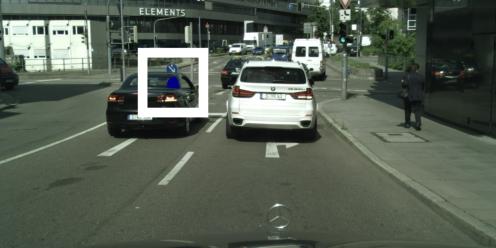}
            \end{subfigure}
            \begin{subfigure}[h!]{0.14\linewidth}
                \centering
                \includegraphics[width=\linewidth, height=0.7\linewidth]{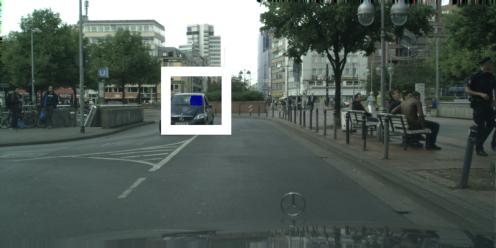}
            \end{subfigure}
            \begin{subfigure}[h!]{0.14\linewidth}
                \centering
                \includegraphics[width=\linewidth, height=0.7\linewidth]{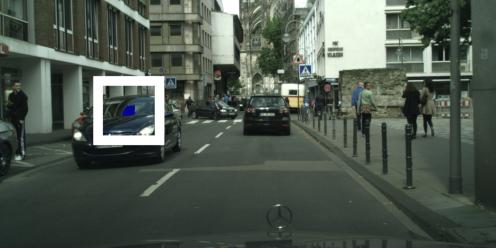}
            \end{subfigure}
        \end{minipage}

        \vspace{0.25em}        

        \begin{minipage}{0.02\textwidth}
            \rotatebox{90}{\small \textbf{\textit{bicycle}}}
        \end{minipage}\hspace{-0.5em}%
        \begin{minipage}{0.98\textwidth}
            \centering
            \begin{subfigure}[h!]{0.14\linewidth}
                \centering
                \includegraphics[width=\linewidth, height=0.7\linewidth]{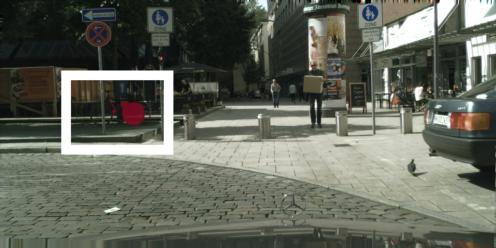}
            \end{subfigure}
            \begin{subfigure}[h!]{0.14\linewidth}
                \centering
                \includegraphics[width=\linewidth, height=0.7\linewidth]{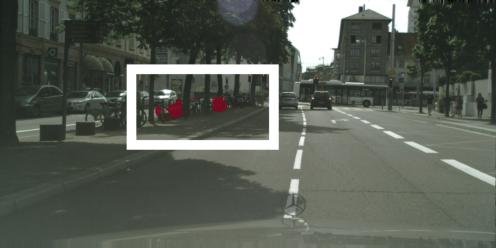}
            \end{subfigure}
            \begin{subfigure}[h!]{0.14\linewidth}
                \centering
                \includegraphics[width=\linewidth, height=0.7\linewidth]{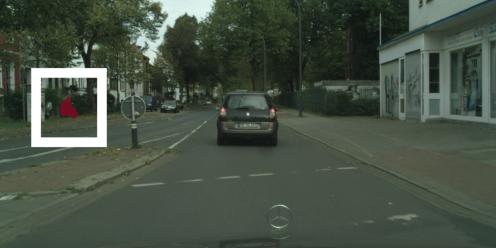}
            \end{subfigure}
            \begin{subfigure}[h!]{0.14\linewidth}
                \centering
                \includegraphics[width=\linewidth, height=0.7\linewidth]{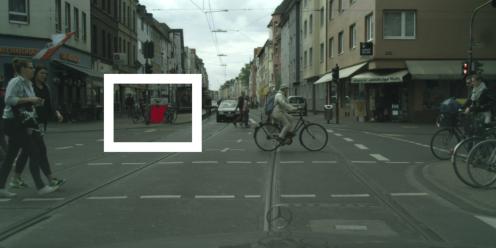}
            \end{subfigure}
            \begin{subfigure}[h!]{0.14\linewidth}
                \centering
                \includegraphics[width=\linewidth, height=0.7\linewidth]{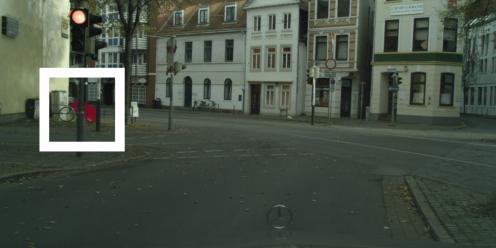}
            \end{subfigure}
            \begin{subfigure}[h!]{0.14\linewidth}
                \centering
                \includegraphics[width=\linewidth, height=0.7\linewidth]{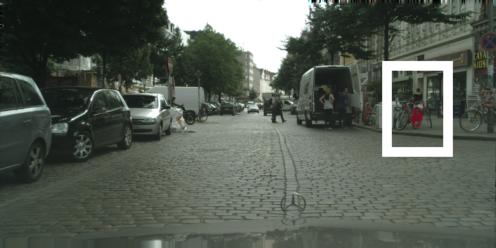}
            \end{subfigure}
        \end{minipage}
        
    \end{minipage}
    
    \vspace{0.8em}
    {\centering \textbf{(a) Cityscapes} \par}
    \vspace{1.2em}
    
    \begin{minipage}{\textwidth}
        \begin{minipage}{0.02\textwidth}
            \rotatebox{90}{\small \textbf{\textit{bus}}}
        \end{minipage}\hspace{-0.5em}%
        \begin{minipage}{0.98\textwidth}
            \centering
            \begin{subfigure}[h!]{0.14\linewidth}
                \centering
                \includegraphics[width=\linewidth, height=0.7\linewidth]{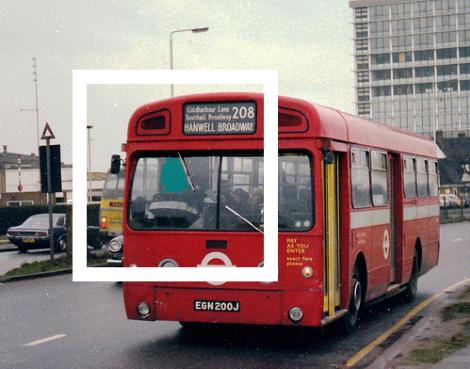}
            \end{subfigure}
            \begin{subfigure}[h!]{0.14\linewidth}
                \centering
                \includegraphics[width=\linewidth, height=0.7\linewidth]{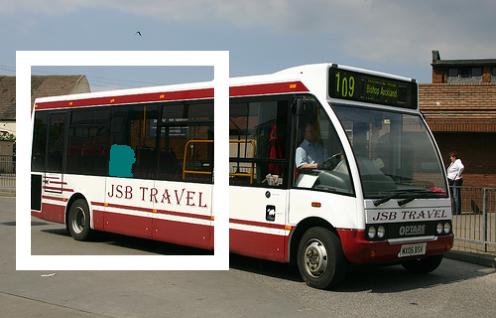}
            \end{subfigure}
            \begin{subfigure}[h!]{0.14\linewidth}
                \centering
                \includegraphics[width=\linewidth, height=0.7\linewidth]{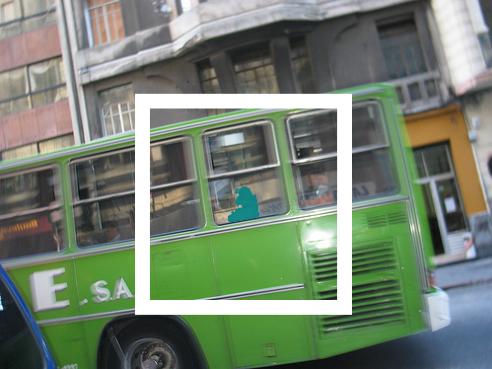}
            \end{subfigure}
            \begin{subfigure}[h!]{0.14\linewidth}
                \centering
                \includegraphics[width=\linewidth, height=0.7\linewidth]{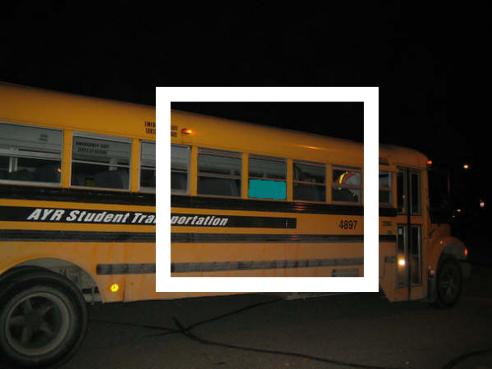}
            \end{subfigure}            
            \begin{subfigure}[h!]{0.14\linewidth}
                \centering
                \includegraphics[width=\linewidth, height=0.7\linewidth]{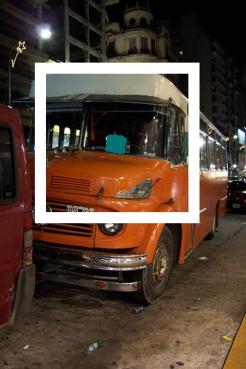}
            \end{subfigure}
            \begin{subfigure}[h!]{0.14\linewidth}
                \centering
                \includegraphics[width=\linewidth, height=0.7\linewidth]{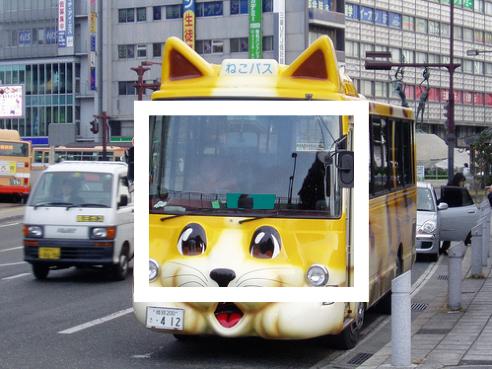}
            \end{subfigure}
        \end{minipage}

        \vspace{0.25em}

        \begin{minipage}{0.02\textwidth}
            \rotatebox{90}{\small \textbf{\textit{car}}}
        \end{minipage}\hspace{-0.5em}%
        \begin{minipage}{0.98\textwidth}
            \centering
            \begin{subfigure}[h!]{0.14\linewidth}
                \centering
                \includegraphics[width=\linewidth, height=0.7\linewidth]{figures/lcm_corr_masks_boxadded_20250514/pascal/car7/2247___2008_000028_obj1_before20_after7_gt7.jpg}
            \end{subfigure}
            \begin{subfigure}[h!]{0.14\linewidth}
                \centering
                \includegraphics[width=\linewidth, height=0.7\linewidth]{figures/lcm_corr_masks_boxadded_20250514/pascal/car7/1563___2007_006900_obj10_before9_after7_gt7.jpg}
            \end{subfigure}
            \begin{subfigure}[h!]{0.14\linewidth}
                \centering
                \includegraphics[width=\linewidth, height=0.7\linewidth]{figures/lcm_corr_masks_boxadded_20250514/pascal/car7/1439___2007_009322_obj13_before20_after7_gt7.jpg}
            \end{subfigure}
            \begin{subfigure}[h!]{0.14\linewidth}
                \centering
                \includegraphics[width=\linewidth, height=0.7\linewidth]{figures/lcm_corr_masks_boxadded_20250514/pascal/car7/1100___2007_009052_obj3_before9_after7_gt7.jpg}
            \end{subfigure}
            \begin{subfigure}[h!]{0.14\linewidth}
                \centering
                \includegraphics[width=\linewidth, height=0.7\linewidth]{figures/lcm_corr_masks_boxadded_20250514/pascal/car7/1033___2010_002363_obj67_before9_after7_gt7.jpg}
            \end{subfigure}
            \begin{subfigure}[h!]{0.14\linewidth}
                \centering
                \includegraphics[width=\linewidth, height=0.7\linewidth]{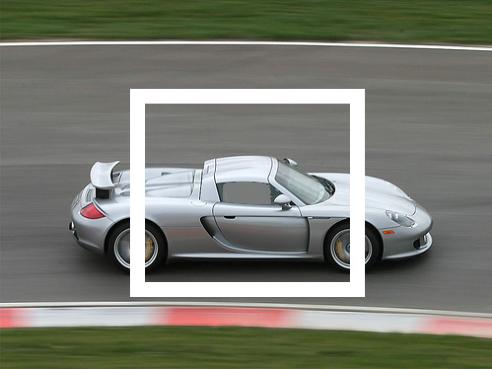}
            \end{subfigure}
        \end{minipage}

        \vspace{0.25em}

        \begin{minipage}{0.02\textwidth}
            \rotatebox{90}{\small \textbf{\textit{table.}}}
        \end{minipage}\hspace{-0.5em}%
        \begin{minipage}{0.98\textwidth}
            \centering
            \begin{subfigure}[h!]{0.14\linewidth}
                \centering
                \includegraphics[width=\linewidth, height=0.7\linewidth]{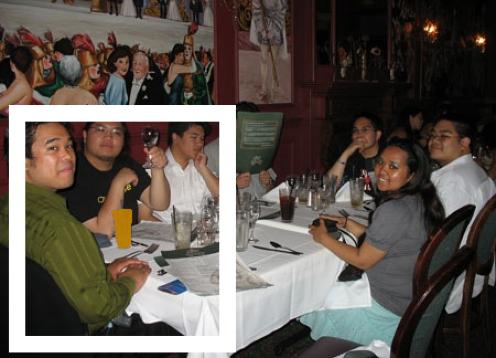}
            \end{subfigure}
            \begin{subfigure}[h!]{0.14\linewidth}
                \centering
                \includegraphics[width=\linewidth, height=0.7\linewidth]{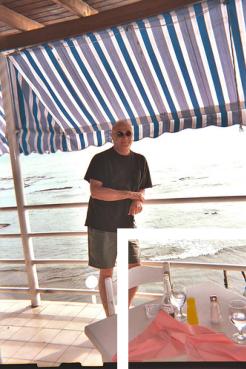}
            \end{subfigure}
            \begin{subfigure}[h!]{0.14\linewidth}
                \centering
                \includegraphics[width=\linewidth, height=0.7\linewidth]{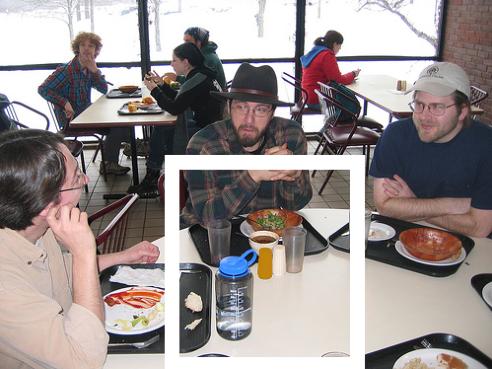}
            \end{subfigure}
            \begin{subfigure}[h!]{0.14\linewidth}
                \centering
                \includegraphics[width=\linewidth, height=0.7\linewidth]{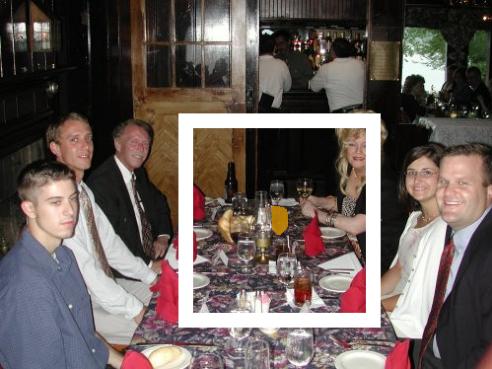}
            \end{subfigure}
            \begin{subfigure}[h!]{0.14\linewidth}
                \centering
                \includegraphics[width=\linewidth, height=0.7\linewidth]{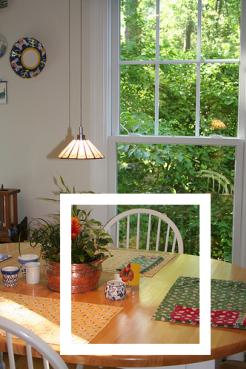}
            \end{subfigure}
            \begin{subfigure}[h!]{0.14\linewidth}
                \centering
                \includegraphics[width=\linewidth, height=0.7\linewidth]{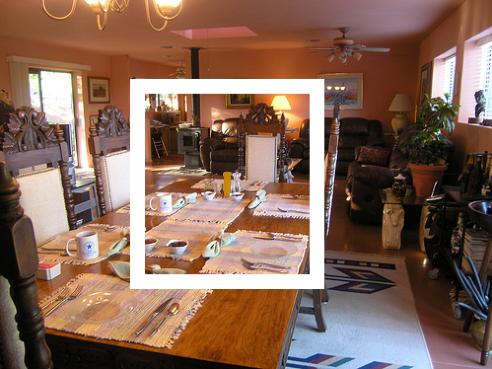}
            \end{subfigure}
        \end{minipage}

        \vspace{0.25em}

        \begin{minipage}{0.02\textwidth}
            \rotatebox{90}{\small \textbf{\textit{motor.}}}
        \end{minipage}\hspace{-0.5em}%
        \begin{minipage}{0.98\textwidth}
            \centering
            \begin{subfigure}[h!]{0.14\linewidth}
                \centering
                \includegraphics[width=\linewidth, height=0.7\linewidth]{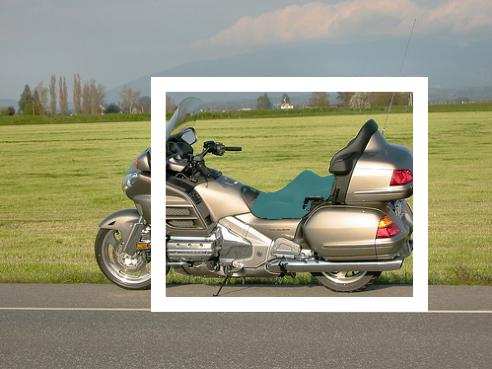}
            \end{subfigure}
            \begin{subfigure}[h!]{0.14\linewidth}
                \centering
                \includegraphics[width=\linewidth, height=0.7\linewidth]{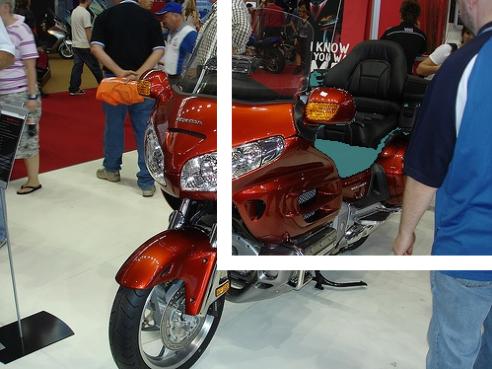}
            \end{subfigure}
            \begin{subfigure}[h!]{0.14\linewidth}
                \centering
                \includegraphics[width=\linewidth, height=0.7\linewidth]{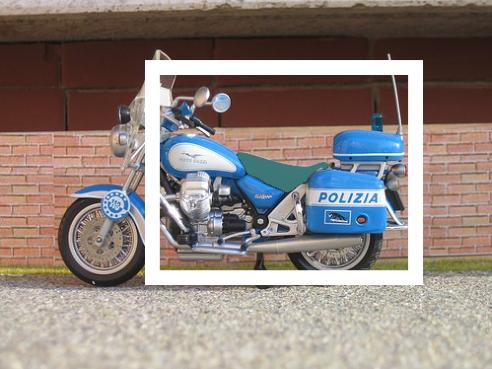}
            \end{subfigure}
            \begin{subfigure}[h!]{0.14\linewidth}
                \centering
                \includegraphics[width=\linewidth, height=0.7\linewidth]{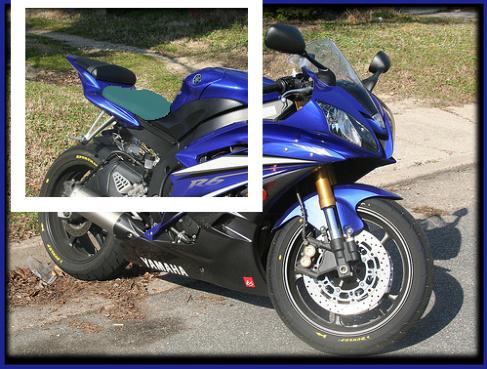}
            \end{subfigure}
            \begin{subfigure}[h!]{0.14\linewidth}
                \centering
                \includegraphics[width=\linewidth, height=0.7\linewidth]{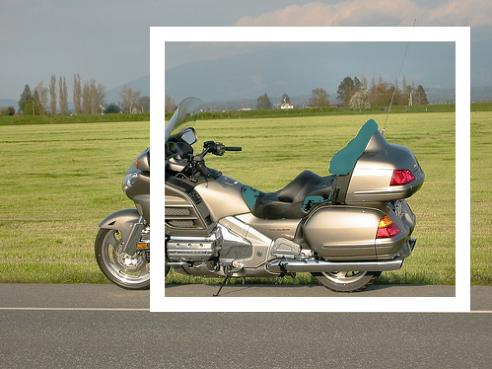}
            \end{subfigure}
            \begin{subfigure}[h!]{0.14\linewidth}
                \centering
                \includegraphics[width=\linewidth, height=0.7\linewidth]{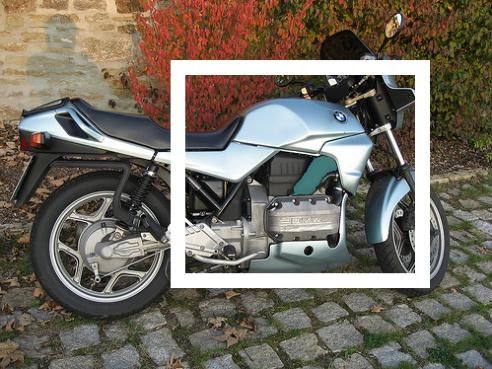}
            \end{subfigure}
        \end{minipage}

        \vspace{0.25em}

        \begin{minipage}{0.02\textwidth}
            \rotatebox{90}{\small \textbf{\textit{person}}}
        \end{minipage}\hspace{-0.5em}%
        \begin{minipage}{0.98\textwidth}
            \centering
            \begin{subfigure}[h!]{0.14\linewidth}
                \centering
                \includegraphics[width=\linewidth, height=0.7\linewidth]{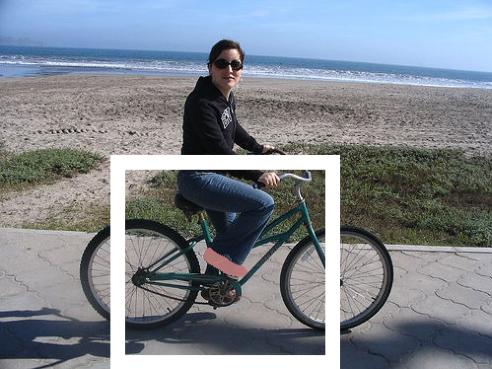}
            \end{subfigure}
            \begin{subfigure}[h!]{0.14\linewidth}
                \centering
                \includegraphics[width=\linewidth, height=0.7\linewidth]{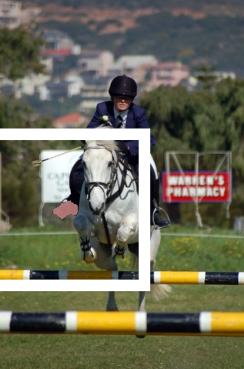}
            \end{subfigure}
            \begin{subfigure}[h!]{0.14\linewidth}
                \centering
                \includegraphics[width=\linewidth, height=0.7\linewidth]{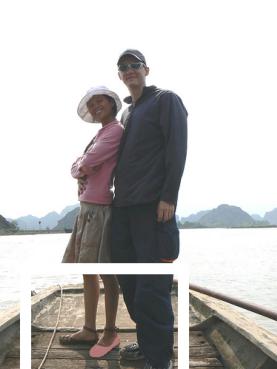}
            \end{subfigure}
            \begin{subfigure}[h!]{0.14\linewidth}
                \centering
                \includegraphics[width=\linewidth, height=0.7\linewidth]{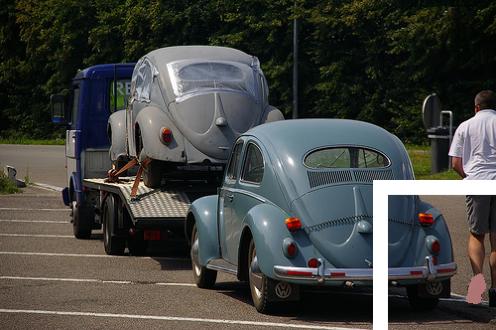}
            \end{subfigure}
            \begin{subfigure}[h!]{0.14\linewidth}
                \centering
                \includegraphics[width=\linewidth, height=0.7\linewidth]{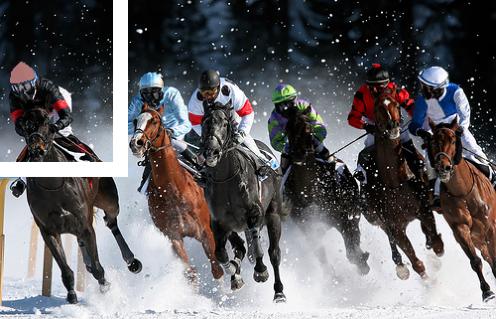}
            \end{subfigure}
            \begin{subfigure}[h!]{0.14\linewidth}
                \centering
                \includegraphics[width=\linewidth, height=0.7\linewidth]{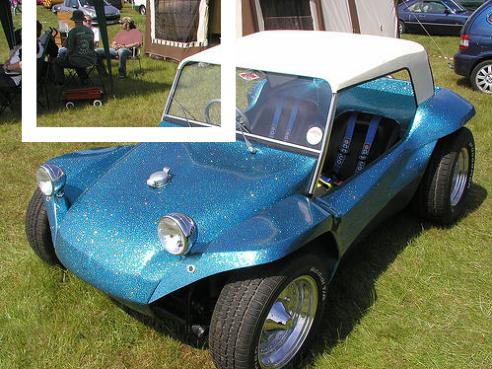}
            \end{subfigure}
        \end{minipage}

        \vspace{0.25em}
        
        \begin{minipage}{0.02\textwidth}
            \rotatebox{90}{\small \textbf{\textit{train}}}
        \end{minipage}\hspace{-0.5em}%
        \begin{minipage}{0.98\textwidth}
            \centering
            \begin{subfigure}[h!]{0.14\linewidth}
                \centering
                \includegraphics[width=\linewidth, height=0.7\linewidth]{figures/lcm_corr_masks_boxadded_20250514/pascal/train19/9080___2011_000221_obj7_before20_after19_gt19.jpg}
            \end{subfigure}
            \begin{subfigure}[h!]{0.14\linewidth}
                \centering
                \includegraphics[width=\linewidth, height=0.7\linewidth]{figures/lcm_corr_masks_boxadded_20250514/pascal/train19/4976___2007_007947_obj3_before20_after19_gt19.jpg}
            \end{subfigure}
            \begin{subfigure}[h!]{0.14\linewidth}
                \centering
                \includegraphics[width=\linewidth, height=0.7\linewidth]{figures/lcm_corr_masks_boxadded_20250514/pascal/train19/4874___2007_002462_obj3_before1_after19_gt19.jpg}
            \end{subfigure}
            \begin{subfigure}[h!]{0.14\linewidth}
                \centering
                \includegraphics[width=\linewidth, height=0.7\linewidth]{figures/lcm_corr_masks_boxadded_20250514/pascal/train19/4015___2009_005084_obj5_before20_after19_gt19.jpg}
            \end{subfigure}
            \begin{subfigure}[h!]{0.14\linewidth}
                \centering
                \includegraphics[width=\linewidth, height=0.7\linewidth]{figures/lcm_corr_masks_boxadded_20250514/pascal/train19/3309___2008_007011_obj4_before20_after19_gt19.jpg}
            \end{subfigure}
            \begin{subfigure}[h!]{0.14\linewidth}
                \centering
                \includegraphics[width=\linewidth, height=0.7\linewidth]{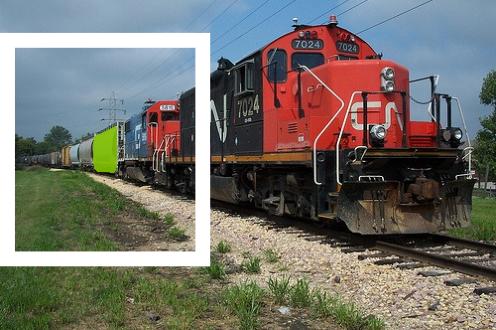}
            \end{subfigure}
        \end{minipage}
    \end{minipage}
    
    \vspace{0.8em}
    {\centering \textbf{(b) PASCAL VOC 2012} \par}
\end{minipage}
\caption{Visualization of masks corrected by LCM. We visualize the masks by applying class colors and overlaying white bounding boxes around them.}
\label{fig:visualize_LCM_masks}
\end{figure*}

Figure~\ref{fig:visualize_LCM_masks} presents qualitative results of masks corrected by LCM on Cityscapes and PASCAL.

\subsection{Constructed Pseudo-Labels}
\begin{figure*}[h]
\centering    
\begin{minipage}{\textwidth}
    \centering   
    \begin{subfigure}[h!]{0.195\linewidth}
        \centering
        \includegraphics[width=\linewidth, height=0.65\linewidth]{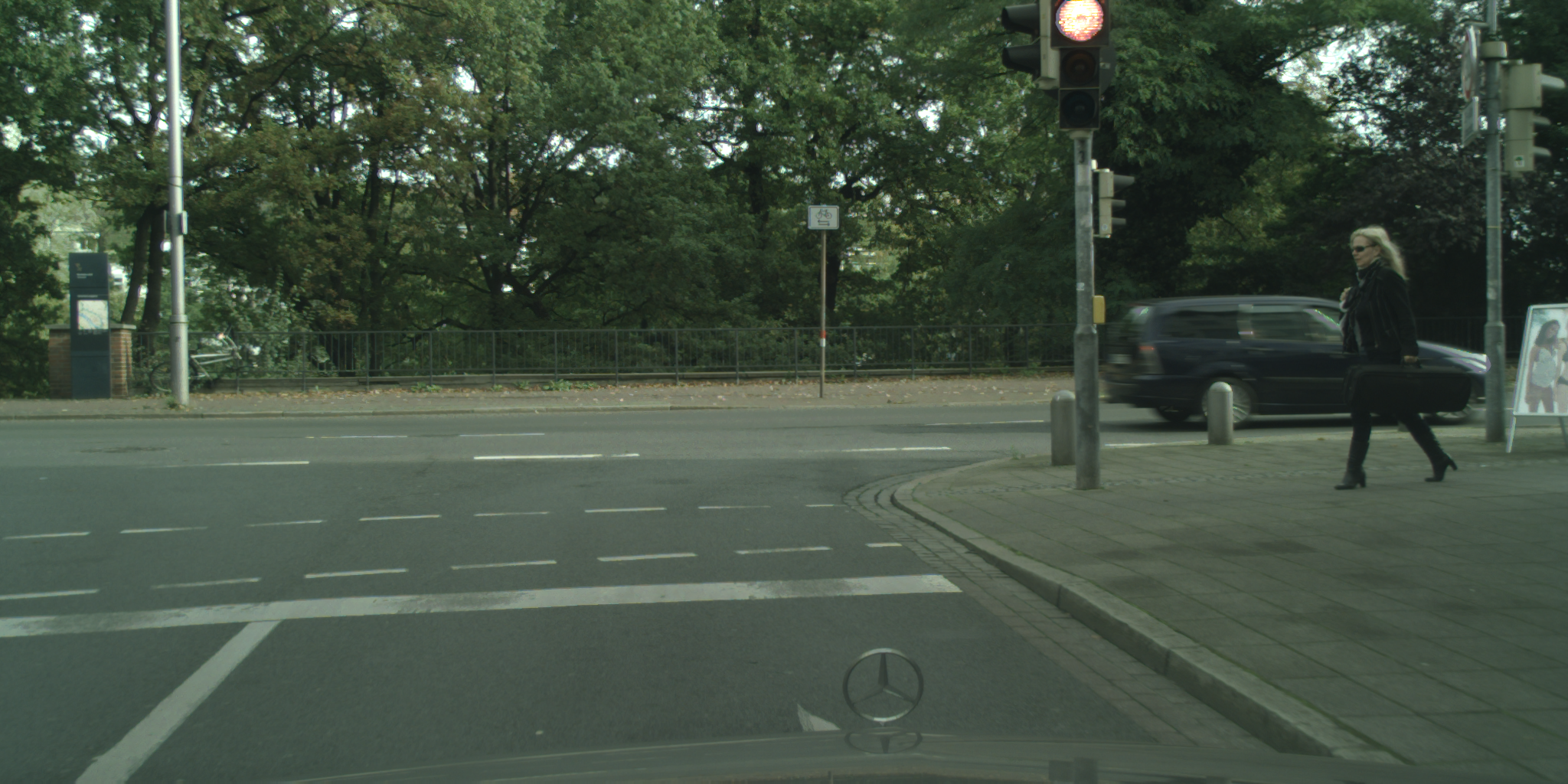}
    \end{subfigure}
    \begin{subfigure}[h!]{0.195\linewidth}
        \centering
        \includegraphics[width=\linewidth, height=0.65\linewidth]{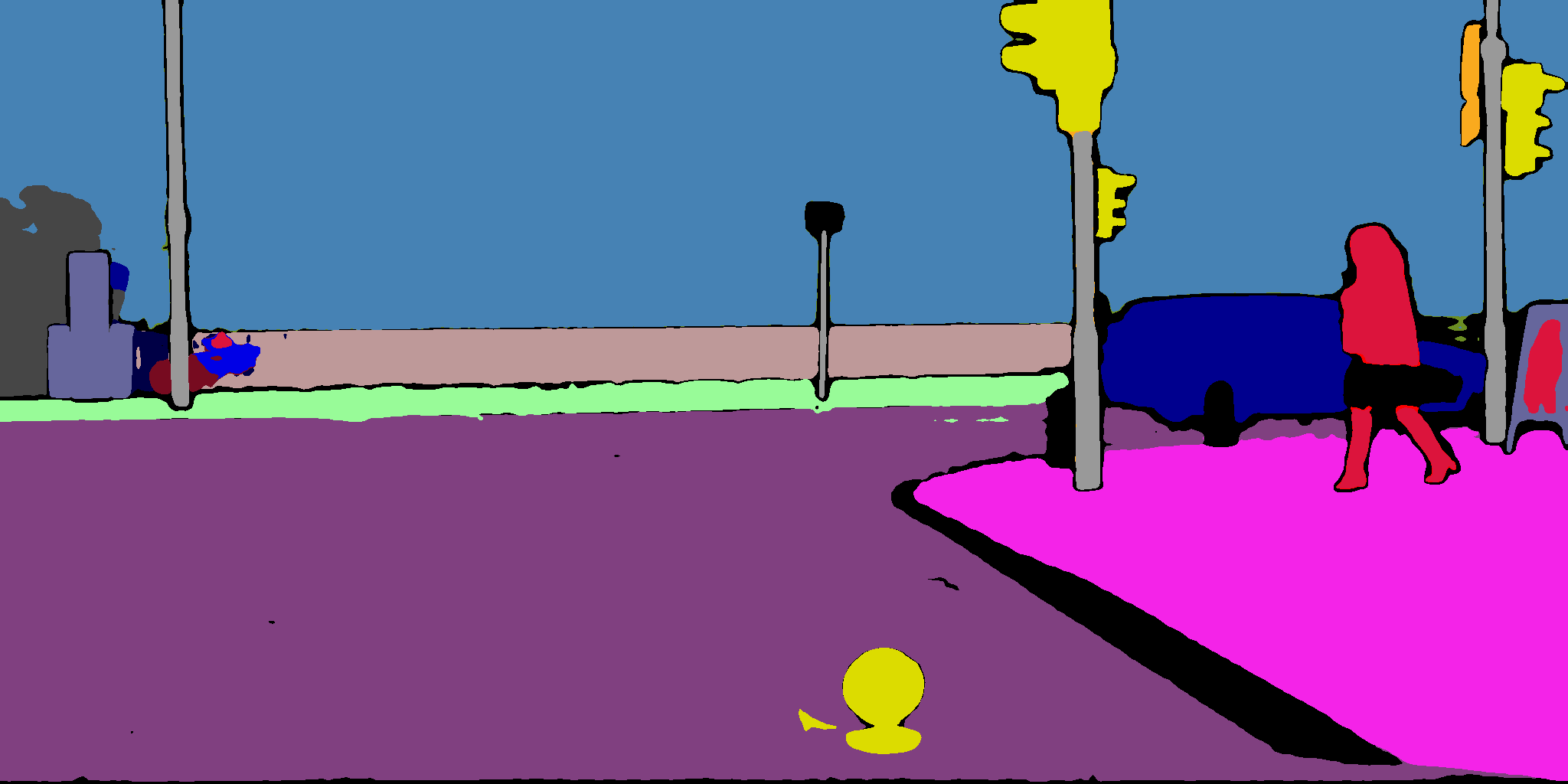}
    \end{subfigure}
    \begin{subfigure}[h!]{0.195\linewidth}
        \centering
        \includegraphics[width=\linewidth, height=0.65\linewidth]{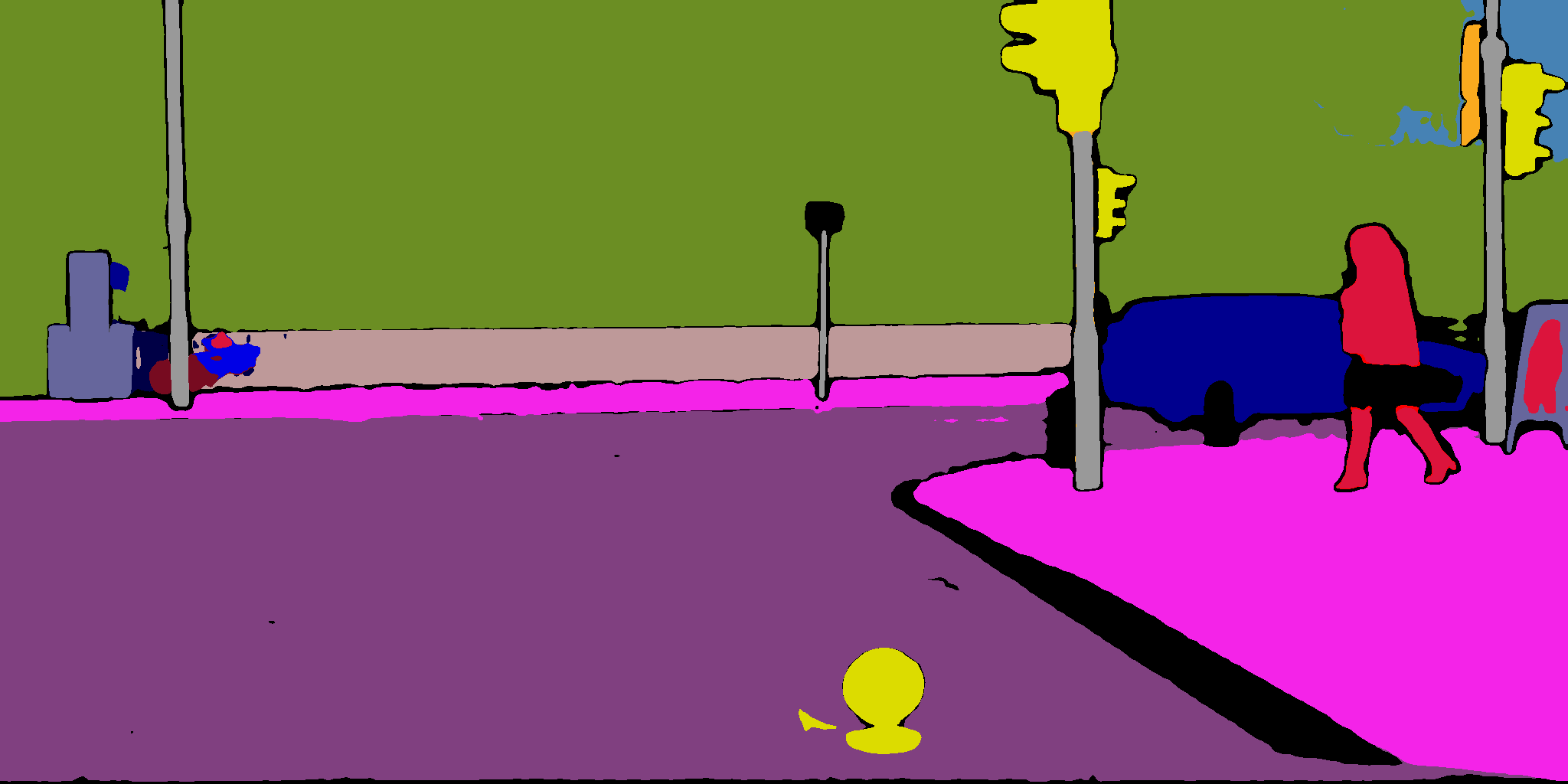}
    \end{subfigure}
    \begin{subfigure}[h!]{0.195\linewidth}
        \centering
        \includegraphics[width=\linewidth, height=0.65\linewidth]{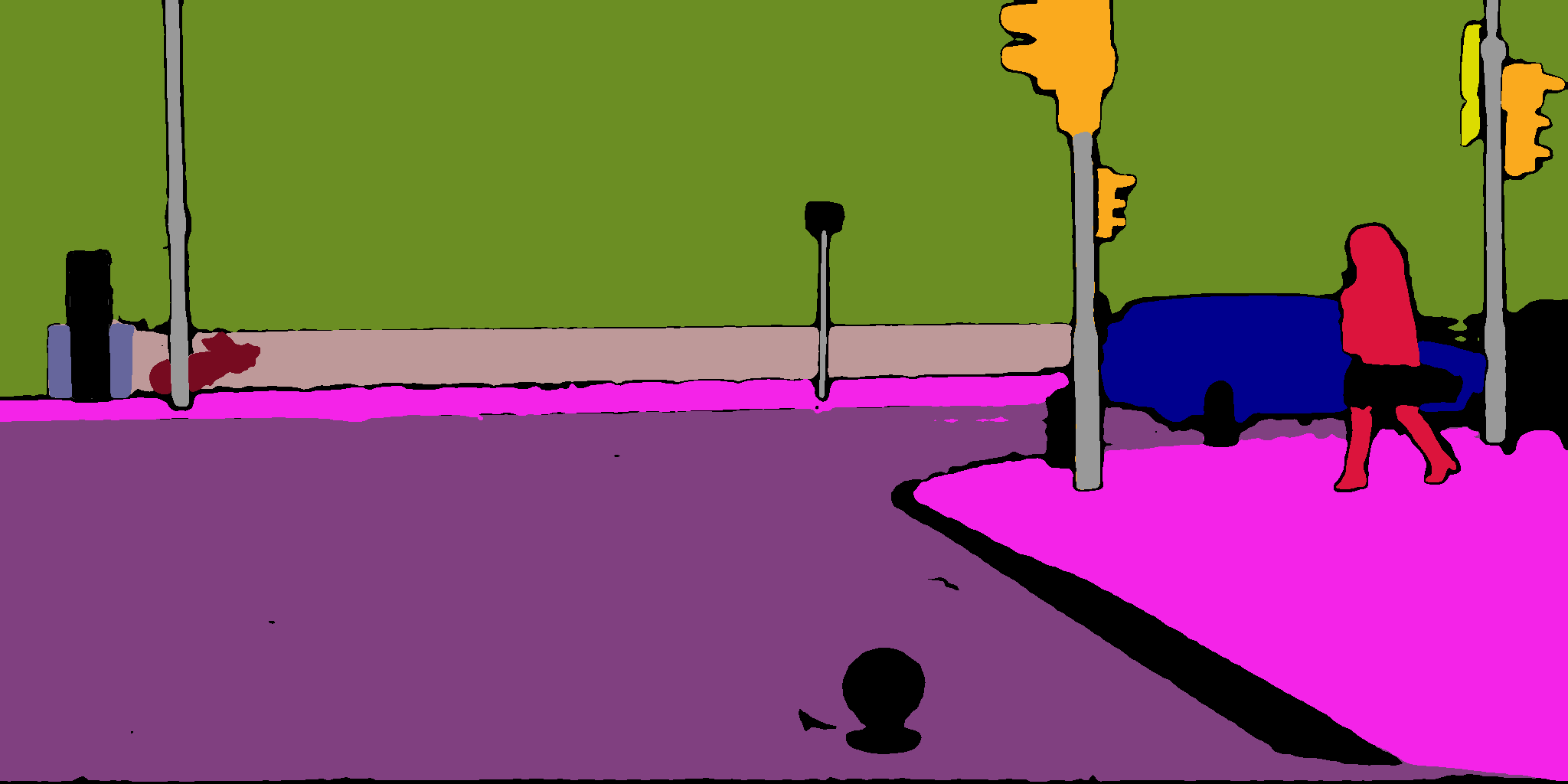}
    \end{subfigure}
    \begin{subfigure}[h!]{0.195\linewidth}
        \centering
        \includegraphics[width=\linewidth, height=0.65\linewidth]{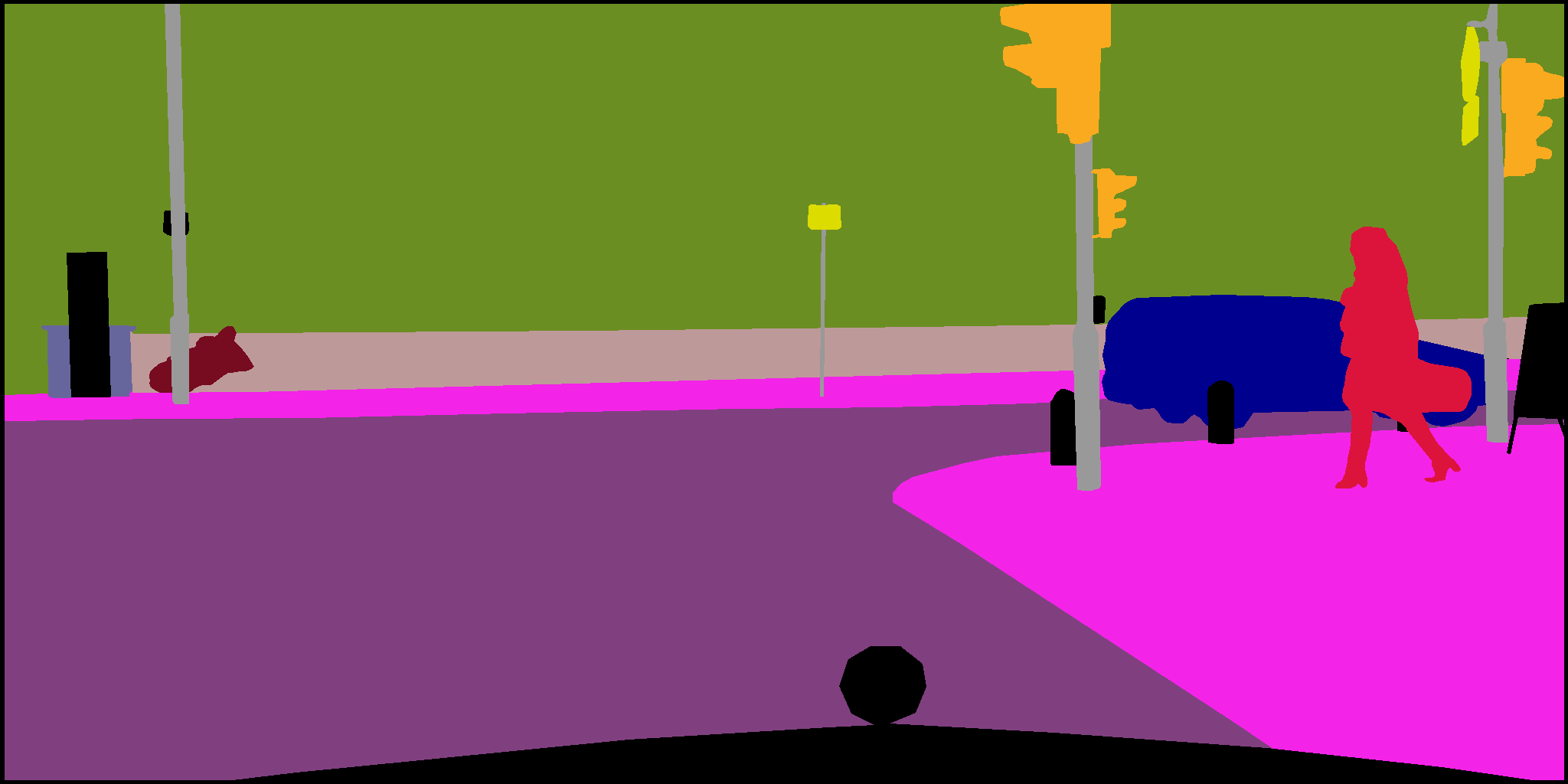}
    \end{subfigure}
    
    \begin{subfigure}[h!]{0.195\linewidth}
        \centering
        \includegraphics[width=\linewidth, height=0.65\linewidth]{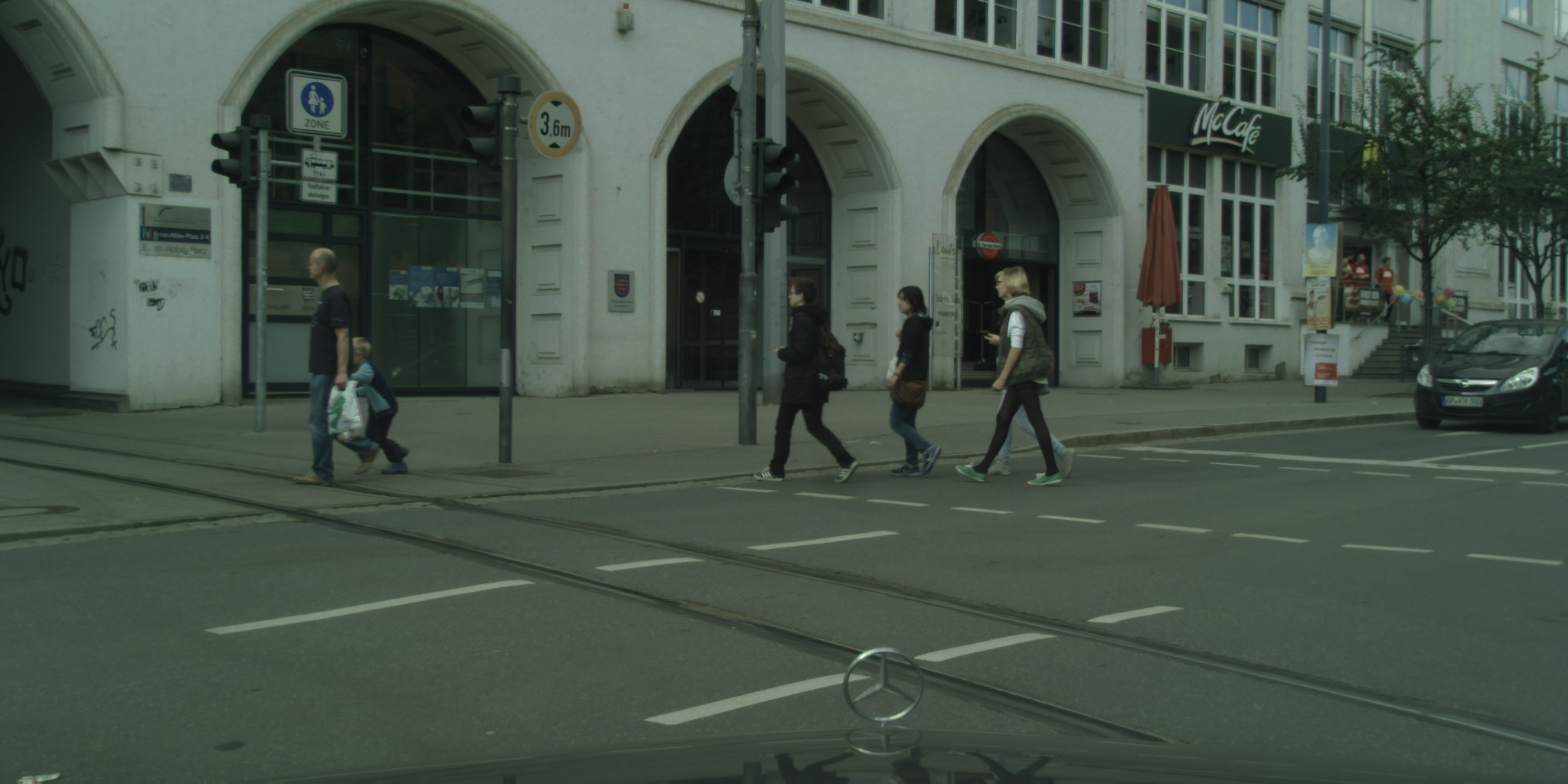}
    \end{subfigure}
    \begin{subfigure}[h!]{0.195\linewidth}
        \centering
        \includegraphics[width=\linewidth, height=0.65\linewidth]{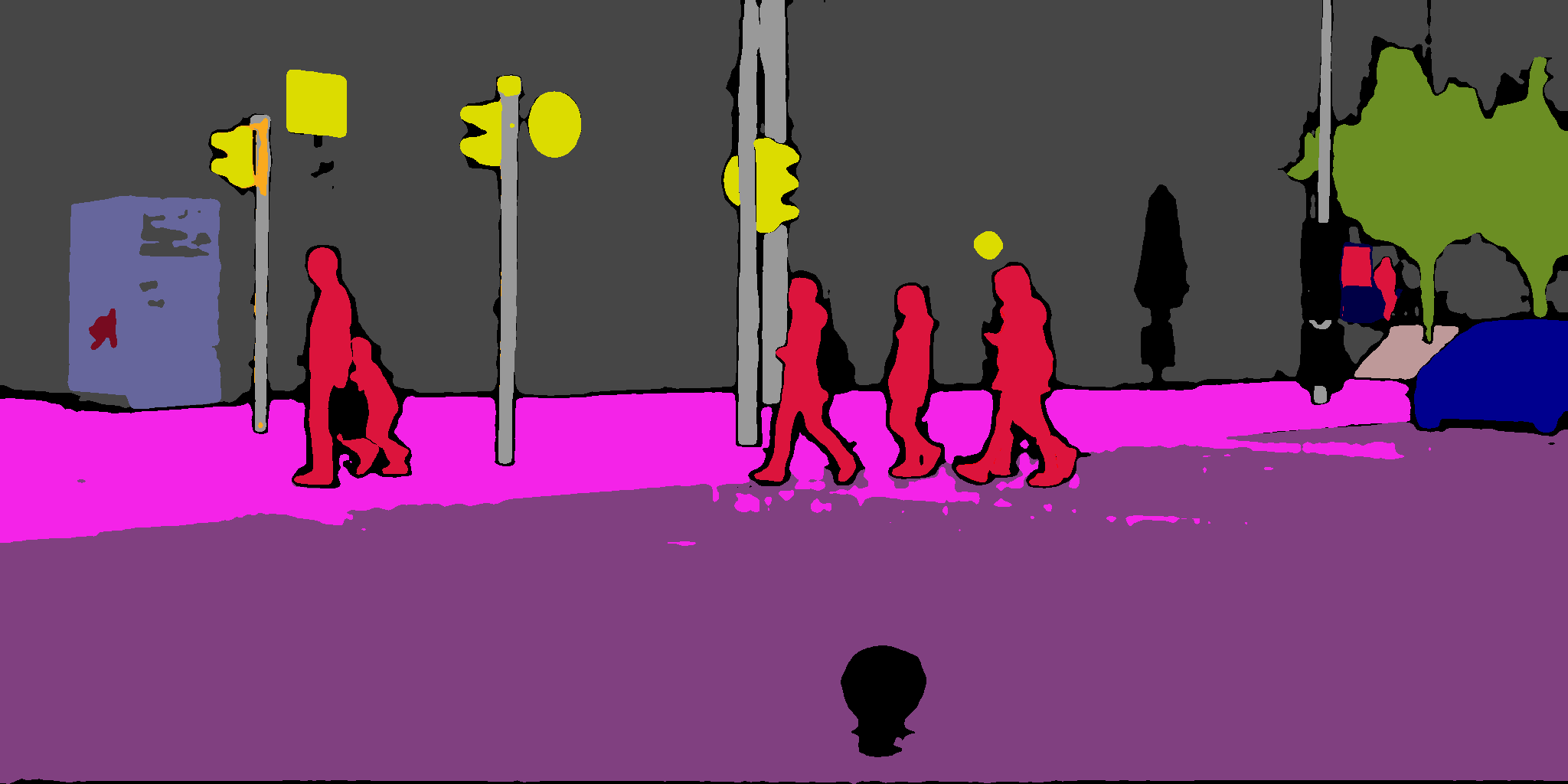}
    \end{subfigure}
    \begin{subfigure}[h!]{0.195\linewidth}
        \centering
        \includegraphics[width=\linewidth, height=0.65\linewidth]{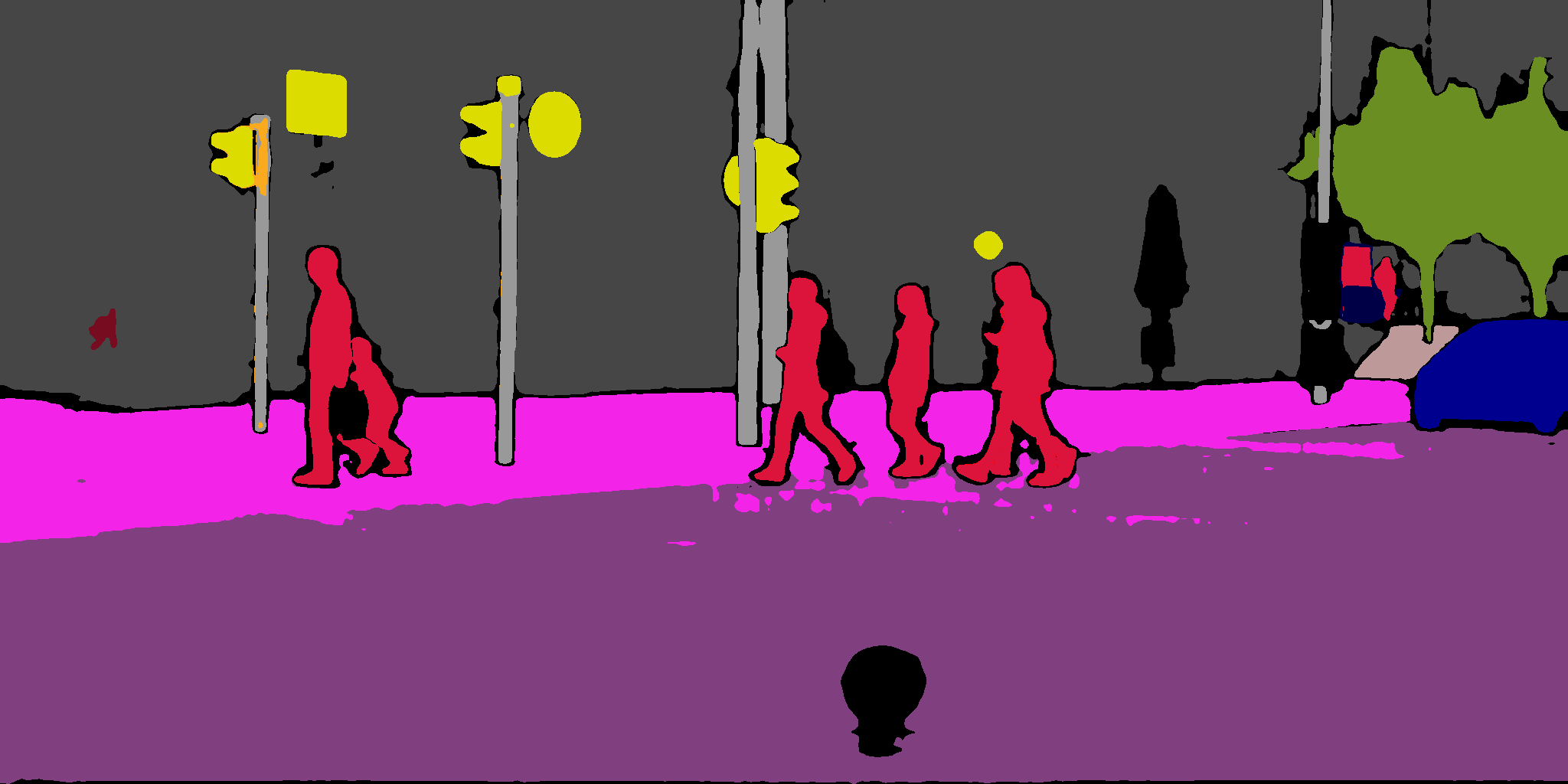}
    \end{subfigure}
    \begin{subfigure}[h!]{0.195\linewidth}
        \centering
        \includegraphics[width=\linewidth, height=0.65\linewidth]{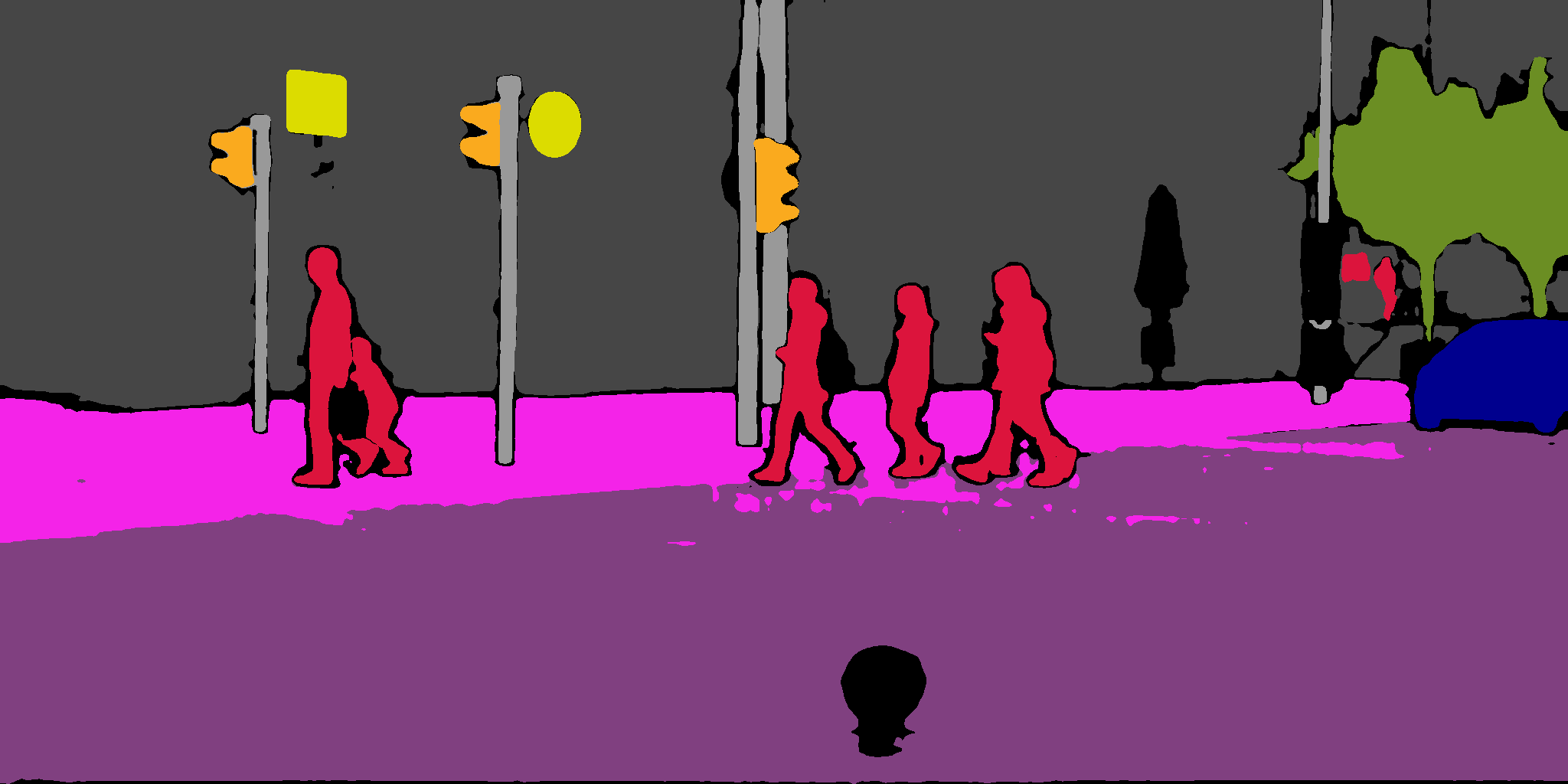}
    \end{subfigure}
    \begin{subfigure}[h!]{0.195\linewidth}
        \centering
        \includegraphics[width=\linewidth, height=0.65\linewidth]{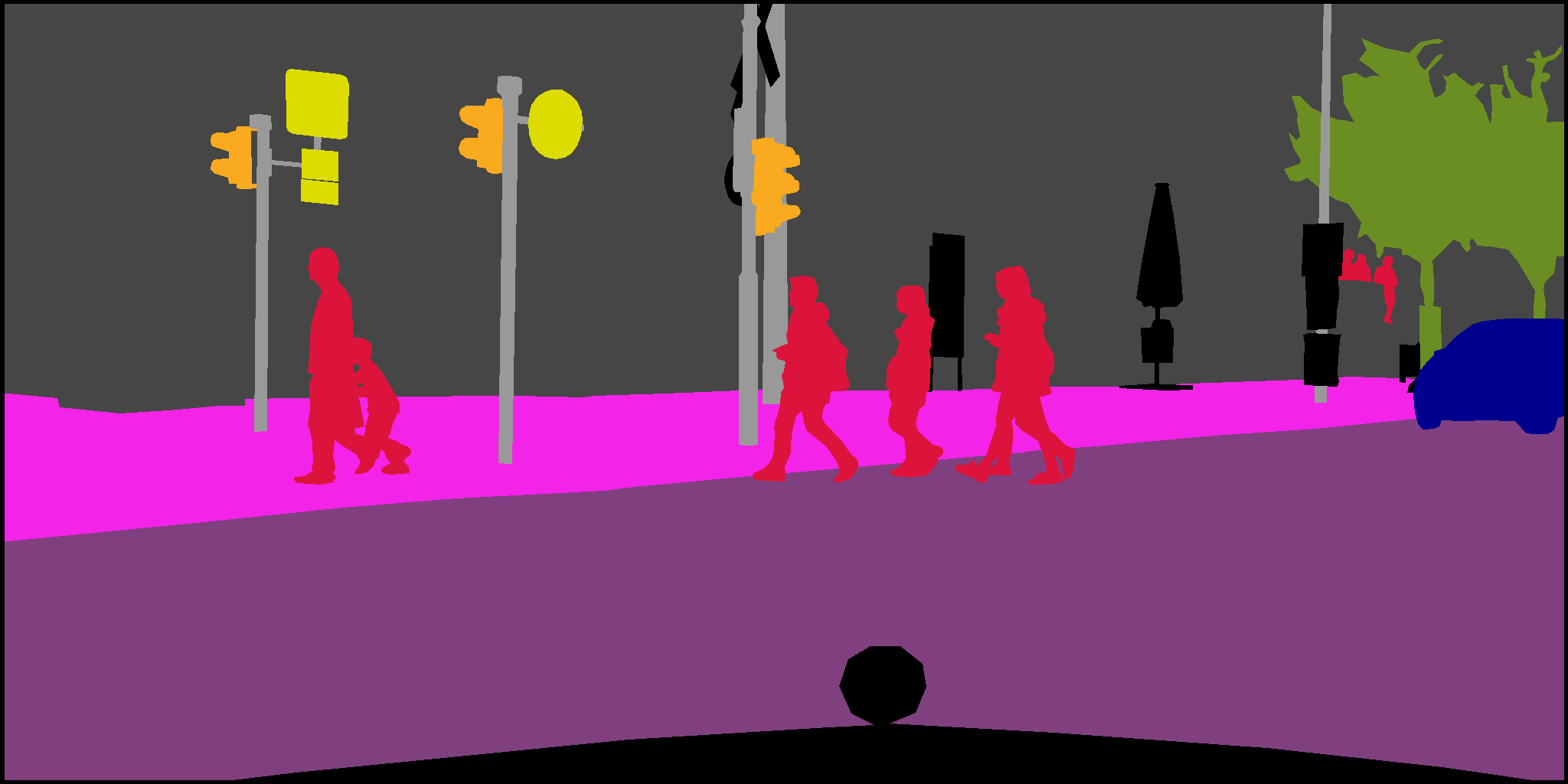}
    \end{subfigure}
    
    \begin{subfigure}[h!]{0.195\linewidth}
        \centering
        \includegraphics[width=\linewidth, height=0.65\linewidth]{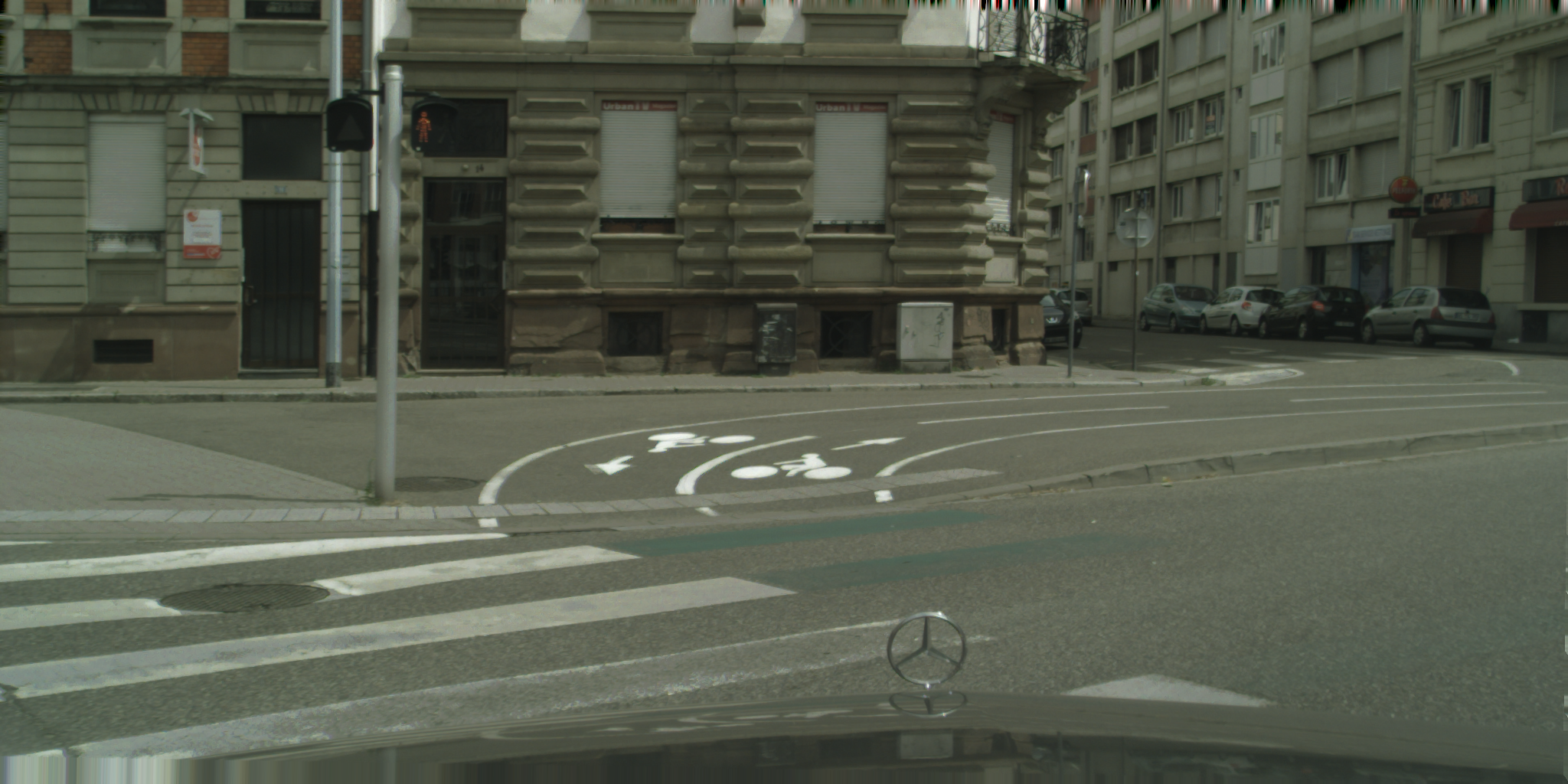}
    \end{subfigure}
    \begin{subfigure}[h!]{0.195\linewidth}
        \centering
        \includegraphics[width=\linewidth, height=0.65\linewidth]{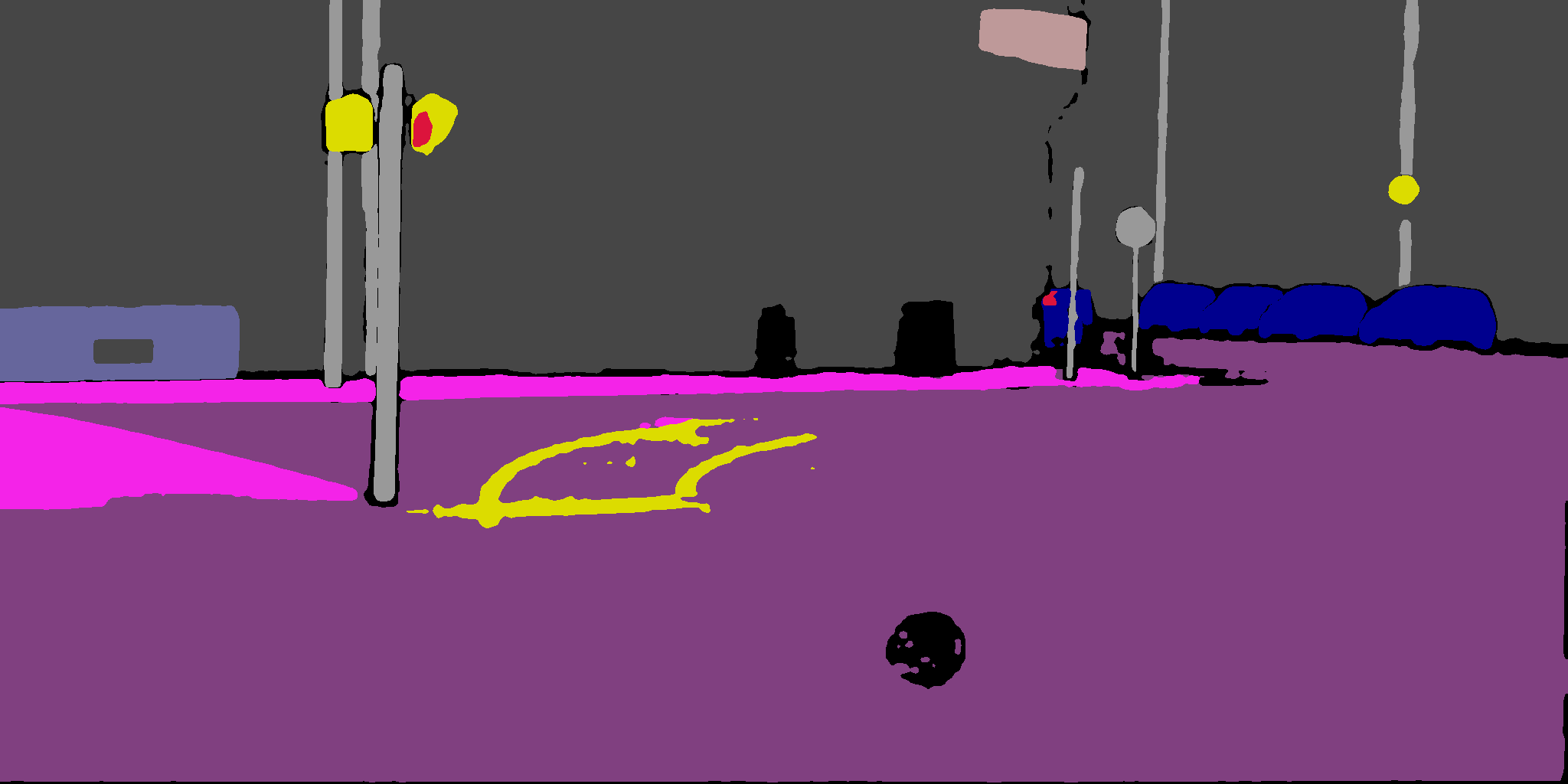}
    \end{subfigure}
    \begin{subfigure}[h!]{0.195\linewidth}
        \centering
        \includegraphics[width=\linewidth, height=0.65\linewidth]{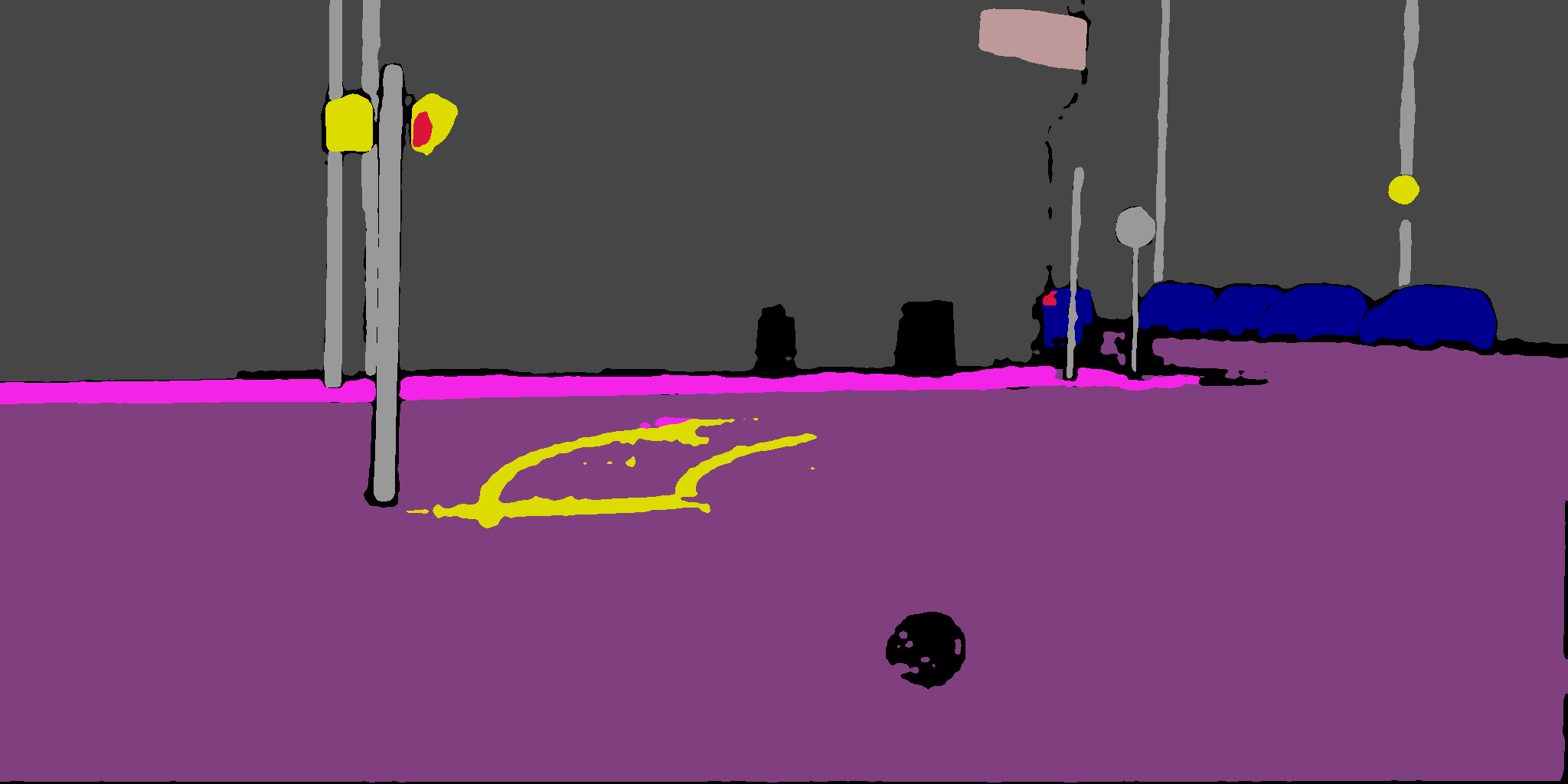}
    \end{subfigure}
    \begin{subfigure}[h!]{0.195\linewidth}
        \centering
        \includegraphics[width=\linewidth, height=0.65\linewidth]{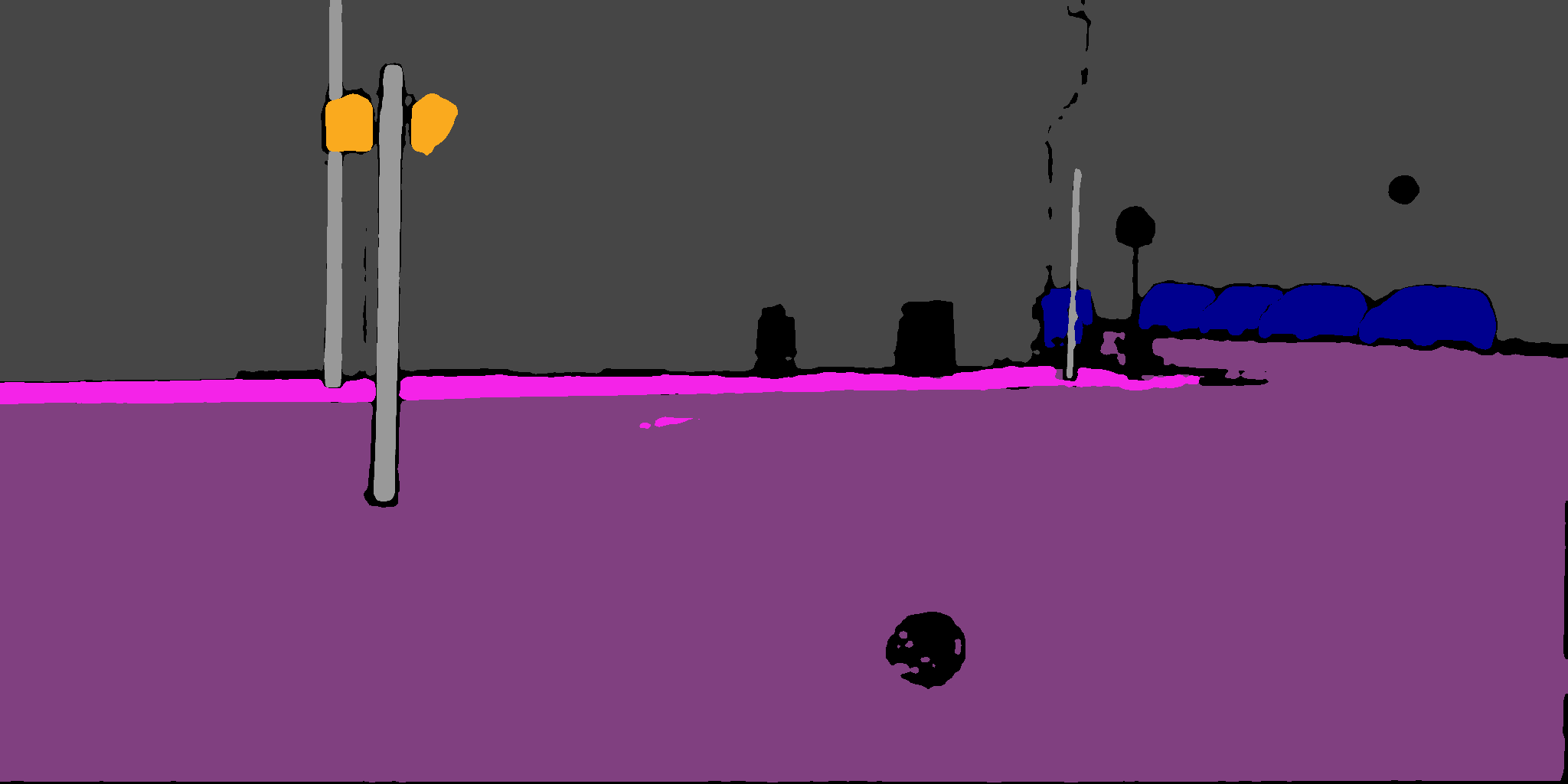}
    \end{subfigure}
    \begin{subfigure}[h!]{0.195\linewidth}
        \centering
        \includegraphics[width=\linewidth, height=0.65\linewidth]{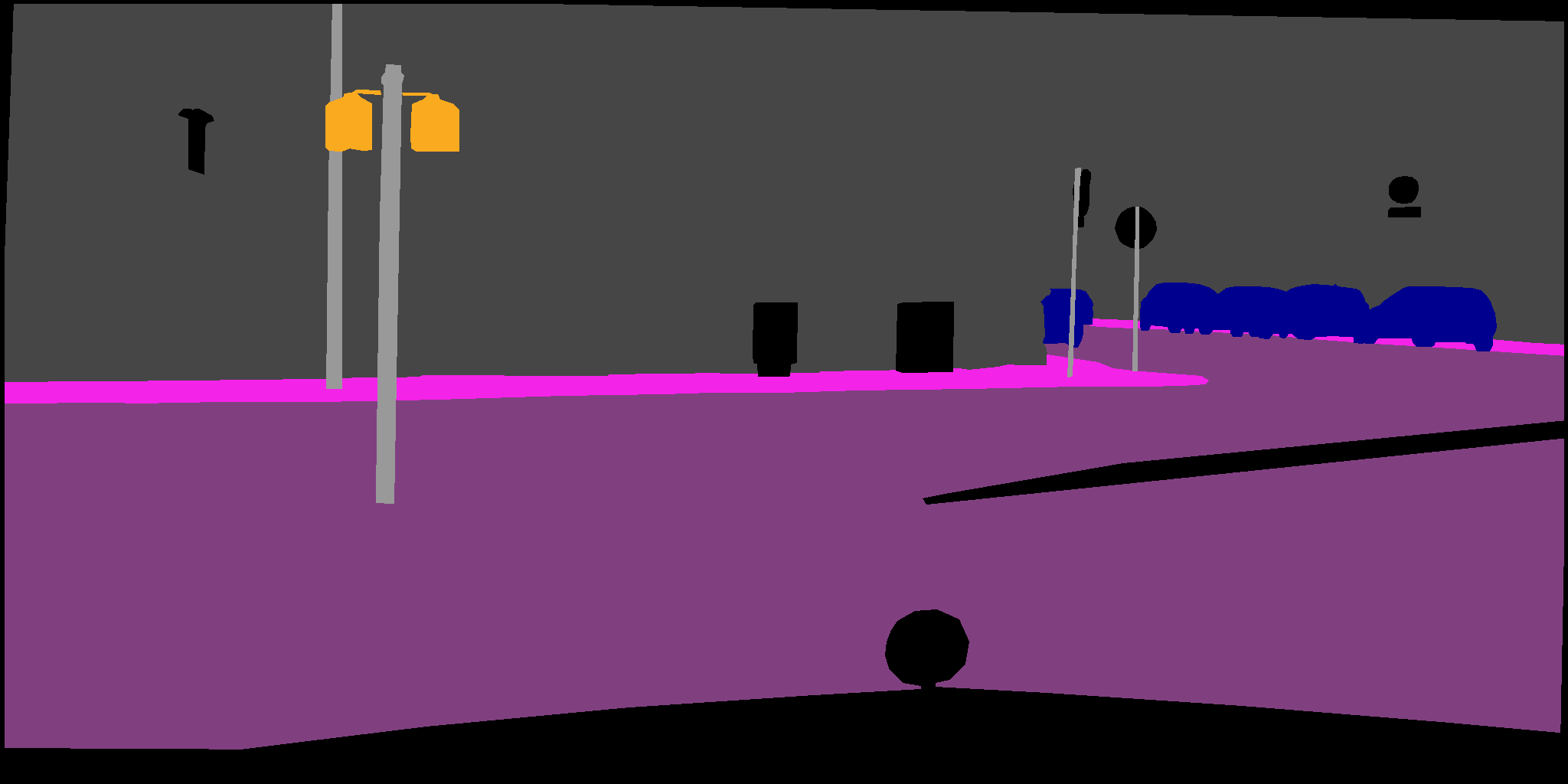}
    \end{subfigure}
    
    \begin{subfigure}[h!]{0.195\linewidth}
        \centering
        \includegraphics[width=\linewidth, height=0.65\linewidth]{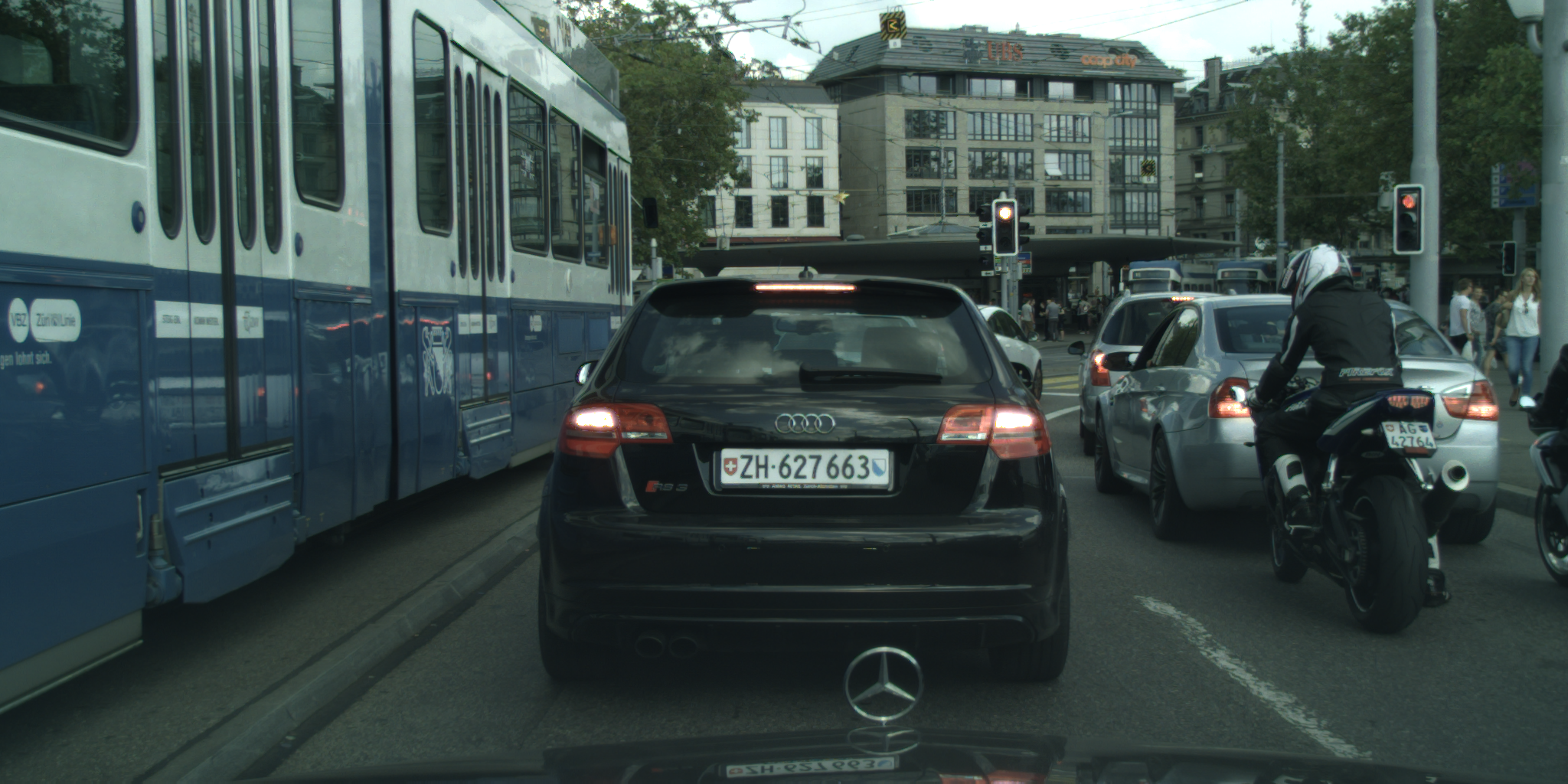}
    \end{subfigure}
    \begin{subfigure}[h!]{0.195\linewidth}
        \centering
        \includegraphics[width=\linewidth, height=0.65\linewidth]{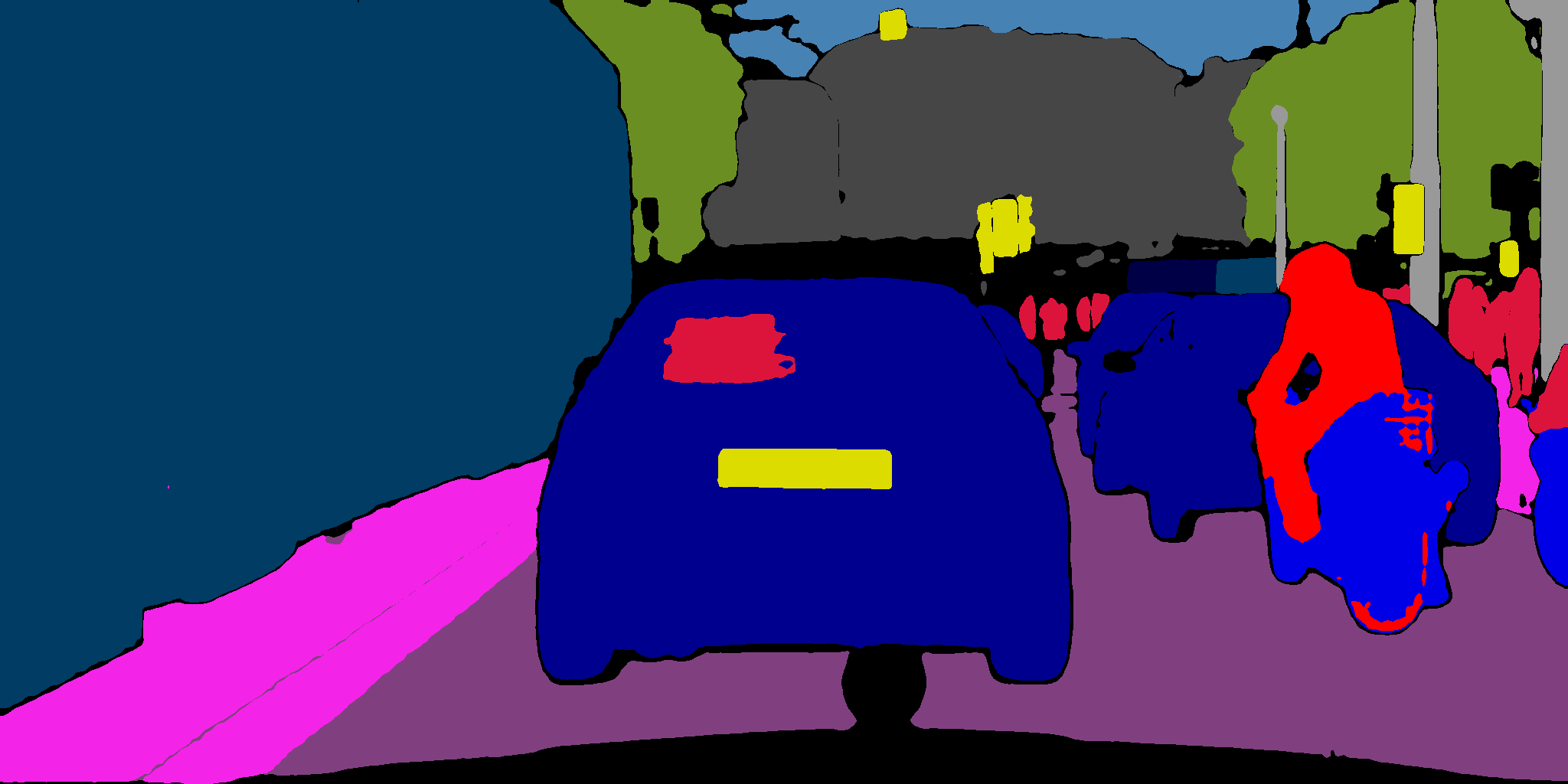}
    \end{subfigure}
    \begin{subfigure}[h!]{0.195\linewidth}
        \centering
        \includegraphics[width=\linewidth, height=0.65\linewidth]{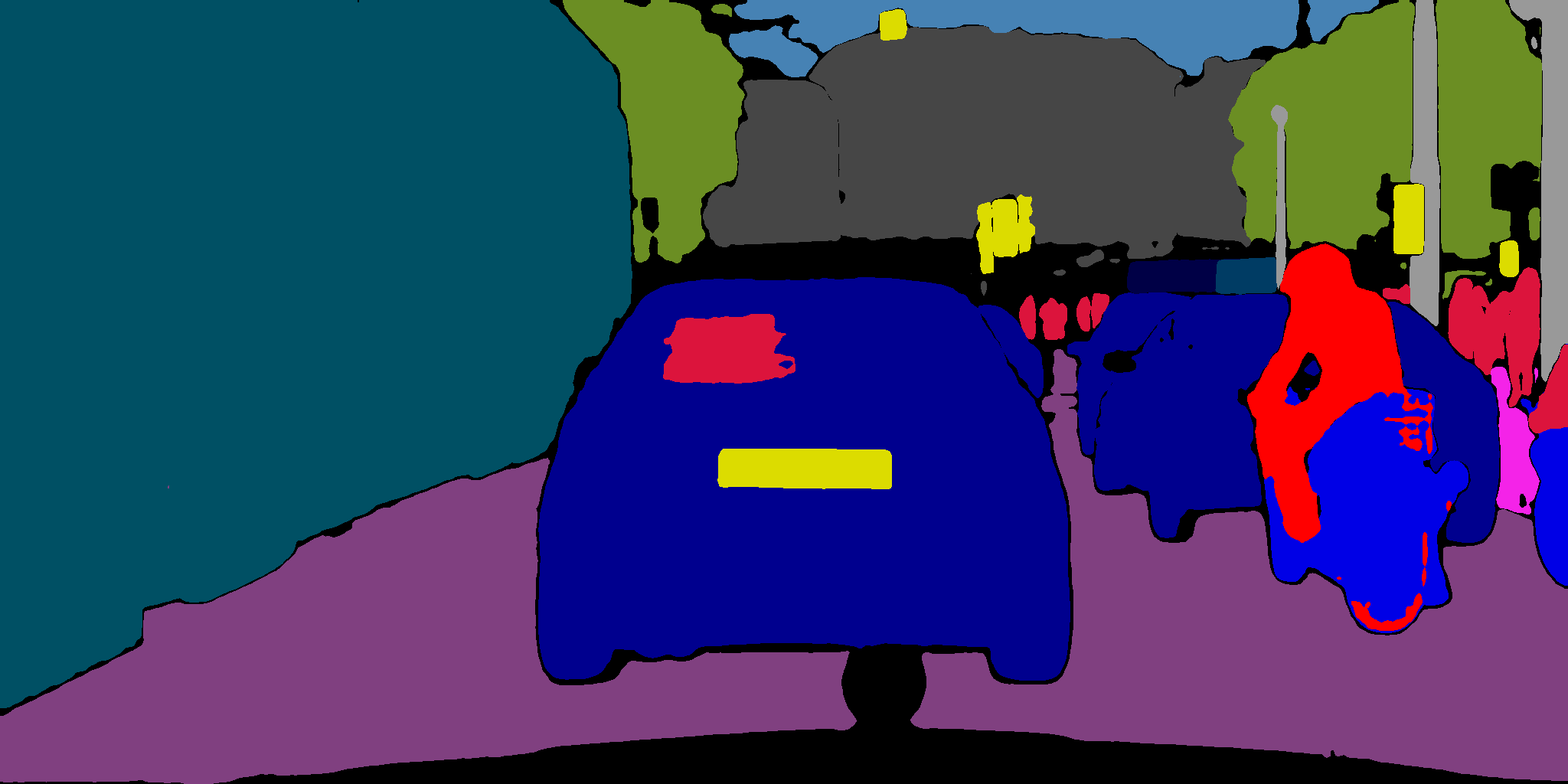}
    \end{subfigure}
    \begin{subfigure}[h!]{0.195\linewidth}
        \centering
        \includegraphics[width=\linewidth, height=0.65\linewidth]{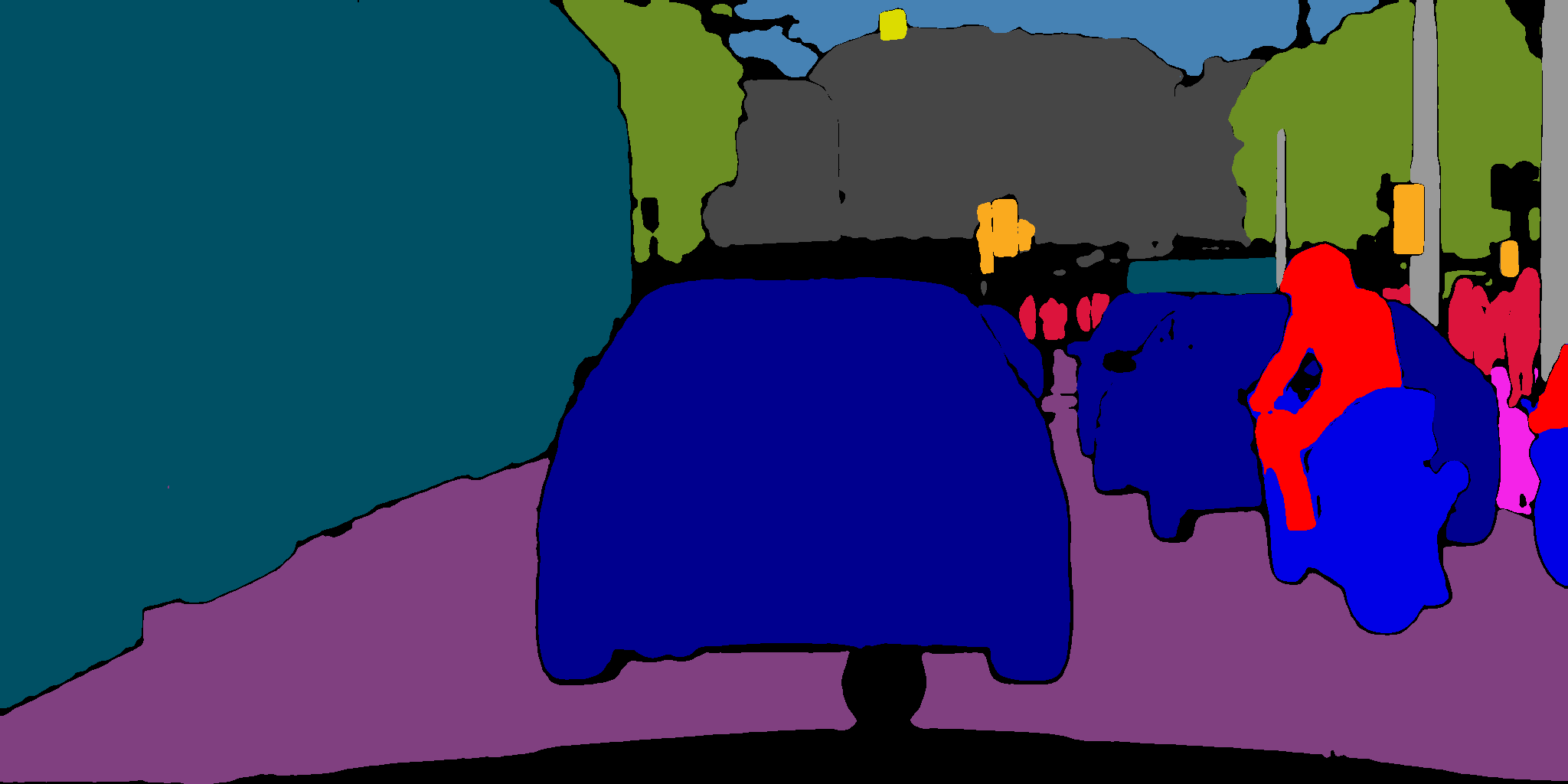}
    \end{subfigure}
    \begin{subfigure}[h!]{0.195\linewidth}
        \centering
        \includegraphics[width=\linewidth, height=0.65\linewidth]{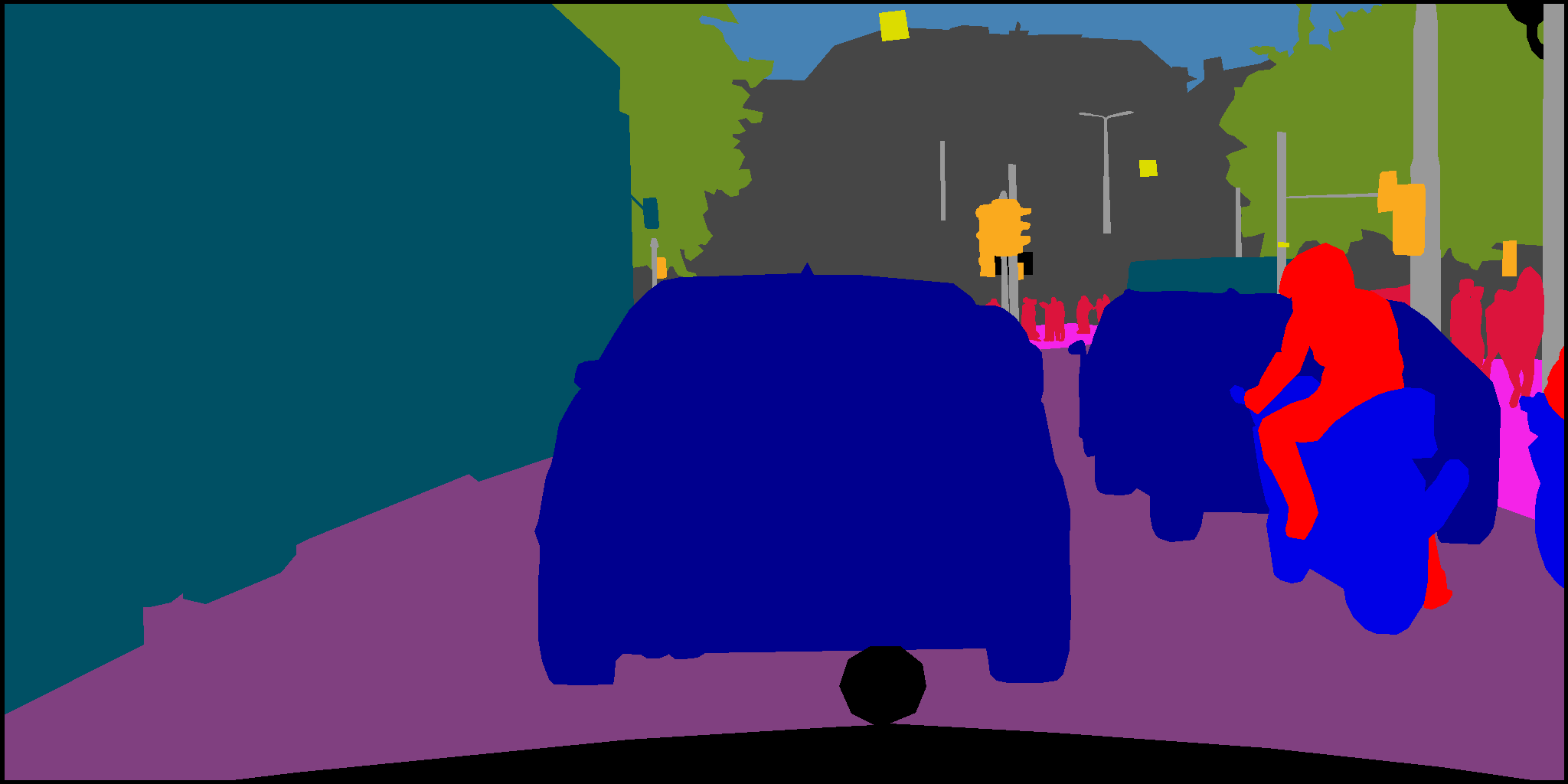}
    \end{subfigure}
    
    \begin{subfigure}[h!]{0.195\linewidth}
        \centering
        \includegraphics[width=\linewidth, height=0.65\linewidth]{figures/pseudo_labels/zurich_000044_000019/rgb.png}
    \end{subfigure}
    \begin{subfigure}[h!]{0.195\linewidth}
        \centering
        \includegraphics[width=\linewidth, height=0.65\linewidth]{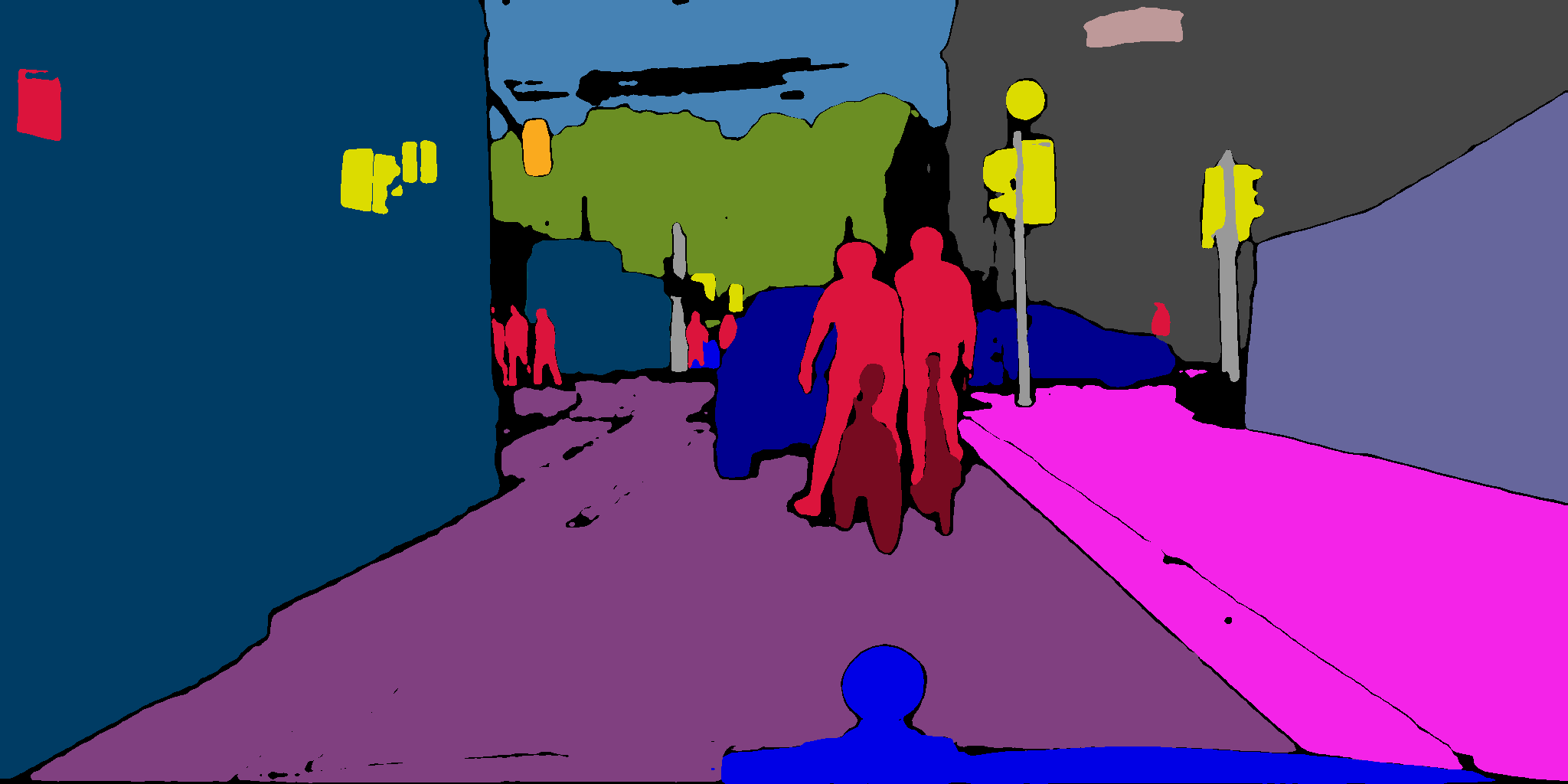}
    \end{subfigure}
    \begin{subfigure}[h!]{0.195\linewidth}
        \centering
        \includegraphics[width=\linewidth, height=0.65\linewidth]{figures/pseudo_labels/zurich_000044_000019/alc.png}
    \end{subfigure}
    \begin{subfigure}[h!]{0.195\linewidth}
        \centering
        \includegraphics[width=\linewidth, height=0.65\linewidth]{figures/pseudo_labels/zurich_000044_000019/ours.png}
    \end{subfigure}
    \begin{subfigure}[h!]{0.195\linewidth}
        \centering
        \includegraphics[width=\linewidth, height=0.65\linewidth]{figures/pseudo_labels/zurich_000044_000019/gt.png}
    \end{subfigure}
    
    \begin{subfigure}[h!]{0.195\linewidth}
        \centering
        \includegraphics[width=\linewidth, height=0.65\linewidth]{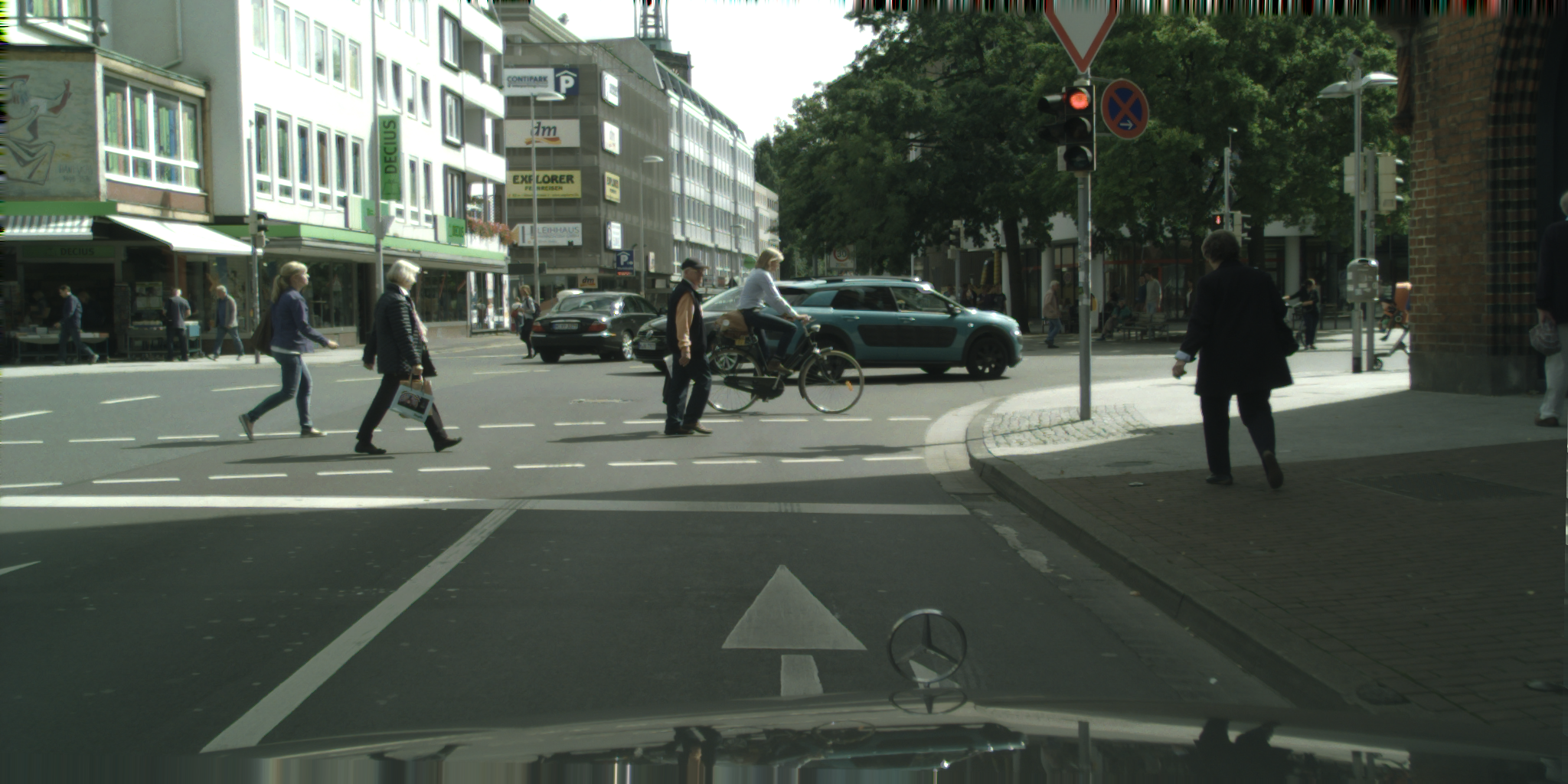}
    \end{subfigure}
    \begin{subfigure}[h!]{0.195\linewidth}
        \centering
        \includegraphics[width=\linewidth, height=0.65\linewidth]{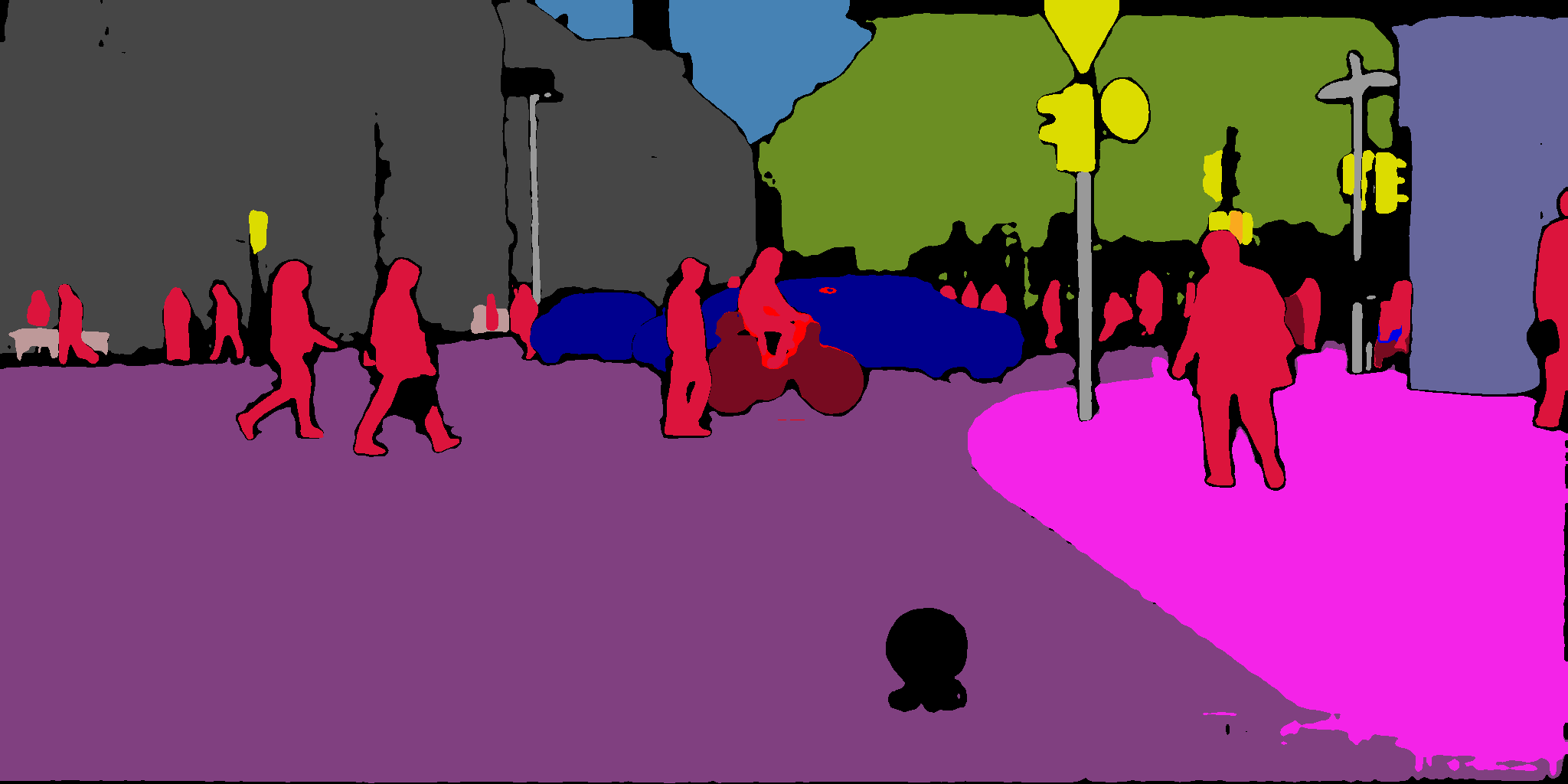}
    \end{subfigure}
    \begin{subfigure}[h!]{0.195\linewidth}
        \centering
        \includegraphics[width=\linewidth, height=0.65\linewidth]{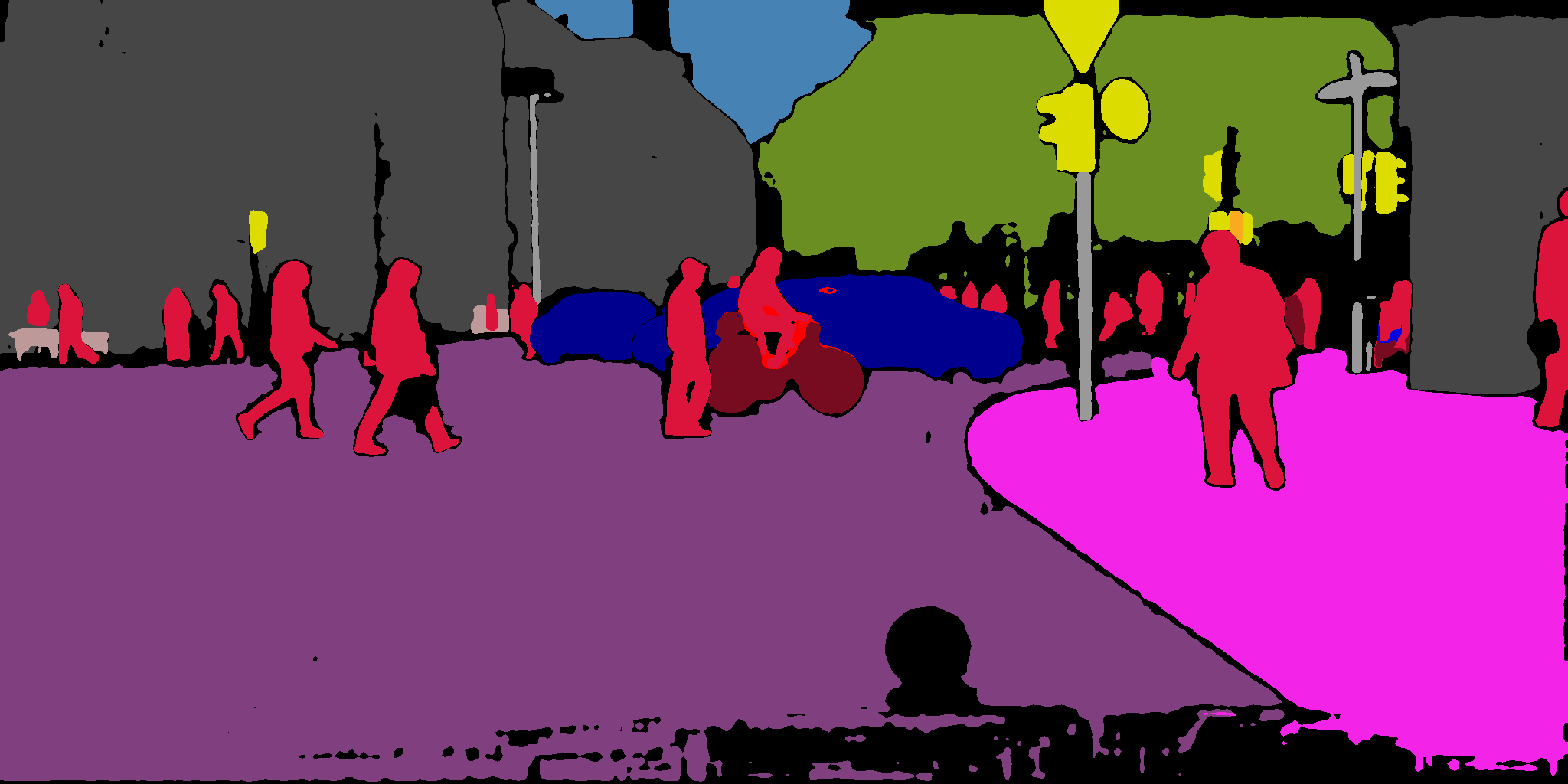}
    \end{subfigure}
    \begin{subfigure}[h!]{0.195\linewidth}
        \centering
        \includegraphics[width=\linewidth, height=0.65\linewidth]{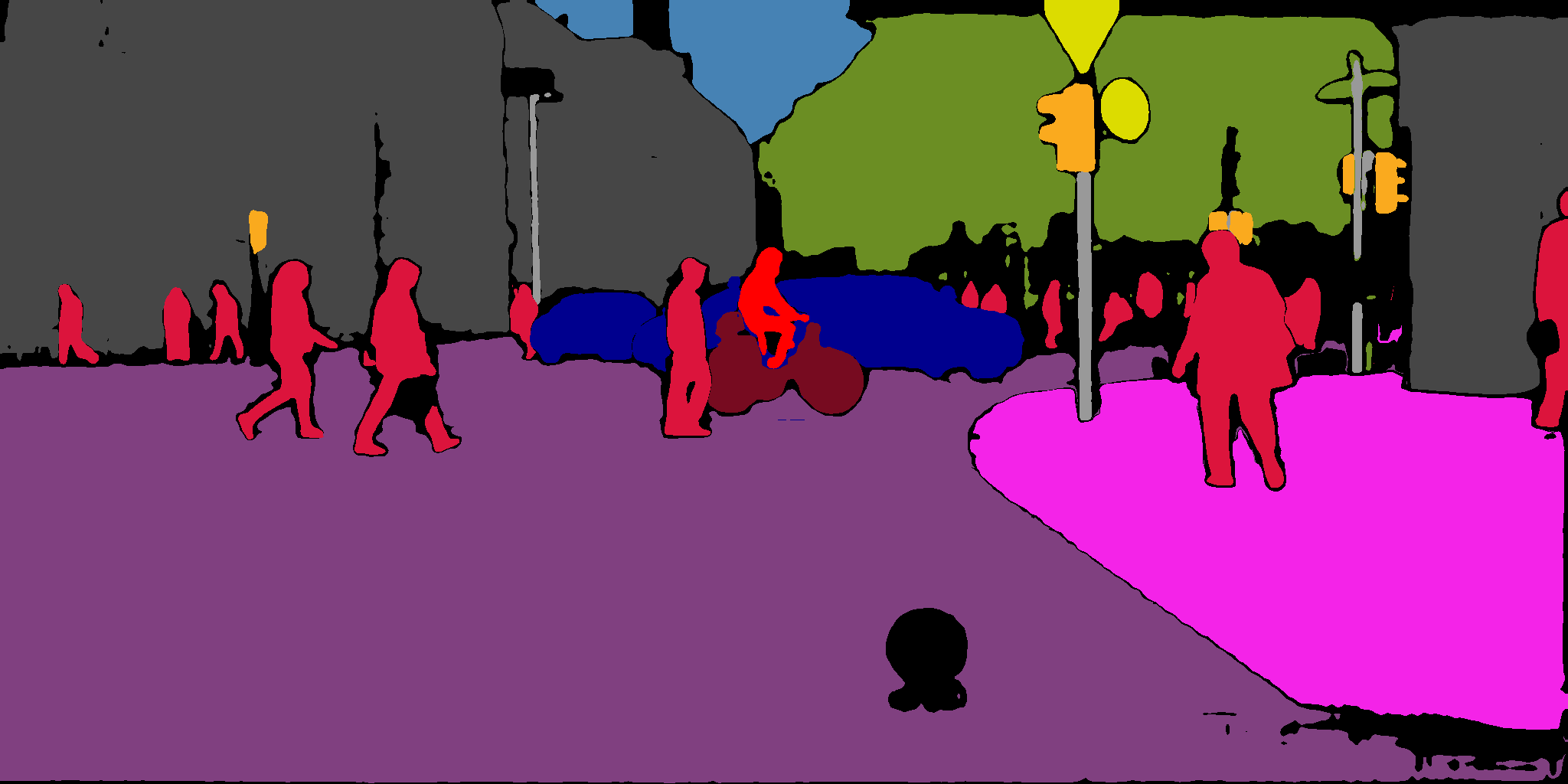}
    \end{subfigure}
    \begin{subfigure}[h!]{0.195\linewidth}
        \centering
        \includegraphics[width=\linewidth, height=0.65\linewidth]{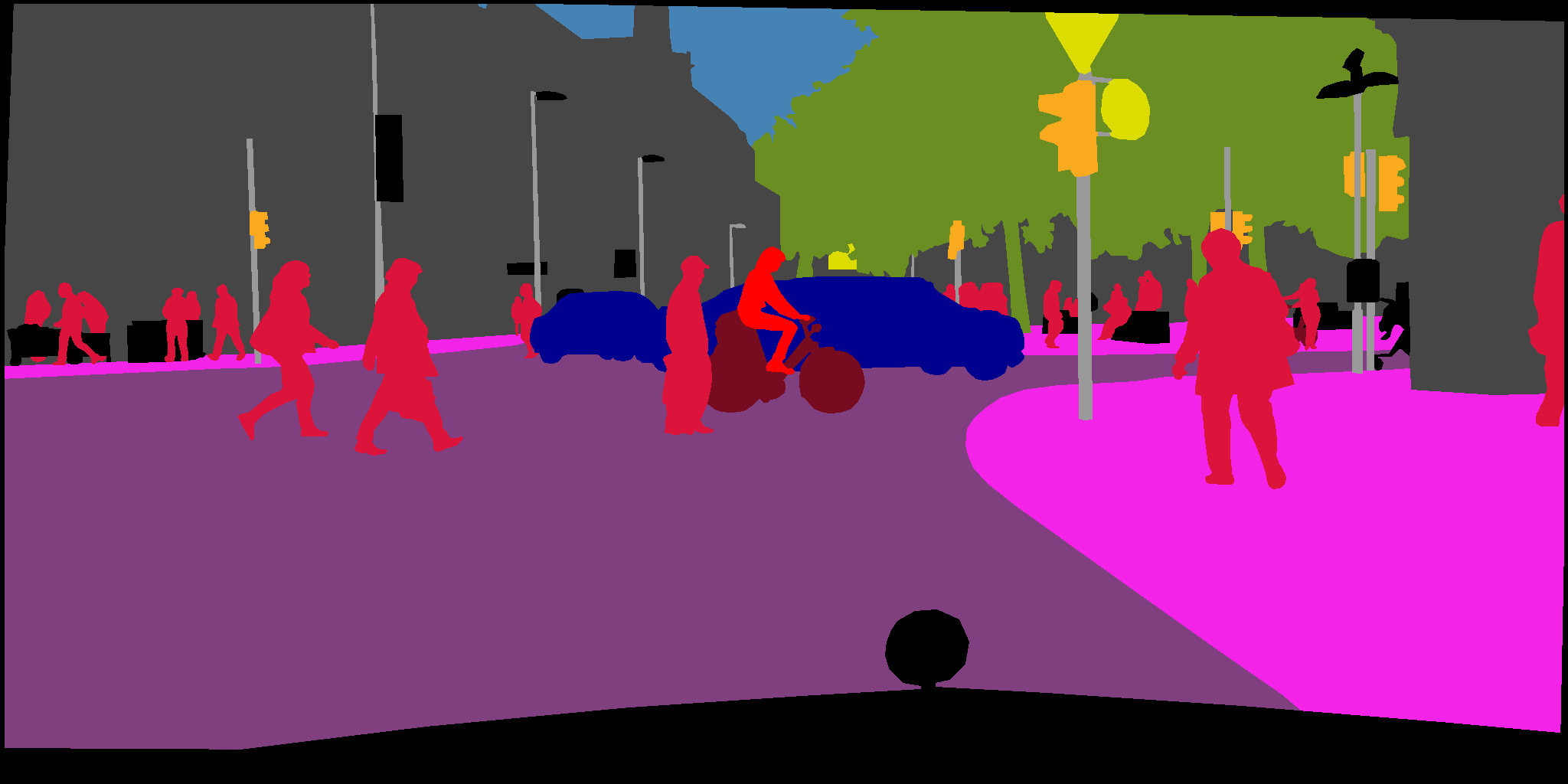}
    \end{subfigure} 
    
    \begin{subfigure}[h!]{0.195\linewidth}
        \centering
        \includegraphics[width=\linewidth, height=0.65\linewidth]{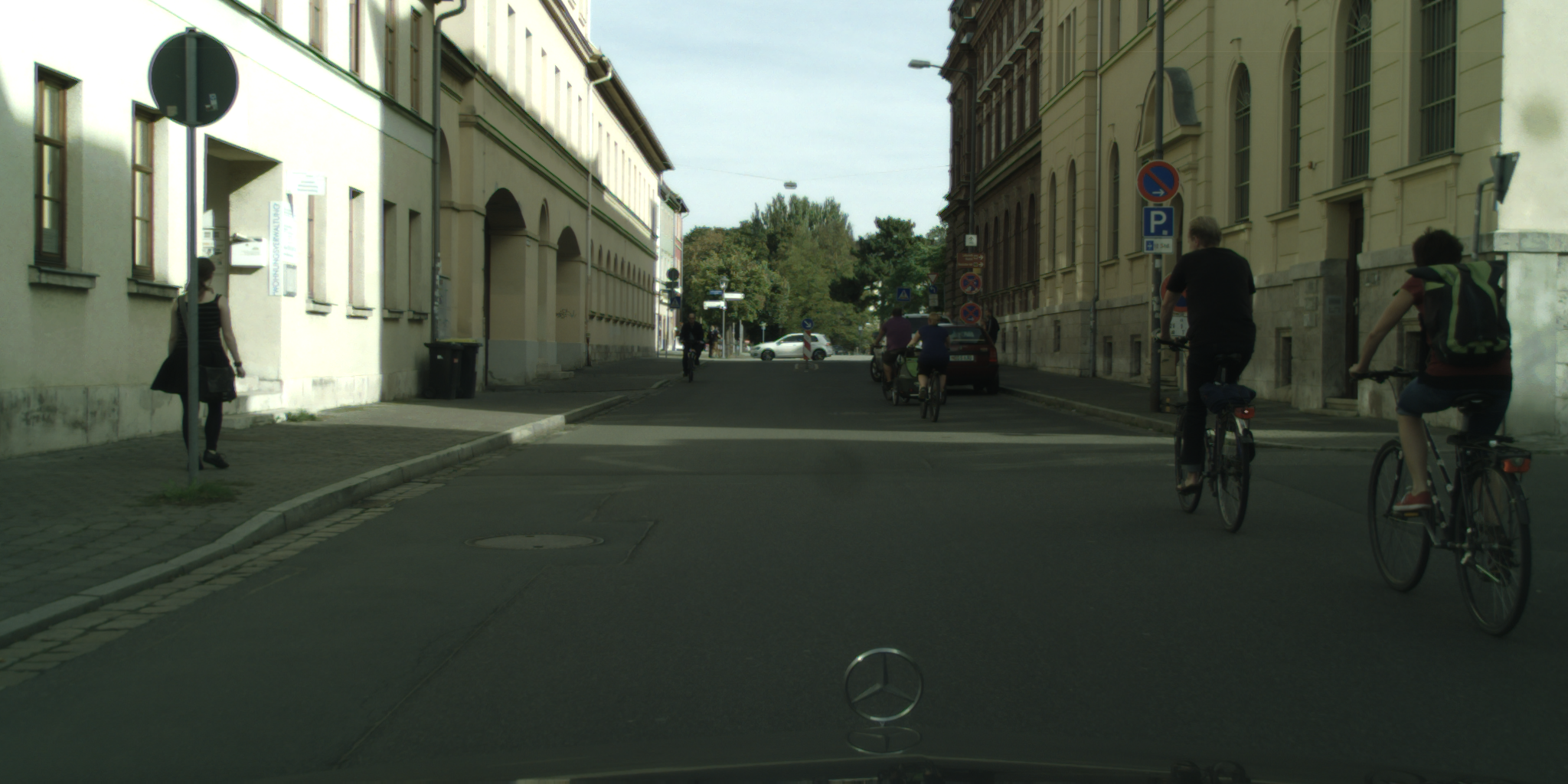}
        \caption{Unlabeled Image}
    \end{subfigure}
    \begin{subfigure}[h!]{0.195\linewidth}
        \centering
        \includegraphics[width=\linewidth, height=0.65\linewidth]{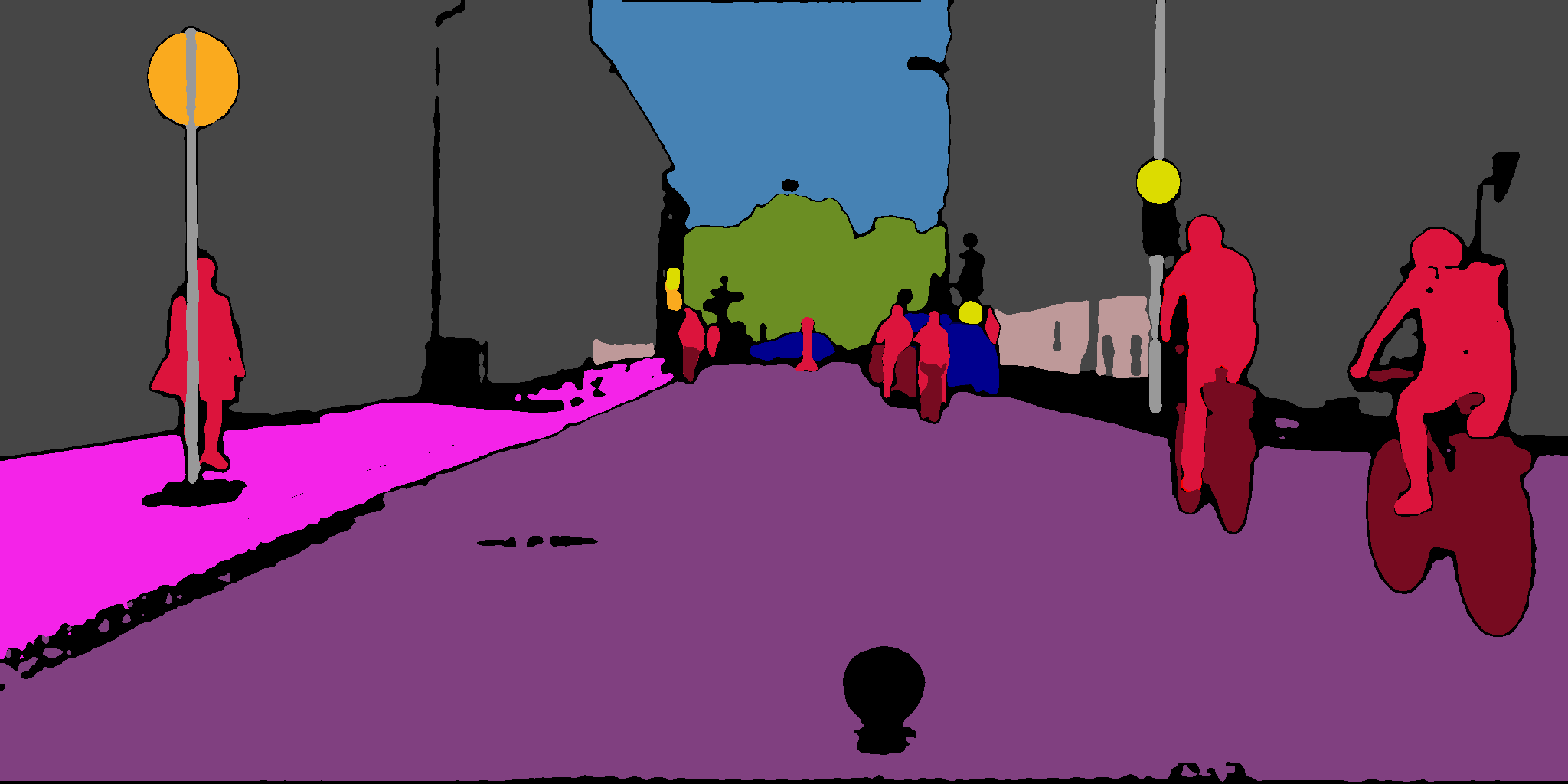}
        \caption{Grounded SAM}
    \end{subfigure}
    \begin{subfigure}[h!]{0.195\linewidth}
        \centering
        \includegraphics[width=\linewidth, height=0.65\linewidth]{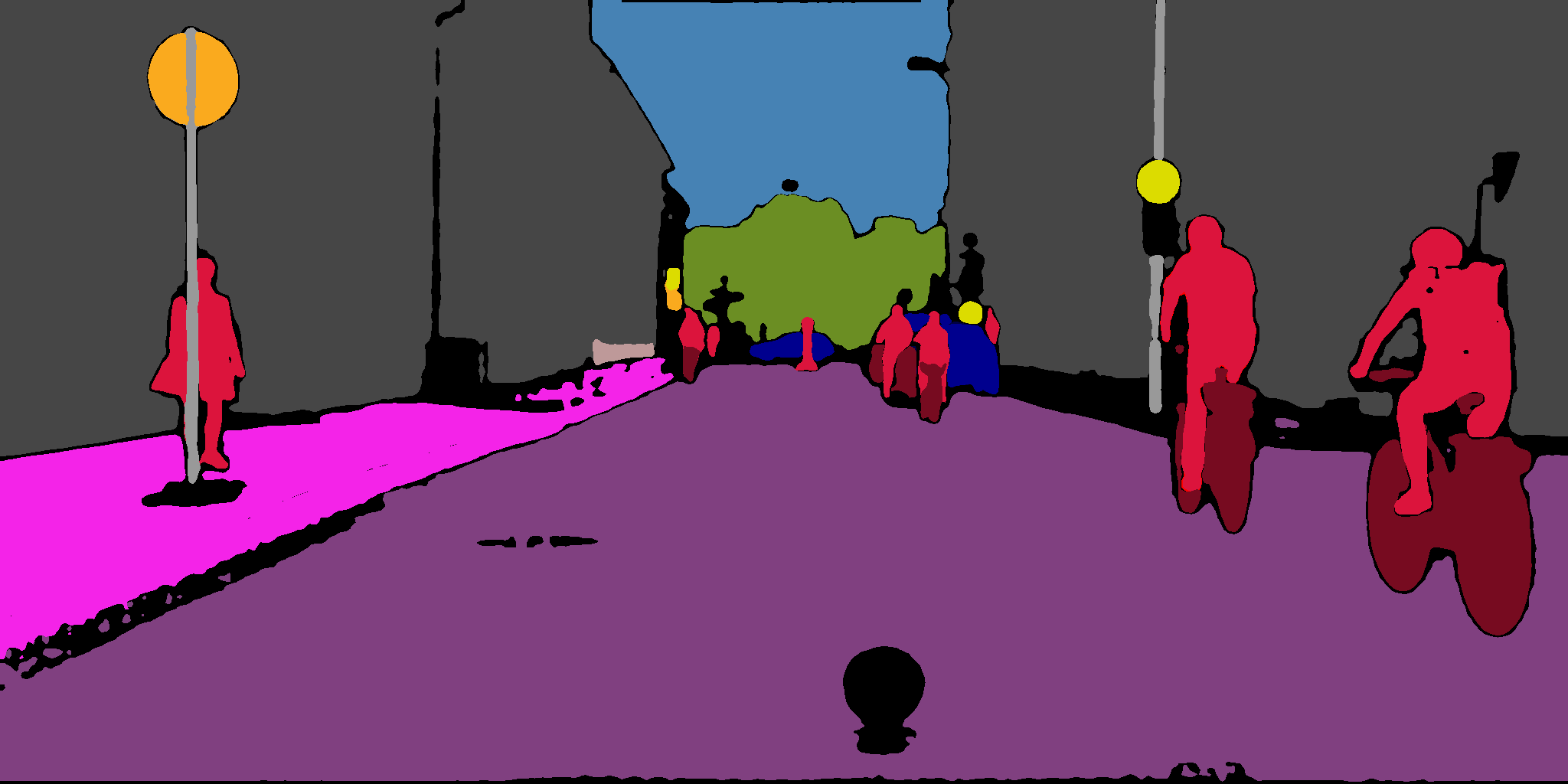}
        \caption{ALC~\cite{kim2024active}}
    \end{subfigure}
    \begin{subfigure}[h!]{0.195\linewidth}
        \centering
        \includegraphics[width=\linewidth, height=0.65\linewidth]{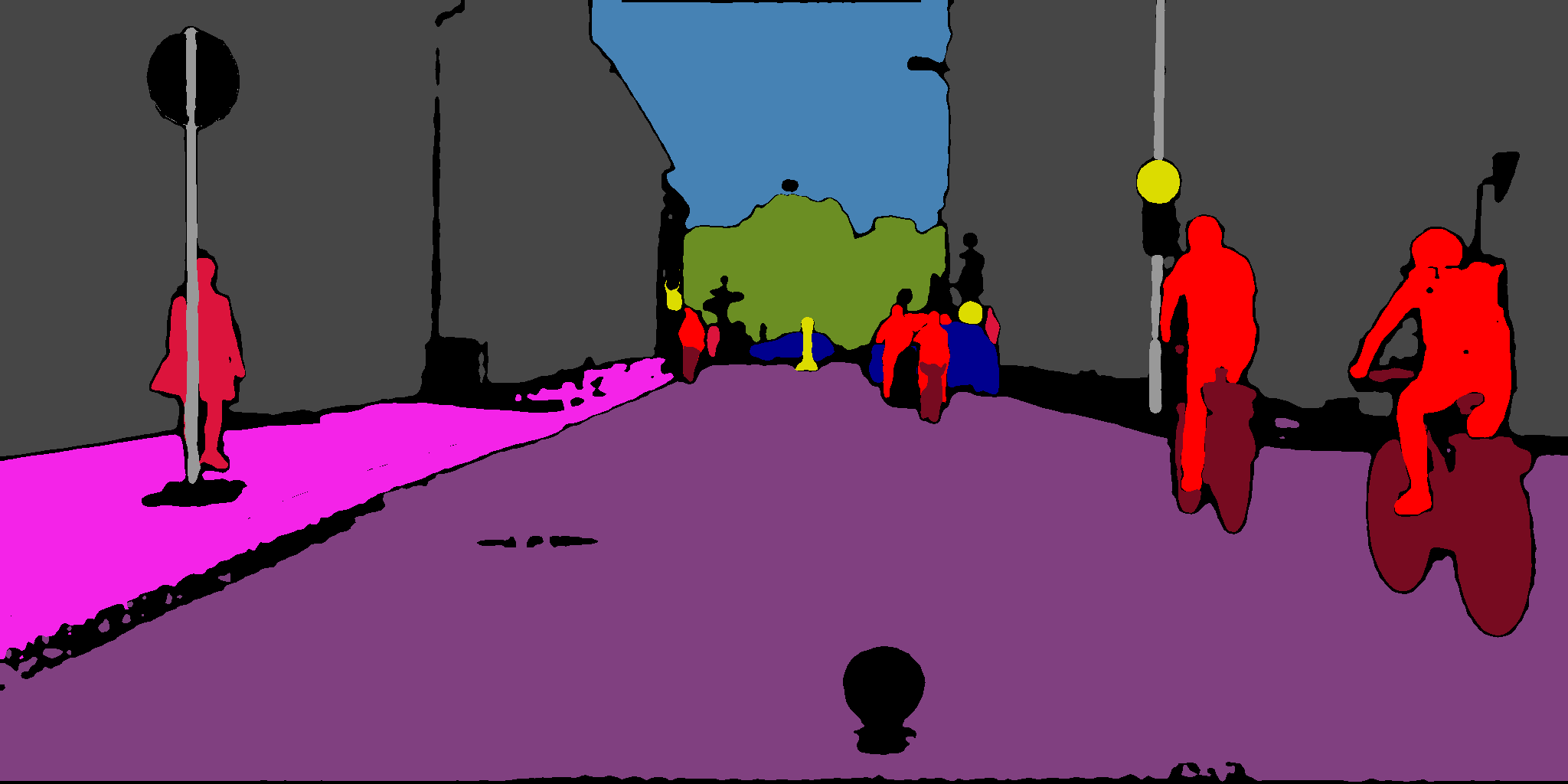}
        \caption{A\(^2\)LC (Ours)}
    \end{subfigure}
    \begin{subfigure}[h!]{0.195\linewidth}
        \centering
        \includegraphics[width=\linewidth, height=0.65\linewidth]{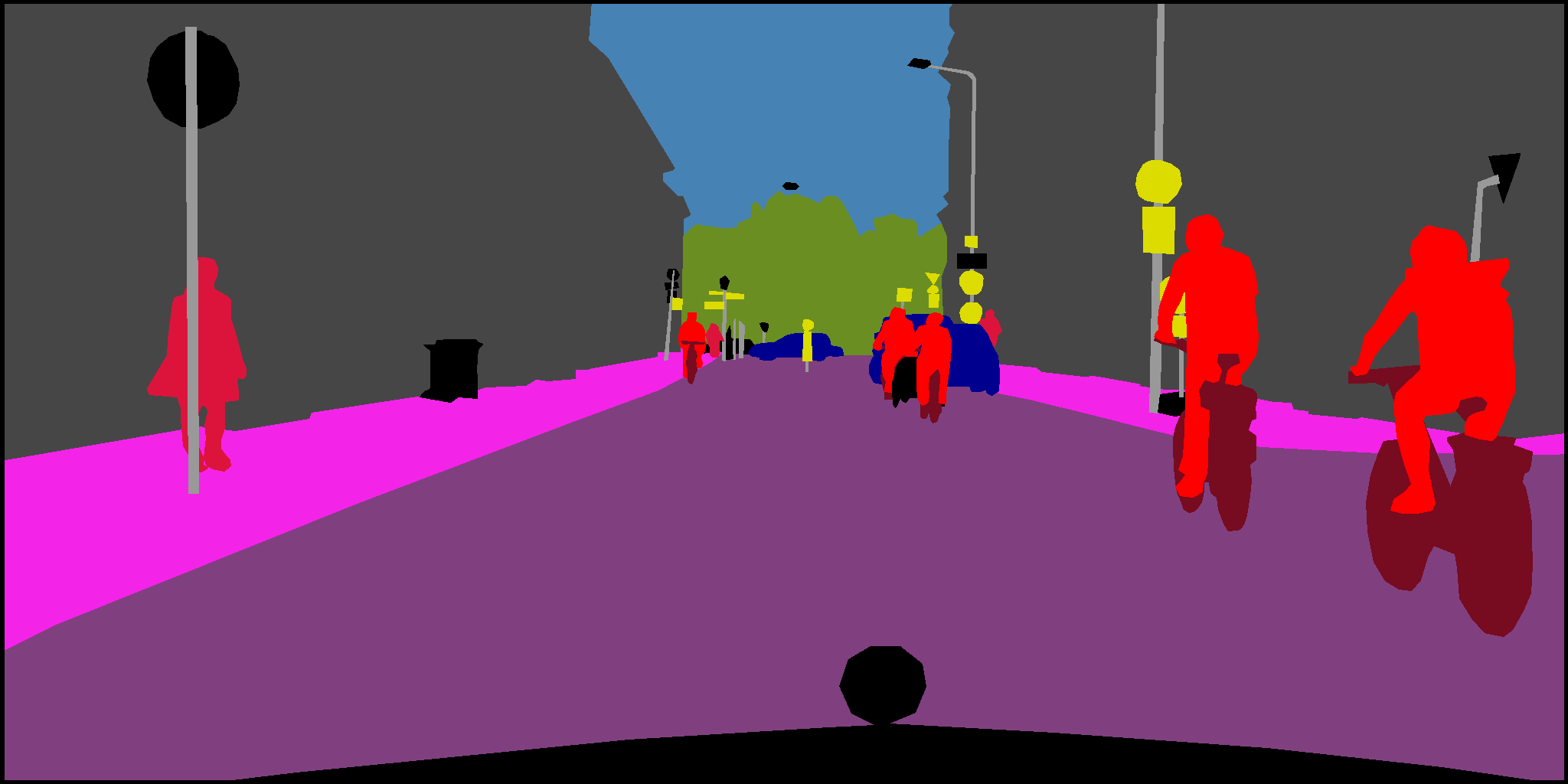}
        \caption{Ground Truth}
    \end{subfigure}
    \caption{Qualitative results of constructed pseudo-labels. A\(^2\)LC achieves significantly superior correction results compared to previous studies. Overall, the \textit{traffic light (orange)} masks that are previously mislabeled as \textit{traffic sign (yellow)} are now properly corrected. Additionally, in the bottom three rows, the masks mislabeled as \textit{person (darker red)} are accurately corrected to \textit{rider (pure red)}. Considering these classes belong to tail classes, this demonstrates that our proposed framework effectively mitigates the previously overlooked issue of class imbalance.}
    \label{fig:qual_1_1}
\end{minipage}
\end{figure*}

\begin{figure*}[t]
\centering    
\begin{minipage}{\textwidth}
    \centering 
    \begin{subfigure}[h!]{0.195\linewidth}
        \centering
        \includegraphics[width=\linewidth, height=0.65\linewidth]{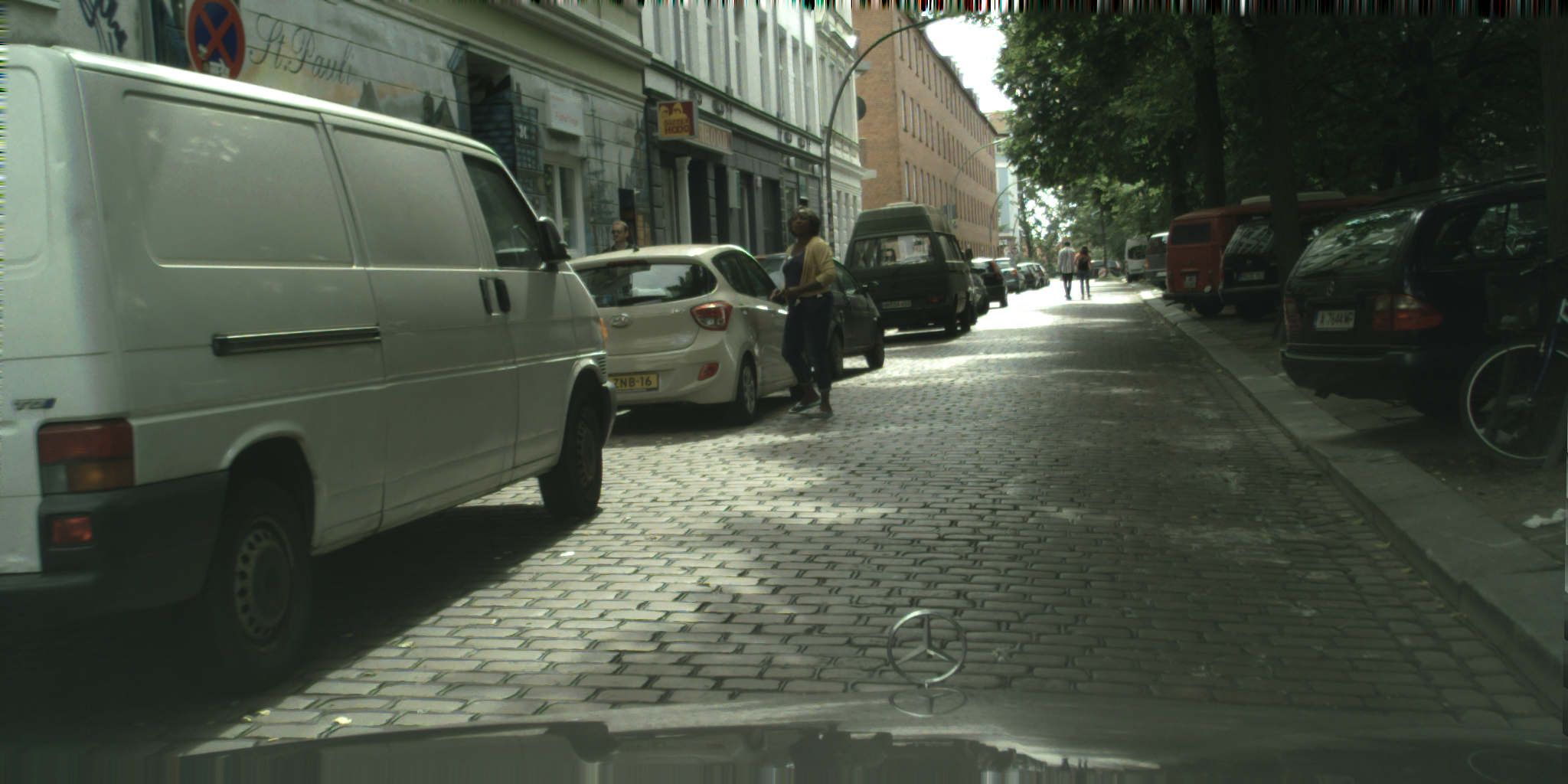}
    \end{subfigure}
    \begin{subfigure}[h!]{0.195\linewidth}
        \centering
        \includegraphics[width=\linewidth, height=0.65\linewidth]{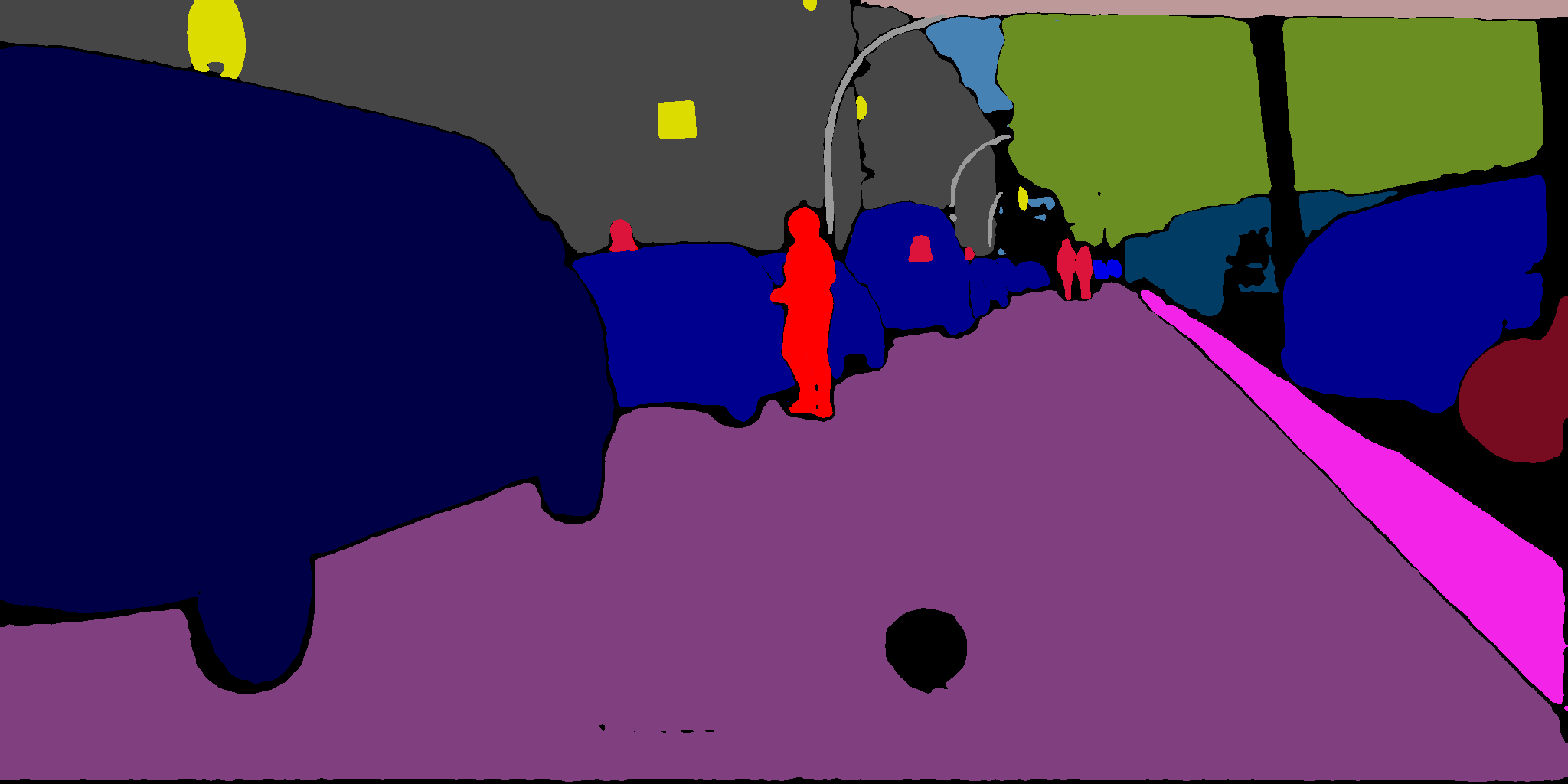}
    \end{subfigure}
    \begin{subfigure}[h!]{0.195\linewidth}
        \centering
        \includegraphics[width=\linewidth, height=0.65\linewidth]{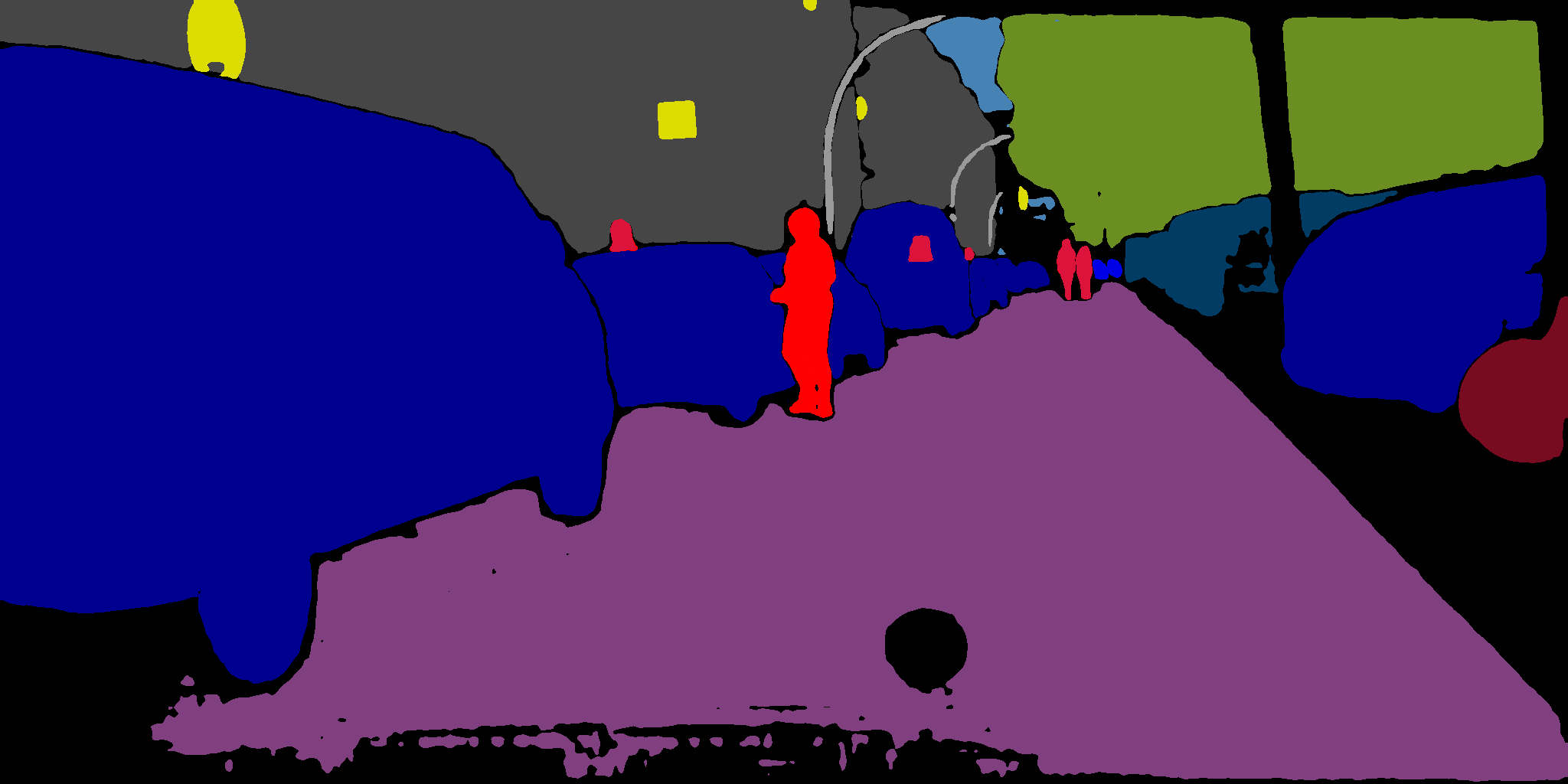}
    \end{subfigure}
    \begin{subfigure}[h!]{0.195\linewidth}
        \centering
        \includegraphics[width=\linewidth, height=0.65\linewidth]{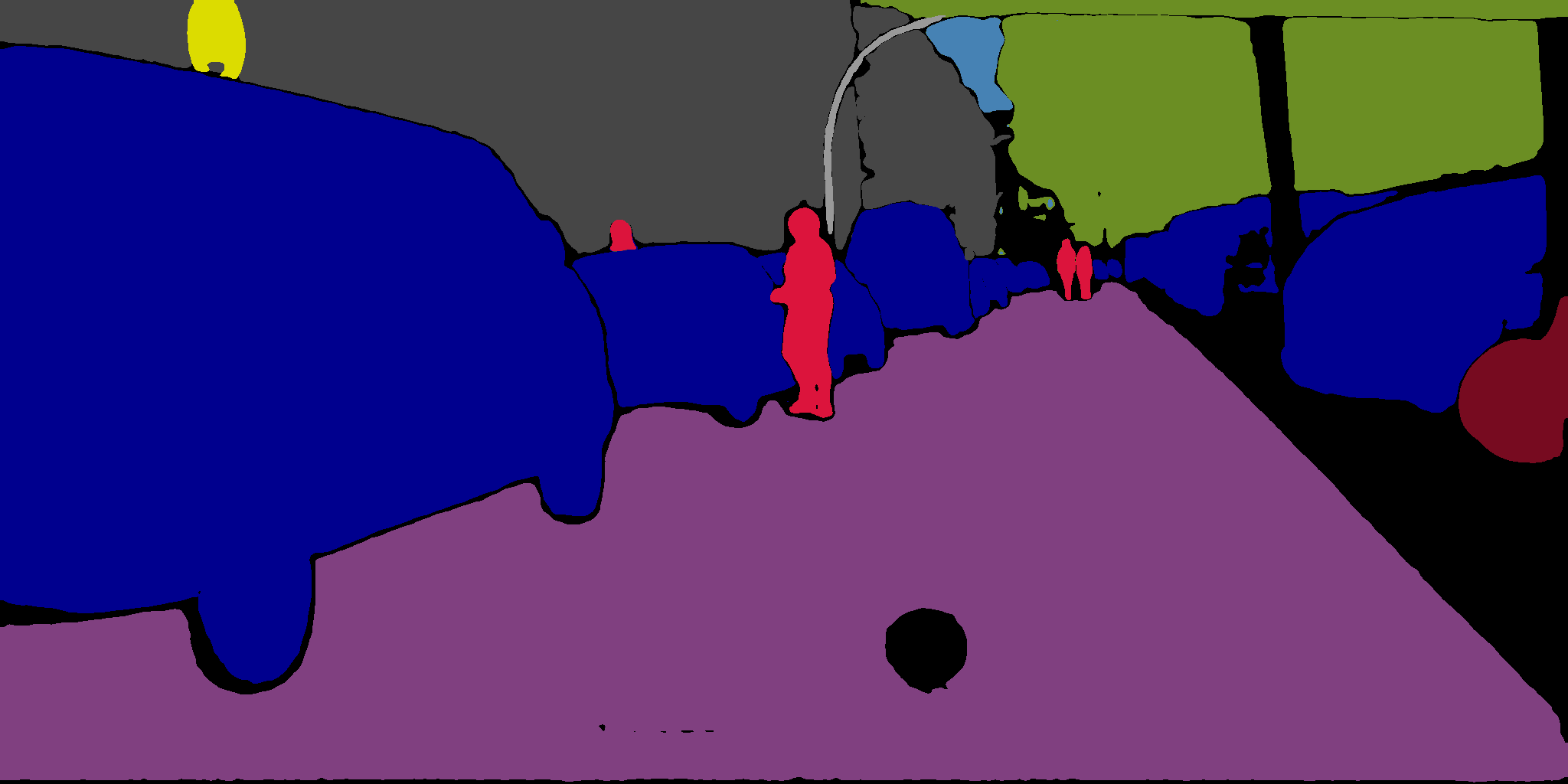}
    \end{subfigure}
    \begin{subfigure}[h!]{0.195\linewidth}
        \centering
        \includegraphics[width=\linewidth, height=0.65\linewidth]{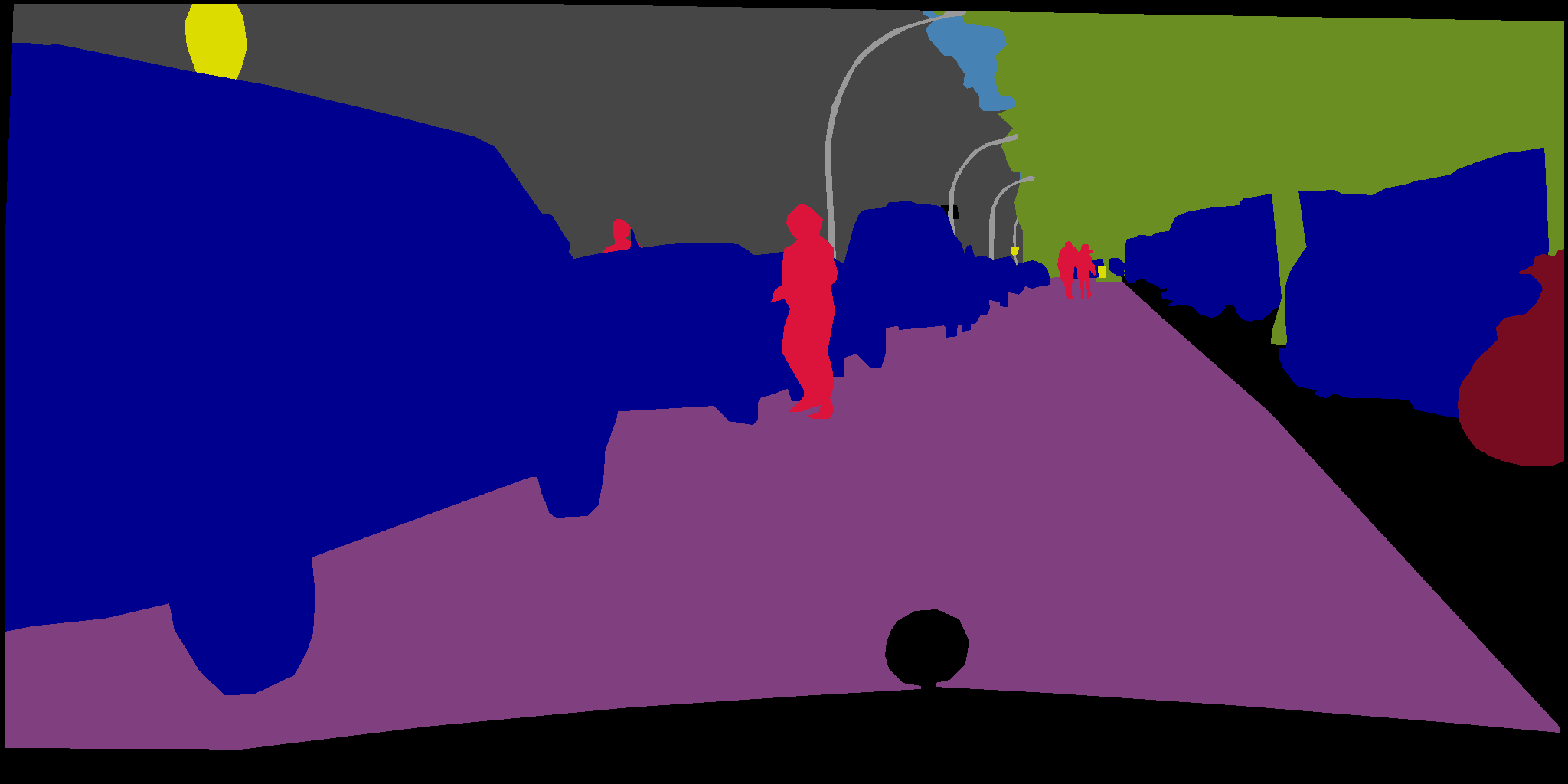}
    \end{subfigure}

    \begin{subfigure}[h!]{0.195\linewidth}
        \centering
        \includegraphics[width=\linewidth, height=0.65\linewidth]{figures/pseudo_labels/zurich_000069_000019/rgb.png}
    \end{subfigure}
    \begin{subfigure}[h!]{0.195\linewidth}
        \centering
        \includegraphics[width=\linewidth, height=0.65\linewidth]{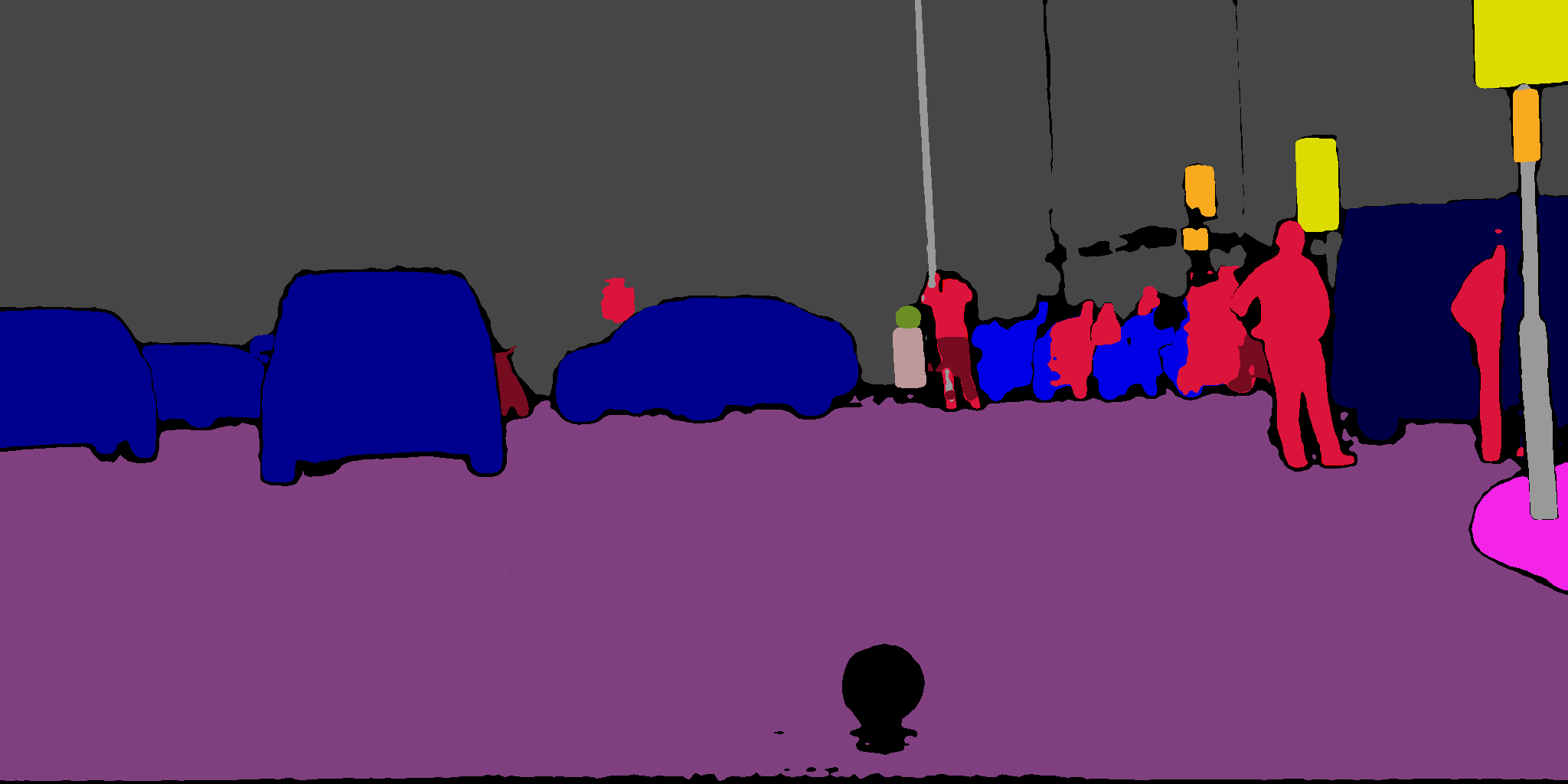}
    \end{subfigure}
    \begin{subfigure}[h!]{0.195\linewidth}
        \centering
        \includegraphics[width=\linewidth, height=0.65\linewidth]{figures/pseudo_labels/zurich_000069_000019/alc.png}
    \end{subfigure}
    \begin{subfigure}[h!]{0.195\linewidth}
        \centering
        \includegraphics[width=\linewidth, height=0.65\linewidth]{figures/pseudo_labels/zurich_000069_000019/ours.png}
    \end{subfigure}
    \begin{subfigure}[h!]{0.195\linewidth}
        \centering
        \includegraphics[width=\linewidth, height=0.65\linewidth]{figures/pseudo_labels/zurich_000069_000019/gt.png}
    \end{subfigure}

    \begin{subfigure}[h!]{0.195\linewidth}
        \centering
        \includegraphics[width=\linewidth, height=0.65\linewidth]{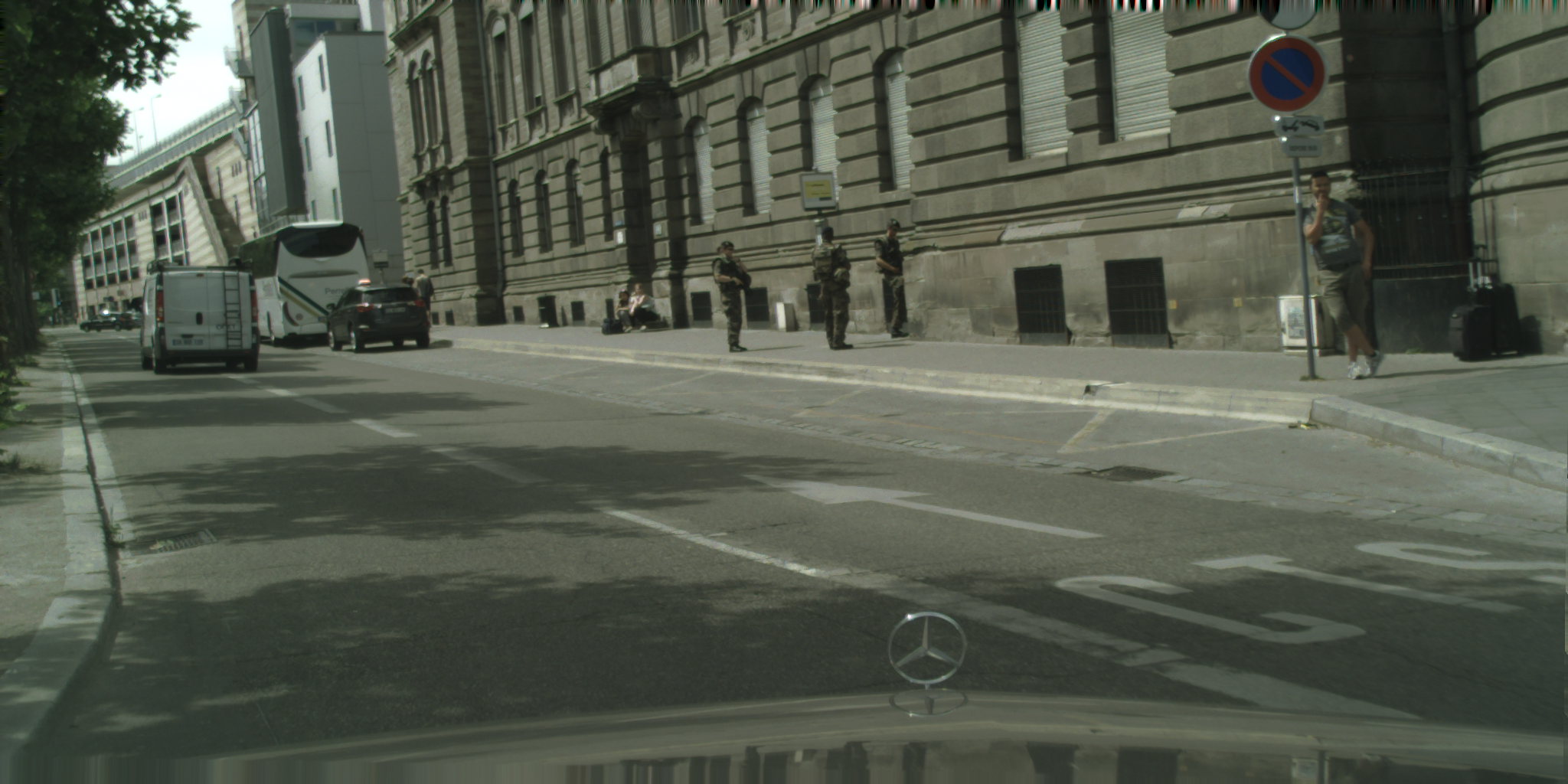}
    \end{subfigure}
    \begin{subfigure}[h!]{0.195\linewidth}
        \centering
        \includegraphics[width=\linewidth, height=0.65\linewidth]{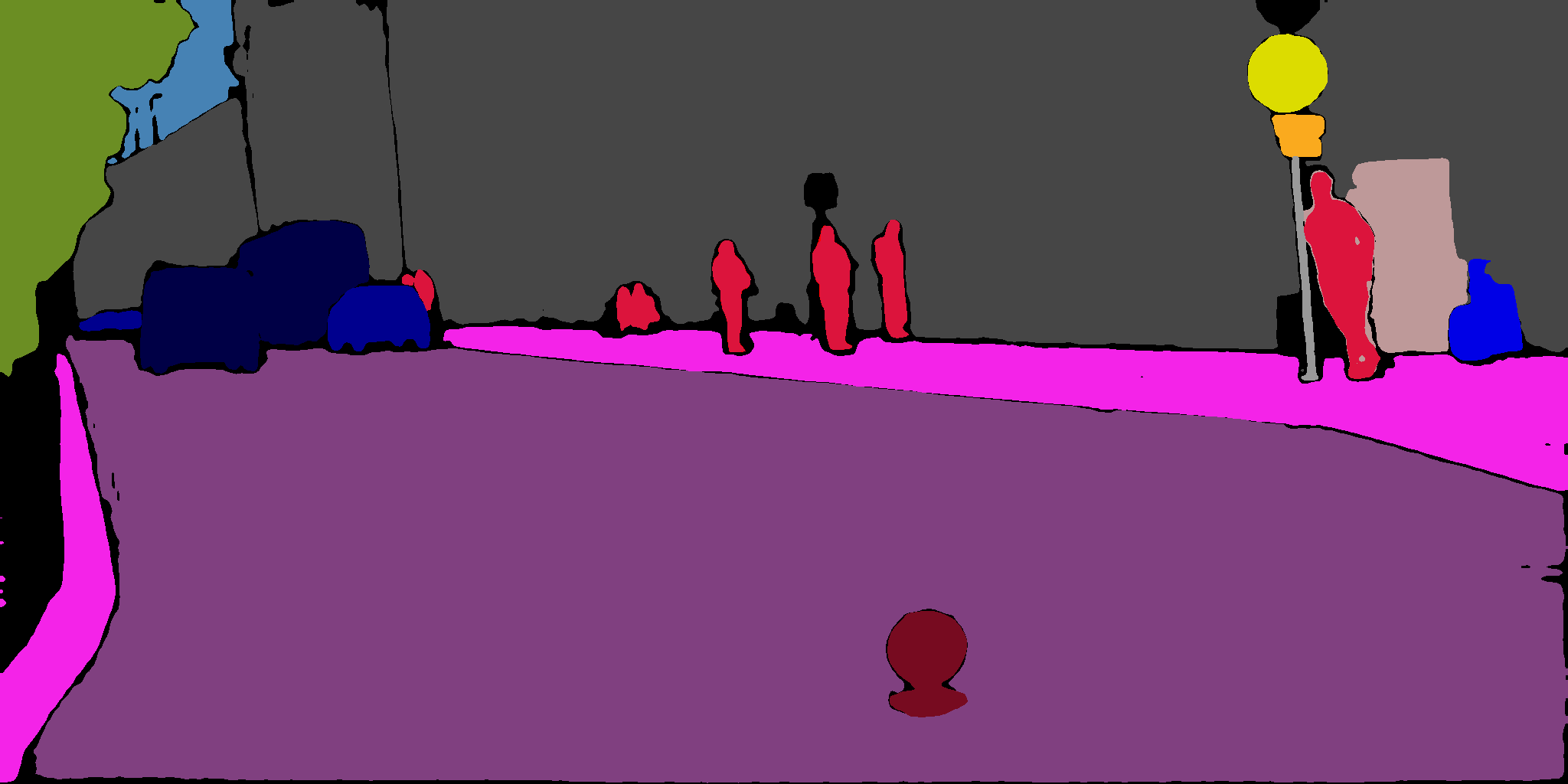}
    \end{subfigure}
    \begin{subfigure}[h!]{0.195\linewidth}
        \centering
        \includegraphics[width=\linewidth, height=0.65\linewidth]{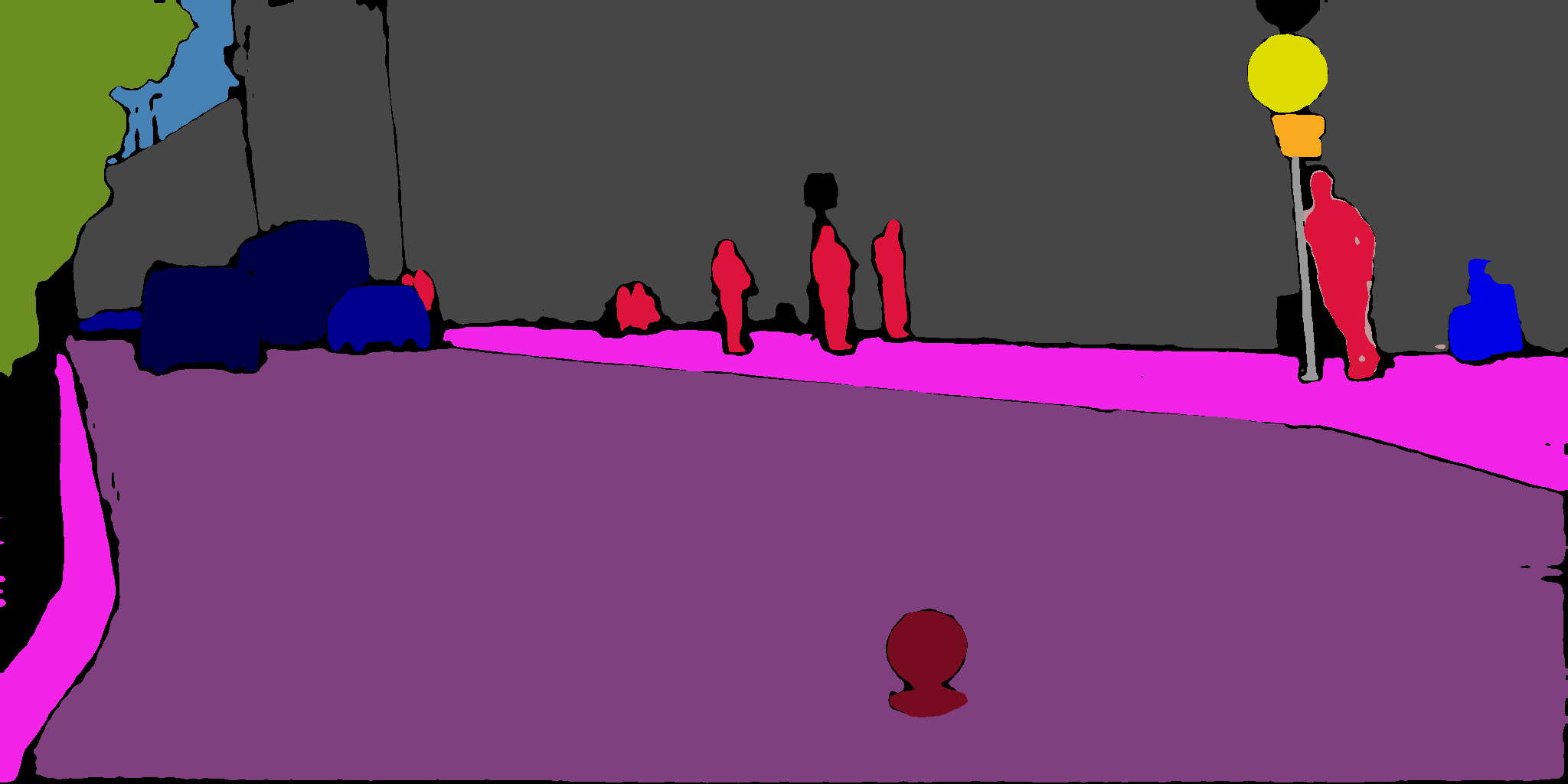}
    \end{subfigure}
    \begin{subfigure}[h!]{0.195\linewidth}
        \centering
        \includegraphics[width=\linewidth, height=0.65\linewidth]{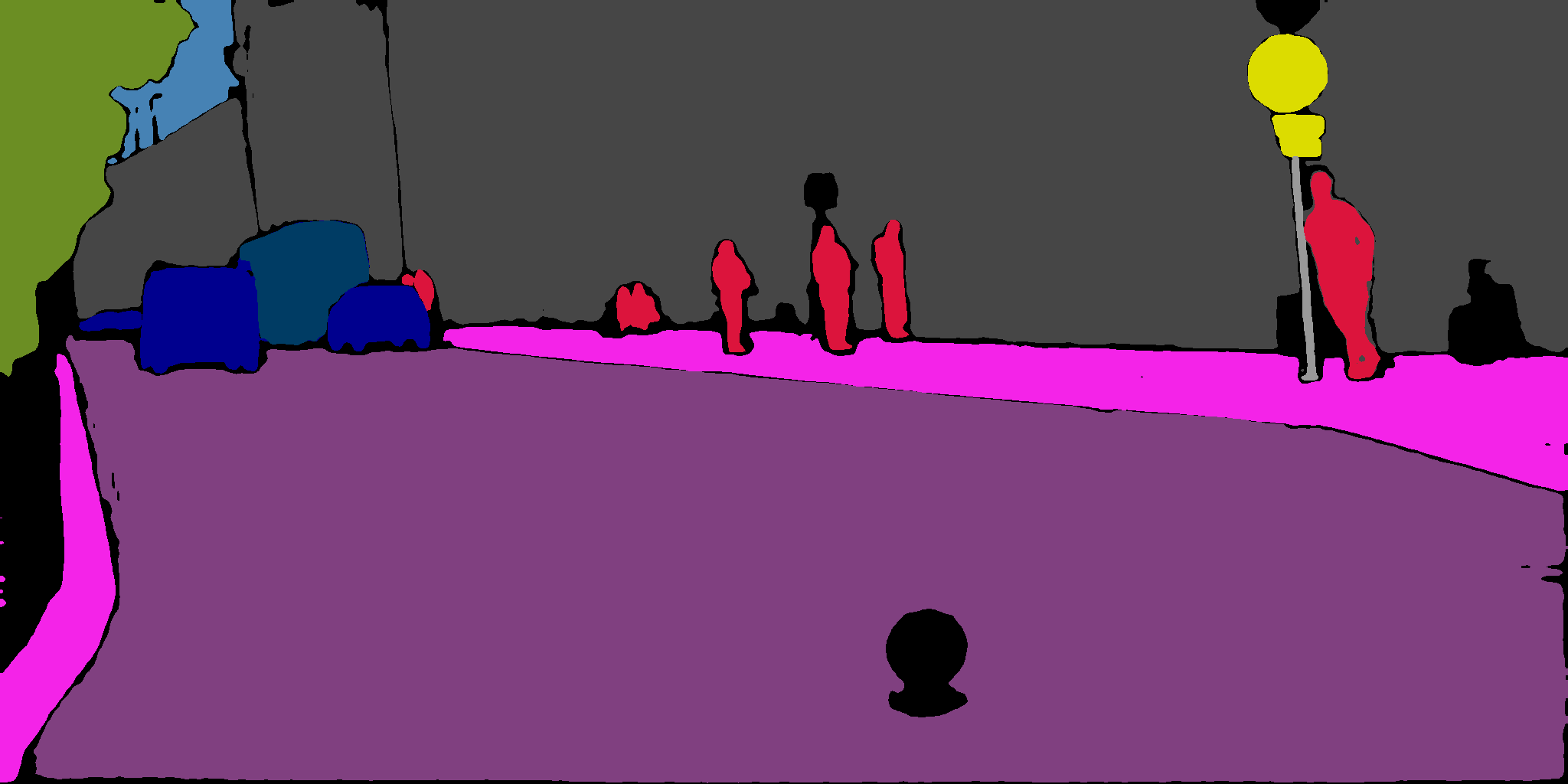}
    \end{subfigure}
    \begin{subfigure}[h!]{0.195\linewidth}
        \centering
        \includegraphics[width=\linewidth, height=0.65\linewidth]{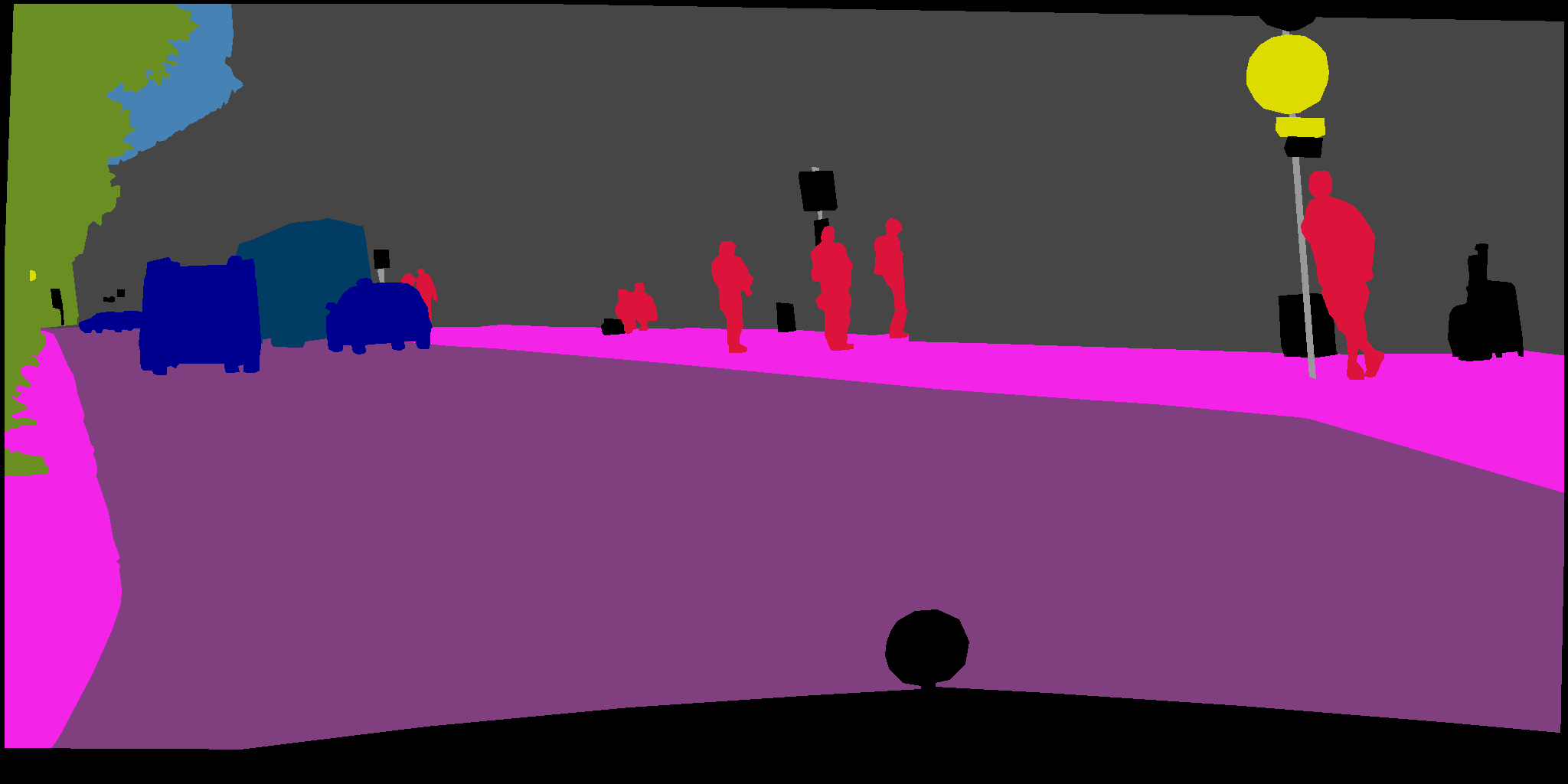}
    \end{subfigure}

    \begin{subfigure}[h!]{0.195\linewidth}
        \centering
        \includegraphics[width=\linewidth, height=0.65\linewidth]{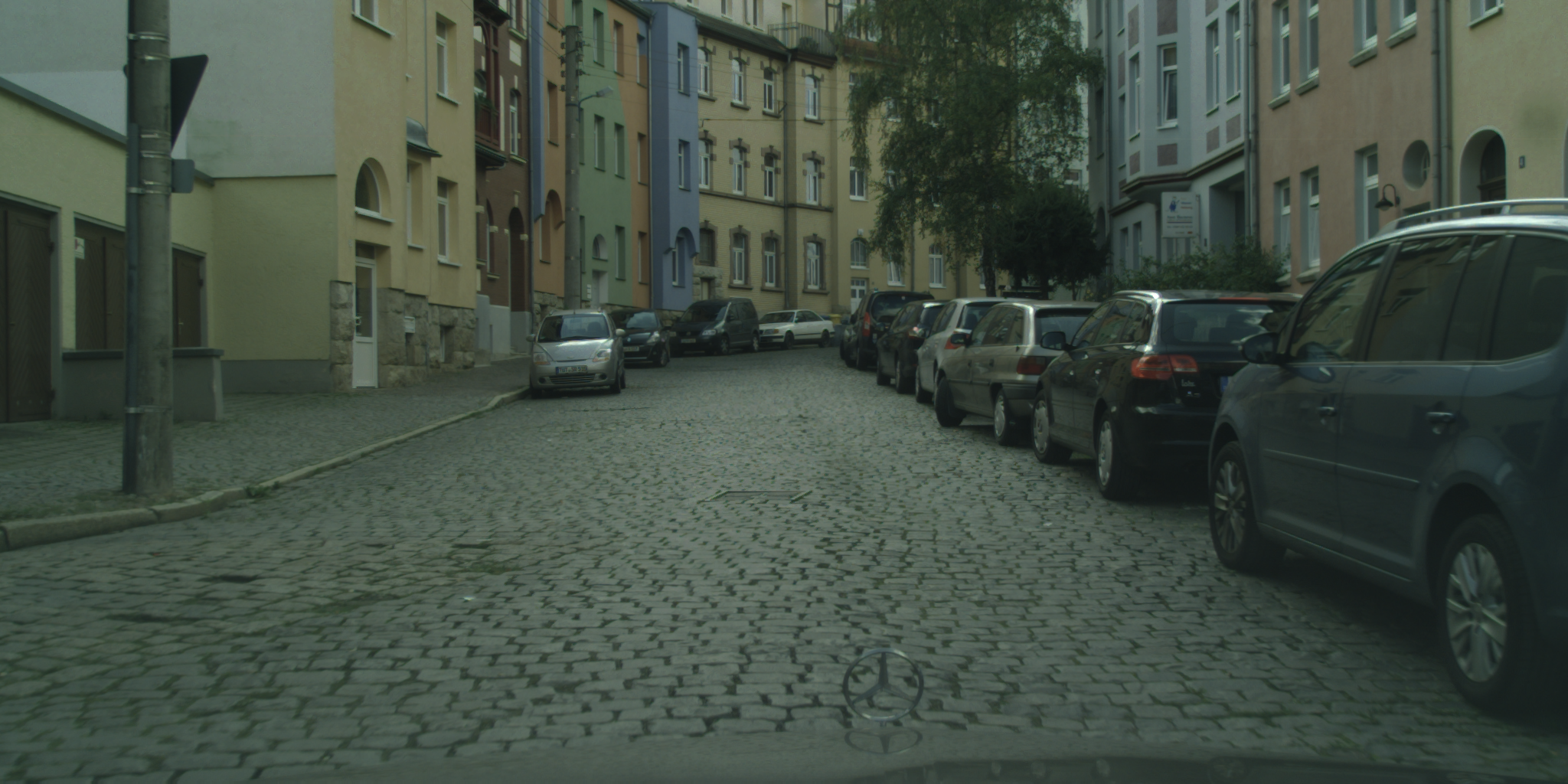}
    \end{subfigure}
    \begin{subfigure}[h!]{0.195\linewidth}
        \centering
        \includegraphics[width=\linewidth, height=0.65\linewidth]{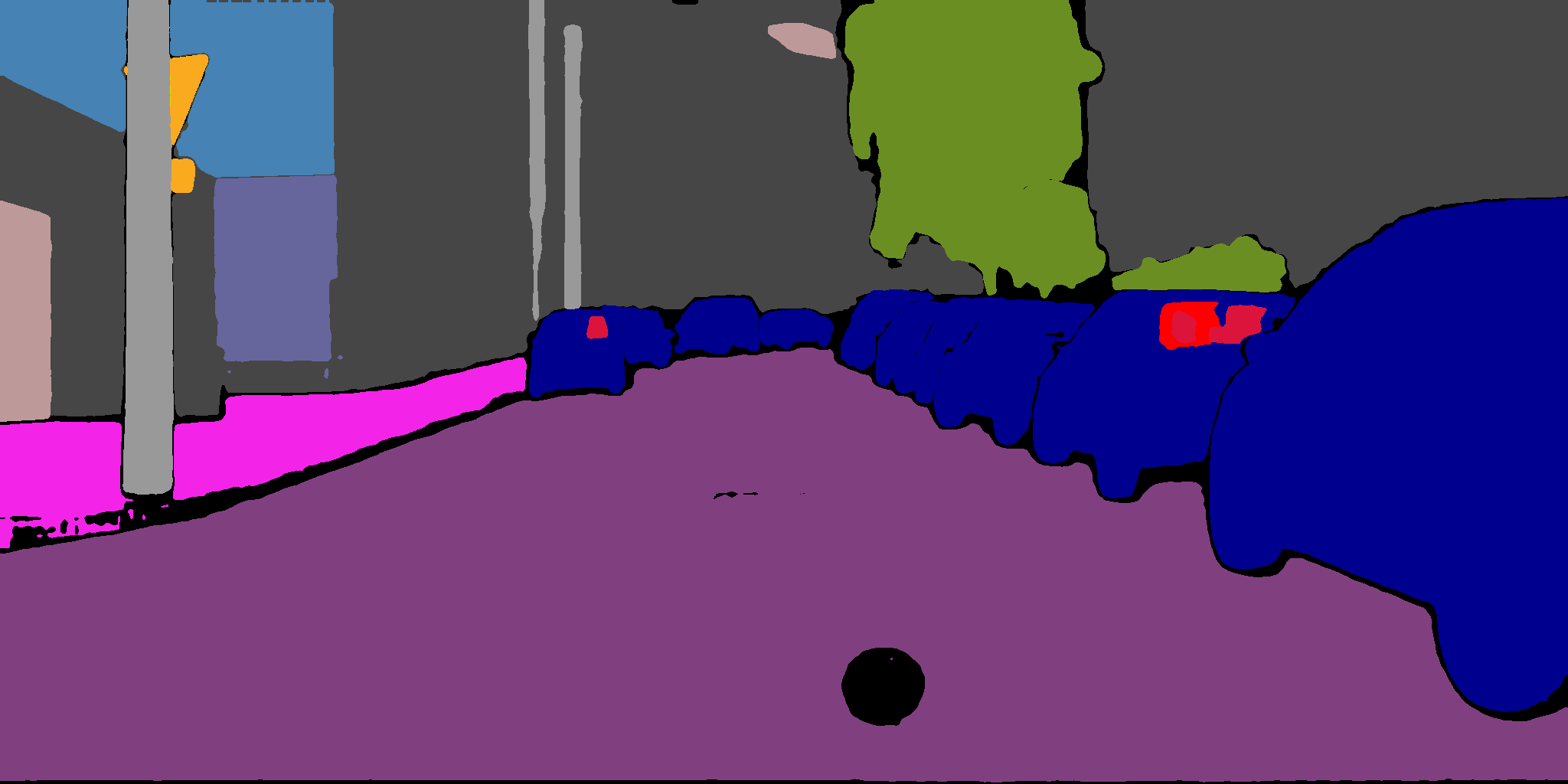}
    \end{subfigure}
    \begin{subfigure}[h!]{0.195\linewidth}
        \centering
        \includegraphics[width=\linewidth, height=0.65\linewidth]{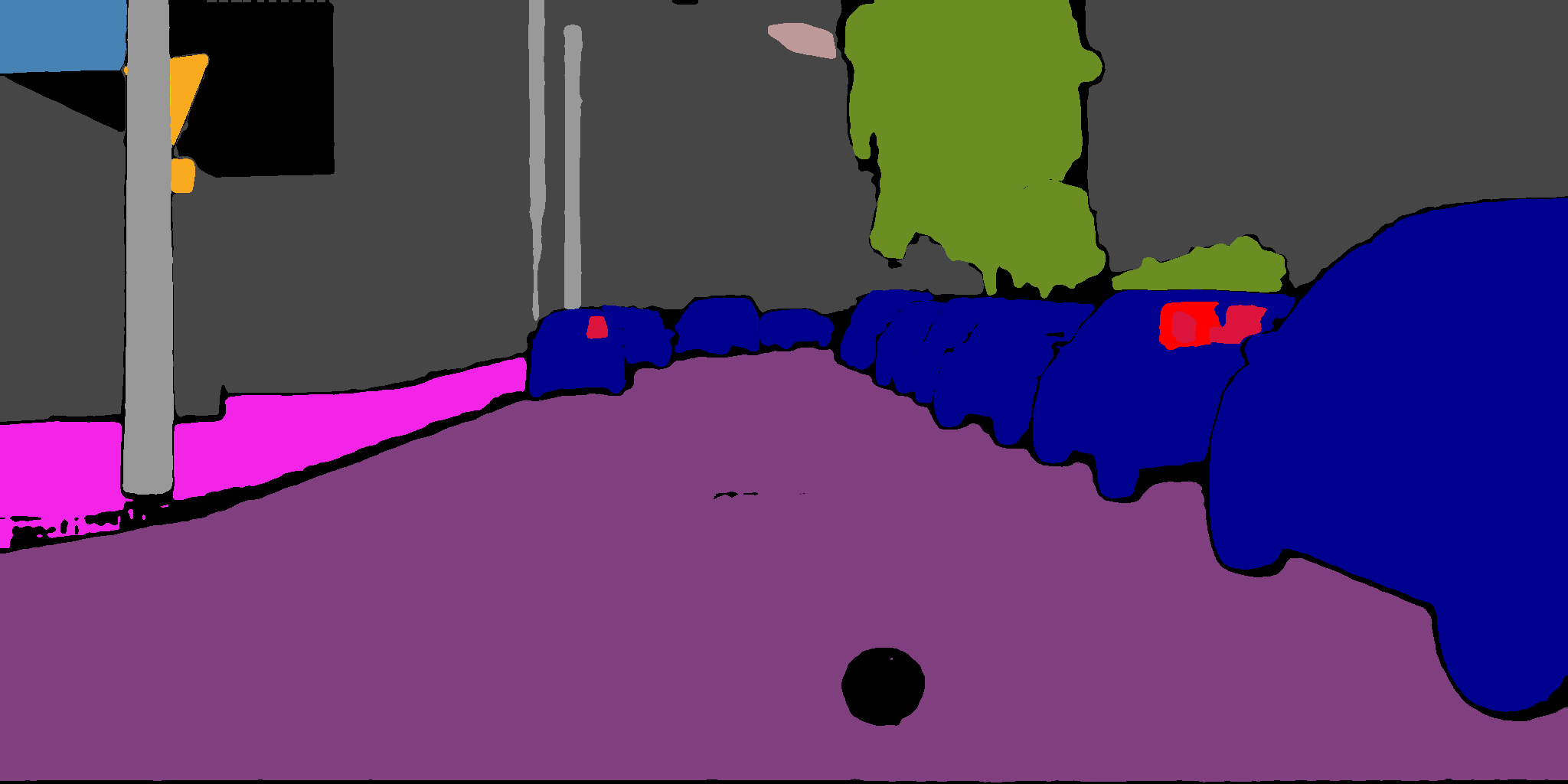}
    \end{subfigure}
    \begin{subfigure}[h!]{0.195\linewidth}
        \centering
        \includegraphics[width=\linewidth, height=0.65\linewidth]{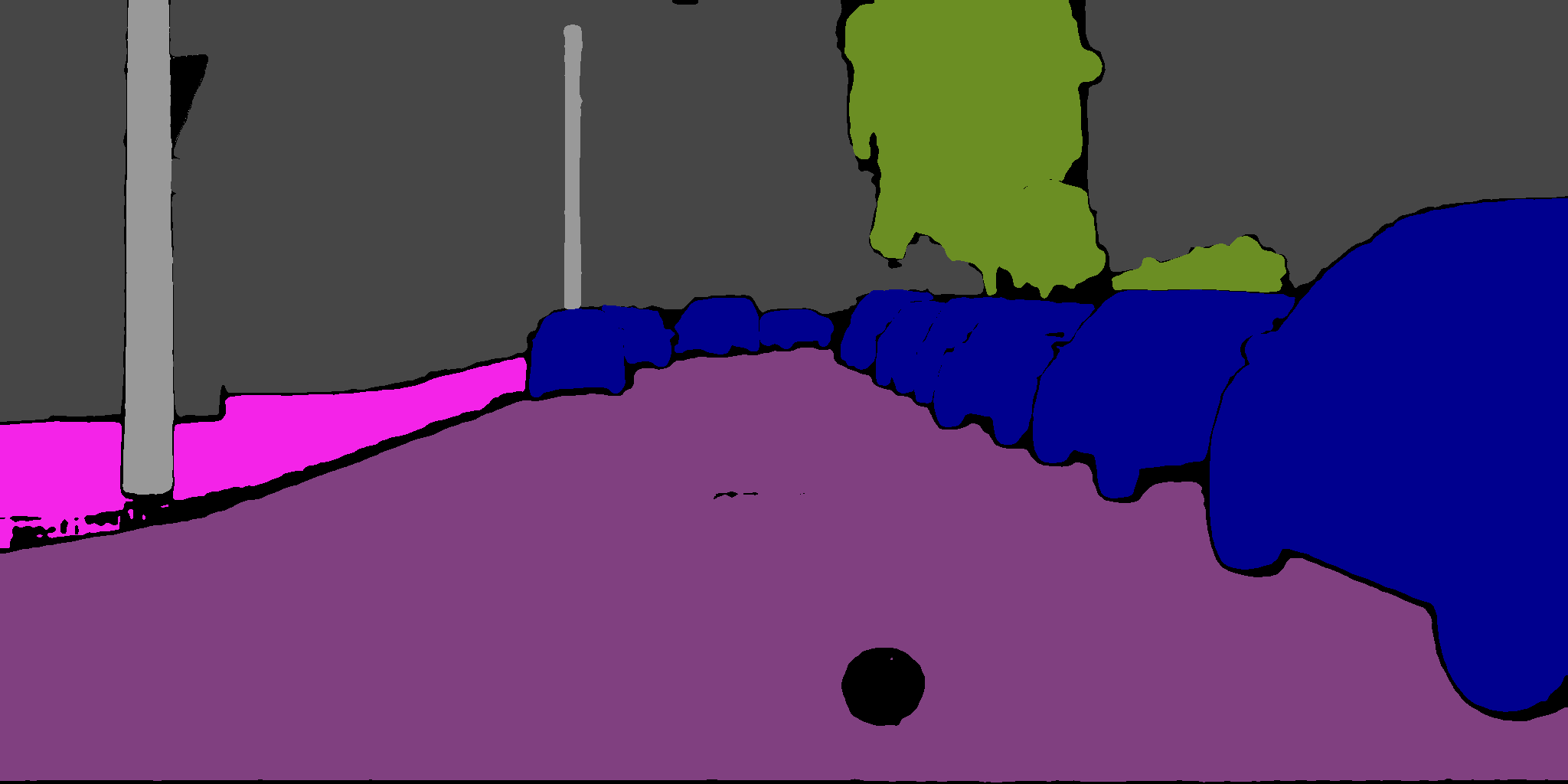}
    \end{subfigure}
    \begin{subfigure}[h!]{0.195\linewidth}
        \centering
        \includegraphics[width=\linewidth, height=0.65\linewidth]{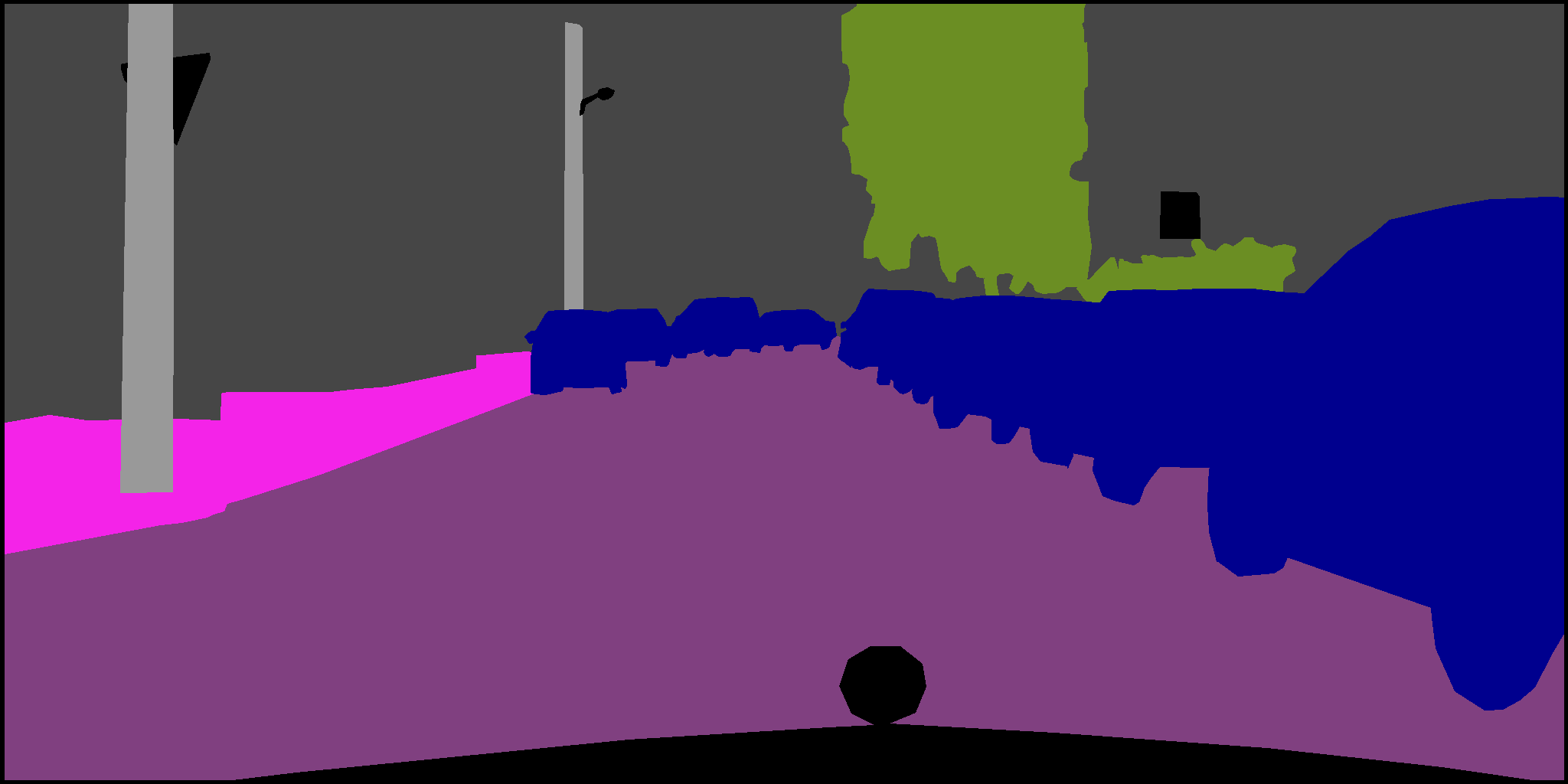}
    \end{subfigure}

    \begin{subfigure}[h!]{0.195\linewidth}
        \centering
        \includegraphics[width=\linewidth, height=0.65\linewidth]{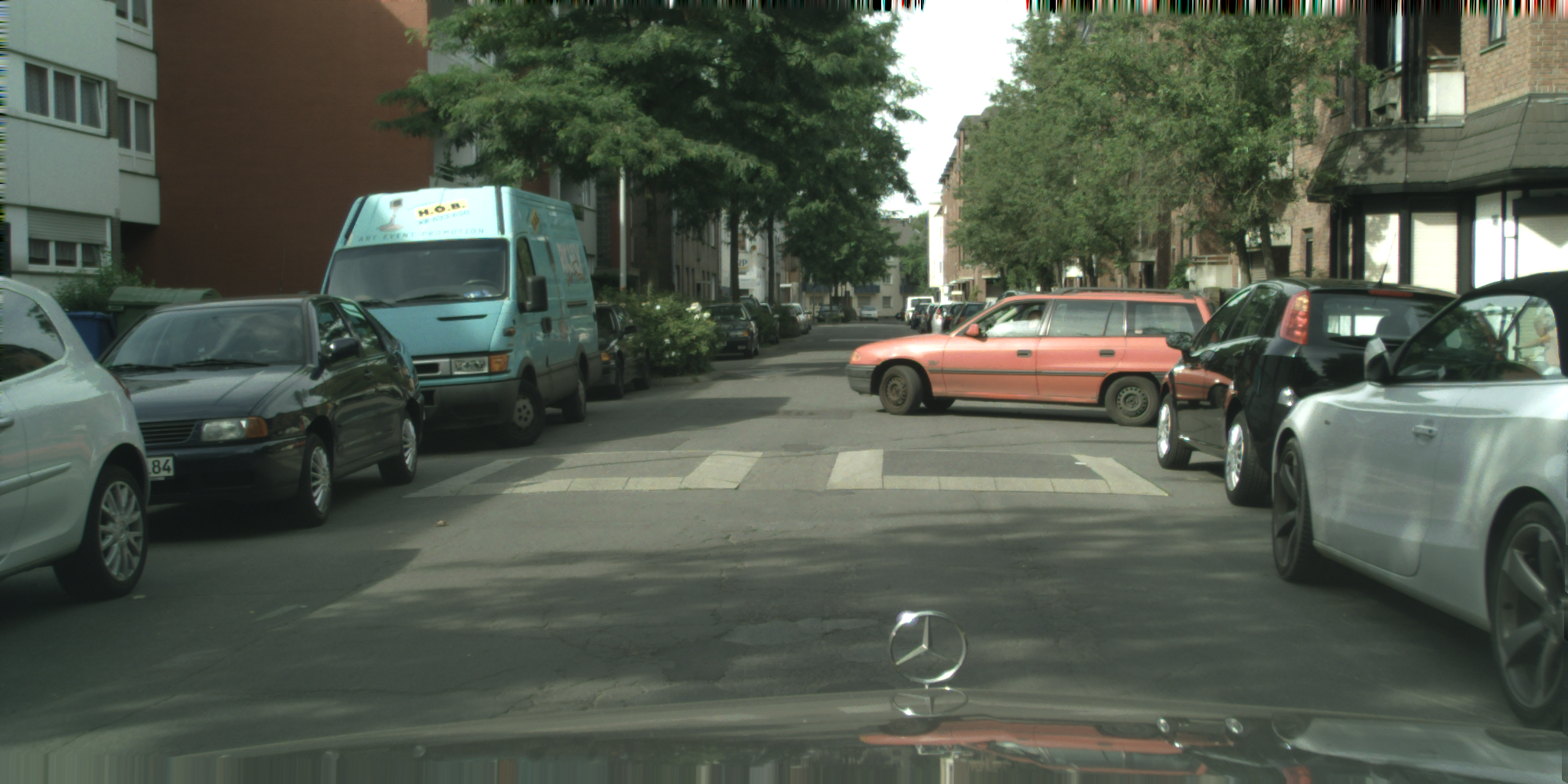}
    \end{subfigure}
    \begin{subfigure}[h!]{0.195\linewidth}
        \centering
        \includegraphics[width=\linewidth, height=0.65\linewidth]{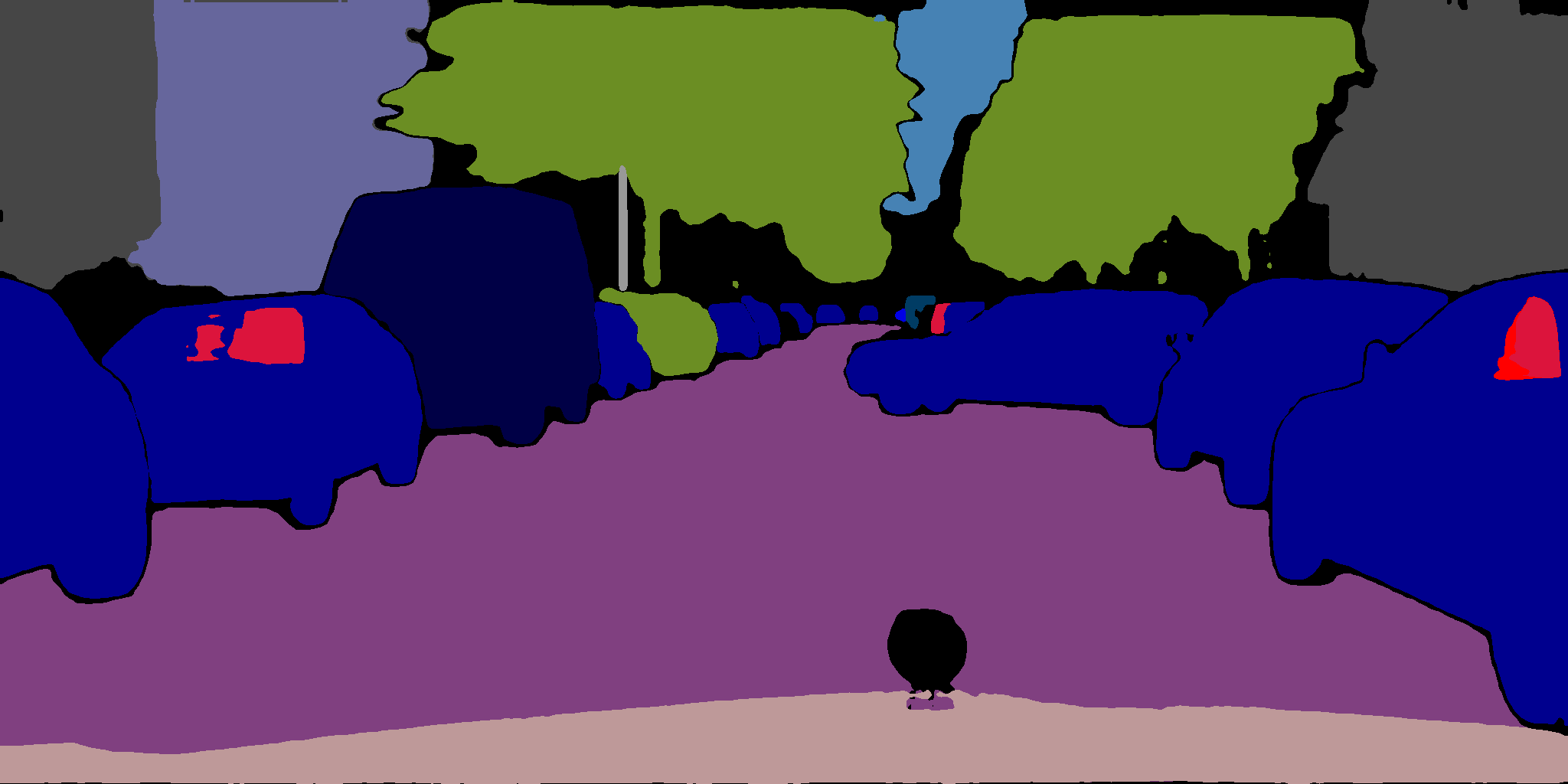}
    \end{subfigure}
    \begin{subfigure}[h!]{0.195\linewidth}
        \centering
        \includegraphics[width=\linewidth, height=0.65\linewidth]{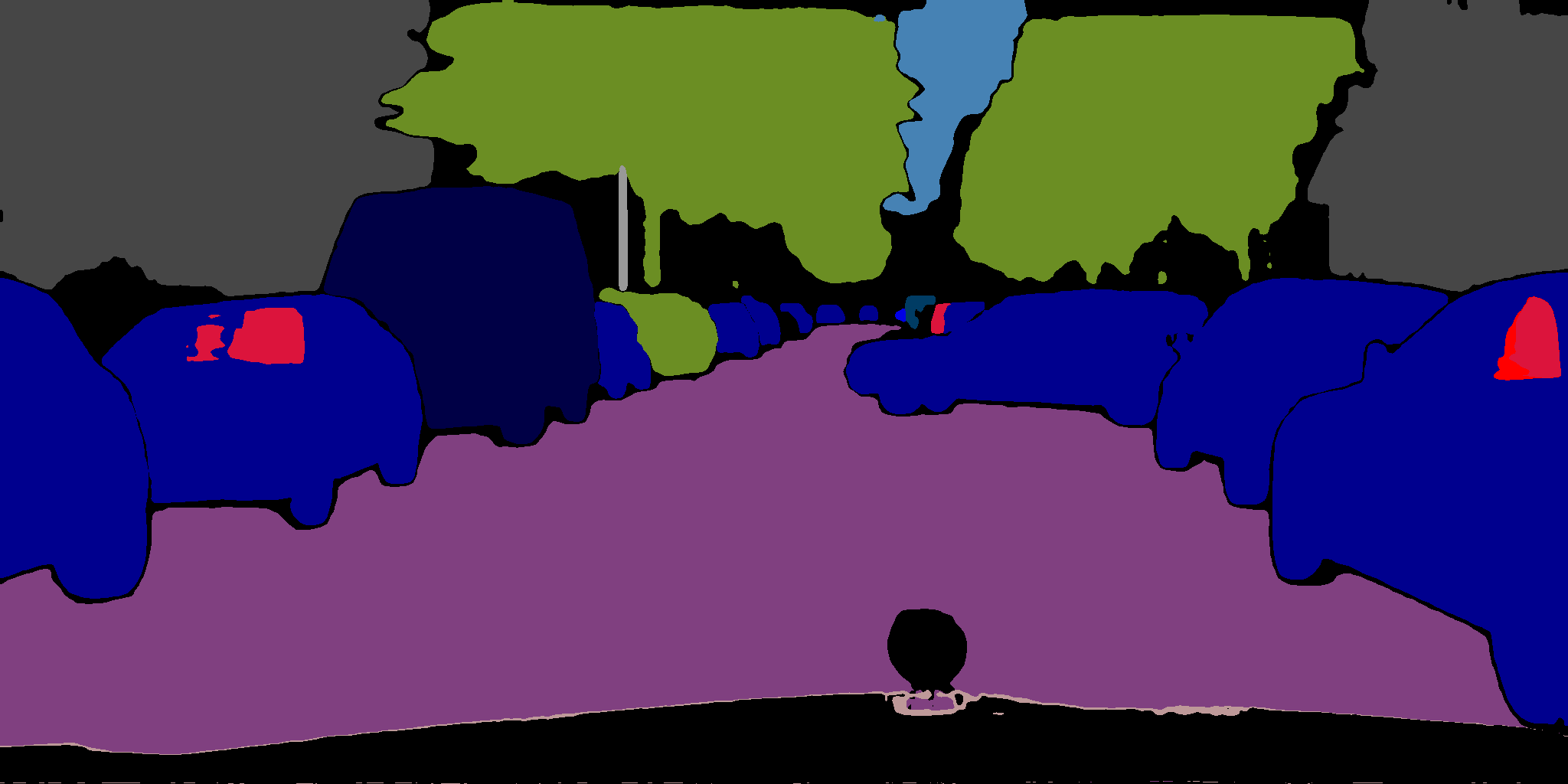}
    \end{subfigure}
    \begin{subfigure}[h!]{0.195\linewidth}
        \centering
        \includegraphics[width=\linewidth, height=0.65\linewidth]{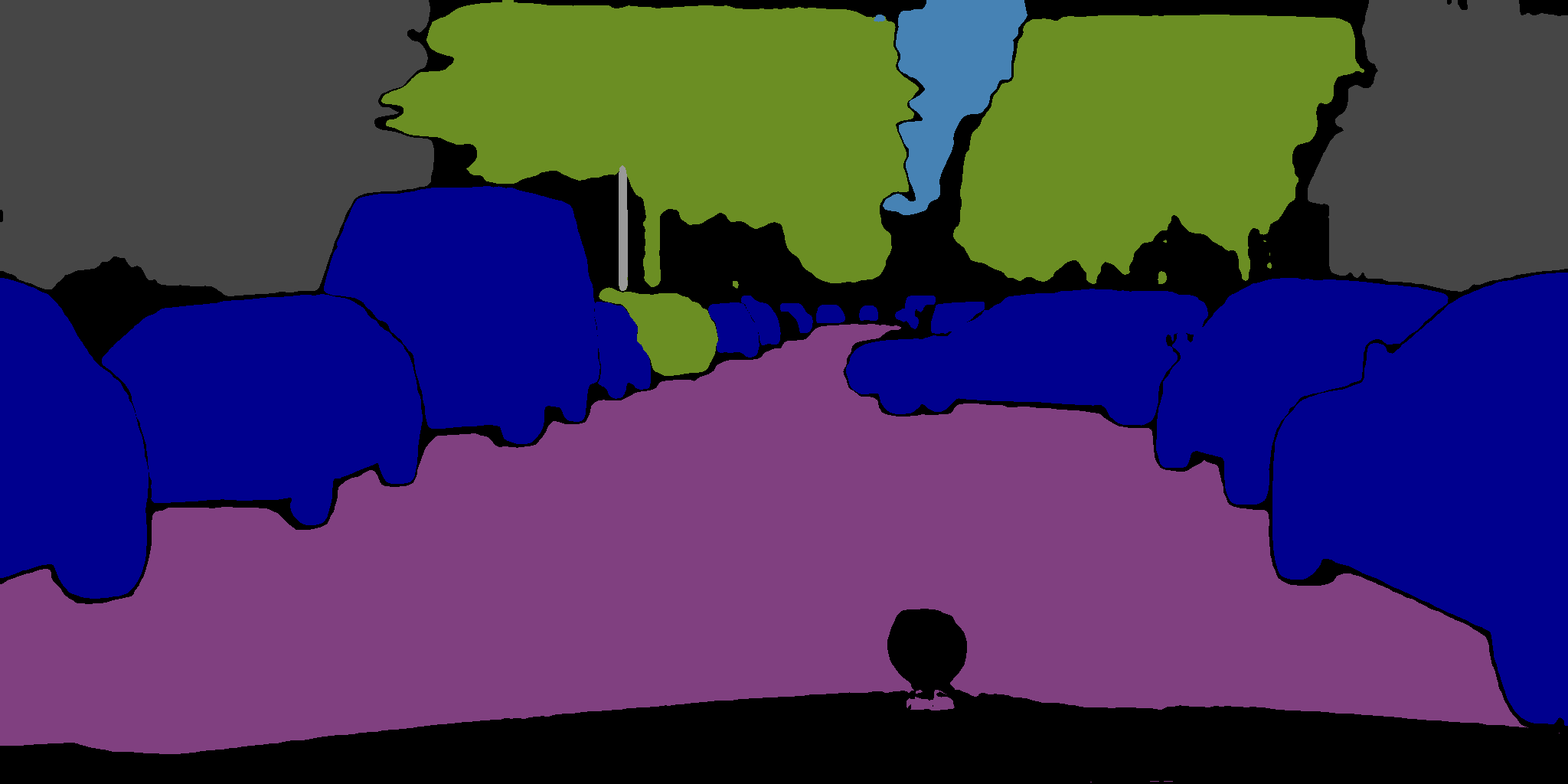}
    \end{subfigure}
    \begin{subfigure}[h!]{0.195\linewidth}
        \centering
        \includegraphics[width=\linewidth, height=0.65\linewidth]{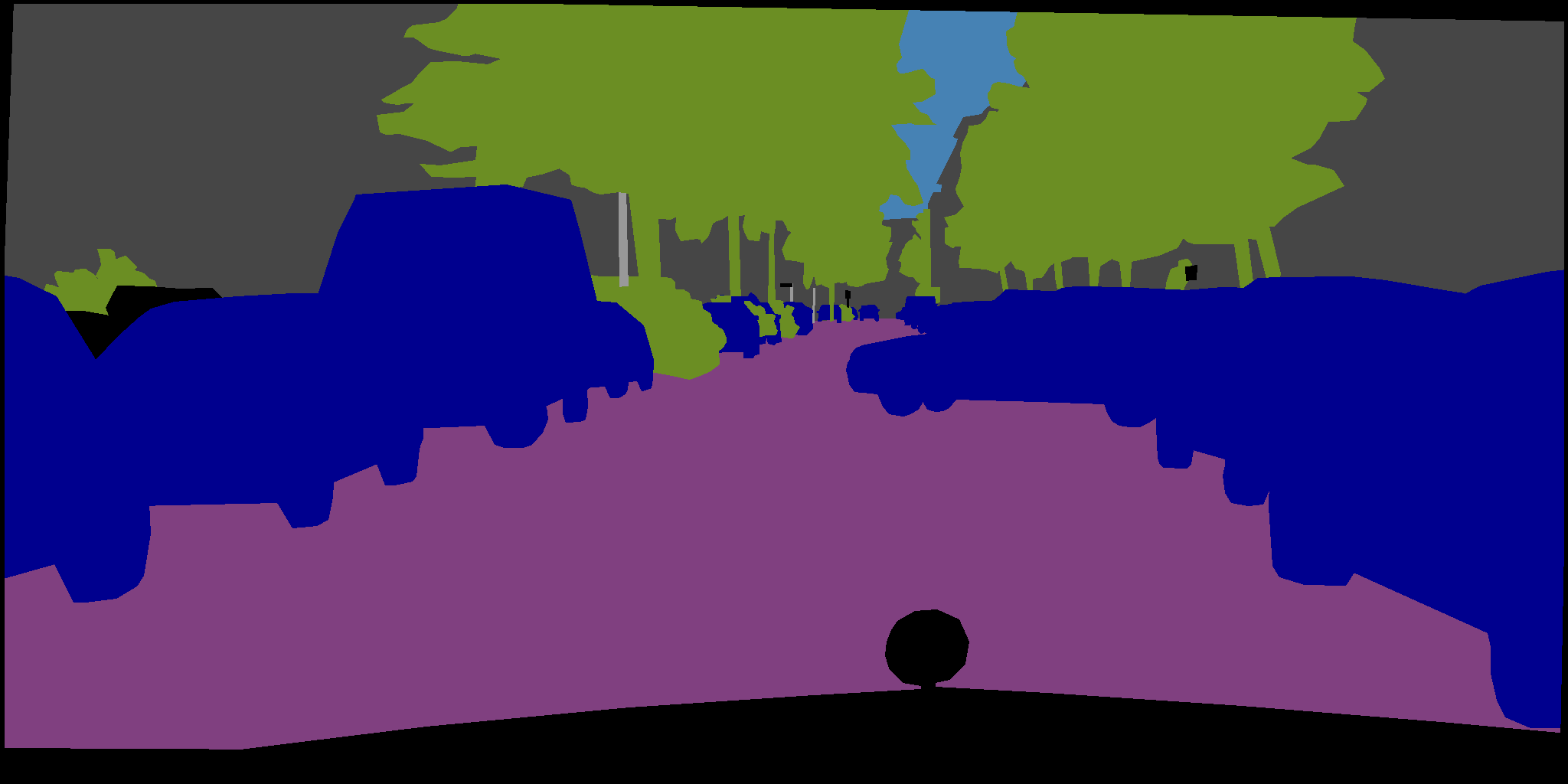}
    \end{subfigure}

    \begin{subfigure}[h!]{0.195\linewidth}
        \centering
        \includegraphics[width=\linewidth, height=0.65\linewidth]{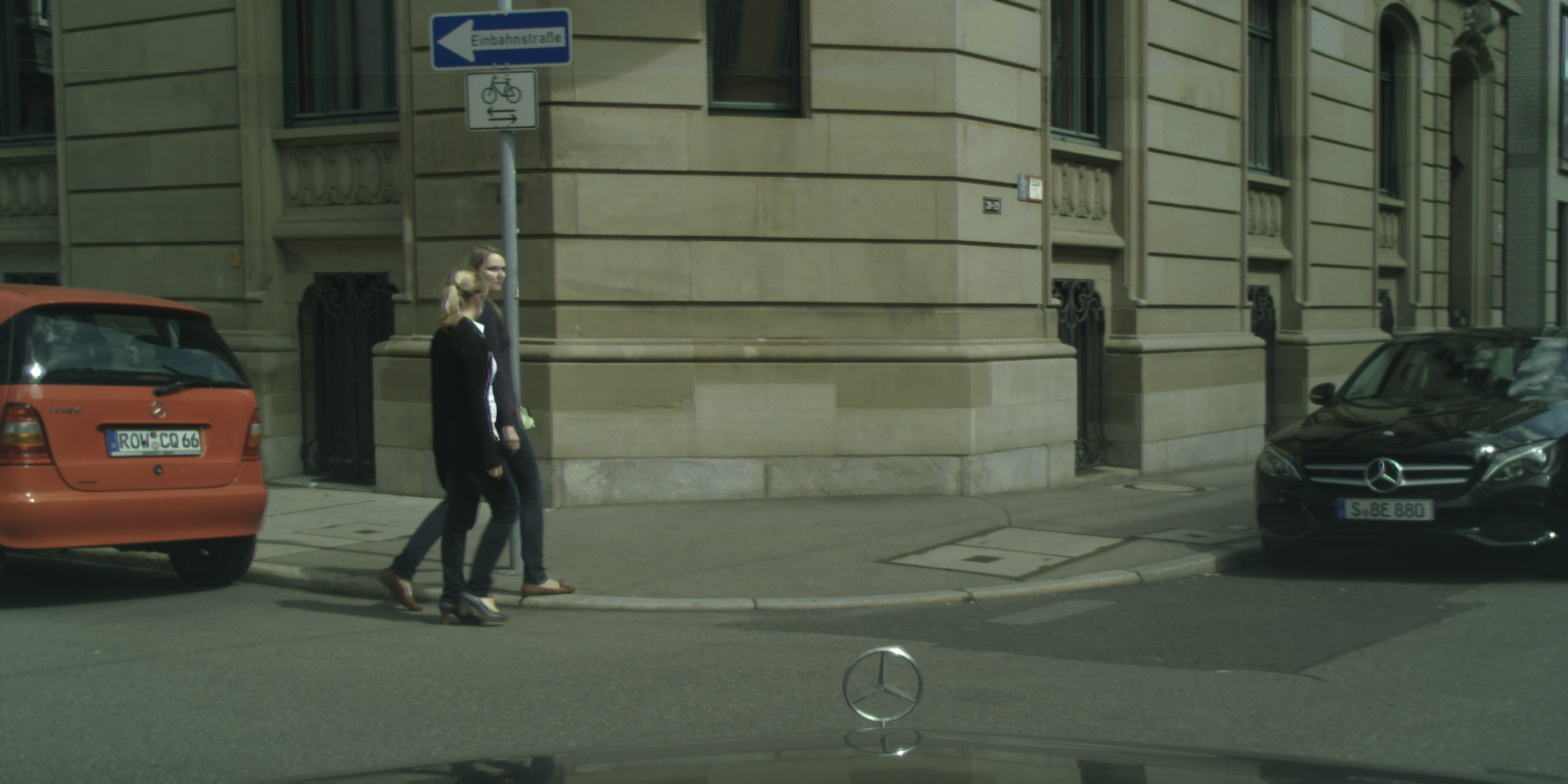}
    \end{subfigure}
    \begin{subfigure}[h!]{0.195\linewidth}
        \centering
        \includegraphics[width=\linewidth, height=0.65\linewidth]{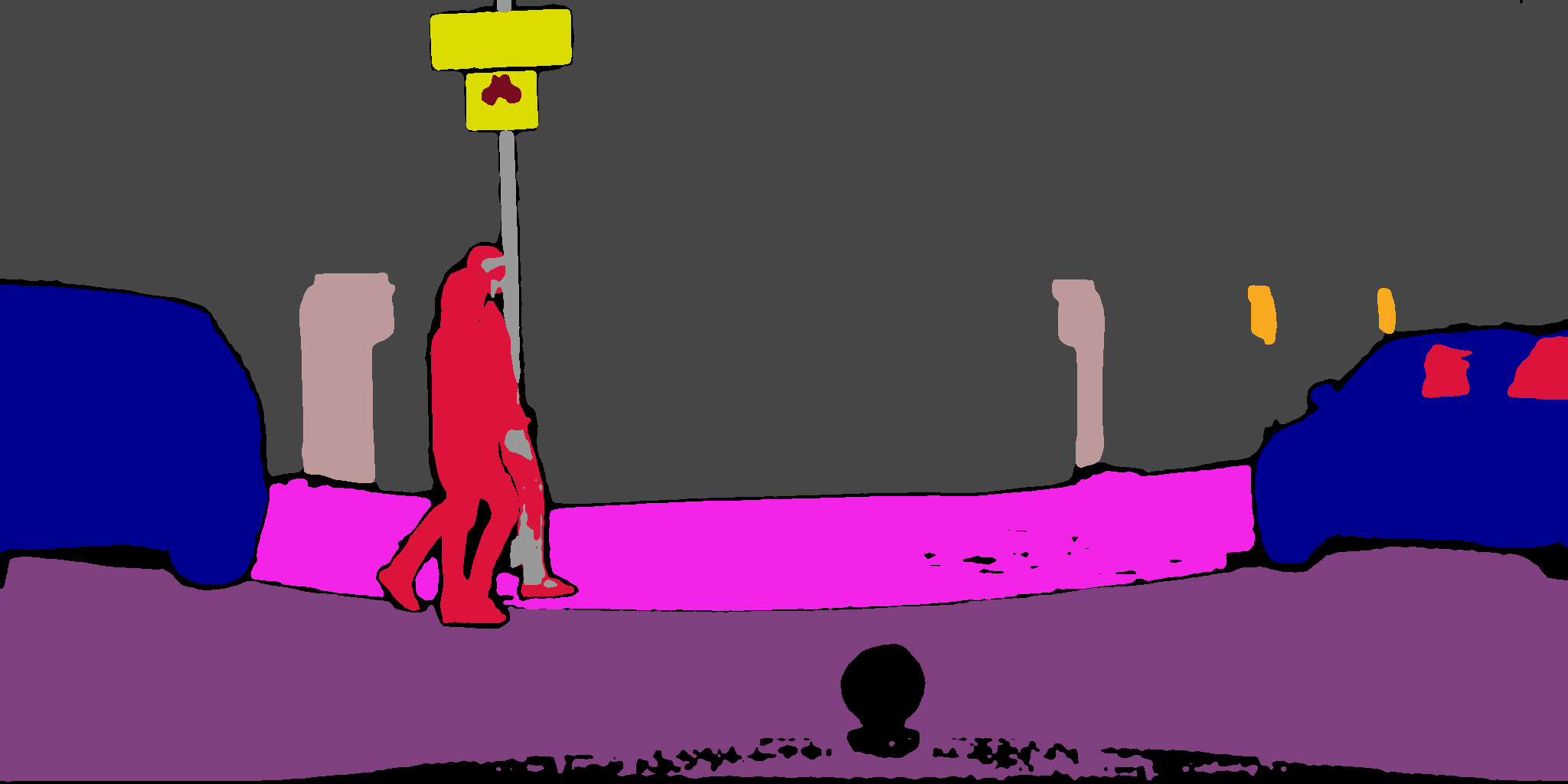}
    \end{subfigure}
    \begin{subfigure}[h!]{0.195\linewidth}
        \centering
        \includegraphics[width=\linewidth, height=0.65\linewidth]{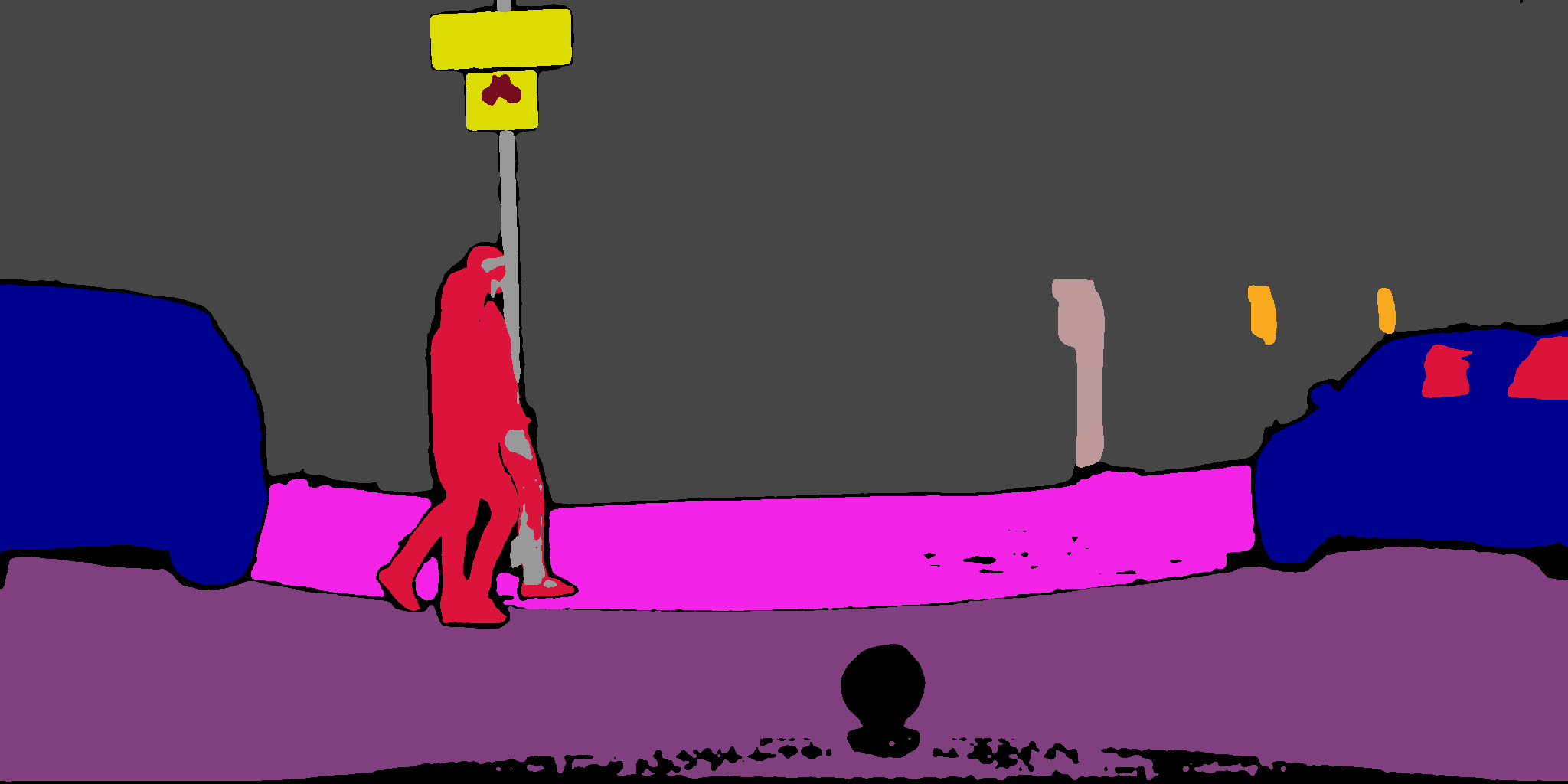}
    \end{subfigure}
    \begin{subfigure}[h!]{0.195\linewidth}
        \centering
        \includegraphics[width=\linewidth, height=0.65\linewidth]{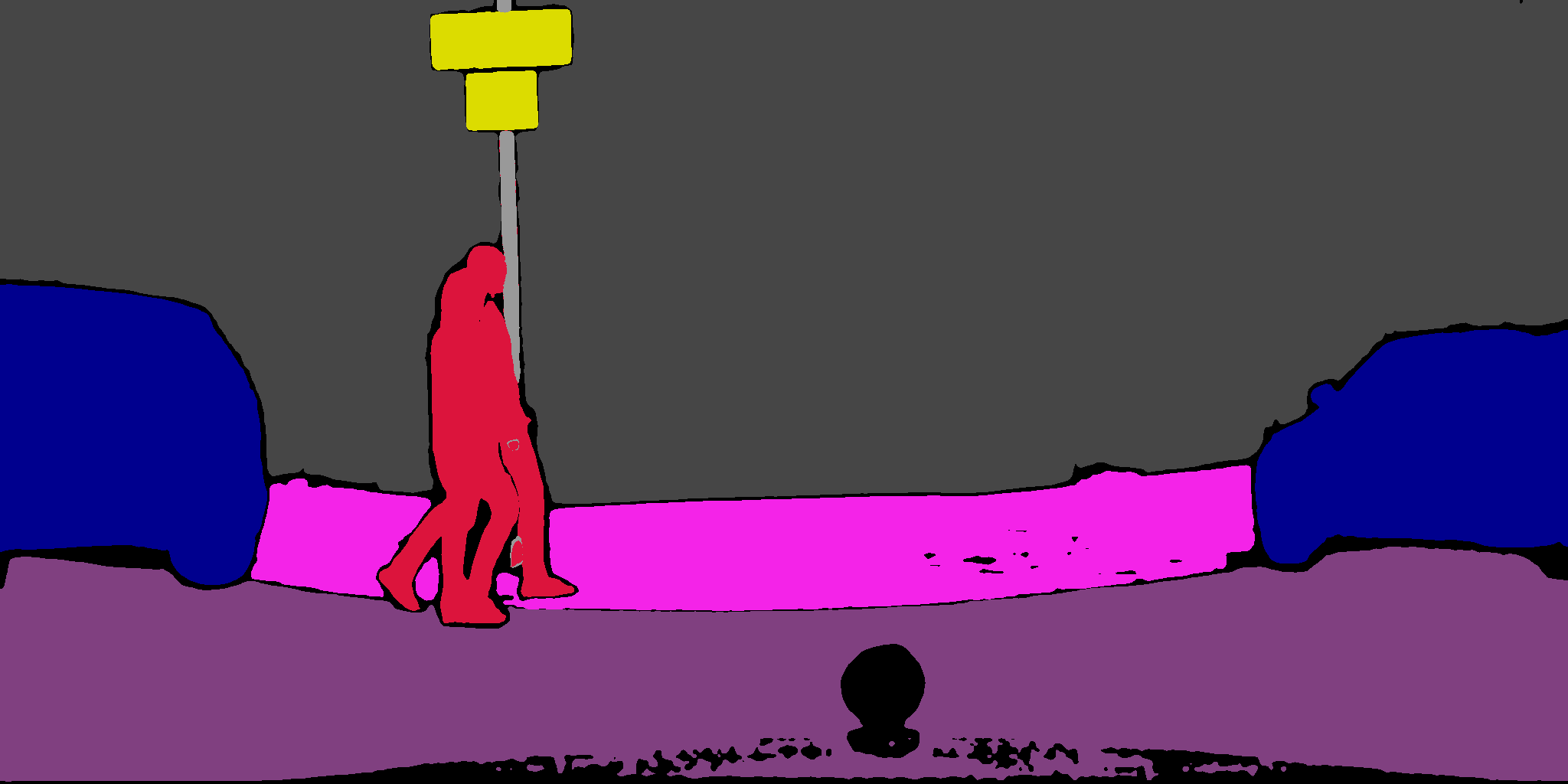}
    \end{subfigure}
    \begin{subfigure}[h!]{0.195\linewidth}
        \centering
        \includegraphics[width=\linewidth, height=0.65\linewidth]{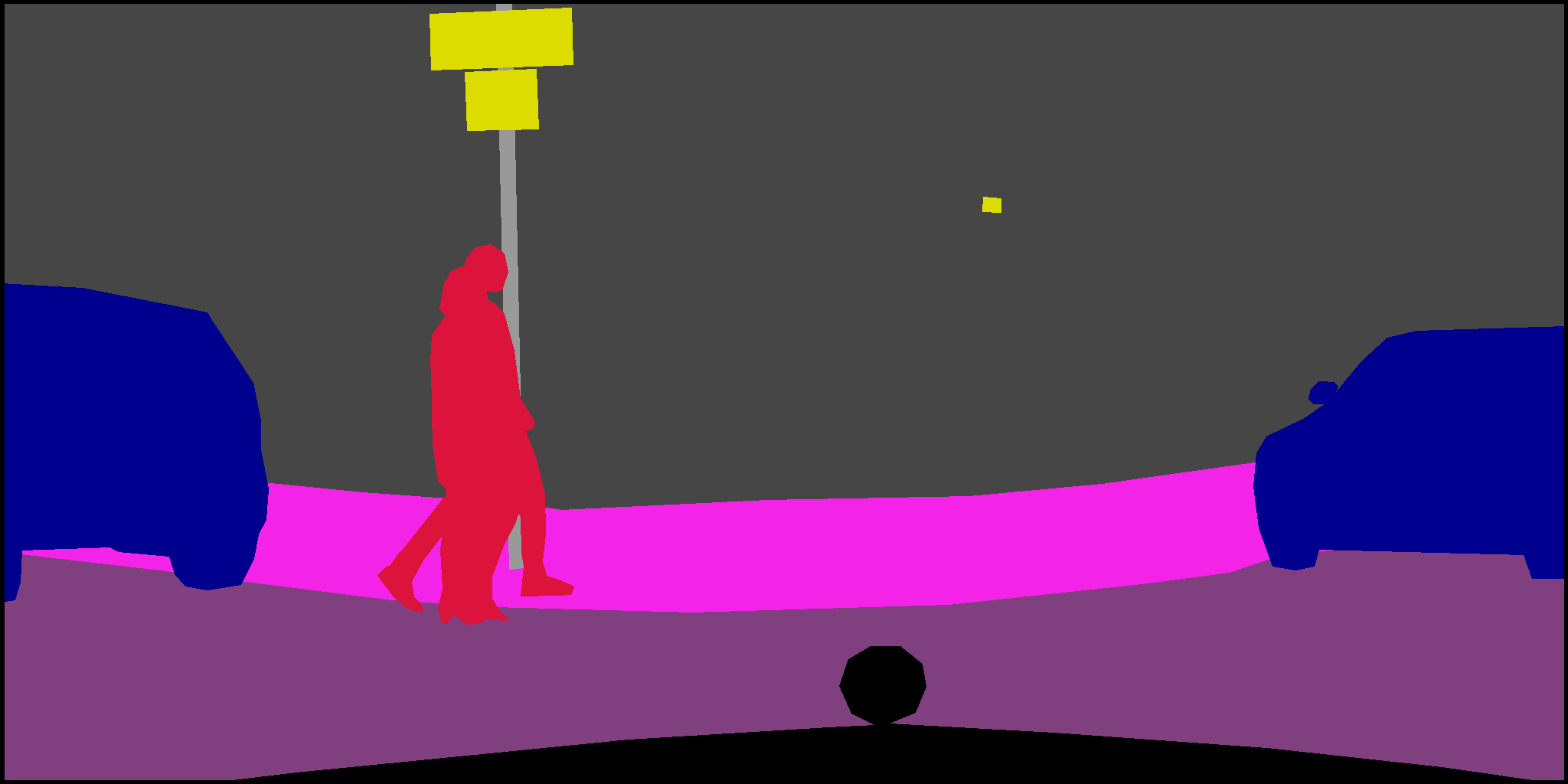}
    \end{subfigure}

    \begin{subfigure}[h!]{0.195\linewidth}
        \centering
        \includegraphics[width=\linewidth, height=0.65\linewidth]{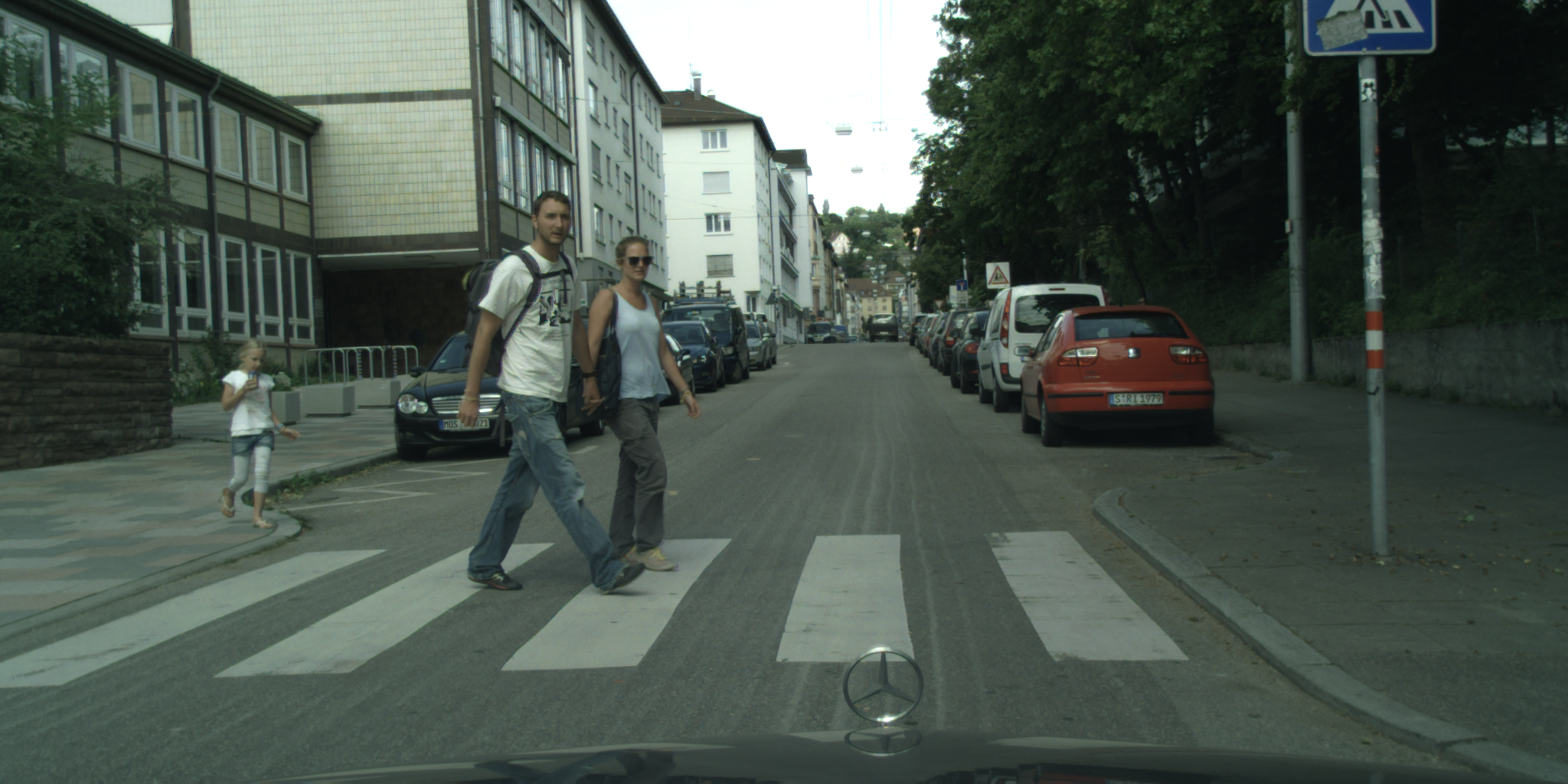}
    \end{subfigure}
    \begin{subfigure}[h!]{0.195\linewidth}
        \centering
        \includegraphics[width=\linewidth, height=0.65\linewidth]{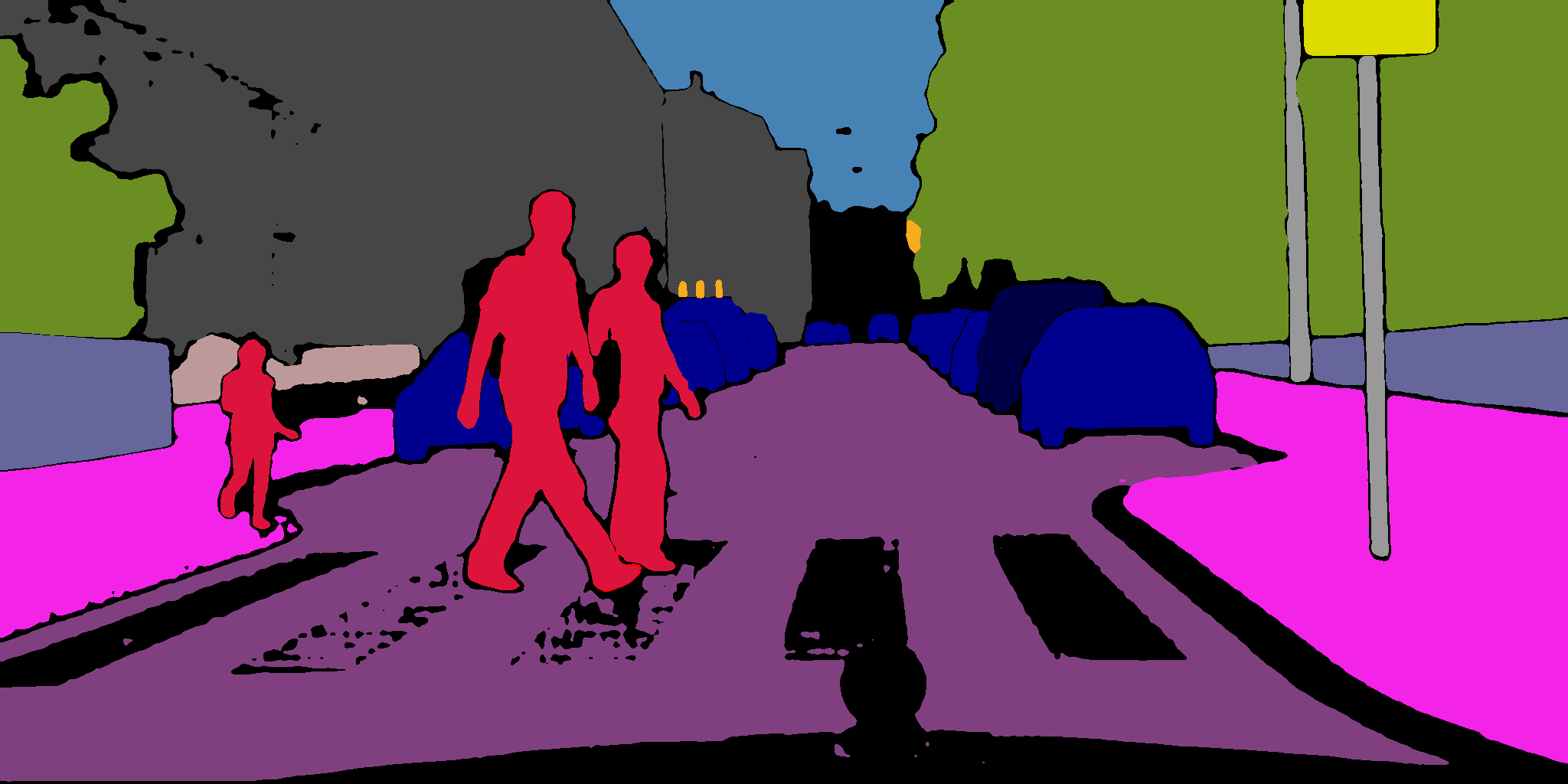}
    \end{subfigure}
    \begin{subfigure}[h!]{0.195\linewidth}
        \centering
        \includegraphics[width=\linewidth, height=0.65\linewidth]{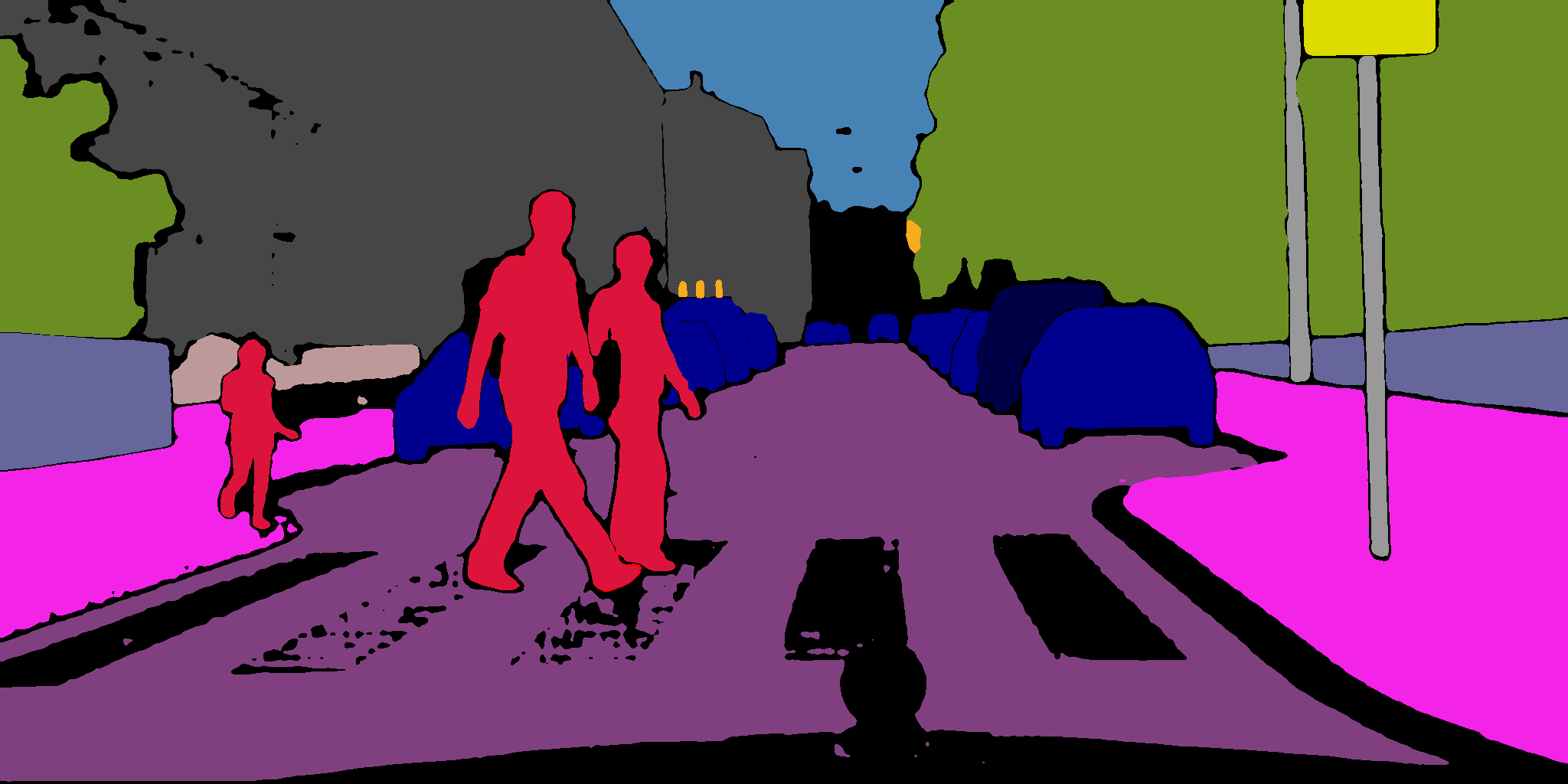}
    \end{subfigure}
    \begin{subfigure}[h!]{0.195\linewidth}
        \centering
        \includegraphics[width=\linewidth, height=0.65\linewidth]{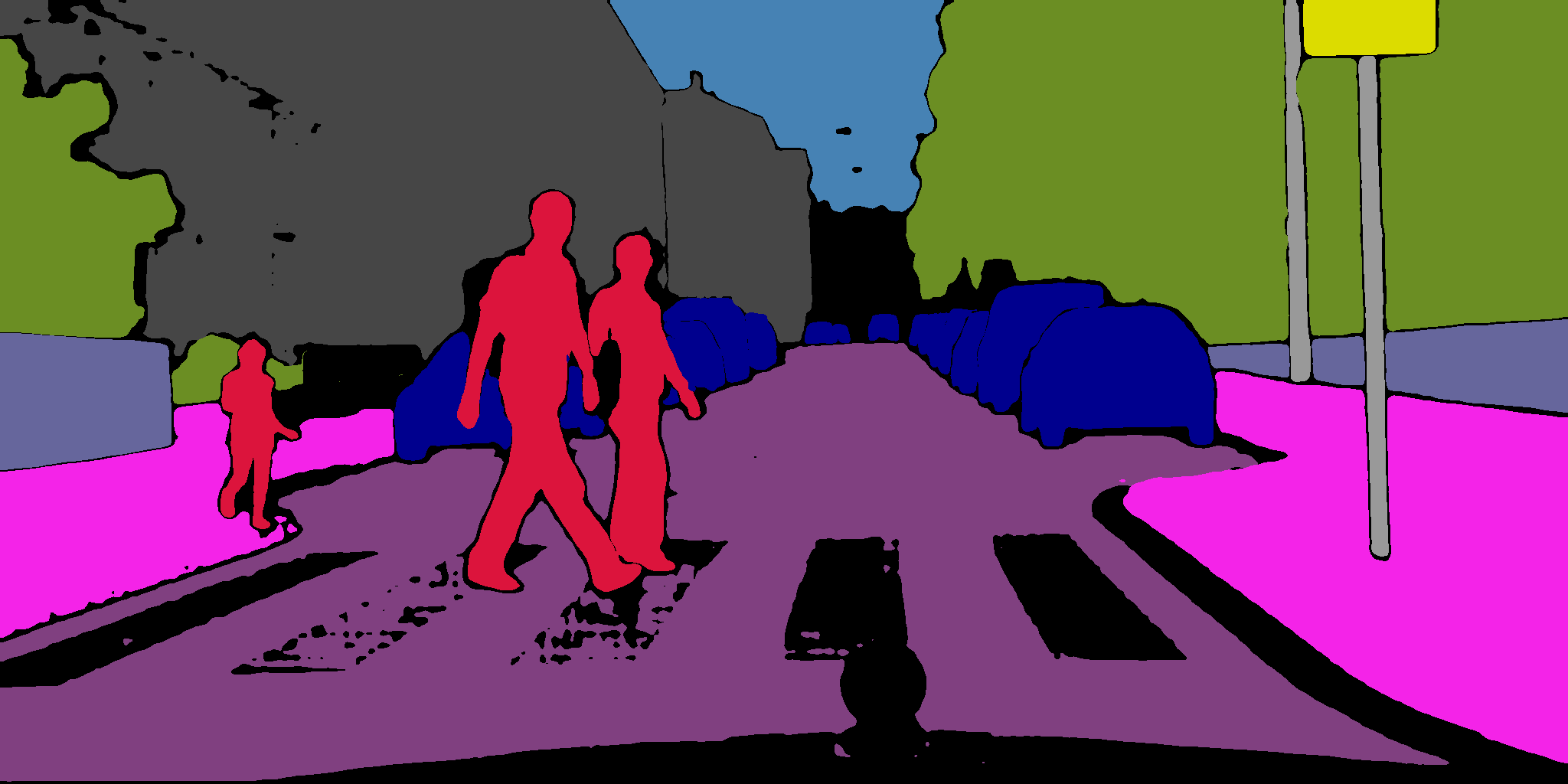}
    \end{subfigure}
    \begin{subfigure}[h!]{0.195\linewidth}
        \centering
        \includegraphics[width=\linewidth, height=0.65\linewidth]{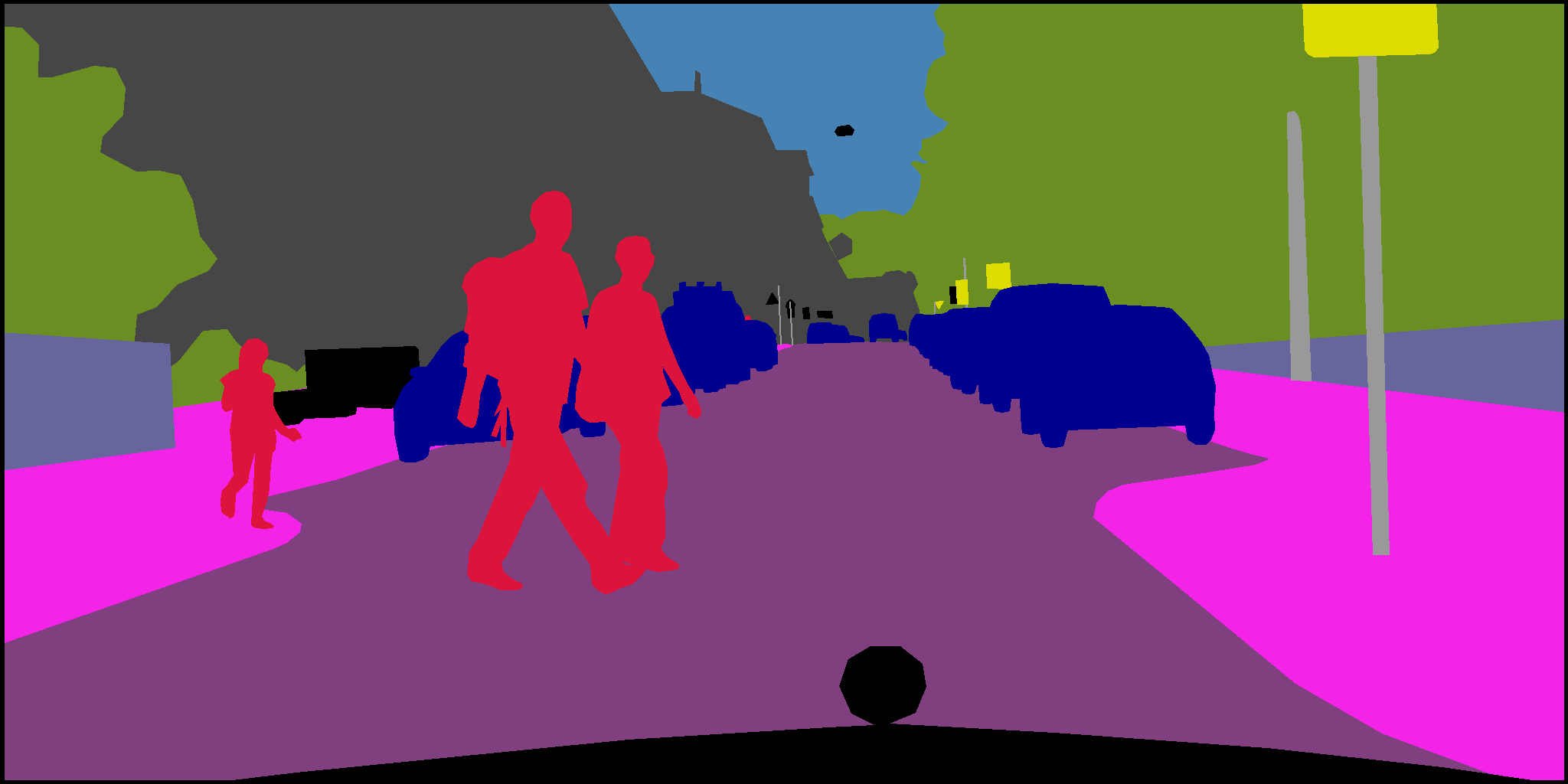}
    \end{subfigure} 
    
    \begin{subfigure}[h!]{0.195\linewidth}
        \centering
        \includegraphics[width=\linewidth, height=0.65\linewidth]{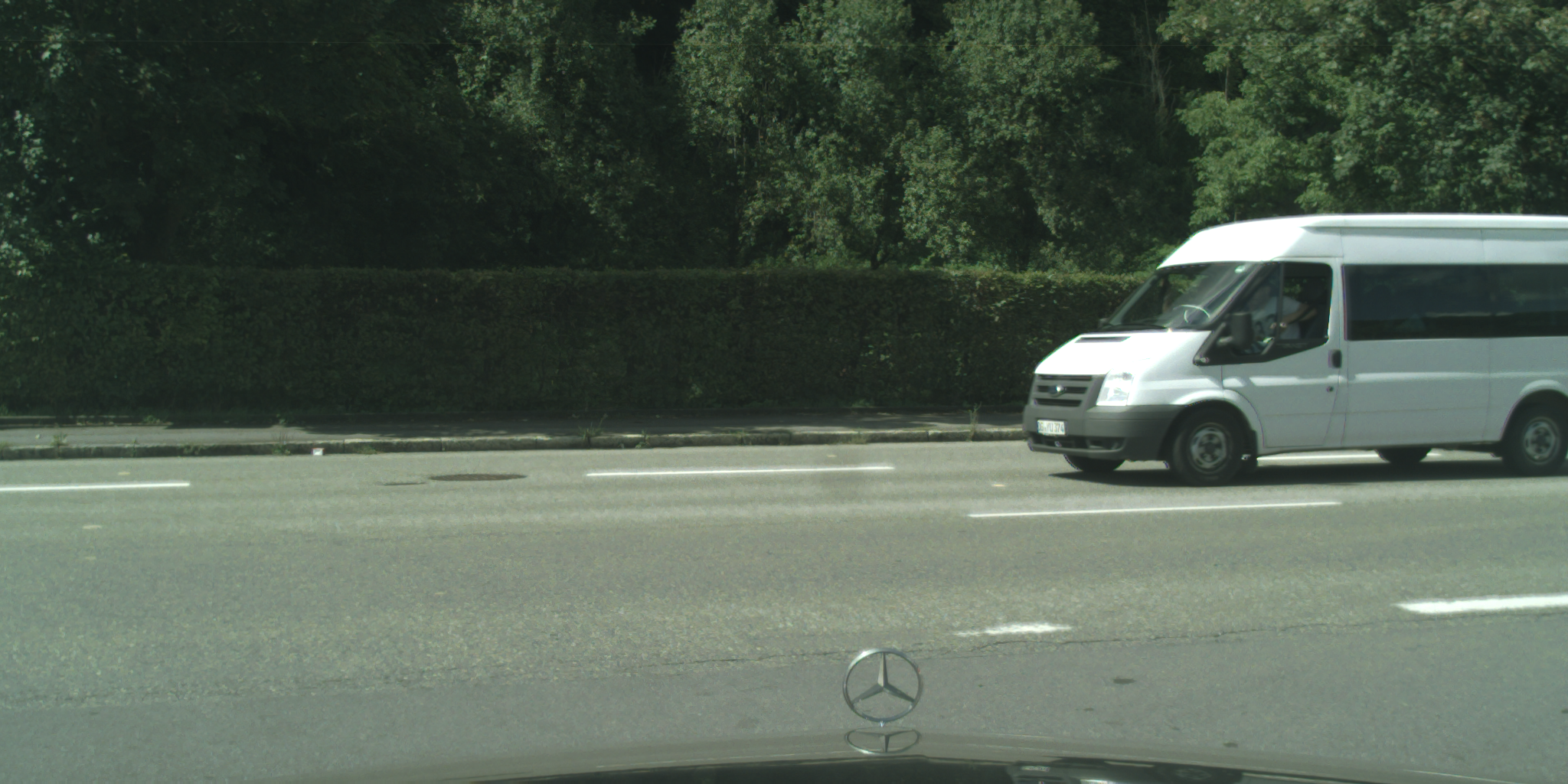}
        \caption{Unlabeled Image}
    \end{subfigure}
    \begin{subfigure}[h!]{0.195\linewidth}
        \centering
        \includegraphics[width=\linewidth, height=0.65\linewidth]{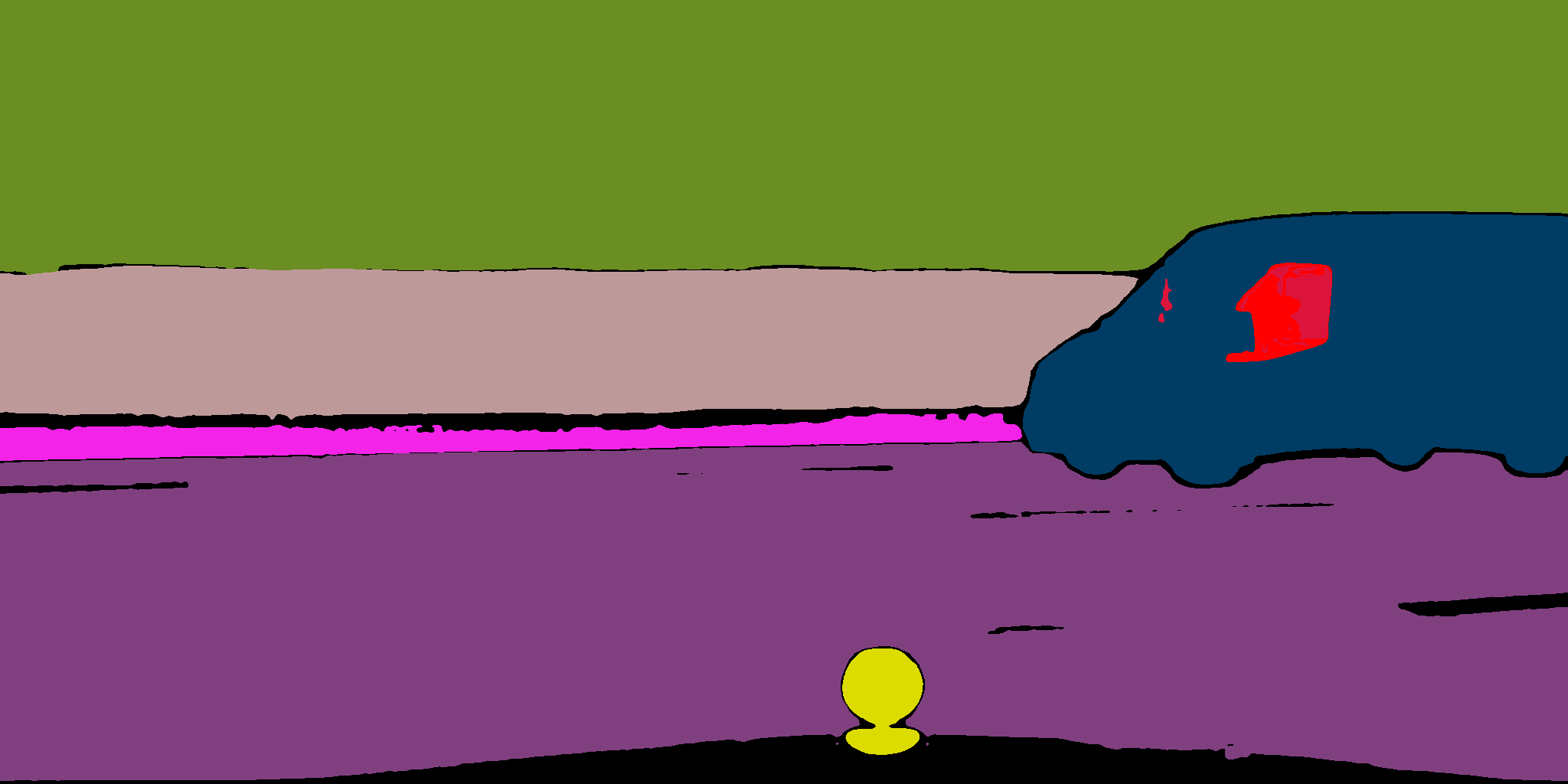}
        \caption{Grounded SAM}
    \end{subfigure}
    \begin{subfigure}[h!]{0.195\linewidth}
        \centering
        \includegraphics[width=\linewidth, height=0.65\linewidth]{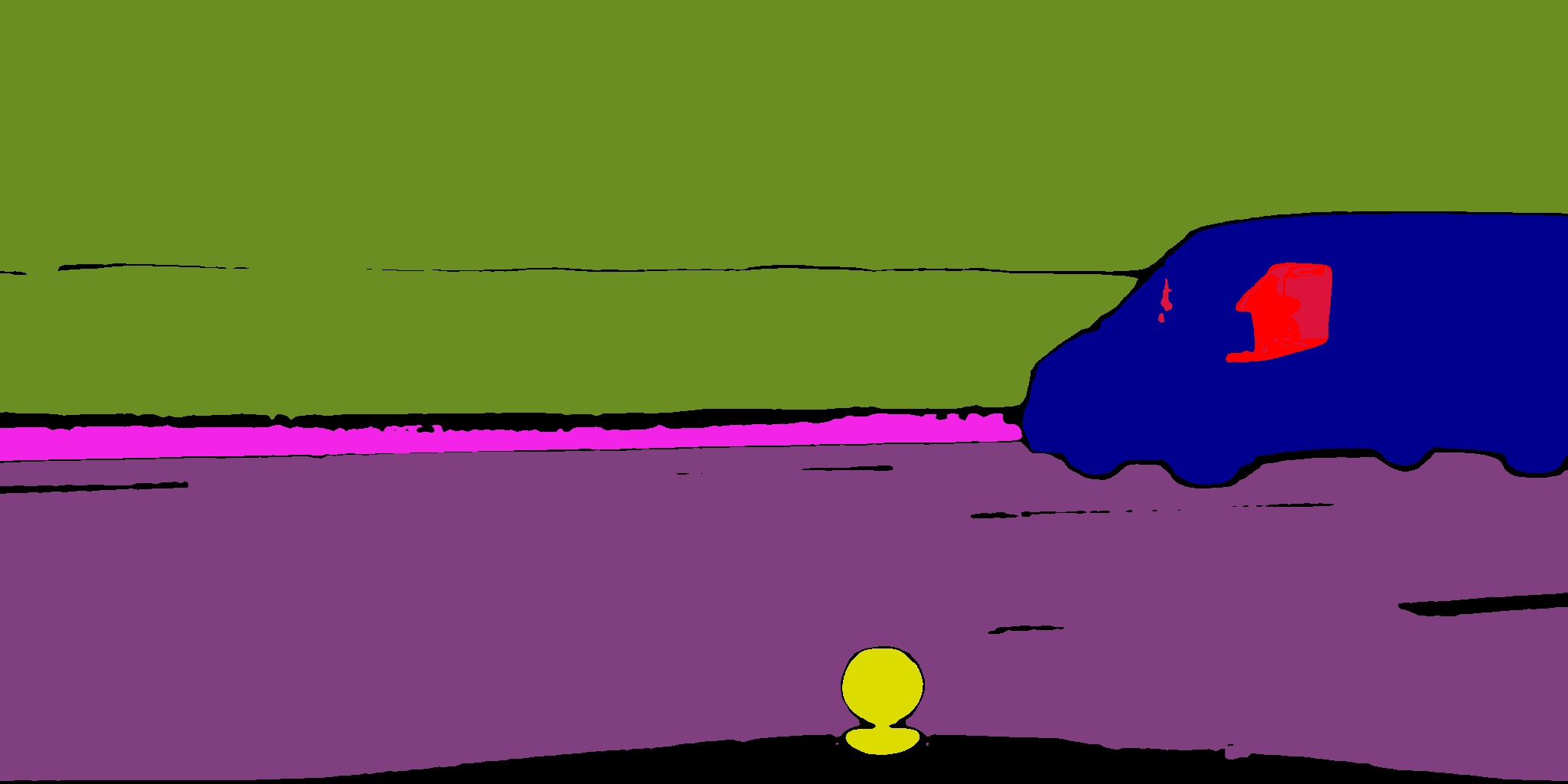}
        \caption{ALC~\cite{kim2024active}}
    \end{subfigure}
    \begin{subfigure}[h!]{0.195\linewidth}
        \centering
        \includegraphics[width=\linewidth, height=0.65\linewidth]{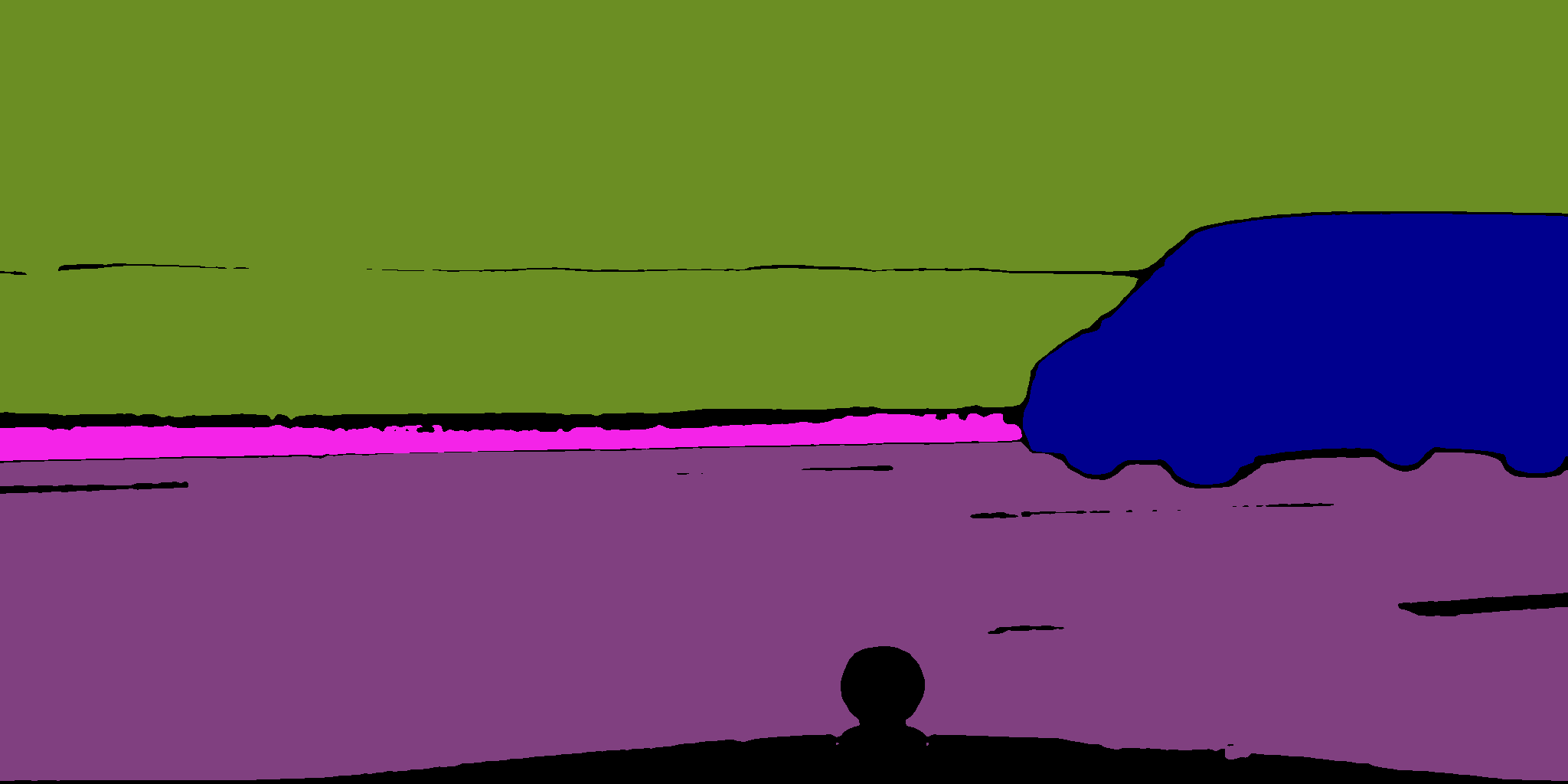}
        \caption{A\(^2\)LC (Ours)}
    \end{subfigure}
    \begin{subfigure}[h!]{0.195\linewidth}
        \centering
        \includegraphics[width=\linewidth, height=0.65\linewidth]{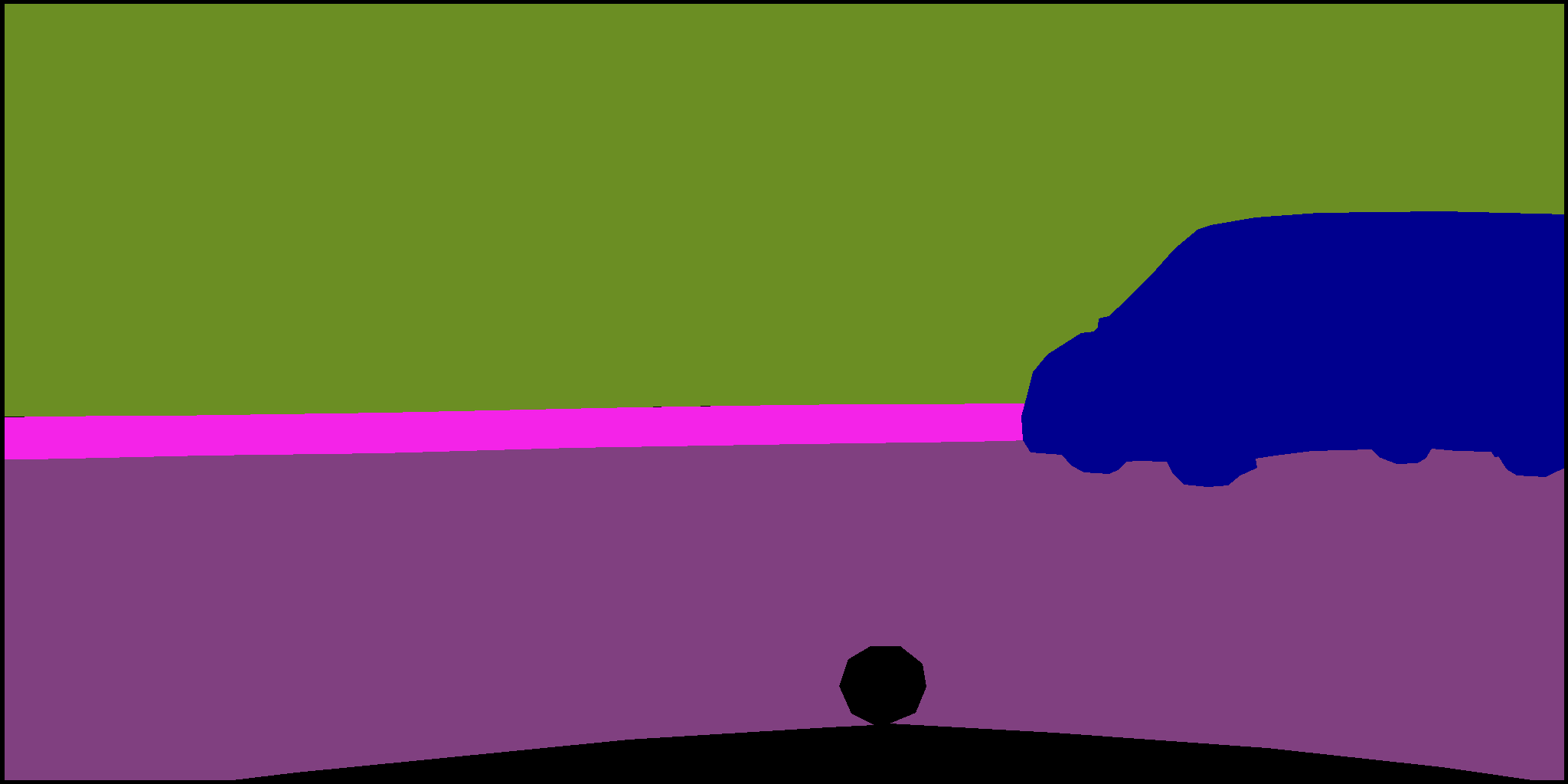}
        \caption{Ground Truth}
    \end{subfigure}
    \caption{Qualitative results of constructed pseudo-labels. In the first row, the masks mislabeled as \textit{rider (pure red)} are accurately corrected to \textit{person (darker red)}. In the second and third row, the masks mislabeled as \textit{traffic light (orange)} are accurately corrected to \textit{traffic sign (yellow)}.}
    \label{fig:qual_1_2}
\end{minipage}
\end{figure*}

Figures~\ref{fig:qual_1_1} and \ref{fig:qual_1_2} present the final pseudo-labels generated by our proposed framework (column 4), in comparison with the initial pseudo-labels from Grounded SAM (column 2) and those generated by previous research (column 3), starting from fully unlabeled images with no ground truth available.

\section{Limitations}
While our proposed framework introduces an automated correction stage that significantly enhances performance with minimal human intervention, it does not entirely eliminate the need for human involvement.

\section{Broader Impacts}
A\(^2\)LC addresses the practical challenge of obtaining labeled dataset by introducing a dual-source label correction framework, promoting maximum automation with minimal human effort. 
This is particularly helpful for researchers who struggle to train deep learning models for downstream tasks in the absence of labeled dataset.
However, unlike manual and fine-grained human correction, model-driven correction based on its internal knowledge cannot guarantee perfect accuracy. 
Therefore, users should be aware of potential inaccuracies when deploying this framework in practice.

\end{document}